\documentclass[applsci,article,submit,pdftex,moreauthors]{Definitions/mdpi} 

\firstpage{1} 
\pubvolume{1}
\issuenum{1}
\articlenumber{0}
\pubyear{2023}
\copyrightyear{2023}
\datereceived{ } 
\daterevised{ } % Comment out if no revised date
\dateaccepted{ } 
\datepublished{ } 
\hreflink{https://doi.org/} % If needed use \linebreak
\usepackage[linesnumbered,ruled,vlined]{algorithm2e}
\usepackage{algorithmic}

\usepackage{amssymb,amsmath,amsfonts,caption}

\usepackage{cases}
\usepackage{float,subfig}
\usepackage{caption}
\usepackage{url}
\usepackage{here}
\usepackage{stfloats} 
\usepackage{braket}
\usepackage{bm}
\usepackage{hhline}
\usepackage{colortbl}

\newcommand{\argmin}{\mathop{\rm \text{arg~min}}\limits}

\newcolumntype{C}[1]{>{\hfil}m{#1}<{\hfil}}
\usepackage{lscape} % figure/table rotation

\DeclareMathAlphabet      {\mathbf}{OT1}{cmr}{bx}{n}

\Title{Class-wise Classifier Design Capable of  Continual Learning using Adaptive Resonance Theory-based Topological Clustering}

\TitleCitation{Class-wise Classifier Design Capable of  Continual Learning using Adaptive Resonance Theory-based Topological Clustering}

\Author{Naoki Masuyama $^{1,*}$\orcidA{}, Yusuke Nojima $^{1}$, Farhan Dawood $^{2}$, and Zongying Liu $^{3}$\orcidD{}}

\AuthorNames{Naoki Masuyama, Yusuke Nojima, Farhan Dawood, and Zongying Liu}

\AuthorCitation{Masuyama, N.; Nojima, Y.; Dawood, F.; Liu, Z.}
\address{%
$^{1}$ \quad Osaka Metropolitan University; {\{masuyama, nojima\}@omu.ac.jp}\\
$^{2}$ \quad University of Central Punjab; farhan.dawood@ucp.edu.pk \\
$^{3}$ \quad Dalian Maritime University; liuzongying@dlmu.edu.cn 
}

\corres{Correspondence: masuyama@omu.ac.jp}

\abstract{This paper proposes a supervised classification algorithm capable of continual learning by utilizing an Adaptive Resonance Theory (ART)-based growing self-organizing clustering algorithm. The ART-based clustering algorithm is theoretically capable of continual learning, and the proposed algorithm independently applies it to each class of training data for generating classifiers. Whenever an additional training data set from a new class is given, a new ART-based clustering will be defined in a different learning space. Thanks to the above-mentioned features, the proposed algorithm realizes continual learning capability. Simulation experiments showed that the proposed algorithm has superior classification performance compared with state-of-the-art clustering-based classification algorithms capable of continual learning.}


\begin{document}

%%%%%%%%%%%%%%%%%%%%%%%%%%%%%%%%%%%%%%%%%%
% \setcounter{section}{-1} %% Remove this when starting to work on the template.
\section{Introduction}
\label{sec:introduction}
With the recent development of IoT technology, a wide variety of big data from different domains have become easily available. In order to make effective use of such big data, many studies have been conducted in various research fields.

One of the major challenges in the field of machine learning is to achieve continual learning like the human brain by computational algorithms. In general, it is difficult for many computational algorithms to avoid catastrophic forgetting, i.e., previously learned knowledge is collapsed due to learning new knowledge \cite{mccloskey89}, which makes the algorithms difficult to realize efficient learning for big data with increasing amount and types of data. Recently, therefore, computational algorithms capable of continual learning attract much attention as a promising approach for continually and efficiently extracting knowledge from big data \cite{parisi19}. 

Continual learning is categorized into three scenarios: domain incremental learning, task incremental learning, and class incremental learning \cite{van19, wiewel19}. This paper focuses on class incremental learning which includes the common real-world problem of incrementally learning new classes of information. In the case of layer-wise neural networks capable of continual learning, their memory capacity is often limited because most part of the network architecture is fixed \cite{kirkpatrick17}. One promising approach to overcome the limitation of memory capacity is to apply a growing self-organizing clustering algorithm as a classifier. Self-Organizing Incremental Neural Networks (SOINN) \cite{furao06} is a well-known growing self-organizing clustering algorithm which is inspired by Growing Neural Gas (GNG) \cite{fritzke95}. Several conventional studies have shown that SOINN-based classifiers capable of continual learning have good classification performance \cite{shen08, parisi17, wiwatcharakoses21}. However, the self-organizing process of SOINN-based algorithms is highly unstable due to the instability of GNG-based algorithms.

This paper proposes a new classification algorithm capable of continual learning by utilizing a growing self-organizing clustering based on Adaptive Resonance Theory (ART) \cite{grossberg87}. Among recent ART-based clustering algorithms, an algorithm that utilizes Correntropy-Induced Metric (CIM) \cite{liu07} to a similarity measure shows faster and more stable self-organizing performance than GNG-based algorithms \cite{chalasani15, masuyama19b, masuyamaFTCA}. We apply CIM-based ART with Edge and Age (CAEA) \cite{masuyama22a} as a base clustering algorithm in the proposed algorithm. Moreover, we also propose two variants of the proposed algorithm by modifying a computation of the CIM for improving the classification performance.

The contributions of this paper are summarized as follows:
\begin{itemize}
	\vspace{-1mm}
	\item[(i)] A new classification algorithm capable of continual learning, called CAEA Classifier (CAEAC), is proposed by applying an ART-based clustering algorithm.
	
	\item[(ii)] Two variants of CAEAC are introduced by modifying a computation of the CIM.
	
	\item[(iii)] Empirical studies show that CAEAC and its variants have superior classification performance than state-of-the-art clustering-based classifiers.
	
	\item[(iv)] The parameter sensitivity of CAEAC (and its variants) is analyzed in detail.
\end{itemize}

The paper is organized as follows. Section \ref{sec:literature} presents literature review for growing self-organizing clustering algorithms and classification algorithms capable of continual learning. Section \ref{sec:proposedAlgorithm} describes the details of mathematical backgrounds for CAEAC and its variants. Section \ref{sec:experiment} presents extensive simulation experiments to evaluate the classification performance of CAEAC and its variants by using real-world datasets. Section \ref{sec:conclusion} concludes this paper.

\section{Literature Review}
\label{sec:literature}

\subsection{Growing Self-organizing Clustering Algorithms}
\label{sec:literature_GC}
In general, the major drawback of classical clustering algorithms such as Gaussian Mixture Model (GMM) \cite{mclachlan19} and $ k $-means \cite{lloyd82} is that the number of clusters/partitions has to be specified in advance. GNG \cite{fritzke95} and ASOINN \cite{shen08} are typical types of growing self-organizing clustering algorithms that can handle the drawback of GMM and $ k $-means. GNG and ASOINN adaptively generate topological networks by generating nodes and edges for representing sequentially given data. However, since these algorithms permanently insert new nodes into topological networks for extracting new knowledge, they have a potential to forget learned knowledge (i.e., catastrophic forgetting). More generally, this phenomena is called the plasticity-stability dilemma \cite{carpenter88}. As a SOINN-based algorithm, SOINN+ \cite{wiwatcharakoses20} can detect clusters of arbitrary shapes in noisy data streams without any predefined parameters. Grow When Required (GWR) \cite{marsland02} is a GNG-based algorithm which can avoid the plasticity-stability dilemma by adding nodes whenever the state of the current network does not sufficiently match to the data. One problem of GWR is that as the number of nodes in the network increases, the cost of calculating a threshold for each node increases, and thus the learning efficiency decreases.

In contrast to GNG-based algorithms, ART-based algorithms can theoretically avoid the plasticity-stability dilemma by utilizing a predefined similarity threshold (i.e., a vigilance parameter) for controlling a learning process. Thanks to this ability, a number of ART-based algorithms and their improvements have been proposed for both supervised learning \cite{tan14, matias18, matias21} and unsupervised learning \cite{carpenter91b, vigdor07, wang19, da20}. Specifically, algorithms which utilize the CIM as a similarity measure have achieved faster and more stable self-organizing performance than GNG-based algorithms \cite{masuyama18, masuyama19a, masuyama19b, masuyamaFTCA}. One drawback of the ART-based algorithms is a specification of data-dependent parameters such as a similarity threshold. Several studies have proposed to solve this drawback by utilizing multiple vigilance levels \cite{da19}, adjusting parameters during a learning process \cite{matias21}, and estimating parameters from given data \cite{masuyama22a}. In particular, CAEA \cite{masuyama22a}, which utilizes the CIM as a similarity measure, has shown superior clustering performance while successfully reducing the effect of data-dependent parameters.

\subsection{Classification Algorithms Capable of Continual Learning}
\label{sec:literature_CL}
In recent years, neural networks have demonstrated high performance in object recognition, speech recognition, and natural language processing. On the other hand, the ability of neural networks to perform continual learning without catastrophic forgetting is not sufficient \cite{belouadah20}. Continual learning is categorized into three scenarios \cite{van19, wiewel19}: domain incremental learning \cite{zenke17, shin2017}, task incremental learning \cite{nguyen2018}, and class incremental learning \cite{kirkpatrick17, nguyen2018, tahir20, kongsorot20}. In general, when new information is given, layer-wise neural networks capable of continual learning use one of the following two learning mechanisms: selective learning of weight coefficients between neurons and sequential addition of neurons in the output layer corresponding to new information. However, the major problem with the above approaches is that the structure of networks is basically fixed, and therefore, there is an upper limit on the memory capacity.

One promising approach to overcome this difficulty caused by the fixed network structur is to apply a growing self-organizing clustering algorithm as a classifier. The growing self-organizing clustering algorithms adaptively and continually generate a node to represent new information. Typical classifiers of this type are Episodic-GWR \cite{parisi18} and ASOINN Classifier (ASC) \cite{shen08}, which utilize GWR and ASOINN, respectively. One state-of-the-art algorithm is SOINN+ with ghost nodes (GSOINN+) \cite{wiwatcharakoses21}. GSOINN+ has successfully improved the classification performance by generating some ghost nodes near a decision boundary of each class.

Another successful approach is ART-based supervised learning algorithms, i.e., ARTMAP \cite{carpenter92, vigdor07, masuyama18, tan14, matias18, matias21}. As mentioned in Section \ref{sec:literature_GC}, ART-based algorithms theoretically realize sequential and class-incremental learning without catastrophic forgetting. However, especially for an algorithm with an ARTMAP architecture, label information does not fully utilize during a supervised learning process. In general, a class label of each node is determined based on the frequency of the label appearance. Therefore, there is a possibility that a decision boundary of each class cannot be learned clearly.

\begin{table*}[bp]
	\centering
	\caption{Summary of notations}
	\renewcommand{\arraystretch}{1.2}
	\label{tab:notations}
	\scalebox{0.9}{
		\begin{tabular}{ll}
			\hhline{--}
			\hhline{--}
			Notation & Description \\
			\hline
			$ \mathcal{X} =$  $ \{ \mathbf{x}_{1}, \mathbf{x}_{2},\ldots, \mathbf{x}_{n}, \ldots \} $   & A set of training data points \\
			$ \mathbf{x}_{n} = $ $ \left( x_{n1}, x_{n2}, \ldots, x_{nd} \right) $  &  $ d $-dimensional training data point (the $ n $th data point)  \\ 
			$ C $  &  The number of classes in $ \mathcal{X} $ \\ 
			$ \mathcal{X}_{c} $ & A set of training data points belongs to the class $ c \in C $ \\
			$ \mathcal{Y} = \{\mathbf{y}_{1}, \mathbf{y}_{2}, \ldots, \mathbf{y}_{K}\} $  &  A set of prototype nodes \\ 
			$ \mathbf{y}_{k} = $ $ \left( y_{k1}, y_{k2}, \ldots, y_{kd} \right) $  & $ d $-dimensional prototype node (the $ k $th node) \\
			$ \mathcal{S} = \{ \sigma_{1},\sigma_{2},\ldots, \sigma_{K}\} $  & A set of bandwidths for a kernel function \\ 
			$ \kappa_{\sigma} $  & Kernel function with a bandwidth $ \sigma $ \\ 
			\textrm{CIM}  &  Correntropy-Induced Metric\\ 
			$ k_{1} $, $ k_{2} $  & Indexes of the 1st and 2nd winner nodes  \\
			$ \mathbf{y}_{k_{1}} $, $ \mathbf{y}_{k_{2}} $ & The 1st and 2nd winner nodes \\
			$ V_{k_{1}} $, $ V_{k_{2}} $  & Similarities between a data point $ \mathbf{x}_{n} $ and winner nodes ($ \mathbf{y}_{k_{1}} $ and $ \mathbf{y}_{k_{2}} $) \\
			$ V_{\text{threshold}} $  &  Similarity threshold (a vigilance parameter) \\
			$ \mathcal{N}_{k_{1}} $ & A set of neighbor nodes of node $ \mathbf{y}_{k_{1}} $ \\
			$ \alpha_{k_{1}} $  &  The number of data points that have accumulated by the node $ \mathbf{y}_{k_{1}} $ \\
			$ \lambda $ & Predefined interval for computing $ \sigma $ and deleting an isolated node \\
			$ e_{\left(k_{1}, k_{2}\right)} $ & Edge connection between nodes $ \mathbf{y}_{k_{1}} $ and $ \mathbf{y}_{k_{2}} $ \\
			$ a_{(k_{1}, k_{2})} $ & Age of edge $ e_{\left(k_{1}, k_{2}\right)} $ \\
			$ a_{\text{max}}$   & Predefined threshold of an age of edge \\
%			$ TEMP $ & TEMP \\
			% \hline\hline
			\hhline{--}
			\hhline{--}
		\end{tabular}
	}
\end{table*}

\section{Proposed Algorithm}
\label{sec:proposedAlgorithm}
In this section, first the overview of CAEAC is introduced. Next, the mathematical backgrounds of the CIM and CAEA are explained in detail. Then, modifications of the CIM computation are introduced for variants of CAEAC. Table \ref{tab:notations} summarizes the main notations used in this paper.

\subsection{Class-wise Classifier Design Capable of Continual Learning}
\begin{figure}[htbp]
	\centering
	\includegraphics[width=2.2in]{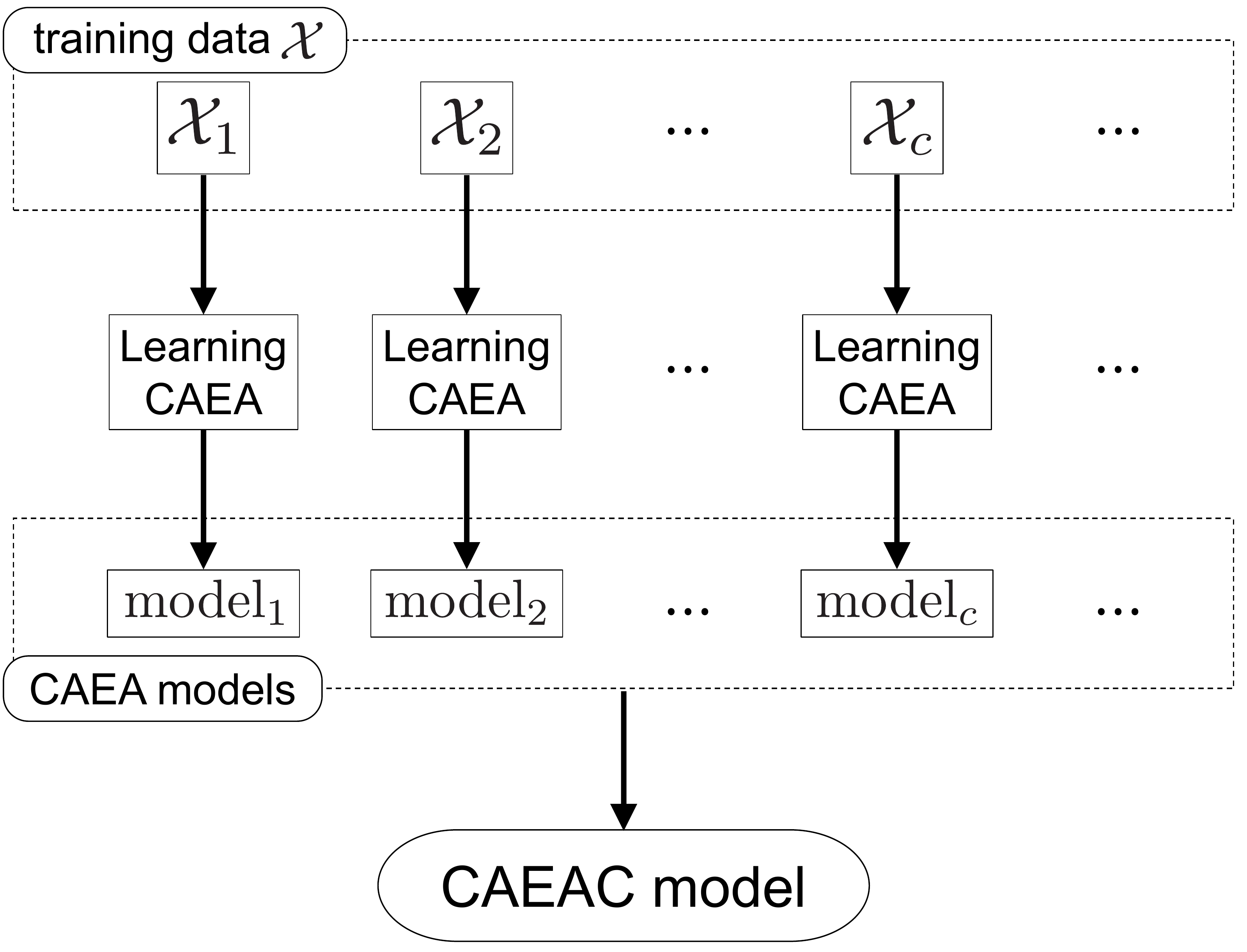}
	\caption{Overview of CAEAC.}
	\label{fig:overviewCAEAC}
\end{figure}

\begin{figure}[htbp]
	\centering
	\includegraphics[width=2.2in]{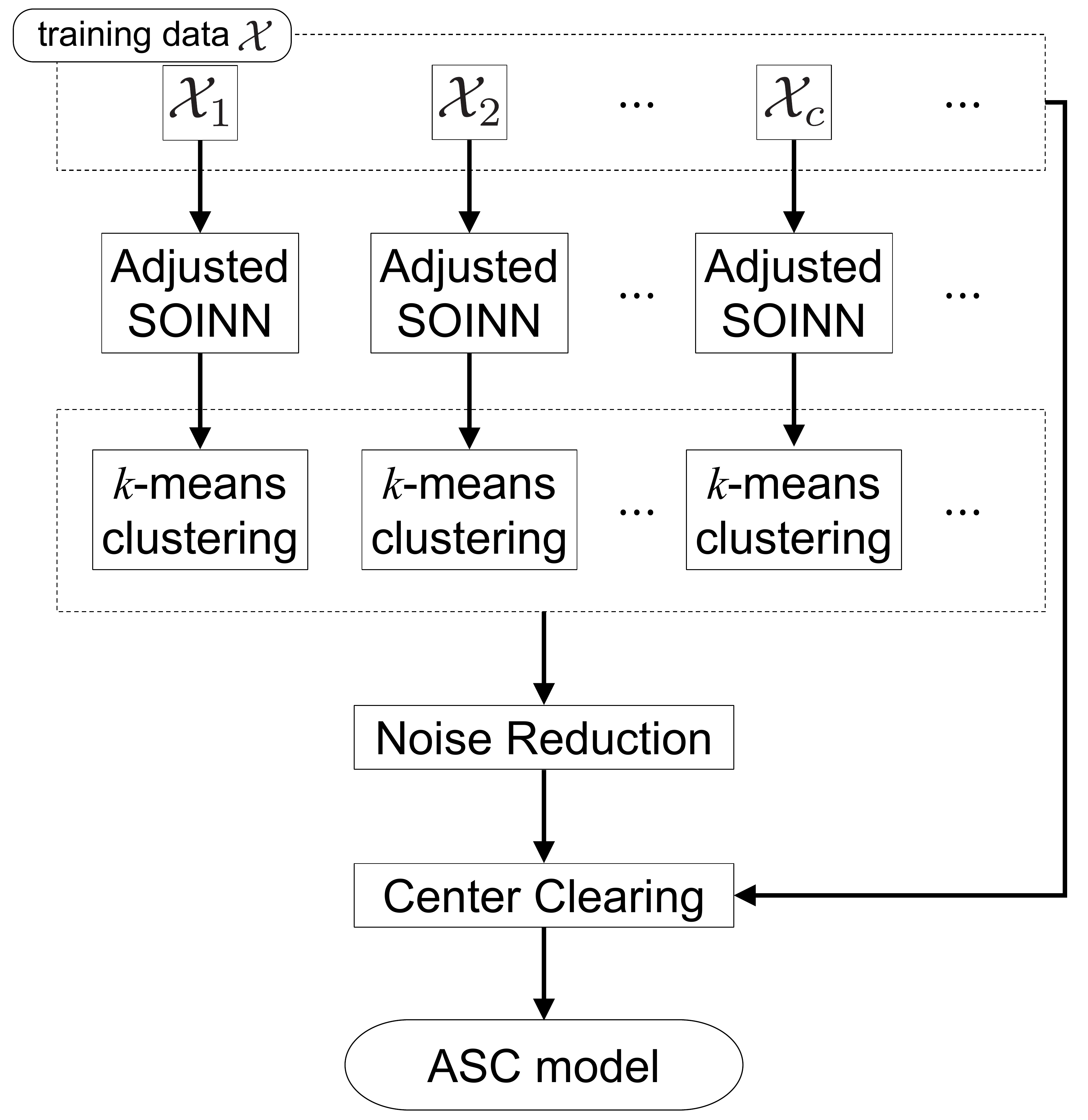}
	\caption{Overview of ASC \cite{shen08}.}
	\label{fig:overviewASC}
\end{figure}

Figure \ref{fig:overviewCAEAC} shows the overview of CAEAC. The architecture of CAEAC is inspired by ASC \cite{shen08}. As shown in Fig. \ref{fig:overviewASC}, ASC is an ASOINN-based supervised classifier incorporating $ k $-means and two node clearance mechanisms after class-wise unsupervised self-organizing processes. The main difference between CAEAC and ASC is that CAEAC does not require $ k $-means and node clearance mechanisms because CAEA has superior clustering performance than ASOINN.

In CAEAC, a training dataset is divided into multiple subsets based on their class labels. The number of the subsets is the same as the number of classes. Each subset is used to generate a classifier (i.e., nodes and edges) through a self-organization process by CAEA. Since CAEA is capable of continual learning, each classifier can be continually updated. Moreover, when a training dataset belongs to a new class, a self-organizing space of CAEA is newly defined. Thus it is possible to learn new knowledge without destroying existing knowledge. When classifying an unknown data point, classifiers (one classifier for each class) are installed in the same space, and the label information of the nearest neighbor node of the unknown data point is output as a classification result. The learning procedure of CAEAC is summarized in Algorithm \ref{al:learnCAEAC}.

In the following subsections, the mathematical backgrounds of the CIM and CAEA are explained in detail.

% CAEAC Algorithm------------------------------------------------
\begin{algorithm}[htbp]
	\small
	\DontPrintSemicolon
	\caption{Learning Algorithm of CAEAC}
	\label{al:learnCAEAC}
	\KwIn{\\
		the training data points: $ \mathcal{X} = \{ {\mathbf{x}}_{1}, {\mathbf{x}}_{2}, \ldots, {\mathbf{x}}_{L} \} (\mathbf{x}_{l} \in \mathbb{R}^{d})$, \\
		%the current level of layer: $h$, \\
		the predefined interval for computing $\sigma$ and deleting an isolated node: $\lambda$, \\
		and the predefined threshold of an age of edge: $ a_{\text{max}}$. \\
	}
	\KwOut{\\
	the CAEA models.
		% \begin{quote}
		% 	\textbf{model contents} \\
		% 	a set of CAEA models:  \\
		% \end{quote}
	}
	\vspace{2mm}
	\SetKwBlock{Begin}{function}{end function}
	\Begin( \text{LearningCAEAC($ \mathcal{X} $, $\lambda$, $a_{\text{max}}$)}){
		\ForAll{ $ c \in {1, 2, \ldots, C} $ }{
			$ \text{model}_{c} = $ \text{LearningCAEA}($ \mathcal{X}_{c} $, $ \lambda $,  $ a_{\text{max}} $). \\
		}
		\Return{ \upshape{CAEA models} }. 
	}
\end{algorithm}

\subsection{Correntropy and Correntropy-induced Metric}
\label{sec:cimDefinition}
Correntropy \cite{liu07} provides a generalized similarity measure between two arbitrary data points $ \mathbf{x} = (x_{1},x_{2},\ldots,x_{d}) $ and $ \mathbf{y} = (y_{1},y_{2},\ldots,y_{d}) $ as follows:
\begin{equation}
	C(\mathbf{x}, \mathbf{y}) = \textbf{E} \left[ \kappa_{\sigma} \left( \mathbf{x}, \mathbf{y} \right) \right],
\end{equation}
where $ \textbf{E} \left[ \cdot \right] $ is the expectation operation, and $ \kappa_{\sigma} \left( \cdot \right) $ denotes a positive definite kernel with a bandwidth $ \sigma $. The correntropy is estimated as follows:
\begin{equation}
	\hat{C}(\mathbf{x}, \mathbf{y}, \sigma) = \frac{1}{d} \sum_{i=1}^{d} \kappa_{\sigma} \left( x_{i}, y_{i} \right).
	\label{eq:correntropy}
\end{equation}

In this paper, we use the following Gaussian kernel in the correntropy:
\begin{equation}
	\kappa_{\sigma} \left( x_{i}, y_{i} \right) = \exp \left[ - \frac{\left( x_{i} - y_{i} \right)^{2} }{2 \sigma^{2}} \right].
	\label{eq:gaussian}
\end{equation}

A nonlinear metric called CIM is derived from the correntropy \cite{liu07}. CIM quantifies the similarity between two data points $ \mathbf{x} $ and $ \mathbf{y} $ as follows:
\begin{equation}
	\mathrm{CIM}\left(\mathbf{x}, \mathbf{y}, \sigma \right) = \left[ 1 - \hat{C}(\mathbf{x}, \mathbf{y}, \sigma) \right]^{\frac{1}{2}},
	\label{eq:defcim}
\end{equation}
here, since the Gaussian kernel in (\ref{eq:gaussian}) does not have the coefficient $ \frac{1}{\sqrt{2\pi}\sigma} $, the range of CIM is limited to $ \left[0,1 \right] $.

In general, the Euclidean distance suffers from the curse of dimensionality. However, CIM reduces this drawback since the correntropy calculates the similarity between two data points by using a kernel function. Moreover, it has also been shown that CIM with the Gaussian kernel has a high outlier rejection ability \cite{liu07}.

\subsection{CIM-based ART with Edge and Age}
\label{sec:caea}
CAEA \cite{masuyama22a} is an ART-based topological clustering algorithm capable of continual learning. In \cite{masuyama22a}, CAEA and its hierarchical approach show comparable clustering performance to recently-proposed clustering algorithms without difficulty of parameter specifications to each dataset. The learning procedure of CAEA is divided into four parts: 1) initialization process for nodes and a bandwidth of a kernel function in the CIM, 2) winner node selection, 3) vigilance test, and 4) node learning and edge construction. Each of them is explained in the following subsections.

In this paper, we use the following notations: A set of training data points is $ \mathcal{X} =$  $ \{ \mathbf{x}_{1}, \mathbf{x}_{2},\ldots, \mathbf{x}_{n}, \ldots \} $ where $ \mathbf{x}_{n} = $ $ ( x_{n1}, x_{n2}, \ldots,$  $ x_{nd} ) $ is a $ d $-dimensional feature vector. A set of prototype nodes in CAEA at the time of the presentation of a data point $ \mathbf{x}_{n} $ is $ \mathcal{Y} = \{\mathbf{y}_{1}, \mathbf{y}_{2},$  $ \ldots, \mathbf{y}_{K}\} $ ($ K \in \mathbb{Z}^{+} $) where a node $ \mathbf{y}_{k} = $ $ \left( y_{k1}, y_{k2}, \ldots, y_{kd} \right) $ has the same dimension as $ \mathbf{x}_{n} $. Furthermore, each node $ \mathbf{y}_{k} $ has an individual bandwidth $ \sigma_{k} $ for the CIM, i.e., $ \mathcal{S} = \{ \sigma_{1},\sigma_{2},\ldots, \sigma_{K}\} $.

\subsubsection{Initialization Process for Nodes and a Bandwidth of a Kernel Function in the CIM}
\label{sec:initProcess}
In the case that CAEA does not have any nodes, i.e., a set of prototype node $ \mathcal{Y} = \varnothing $, the first $ (\lambda/2) $th training data points $ \mathcal{X}_{\text{init}} = \{ \mathbf{x}_{1}, \mathbf{x}_{2},\ldots \mathbf{x}_{\lambda/2} \} $ directly become prototype nodes, i.e., $ \mathcal{Y}_{\text{init}} = \{\mathbf{y}_{1}, \mathbf{y}_{2}, \ldots, \mathbf{y}_{\lambda/2}\} $, where $ \mathbf{y}_{k} = \mathbf{x}_{k} $ $ ( k = 1, 2, \ldots, \lambda/2 ) $ and $ \lambda \in \mathbb{Z}^{+} $ is a predefined parameter of CAEA. This parameter is also used for a node deletion process that is explained in Section \ref{sec:nodeLearning}.

In an ART-based clustering algorithm, a vigilance parameter (i.e., a similarity threshold) plays an important role in a self-organizing process. Typically, the similarity threshold is data-dependent and specified by hand. On the other hand, CAEA uses the minimum pairwise CIM value between each pair of nodes in $ \mathcal{Y}_{\text{init}} = \{\mathbf{y}_{1}, \mathbf{y}_{2}, \ldots, \mathbf{y}_{\lambda/2}\} $, and the average of pairwise CIM values is used as the similarity threshold $ V_{\text{threshold}} $, i.e.,
\begin{equation}
	V_{\text{threshold}} = \frac{1}{\lambda/2} \sum_{i=1}^{\lambda/2} \min_{j \neq i} \left[\mathrm{CIM}\left({\mathbf{y}}_{i}, {\mathbf{y}_{j}}, \sigma) \right)\right],
	\label{eq:pairwiseCIM}
\end{equation}
where $ \sigma $ is a kernel bandwidth in the CIM.

In general, the bandwidth of a kernel function can be estimated from $ \lambda $ instances belonging to a certain distribution \cite{henderson12}, which is defined as follows:
\begin{equation}
\bm{\Sigma} = U(F_{\nu})  \bm{\Gamma} \lambda^{-\frac{1}{2\nu+d}},
\label{eq:sigmaEst1}
\end{equation}
\begin{equation}
U(F_{\nu}) = \left( \frac{\pi ^{d/2} 2^{d+\nu-1}(\nu !)^{2}R(F)^{d}}{\nu \kappa_{\nu}^{2}(F)\left[(2\nu)!!+(d-1)(\nu!!)^{2}\right]} \right)^{\frac{1}{2\nu+d}},
\label{eq:sigmaEst2}
\end{equation}
where $ \bm{\Gamma} $ denotes a rescale operator ($ d $-dimensional vector) which is defined by a standard deviation of each of the $ d $ attributes among $ \lambda $ instances, $ \nu $ is the order of a kernel, the single factorial of $\nu$ is calculated by the product of integer numbers from 1 to $\nu$, the double factorial notation is defined as $ (2\nu)!!  = (2\nu-1) \cdot 5 \cdot 3 \cdot 1 $ (commonly known as the odd factorial), $ R(F) $ is a roughness function, and $ \kappa_{\nu}(F) $ is the moment of a kernel. The details of the derivation of (\ref{eq:sigmaEst1}) and (\ref{eq:sigmaEst2}) can be found in \cite{henderson12}.

In this paper, we use the Gaussian kernel for CIM. Therefore, $ \nu = 2 $, $ R(F) = (2\sqrt{\pi})^{-1} $, and $ \kappa_{\nu}^{2}(F) = 1 $. Then, (\ref{eq:sigmaEst2}) is rewritten as follows:
\begin{equation}
	\mathbf{H} = \left( \frac{4}{2+d} \right)^{\frac{1}{4+d}}  \bm{\Gamma}  \lambda^{-\frac{1}{4+d}}.
	\label{eq:SIGMA}
\end{equation}
% where $ \bm{\Gamma} $ denotes a rescale operator ($ d $-dimensional vector) which is defined by a standard deviation of the $ d $ attributes among $ \lambda $ instances. 

Equation (\ref{eq:SIGMA}) is known as the Silverman's rule \cite{silverman18}. Here, $ \mathbf{H} $ contains the bandwidth of a kernel function in CIM. In (\ref{eq:SIGMA}), $ \mathbf{\Sigma} $ contains the bandwidth of each attribute (i.e., $ \mathbf{\Sigma} $ is a $ d $-dimensional vector). In this paper, the median of the $ d $ elements of $ \mathbf{\Sigma} $ is selected as a representative bandwidth of the Gaussian kernel in the CIM, i.e.,
\begin{equation}
	\sigma = \mathrm{median} \left( \mathbf{\Sigma} \right).
	\label{eq:sigma}
\end{equation}

In CAEA, the initial prototype nodes $ \mathcal{Y}_{\text{init}} = \{\mathbf{y}_{1}, \mathbf{y}_{2}, \ldots,$  $ \mathbf{y}_{\lambda/2}\} $ have a common bandwidth of the Gaussian kernel in the CIM, i.e., $ \mathcal{S}_{\text{init}} = \{ \sigma_{1},\sigma_{2},\ldots, \sigma_{\lambda/2}\} $ where $ \sigma_{1} = \sigma_{2} = \cdots = \sigma_{\lambda/2} $. When a new node $ \mathbf{y}_{K+1} $ is generated from $ \mathbf{x}_{n} $, a bandwidth $ \sigma_{K+1} $ is estimated from the past $ \lambda/2 $ data points, i.e., $  \{ \mathbf{x}_{n-\lambda/2},\ldots, \mathbf{x}_{n-2}, \mathbf{x}_{n-1} \} $ by using (\ref{eq:sigmaEst1}) and (\ref{eq:sigmaEst2}). As a result, each new node has a different bandwidth $ \sigma $ depending on the distribution of training data points. In addition, a set of counters $ \mathcal{M} = \{M_{1}, M_{2},\ldots,M_{\lambda/2} \} $ where $ M_{1} = M_{2} = \cdots = M_{\lambda/2} = 1 $ is defined. Although the similarity threshold $ V_{\text{threshold}} $ depends on the distribution of the initial $ \lambda/2 $ training data points, we regard that an adaptive $ V_{\text{threshold}} $ estimation is realized by assigning a different bandwidth $ \sigma $, which affects the CIM value, to each node in response to the changes in the data distribution.

\subsubsection{Winner Node Selection}
\label{sec:winnerNS}
Once a data point $ \mathbf{x}_{n} $ is presented to CAEA with the prototype node set $ \mathcal{Y} = \{\mathbf{y}_{1}, \mathbf{y}_{2}, \ldots, $ $ \mathbf{y}_{K}\} $, two nodes which have a similar state to the data point $ \mathbf{x}_{n} $ are selected, namely, winner nodes $ \mathbf{y}_{k_{1}} $ and $ \mathbf{y}_{k_{2}} $. The winner nodes are determined based on the state of the CIM as follows:
\begin{equation}
	k_{1} = \argmin_{\mathbf{y}_{i} \in \mathcal{Y}}\left[ \mathrm{CIM}\left(\mathbf{x}_{n}, \mathbf{y}_{i}, \mathrm{mean}(\mathcal{S}) \right) \right],
	\label{eq:winnerCIM1}
\end{equation}
\vspace{-5pt}
\begin{equation}
	k_{2} = \argmin_{\mathbf{y}_{i} \in \mathcal{Y} \backslash \{\mathbf{y}_{k_{1}}\}}\left[ \mathrm{CIM}\left(\mathbf{x}_{n}, \mathbf{y}_{i}, \mathrm{mean}(\mathcal{S}) \right) \right],
	\label{eq:winnerCIM2}
\end{equation}
\noindent where $ k_{1} $ and $ k_{2} $ denote the indexes of the 1st and 2nd winner nodes, i.e., $ \mathbf{y}_{k_{1}} $ and $ \mathbf{y}_{k_{2}} $, respectively. $ \mathcal{S} $ is bandwidths of the Gaussian kernel in the CIM for each node.

\subsubsection{Vigilance Test}
\label{sec:vigilanceTest}
Similarities between the data point $ \mathbf{x}_{n} $ and the 1st and 2nd winner nodes are defined as follows:
\begin{equation}
	V_{k_{1}} =  \mathrm{CIM}\left(\mathbf{x}_{n}, \mathbf{y}_{k_{1}}, \mathrm{mean}(\mathcal{S}) \right),
	\label{eq:cim1}
\end{equation}
\vspace{-5pt}
\begin{equation}
	V_{k_{2}} = \mathrm{CIM}\left(\mathbf{x}_{n}, \mathbf{y}_{k_{2}}, \mathrm{mean}(\mathcal{S}) \right).
	\label{eq:cim2}
\end{equation}

The vigilance test classifies the relationship between a data point and a node into three cases by using a predefined similarity threshold $ V_{\text{threshold}} $, i.e.,

\begin{itemize}
	% [ 
	% \setlength{\IEEElabelindent}{\dimexpr-\labelwidth-\labelsep}% Wrapping of text beyond first line of \item 
	% \setlength{\itemindent}{\dimexpr\labelwidth+\labelsep}% identation for each new \item 
	% \setlength{\listparindent}{\parindent}% Restore regular paragraph indentation 
	% ]
	\item Case I \\
	\indent The similarity between the data point $ \mathbf{x}_{n} $ and the 1st winner node $ \mathbf{y}_{k_{1}} $ is larger (i.e., less similar) than $ V_{\text{threshold}} $, namely:
	\begin{equation}
		V_{\text{threshold}} < V_{k_{1}} \leq V_{k_{2}}.
		\label{eq:case1}
	\end{equation}
	
	\vspace{0.8mm}
	
	\item Case I\hspace{-.1em}I \\
	\indent The similarity between the data point $ \mathbf{x}_{n} $ and the 1st winner node $ \mathbf{y}_{k_{1}} $ is smaller (i.e., more similar) than $ V_{\text{threshold}} $, and the similarity between the data point $ \mathbf{x}_{n} $ and the 2nd winner node $ \mathbf{y}_{k_{2}} $ is larger (i.e., less similar) than $ V_{\text{threshold}} $, namely:
	\begin{equation}
		V_{k_{1}} \leq V_{\text{threshold}} < V_{k_{2}}.
		\label{eq:case2}
	\end{equation}
	
	\vspace{0.8mm}
	
	\item Case I\hspace{-.1em}I\hspace{-.1em}I \\
	\indent The similarities between the data point $ \mathbf{x}_{n} $ and the 1st and 2nd winner nodes are both smaller (i.e., more similar) than $ V_{\text{threshold}} $, namely:
	\begin{equation}
		V_{k_{1}} \leq V_{k_{2}} \leq V_{\text{threshold}}.
		\label{eq:case3}
	\end{equation}
	
\end{itemize}

\subsubsection{Node Learning and Edge Construction}
\label{sec:nodeLearning}

Depending on the result of the vigilance test, a different operation is performed.

If $ \mathbf{x}_{n} $ is classified as Case I by the vigilance test (i.e., (\ref{eq:case1}) is satisfied), a new node $ \mathbf{y}_{K+1} = \mathbf{x}_{n} $ is added to the prototype node set $ \mathcal{Y} = \{\mathbf{y}_{1}, \mathbf{y}_{2}, \ldots, \mathbf{y}_{K}\} $. A bandwidth $ \sigma_{K+1} $ for node $ \mathbf{y}_{K+1} $ is calculated by (\ref{eq:sigma}). In addition, the number of data points that have been accumulated by the node $ \mathbf{y}_{K+1} $ is initialized as $ M_{K+1} = 1 $.

If $ \mathbf{x}_{n} $ is classified as Case I\hspace{-1pt}I by the vigilance test (i.e., (\ref{eq:case2}) is satisfied), first, the age of each edge connected to the first winner node $ \mathbf{y}_{k_{1}} $ is updated as follows:
\begin{equation}
	a_{\left(k_{1}, j\right)} \gets a_{\left(k_{1}, j\right)} + 1 \quad \left(\forall j \in \mathcal{N}_{k_{1}}\right),
	\label{eq:edd_age}
\end{equation}
where $ \mathcal{N}_{k_{1}} $ is a set of all neighbor nodes of the node $ \mathbf{y}_{k_{1}} $. After updating the age of each of those edges, an edge whose age is greater than a predefined threshold $ a_{\text{max}} $ is deleted. In addition, a counter $ M $ for the number of data points that have been accumulated by $ \mathbf{y}_{k_{1}} $ is also updated as follows:
\begin{equation}
	M_{k_{1}} \leftarrow M_{k_{1}} + 1.
	\label{eq:countNode}
\end{equation}

Then, $ \mathbf{y}_{k_{1}} $ is updated as follows:
\begin{equation}
	\mathbf{y}_{k_{1}} \leftarrow \mathbf{y}_{k_{1}} + \frac{1}{M_{k_{1}}} \left( \mathbf{x}
	_{n} - \mathbf{y}_{k_{1}} \right).
	\label{eq:updateNodeWeight1}
\end{equation}

When updating the node, the difference between $ \mathbf{x}_{n} $ and $ \mathbf{y}_{n} $ is divided by $ M_{k_{1}} $. Thus, the changes of the node position is smaller when $ M_{k_{1}} $ is larger. This is based on the idea that the information around a node, where data points are frequently given, is important and should be held by the node.

If $ {\mathbf{x}}_{n} $ is classified as Case I\hspace{-1pt}I\hspace{-1pt}I by the vigilance test (i.e., (\ref{eq:case3}) is satisfied), the same operations as Case I\hspace{-1pt}I (i.e., (\ref{eq:edd_age}), (\ref{eq:updateNodeWeight1}), and (\ref{eq:countNode})) are performed. In addition, the neighbor nodes of $ \mathbf{y}_{k_{2}} $ are updated as follows:
\begin{equation}
	{\mathbf{y}}_{j} \gets {\mathbf{y}}_{j} + \frac{1}{10 M_{j}} \left({\mathbf{x}}_{n}-{\mathbf{y}}_{j}\right) \quad \left(\forall j \in \mathcal{N}_{k_{2}}\right).
	\label{eq:updateNodeWeight2}
\end{equation}

In Case I\hspace{-1pt}I\hspace{-1pt}I, moreover, if there is no edge between $ \mathbf{y}_{k_{1}} $ and $ \mathbf{y}_{k_{2}} $, a new edge $ e_{\left(k_{1}, k_{2}\right)} $ is defined and its age is initialized as follows:
\begin{equation}
	a_{(k_{1}, k_{2})} \leftarrow 0.
	\label{eq:ageInit}
\end{equation} 

In the case that there is an edge between nodes $ \mathbf{y}_{k_{1}} $ and $ \mathbf{y}_{k_{2}} $, its age is also reset by (\ref{eq:ageInit}).

Apart from the above operations in Cases I-I\hspace{-1pt}I\hspace{-1pt}I, as a noise reduction purpose, the nodes without edges are deleted every $ \lambda $ training data points (e.g., the node deletion interval is the presentation of $ \lambda $ training data points). 

The learning procedure of CAEA is summarized in Algorithm \ref{al:trainCAEA}. Note that, in CAEAC, the classification process of an unknown data point is similar to ASC. That is, the unknown data point is assigned to the class of its nearest neighbor node.

% CAEA Algorithm------------------------------------------------
\begin{algorithm}[htbp]
	\small
	\DontPrintSemicolon
	\caption{Learning Algorithm of CAEA \cite{masuyama22a}}
	\label{al:trainCAEA}
	\KwIn{\\
		a set of training data points: $ \mathcal{X} =$  $ \{ \mathbf{x}_{1}, \mathbf{x}_{2},\ldots, \mathbf{x}_{n}, \ldots \} $ where $ \mathbf{x}_{n} = $ $ \left( x_{n1}, x_{n2}, \ldots, x_{nd} \right) $ $ (\bm{x}_{l} \in \Re^{d})$, \\
		%the current level of layer: $h$, \\
		the interval for computing $ \sigma $ and deleting an isolated node: $\lambda$, \\
		and the threshold of an age of edge: $a_{\text{max}}$.
	}
	\KwOut{\\
		the CAEA model.
		\begin{quote}
			\textbf{model contents} \\
			a set of generated nodes: $ \mathcal{Y} = \{ \mathbf{y}_{1}, \mathbf{y}_{2}, \ldots, \mathbf{y}_{K} \} $ $ \left( K \in \mathbb{Z}^{+} \right) $, \\
			a set of bandwidths for a kernel function: $ \mathcal{S} = \{ \sigma_{1},\sigma_{2},\ldots, \sigma_{K}\} $ \\ 
			a set of counters: $ \mathcal{M} = \{ M_{1}, M_{2}, \ldots, M_{K} \} $, \\
			the matrix of edge connections: $\mathbf{e}$, \\
			and the matrix of edge age: $\mathbf{a}$.
		\end{quote}
	}
	\vspace{2mm}
	\SetKwBlock{Begin}{function}{end function}
	\Begin( \text{LearningCAEA($ \mathcal{X} $, $\lambda$, $a_{\text{max}}$)}){
		\ForAll{ $ l \in {1, 2, \ldots, L} $ }{
			
			\uIf{ $ K < \lambda/2 $ }{
				Initialize a counter $ M_{K+1} = 1 $. \\
				Update a set of counter $ \mathcal{M} \leftarrow \mathcal{M} \cup M_{K+1}$. \\
				Create the new node as $ {\mathbf{y}}_{K+1} = {\mathbf{x}}_{l} $. \\
				Calculate the kernel bandwidth $ \sigma_{K+1} $ by (\ref{eq:SIGMA}) and (\ref{eq:sigma}). \\
				\If{ $ K = \lambda/2 $ }{
					Calculate the vigilance parameter $ V_\text{threshold} $ by (\ref{eq:pairwiseCIM}). \\
				}
			}\Else{
				Search the indexes of winner nodes $ k_{1} $ and $ k_{2} $ by (\ref{eq:winnerCIM1}) and (\ref{eq:winnerCIM2}), respectively. \\
				Update the edge age $a_{\left(k_{1}, j\right)}$ by (\ref{eq:edd_age}). \\
				\If{ $a_{\left(k_{1}, j\right)} > a_{\text{\upshape{max}}}$ }{
					Delete the edge. \\
				}
				\uIf{ \upshape{$ V_{k_{1}} > V_\text{threshold} $} }{
					Initialize a counter $ M_{K+1} = 1 $. \\
					Update a set of counter $ \mathcal{M} \leftarrow \mathcal{M} \cup M_{K+1}$. \\
					Create the new node as $ {\mathbf{y}}_{K+1} = {\bm{x}}_{l} $.\\
					Calculate the kernel bandwidth $ \sigma_{k+1} $ by (\ref{eq:SIGMA}) and (\ref{eq:sigma}). \\
				}\Else{
					Update the state of $ M_{k_{1}} $ by (\ref{eq:countNode}). \\
					Update the state of $ {\bm{y}}_{k_{1}} $ by (\ref{eq:updateNodeWeight1}). \\
					\If{ \upshape{$ V_{k_{2}} \leq V_\text{threshold} $} }{
						Update the state of neighbor nodes $ {\mathbf{y}}_{j} $ by (\ref{eq:updateNodeWeight2}). \\
						Create a new edge $e_{\left(k_{1}, k_{2}\right)}$ between $ {\mathbf{y}}_{k_{1}} $ and $ {\mathbf{y}}_{k_{2}} $. \\
					}
				}
			}
			%% start Topology Construction ---------------
			\If{ \upshape{the number of data point inputs $ l $ is multiple of a topology adjustment cycle $ \lambda $} }{
				\ForAll{ $ k \in {1, 2, \ldots, K} $ }{
					\If{ \upshape{$ {\mathbf{y}}_{k} $ does not have any edge} }{
						Remove $ {\mathbf{y}}_{k} $ from $ \mathcal{Y} $.
					}
				}
			}
		} %\ForALL
		\Return{ \upshape{the CAEA model} }.
	} %\SetKwBlock
\end{algorithm}

\subsection{Modifications of the CIM Computation}
\label{sec:CIM_Individual_Clustering}
As shown in (\ref{eq:correntropy}) and (\ref{eq:gaussian}), the CIM in CAEAC uses a common bandwidth $ \sigma $ to all attributes. Thus, a specific attribute may have a large impact on the value of the CIM if the common bandwidth $ \sigma $ is not appropriate for the attribute.

In this paper, two modifications of the CIM computation \cite{masuyama20a} are integrated into CAEAC in order to mitigate the above-mentioned effects: 1) one is to compute the CIM by using each individual attribute separately, and the average CIM value is used for similarity measurement, and 2) the other is to apply a clustering algorithm to attribute values, then attributes with similar value ranges are grouped. The CIM is computed by using each attribute group, and the average CIM value is used for similarity measurement.

\subsubsection{Individual-based Approach}
\label{sec:CIM_Individual}
In this approach, the CIM is computed by using each individual attribute separately, and the average CIM value is used for similarity measurement. The similarity between a data point $ \mathbf{x}_{n} $ and a node $ \mathbf{y}_{k} $ is defined by the $ \mathrm{CIM}^{\text{I}} $ as follows:
\begin{equation}
	\mathrm{CIM}^{\text{I}}  \left(\mathbf{x}_{n}, \mathbf{y}_{k}, \boldsymbol{\sigma}_{k} \right) = \frac{1}{d}  \sum_{i=1}^{d} \left[ \kappa_{\boldsymbol{\sigma}_{k, i}}  (0) - \mathrm{\hat{C}_{E}}^{\text{I}}(x_{ni}, y_{ki}) \right]^{  \frac{1}{2}} ,
	\label{eq:defcim_Individual}
\end{equation}
\begin{equation}
	\mathrm{\hat{C}_{E}}^{\text{I}}(x_{ni}, y_{ki}) = \kappa_{\sigma_{ki}} \left( x_{ni} - y_{ki} \right),
\end{equation}
where $ \boldsymbol{\sigma}_{k} = \left( \sigma_{k1}, \sigma_{k2},\ldots, \sigma_{kd} \right) $ is a bandwidth of a node $ \mathbf{y}_{k} $. A bandwidth for the $ i $th attribute of $ \mathbf{y}_{k} $ (i.e., $ \sigma_{ki} $) is defined as follows:
\begin{equation}
	\sigma_{ki} = \left( \frac{4}{2+d} \right)^{\frac{1}{4+d}}  \Gamma_{i}  \lambda^{-\frac{1}{4+d}},
	\label{eq:sigma_Individual}
\end{equation}
where $ \Gamma_{i} $ denotes a rescale operator which is defined by a standard deviation of the $ i $th attribute values among the $ \lambda $ data points.

In this paper, CAEAC with the individual-based approach is called CAEAC-Individual (CAEAC-I).

\subsubsection{Clustering-based Approach}
\label{sec:CIM_Clustering}
In this approach, for every $ \lambda $ data points, the clustering algorithm presented in \cite{masuyama20a} is applied to the attribute values. Each attribute value of $ \lambda $ data points is regarded as a one-dimensional vector and used as an input to the clustering algorithm. As a result, attributes with similar value ranges are grouped together. By using this grouping information, the similarity is calculated for each attribute group, and their average is defined as the similarity between a data point $ \mathbf{x}_{n} $ and a node $ \mathbf{y}_{k} $. Specifically, by using the grouping information, the data point $ \mathbf{x}_{n} = $ $ \left( x_{n1}, x_{n2}, \ldots, x_{nd} \right) $ is transformed into $ \mathbf{x}_{n}^{\text{C}} = \left( \mathbf{u}_{n1}, \mathbf{u}_{n2},\ldots, \mathbf{u}_{nJ} \right) $ $\left( J \leq d \right)$ by the clustering algorithm, where $ \mathbf{u}_{j} $ represents the $ j $th attribute group. Similarly, the node $ \mathbf{y}_{k} $ is transformed into $ \mathbf{y}_{k}^{\text{C}} = \left( \mathbf{v}_{k1}, \mathbf{v}_{k2},\ldots, \mathbf{v}_{kJ} \right) $ $\left( J \leq d \right)$ where $ \mathbf{v}_{j} $ represents the $ j $th attribute group. The dimensionality of each attribute group is represented as $ d^{\text{C}} = \{ d_{1}, d_{2},\ldots, d_{J} \} $ where $ d_{j} $ is the dimensionality of the $ j $th attribute group (i.e., the number of attributes in the $ j $th attribute group).

Here, the similarity between the data point $ \mathbf{x}_{n}^{\text{C}} $ and the node $ \mathbf{y}_{k} $ is defined by the $ \mathrm{CIM}^{\text{C}} $ as follows:
\begin{equation}
	\mathrm{CIM}^{\text{C}}  \left(\mathbf{x}_{n}^{\text{C}}, \mathbf{y}_{k}^{\text{C}}, \mathbf{\sigma}_{k}^{\text{C}} \right) = \frac{1}{J} \sum_{j=1}^{J} \left[ \kappa_{\sigma_{j}} (0) - \mathrm{\hat{C}_{E}}^{\text{C}}(\mathbf{u}_{j}, \mathbf{v}_{j})  \right]^{\frac{1}{2}},
	\label{eq:defcim_Cluster}
\end{equation}
\begin{equation}
	\mathrm{\hat{C}_{E}}^{\text{E}}(\mathbf{u}_{j}, \mathbf{v}_{j}) = \frac{1}{d_{j}} \sum_{i=1}^{d_{j}} \kappa_{\sigma_{j}} \left( u_{i} - v_{i} \right),
\end{equation}
%\begin{IEEEeqnarray}{Cr}
%	\!\!\!\! \mathrm{CIM}^{\text{C}} \! \left(\mathbf{x}_{n}^{\text{C}}, \mathbf{y}_{k}^{\text{C}}, \mathbf{\sigma}_{k}^{\text{C}} \right) \!=\! \frac{1}{J}\! \sum_{j=1}^{J}\! \left[ \!\left\{ \kappa_{\sigma_{j}} \!(0) \!-\! \hat{C}_{\text{C}}(\mathbf{u}_{j}, \mathbf{v}_{j}) \right\}^{\frac{1}{2}} \!\right],
%	\label{eq:defcim_Cluster}
%	\\
%	\hat{C}_{\text{C}}(\mathbf{u}_{j}, \mathbf{v}_{j}) = \frac{1}{d_{j}} \sum_{i=1}^{d_{j}} \kappa_{\sigma_{j}} \left( u_{i} - v_{i} \right),
%\end{IEEEeqnarray}

\noindent where $ \mathbf{y}_{k}^{\text{C}} = \left( \mathbf{v}_{1}, \mathbf{v}_{2},\ldots, \mathbf{v}_{J} \right) $ is a node $ \mathbf{y}_{k} $, but its attributes are grouped based on the attribute grouping of $ \mathbf{x}_{n}^{\text{C}} $. A bandwidth $ \sigma_{j} $ is defined as follows:
\begin{equation}
	\sigma_{j} = \frac{1}{d_{j}} \sum_{i=1}^{d_{j}} \left[ \left( \frac{4}{2+d_{j}} \right)^{\frac{1}{4+d_{j}}}  \Gamma_{i}  \lambda^{-\frac{1}{4+d_{j}}} \right],
	\label{eq:sigma_Clustering}
\end{equation}
where $ \Gamma_{i} $ denotes a rescale operator which is defined by the standard deviation of the $ i $th attribute value in the $ j $th attribute group among the $ \lambda $ data points.

In this paper, CAEAC with the clustering-based approach is called CAEAC-Clustering (CAEAC-C).

\begin{figure}[htbp]
	%	\includegraphics[width=1.1in, left]{figures/legend_cimCalc.pdf}
	%	\\
	\centering
	\subfloat[General Approach (CAEAC)]{
		\includegraphics[width=2.2in]{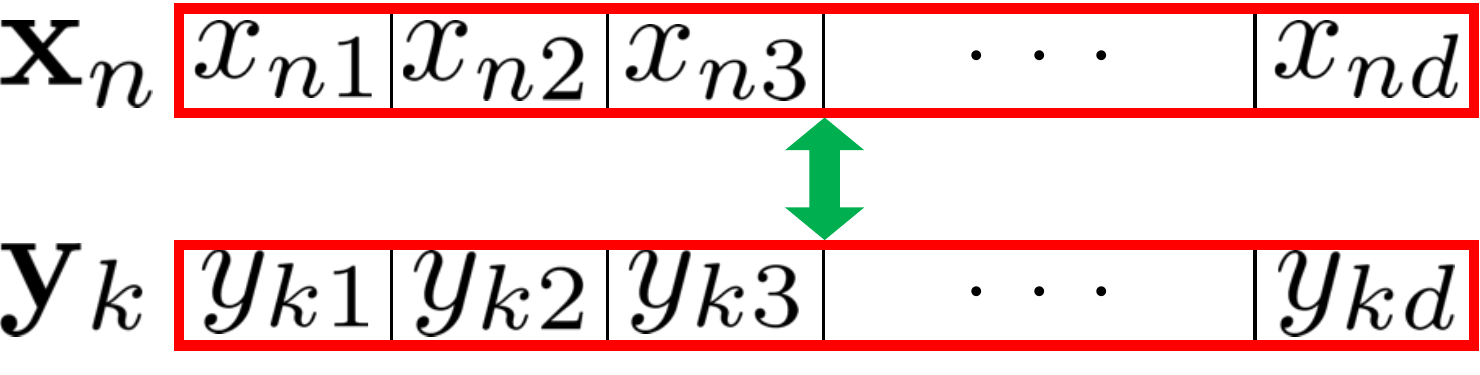}
		\label{fig:cim_original}
	}
	\vspace{5pt}
	\hfil
	\subfloat[Individual-based Approach (CAEAC-I)]{
		\includegraphics[width=2.2in]{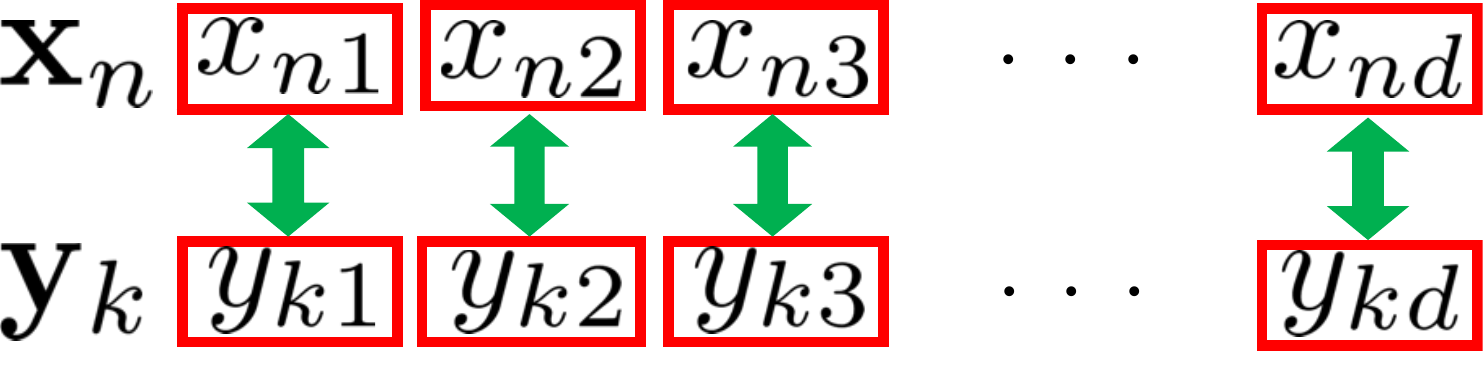}
		\label{fig:cim_individual}
	}
	\\
	\vspace{5pt}
	\subfloat[Clustering-based Approach (CAEAC-C)]{
		\includegraphics[width=3.3in]{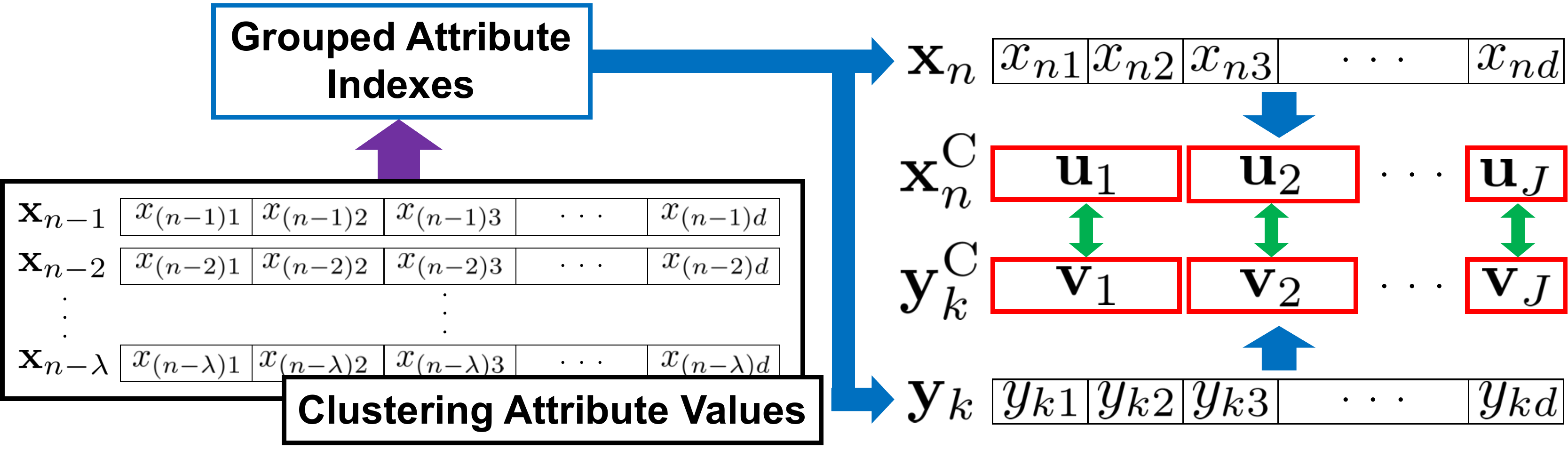}
		\label{fig:cim_clustering}
	}
	\\
	\vspace{5pt}
	\includegraphics[width=1.2in]{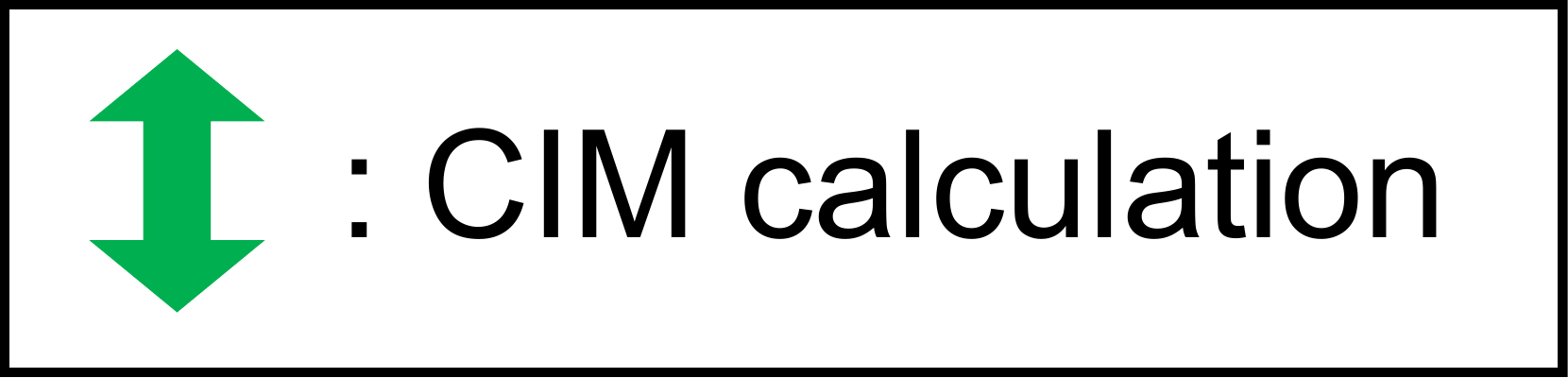}
	\caption{Differences in the CIM calculation.}
	\label{fig:cim_individual_clustering}
\end{figure}

The differences in attribute processing among the general approach, the individual-based approach (CAEAC-I), and the clustering-based approach (CAEAC-C) are shown in Fig. \ref{fig:cim_individual_clustering}. The source codes of CAEAC, CAEAC-I, and CAEAC-C are available on GitHub\footnote{\url{https://github.com/Masuyama-lab/CAEAC}}.

\section{Simulation Experiments}
\label{sec:experiment}
This section presents quantitative comparisons for classification performance of ASC \cite{shen08}, FTCAC, SOINN+C, GSOINN+ \cite{wiwatcharakoses21}, CAEAC, CAEAC-I, and CAEAC-C. The source code of ASC\footnote{\url{https://cs.nju.edu.cn/rinc/Soinn.html}}, FTCA\footnote{\url{https://github.com/Masuyama-lab/FTCA}}, SOINN+\footnote{\url{https://osf.io/6dqu9/}}, and GSOINN+\footnote{\url{https://osf.io/gqxya/}} is provided by the authors of the related papers. 

ASC is a similar approach of CAEAC. GSOINN+ is the state-of-the-art SOINN-based classification algorithm capable of continual learning. FTCAC and SOINN+C are the classifiers based on FTCA \cite{masuyamaFTCA} and SOINN+ \cite{wiwatcharakoses20}, respectively, which have the same architecture as CAEAC (i.e., Fig. \ref{fig:overviewCAEAC}). FTCA is a state-of-the-art ART-based clustering algorithm while SOINN+ is a state-of-the-art GNG-based clustering algorithm. Because those algorithms have similar characteristics to CAEA, we can construct the related classifiers (i.e., FTCAC and SOINN+C) and use them in our computational experiments.

\subsection{Datasets}
\label{sec:datasets}
We utilize five synthetic datasets and nine real-world datasets selected from the commonly used clustering benchmarks \cite{franti18} and public repositories \cite{liu17, dua19}. Table \ref{tab:datasets} summarizes statistics of the datasets.

\begin{table}[htbp]
	\renewcommand{\arraystretch}{1.0}
	\caption{
		Statistics of datasets for classification tasks
	}
	\label{tab:datasets}
	\scalebox{0.91}{ 
		\begin{tabular}{llrrr}
			% \hline\hline
			\hhline{-----}
			\hhline{-----}
			\multirow{2}{*}{Type} & \multirow{2}{*}{Dataset} & Number of  & Number of & Number of \\
			&                          & Attributes & Classes   & Instances \\
			\hline
			Synthetic             & Aggregation              & 2          & 7         & 788       \\
			& Compound                 & 2          & 6         & 399       \\
			& Hard Distribution         & 2          & 3         & 1,500     \\
			& Jain                     & 2          & 2         & 373       \\
			& Pathbased                & 2          & 3         & 300       \\
			\hline
			Real-world            & ALLAML                   & 7,129      & 2         & 72        \\
			& COIL20                   & 1,024      & 20        & 1,440     \\
			& Iris                     & 4          & 3         & 150       \\
			& Isolet                   & 617        & 26        & 7,797     \\
			& OptDigits                & 64         & 10        & 5,620     \\
			& Seeds                    & 7          & 3         & 210       \\
			& Semeion                  & 256        & 10        & 1,593     \\
			& Sonar                    & 60         & 2         & 208       \\
			& TOX171                   & 5,748      & 4         & 171       \\
			% \hline\hline
			\hhline{-----}
			\hhline{-----}
		\end{tabular}
	}
\end{table}

\subsection{Parameter Specifications}
\label{sec:palamSpec}

This section describes parameter settings of each algorithm for classification tasks in Section \ref{sec:classification}. ASC, FTCA, GSOINN+, CAEAC, CAEAC-I, and CAEAC-C have parameters that effect to the classification performance while SOINN+C does not have any parameters.

Table \ref{tab:paramAlgorithms} summarizes parameters of each algorithm. Because SOINN+C does not have any parameters, it is not listed in Table \ref{tab:paramAlgorithms}. In each algorithm, two parameters are specified by grid search. The ranges of grid search are the same as in the original paper of each algorithm, or wider. In ASC, a parameter $ k $ for noise reduction is the same setting as in \cite{shen08}. During grid search in our experiments, the training data points in each dataset are presented to each algorithm only once without pre-processing. For each parameter specification, we repeat the evaluation 20 times. In each of the 20 runs, first, training data points with no pre-processing are randomly ordered using a different random seed. Then, the re-ordered training data points are used for all algorithms.

Table \ref{tab:paramClassificationGrid} summarizes parameter values which are specified by grid search. N/A indicates that the corresponding algorithm could not build a predictive model. Using the parameter specifications in Tables \ref{tab:paramAlgorithms} and \ref{tab:paramClassificationGrid}, each algorithm shows the highest Accuracy for each dataset.

\begin{table}[htbp]
	\vspace{5mm}
	\caption{Parameter settings of each algorithm}
	\label{tab:paramAlgorithms}
	\begin{adjustwidth}{-4.1cm}{0cm}
		\newcolumntype{C}{>{\centering\arraybackslash}X}
		\scalebox{0.85}{
		\begin{tabular}{lllll}
			\hhline{-----}
			\hhline{-----}
			Algorithm & Parameter & Value & Grid Range & Note \\
			\hline
			ASC & $\lambda$ & grid search & \{20, 40, …, 400\} & a node insertion cycle \\
			& $ a_\text{max}$ & grid search & \{2, 4, …, 40\} & a maximum age of edge\\
			& $ k $ & 1 & --- & a parameter for noise reduction                    \\
			\hline
			FTCAC & $\lambda$ & grid search & \{20, 40, …, 400\} & a topology construction cycle \\
			& \textit{V} & grid search & \{0.01, 0.05, 0.10, …, 0.95\} & similarity threshold \\
			\hline
			GSOINN+ & \textit{p} & grid search & \{0.01, 0.05, 0.1, 0.3, 0.5, 0.7, 0.9, 1, 2, 5, 10\} & a parameter for fractional distance \\
			& $ k $ & grid search & \{1, 2, …, 10\} & the number of nearest neighbors for classification \\
			\hline
			CAEAC & $\lambda$ & grid search & \{10, 20, …, 100\} & an interval for adapting $\sigma$ \\
			& $ a_\text{max}$ & grid search & \{2, 4, …, 20\} & a maximum age of edge \\
			\hline
			CAEAC-I & $\lambda$ & grid search & \{10, 20, …, 100\} & an interval for adapting $\sigma$ \\
			& $ a_\text{max}$ & grid search & \{2, 4, …, 20\} & a maximum age of edge \\
			\hline
			CAEAC-C & $\lambda$ & grid search & \{10, 20, …, 100\} & an interval for adapting $\sigma$ \\
			& $ a_\text{max}$ & grid search & \{2, 4, …, 20\} & a maximum age of edge \\
			\hhline{-----}
			\hhline{-----}
		\end{tabular}
		}
	\end{adjustwidth}
\end{table}

\begin{table}[htbp]
	\caption{Parameters specified by grid search}
	\label{tab:paramClassificationGrid}
	\begin{adjustwidth}{-1.3cm}{0cm}
		\newcolumntype{C}{>{\centering\arraybackslash}X}
		\scalebox{0.85}{
			\begin{tabular}{ll|cc|cc|cc|cc|cc|cc}
				\hhline{--------------}
				\hhline{--------------}
				\multirow{2}{*}{Type} & \multirow{2}{*}{Dataset} & \multicolumn{2}{c|}{ASC}            & \multicolumn{2}{c|}{FTCAC}              & \multicolumn{2}{c|}{GSOINN+}                              & \multicolumn{2}{c|}{CAEAC}          & \multicolumn{2}{c|}{CAEAC-I}        & \multicolumn{2}{c}{CAEAC-C}        \\
				&                          & $\lambda$ & $\textit{a}_\text{max}$ & $\lambda$ & \textit{V} & \textit{p} & \textit{k} & $\lambda$ & $ a_\text{max}$ & $\lambda$ & $ a_\text{max}$ & $\lambda$ & $ a_\text{max}$ \\ \hline
				Synthetic             & Aggregation              & 140       & 34                      & 360       & 0.90                        & 0.50                        & 4                           & 90        & 4                       & 100       & 20                      & 50        & 14                      \\
				& Compound                 & 400       & 4                       & 320       & 0.95                        & 0.70                        & 1                           & 70        & 12                      & 60        & 18                      & 70        & 10                      \\
				& Hard Distribution         & 20        & 10                      & 100       & 0.35                        & 0.05                        & 7                           & 90        & 6                       & 100       & 18                      & 30        & 8                       \\
				& Jain                     & 20        & 34                      & 380       & 0.85                        & 0.01                        & 1                           & 40        & 6                       & 30        & 16                      & 20        & 10                      \\
				& Pathbased                & 240       & 6                       & 240       & 0.85                        & 0.05                        & 2                           & 60        & 14                      & 60        & 6                       & 70        & 2                       \\ \hline
				Real-world            & ALLAML                   & 20        & 14                      & 160       & 0.95                        & 0.70                        & 2                           & 70        & 10                      & 80        & 4                       & 80        & 10                      \\
				& COIL20                   & 240       & 38                      & 100       & 0.40                        & 0.50                        & 2                           & 90        & 12                      & 90        & 12                      & 100       & 4                       \\
				& Iris                     & 280       & 30                      & 400       & 0.70                        & 0.05                        & 3                           & 90        & 12                      & 90        & 12                      & 10        & 20                      \\
				& Isolet                   & 40        & 8                       & 360       & 0.70                        & 10.00                       & 2                           & 100       & 6                       & 70        & 12                      & 100       & 20                      \\
				& OptDigits                & 360       & 10                      & 320       & 0.70                        & 0.70                        & 2                           & 90        & 16                      & 90        & 4                       & 90        & 20                      \\
				& Seeds                    & 400       & 14                      & 320       & 0.65                        & 5.00                        & 7                           & 90        & 18                      & 90        & 2                       & 80        & 2                       \\
				& Semeion                  & 340       & 12                      & 260       & 0.50                        & 0.01                        & 2                           & 80        & 10                      & 80        & 18                      & 80        & 16                      \\
				& Sonar                    & 160       & 4                       & 360       & 0.55                        & 0.10                        & 1                           & 100       & 4                       & 60        & 10                      & 60        & 8                       \\
				& TOX171                   & 40        & 38                     &  \multicolumn{2}{c|}{N/A}                        & 0.70                        & 1                           & 70        & 14                      & 70        & 14                      & 100       & 12                      \\
				\hhline{--------------}
				\hhline{--------------}
			\end{tabular}
		}
	\end{adjustwidth}
\end{table}

\subsection{Classification Tasks}
\label{sec:classification}

\subsubsection{Conditions}
\label{sec:condition}
By using parameters in Tables \ref{tab:paramAlgorithms} and \ref{tab:paramClassificationGrid}, we repeat the evaluation 20 times. Similar to Section \ref{sec:palamSpec}, first, training data points with no pre-processing are randomly ordered using a different random seed. Then, the re-ordered training data points are used for all algorithms. The classification performance is evaluated by Accuracy, NMI \cite{strehl02}, and Adjusted Rand Index (ARI) \cite{hubert85}.

As a statistical analysis, the Friedman test and Nemenyi post-hoc analysis \cite{demvsar06} are utilized. The Friedman test is used to test the null hypothesis that all algorithms perform equally. If the null hypothesis is rejected, the Nemenyi post-hoc analysis is conducted. The Nemenyi post-hoc analysis is used for all pairwise comparisons based on the ranks of results over all the evaluation metrics for all datasets. Here, the null hypothesis is rejected at the significance level of $ 0.05 $ both in the Friedman test and the Nemenyi post-hoc analysis. All computations are carried out on Matlab 2020a with a 2.2GHz Xeon Gold 6238R processor and 768GB RAM.

\subsubsection{Results}
\label{sec:nonstationary}
Table \ref{tab:Classification} shows the results of the classification performance. The best value in each metric is indicated by bold for each dataset. The standard deviation is indicated in parentheses. N/A indicates that the corresponding algorithm could not build a predictive model. Training time is in [sec]. The brighter the cell color, the better the performance. In ASC, $ k $-means is applied to adjust the position of the generated nodes in order to achieve a good approximation of the distribution of data points. In addition, some generated nodes and edges are deleted during the learning process. Therefore, the number of clusters is not shown for ASC in Table \ref{tab:Classification}.

\begin{table}[htbp]
	\caption{Results of the classification performance.}
	\label{tab:Classification}
	\begin{adjustwidth}{-2.5cm}{0cm}
		\scalebox{0.68}{
		\newcolumntype{C}{>{\centering\arraybackslash}X}
		\begin{tabularx}{\fulllength}{lll|c|c|c|c|c|c|c}
			% \toprule
			\hhline{----------}
			\hhline{----------}
			Type       & Dataset          & Metric         & ASC                                     & FTCAC                                   & SOINN+C                                 & GSOINN+                                 & CAEAC                                   & CAEAC-I                                 & CAEAC-C                                 \\ \hhline{|-|-|-|-|-|-|-|-|-|-|}
				Synthetic  & Aggregation      & Accuracy       & \cellcolor[gray]{0.95} 0.999 (0.003)                  & \cellcolor[gray]{0.6} 0.997 (0.007)                           & \cellcolor[gray]{0.55} 0.997 (0.006)                           & \cellcolor[gray]{0.7} 0.997 (0.005)                           & \cellcolor[gray]{0.9} 0.998 (0.006)                           & \cellcolor[gray]{0.8} 0.998 (0.006)                           & \cellcolor[gray]{1.0} 0.999 (0.004)      \\
				&                  & NMI            & \cellcolor[gray]{1.0} 0.999 (0.006)                  & \cellcolor[gray]{0.7} 0.995 (0.012)                           & \cellcolor[gray]{0.55} 0.993 (0.012)                           & \cellcolor[gray]{0.6} 0.994 (0.011)                           & \cellcolor[gray]{0.9} 0.997 (0.011)                           & \cellcolor[gray]{0.8} 0.996 (0.013)                           & \cellcolor[gray]{0.95} 0.997 (0.008)                           \\
				&                  & ARI            & \cellcolor[gray]{1.0} 0.998 (0.007) &  \cellcolor[gray]{0.55} 0.994 (0.015)                           & \cellcolor[gray]{0.7} 0.994 (0.012)                           & \cellcolor[gray]{0.7} 0.994 (0.012)                           & \cellcolor[gray]{0.8} 0.995 (0.015)                           & \cellcolor[gray]{0.9} 0.997 (0.012)                           & \cellcolor[gray]{0.95} 0.998 (0.008)      \\
				&                  & Training Time  & \cellcolor[gray]{0.9} 0.100 (0.106)                           & \cellcolor[gray]{1.0} 0.013 (0.000)                           & \cellcolor[gray]{0.55} 2.385 (0.150)                           & \cellcolor[gray]{0.6} 2.019 (1.183)                           & \cellcolor[gray]{0.95} 0.031 (0.033)                           & \cellcolor[gray]{0.7} 0.252 (0.430)                           & \cellcolor[gray]{0.8} 0.242 (0.506)                           \\
				&                  & \# of Nodes    & 38.8 (3.4)                              & 151.8 (5.7)                             & 113.0 (11.3)                            & 58.7 (14.6)                             & 301.0 (23.2)                            & 310.9 (8.6)                             & 217.4 (8.4)                             \\
				&                  & \# of Clusters & ---                                     & 24.5 (3.2)                              & 31.0 (6.4)                              & 10.1 (2.5)                              & 240.7 (22.7)                            & 270.9 (6.0)                             & 92.5 (9.5)                              \\ \hhline{~|-|-|-|-|-|-|-|-|-|}
				& Compound         & Accuracy       & \cellcolor[gray]{0.7} 0.975 (0.026)                           & \cellcolor[gray]{0.6} 0.951 (0.032)                           & \cellcolor[gray]{0.8} 0.976 (0.024)                           & \cellcolor[gray]{0.55} 0.937 (0.054)                           & \cellcolor[gray]{0.95} 0.986 (0.021)                           & \cellcolor[gray]{1.0} 0.986 (0.015) & \cellcolor[gray]{0.9} 0.983 (0.020)                           \\
				&                  & NMI            & \cellcolor[gray]{0.8} 0.963 (0.034)                           & \cellcolor[gray]{0.6} 0.927 (0.042)                           & \cellcolor[gray]{0.8} 0.963 (0.034)                           & \cellcolor[gray]{0.55} 0.922 (0.059)                           & \cellcolor[gray]{1.0} 0.978 (0.030) & \cellcolor[gray]{0.95} 0.974 (0.028)                           & \cellcolor[gray]{0.9} 0.968 (0.035)                           \\
				&                  & ARI            & \cellcolor[gray]{0.7} 0.957 (0.044)                           & \cellcolor[gray]{0.6} 0.912 (0.054)                           & \cellcolor[gray]{0.8} 0.959 (0.042)                           & \cellcolor[gray]{0.55} 0.898 (0.093)                           & \cellcolor[gray]{0.95} 0.975 (0.037)                           & \cellcolor[gray]{1.0} 0.976 (0.026) & \cellcolor[gray]{0.9} 0.966 (0.039)                           \\
				&                  & Training Time  & \cellcolor[gray]{0.9} 0.052 (0.002)                           & \cellcolor[gray]{1.0} 0.008 (0.002)                           & \cellcolor[gray]{0.6} 0.559 (0.101)                           & \cellcolor[gray]{0.55} 0.658 (0.136)                           & \cellcolor[gray]{0.95} 0.012 (0.000)                           & \cellcolor[gray]{0.8} 0.098 (0.155)                           & \cellcolor[gray]{0.7} 0.131 (0.280)                           \\
				&                  & \# of Nodes    & 45.9 (2.9)                              & 71.9 (2.8)                              & 79.5 (10.8)                             & 42.1 (11.0)                             & 188.5 (11.5)                            & 175.4 (3.1)                             & 198.2 (4.7)                             \\
				&                  & \# of Clusters & ---                                     & 13.7 (2.3)                              & 32.1 (9.6)                              & 11.6 (6.3)                              & 154.6 (9.2)                             & 156.0 (5.4)                             & 130.7 (4.8)                             \\ \hhline{~|-|-|-|-|-|-|-|-|-|}
				& Hard Distribution & Accuracy       & \cellcolor[gray]{0.6} 0.991 (0.007)                           & \cellcolor[gray]{1.0} 0.996 (0.005) & \cellcolor[gray]{0.95} 0.993 (0.009)                           & \cellcolor[gray]{0.6} 0.991 (0.007)                           & \cellcolor[gray]{0.8} 0.992 (0.006)                           & \cellcolor[gray]{0.8} 0.992 (0.006)                           & \cellcolor[gray]{0.9} 0.992 (0.005)                           \\
				&                  & NMI            & 0.962 (0.028)                           & \cellcolor[gray]{1.0} 0.982 (0.023) & \cellcolor[gray]{0.95} 0.970 (0.035)                           & \cellcolor[gray]{0.55} 0.962 (0.027)                           & \cellcolor[gray]{0.7} 0.967 (0.026)                           & \cellcolor[gray]{0.6} 0.966 (0.025)                           & \cellcolor[gray]{0.8} 0.967 (0.021)                           \\
				&                  & ARI            & \cellcolor[gray]{0.6} 0.974 (0.019)                           & \cellcolor[gray]{1.0} 0.988 (0.015) & \cellcolor[gray]{0.95} 0.978 (0.026)                           & \cellcolor[gray]{0.55} 0.973 (0.020)                           & \cellcolor[gray]{0.7} 0.977 (0.018)                           & \cellcolor[gray]{0.8} 0.977 (0.017)                           & \cellcolor[gray]{0.9} 0.977 (0.016)                           \\
				&                  & Training Time  & \cellcolor[gray]{0.9} 0.203 (0.113)                           & \cellcolor[gray]{1.0} 0.023 (0.020)                           & \cellcolor[gray]{0.55} 3.134 (0.419)                           & \cellcolor[gray]{0.6} 3.036 (0.355)                           & \cellcolor[gray]{0.95} 0.066 (0.046)                           & \cellcolor[gray]{0.7} 0.905 (1.668)                           & \cellcolor[gray]{0.8} 0.556 (0.663)                           \\
				&                  & \# of Nodes    & 10.3 (1.1)                              & 47.3 (3.7)                              & 103.4 (24.2)                            & 63.4 (18.0)                             & 197.8 (21.0)                            & 219.9 (8.0)                             & 80.1 (5.3)         \\
				&                  & \# of Clusters & ---                                     & 3.8 (1.0)                               & 13.8 (5.8)                              & 6.4 (2.5)                               & 75.8 (21.1)                             & 97.9 (11.5)                             & 12.7 (4.2)                              \\ \hhline{~|-|-|-|-|-|-|-|-|-|}
				& Jain             & Accuracy       & \cellcolor[gray]{0.55} 0.883 (0.127)                           & \cellcolor[gray]{0.6} 0.988 (0.016)                           & \cellcolor[gray]{1.0} 1.000 (0.000) & \cellcolor[gray]{1.0} 1.000 (0.000) & \cellcolor[gray]{1.0} 1.000 (0.000) & \cellcolor[gray]{1.0} 1.000 (0.000) & \cellcolor[gray]{1.0} 1.000 (0.000) \\
				&                  & NMI            & \cellcolor[gray]{0.55} 0.612 (0.283)                           & \cellcolor[gray]{0.6} 0.921 (0.106)                           & \cellcolor[gray]{1.0} 1.000 (0.000) & \cellcolor[gray]{1.0} 1.000 (0.000) & \cellcolor[gray]{1.0} 1.000 (0.000) & \cellcolor[gray]{1.0} 1.000 (0.000) & \cellcolor[gray]{1.0} 1.000 (0.000) \\
				&                  & ARI            & \cellcolor[gray]{0.55} 0.627 (0.326)                           & \cellcolor[gray]{0.6} 0.948 (0.071)                           & \cellcolor[gray]{1.0} 1.000 (0.000) & \cellcolor[gray]{1.0} 1.000 (0.000) & \cellcolor[gray]{1.0} 1.000 (0.000)                 & \cellcolor[gray]{1.0} 1.000 (0.000) & \cellcolor[gray]{1.0} 1.000 (0.000) \\
				&                  & Training Time  & \cellcolor[gray]{0.9} 0.038 (0.001)                           & \cellcolor[gray]{1.0} 0.007 (0.001)                           & \cellcolor[gray]{0.55} 0.773 (0.210)                           & \cellcolor[gray]{0.6} 0.624 (0.105)                           & \cellcolor[gray]{0.95} 0.010 (0.001)                           & \cellcolor[gray]{0.8} 0.182 (0.355)                           & \cellcolor[gray]{0.7} 0.484 (0.800)                           \\
				&                  & \# of Nodes    & 6.0 (0.8)                               & 76.2 (4.1)                              & 49.4 (9.6)                              & 39.5 (13.6)                             & 46.1 (4.0)                              & 42.0 (6.5)                              & 31.1 (3.1)                              \\
				&                  & \# of Clusters & ---                                     & 14.5 (2.6)                              & 14.7 (6.1)                              & 10.4 (5.4)                              & 20.4 (3.0)                              & 20.9 (6.9)                              & 8.5 (2.8)                               \\ \hhline{~|-|-|-|-|-|-|-|-|-|}
				& Pathbased        & Accuracy       & \cellcolor[gray]{1.0} 0.993 (0.014) & \cellcolor[gray]{0.55} 0.937 (0.043)                           & \cellcolor[gray]{0.7} 0.970 (0.036)                           & \cellcolor[gray]{0.6} 0.967 (0.046)                           & \cellcolor[gray]{0.95} 0.992 (0.015)                           & \cellcolor[gray]{0.9} 0.990 (0.022)                           & \cellcolor[gray]{0.8} 0.987 (0.020)                           \\
				&                  & NMI            & \cellcolor[gray]{1.0} 0.980 (0.042) & \cellcolor[gray]{0.55} 0.830 (0.108)                           & \cellcolor[gray]{0.6} 0.912 (0.104)                           & \cellcolor[gray]{0.7} 0.913 (0.097)                           & \cellcolor[gray]{0.95} 0.975 (0.045)                           & \cellcolor[gray]{0.9} 0.969 (0.066)                           & \cellcolor[gray]{0.8} 0.958 (0.062)                           \\
				&                  & ARI            & \cellcolor[gray]{1.0} 0.980 (0.042) & \cellcolor[gray]{0.55} 0.816 (0.123)                           & \cellcolor[gray]{0.7} 0.909 (0.108)                           & \cellcolor[gray]{0.6} 0.904 (0.120)                           & \cellcolor[gray]{0.95} 0.976 (0.043)                           & \cellcolor[gray]{0.9} 0.970 (0.064)                           & \cellcolor[gray]{0.8} 0.958 (0.062)                           \\
				&                  & Training Time  & \cellcolor[gray]{0.9} 0.033 (0.002)                           & \cellcolor[gray]{1.0} 0.006 (0.000)                           & \cellcolor[gray]{0.6} 0.420 (0.081)                           & \cellcolor[gray]{0.55} 0.463 (0.096)                           & \cellcolor[gray]{0.95} 0.010 (0.000)                           & \cellcolor[gray]{0.7} 0.192 (0.263)                           & \cellcolor[gray]{0.8} 0.095 (0.202)                           \\
				&                  & \# of Nodes    & 30.3 (2.8)                              & 64.7 (5.9)                              & 51.9 (9.8)                              & 30.8 (6.6)                              & 104.9 (5.9)                             & 102.4 (3.1)                             & 114.9 (3.3)                             \\
				&                  & \# of Clusters & ---                                     & 17.6 (3.2)                              & 18.6 (7.9)                              & 7.4 (3.9)                               & 76.9 (4.3)                              & 80.6 (3.9)                              & 63.8 (4.0)                              \\ \hhline{-|-|-|-|-|-|-|-|-|-|}
				Real-world & ALLAML           & Accuracy       & \cellcolor[gray]{1.0} 0.915 (0.134) & \cellcolor[gray]{0.95} 0.911 (0.112)                           & \cellcolor[gray]{0.55} 0.770 (0.180)                           & \cellcolor[gray]{0.6} 0.799 (0.146)                           & \cellcolor[gray]{0.9} 0.890 (0.096)                           & \cellcolor[gray]{0.7} 0.887 (0.110)                           & \cellcolor[gray]{0.8} 0.888 (0.107)                           \\
				&                  & NMI            & \cellcolor[gray]{1.0} 0.721 (0.372) & \cellcolor[gray]{0.9} 0.720 (0.339)                           & \cellcolor[gray]{0.6} 0.407 (0.388)                           & \cellcolor[gray]{0.55} 0.304 (0.378)                           & \cellcolor[gray]{0.7} 0.586 (0.356)                           & \cellcolor[gray]{0.8} 0.591 (0.347)                           & \cellcolor[gray]{0.9} 0.595 (0.374)                           \\
				&                  & ARI            & \cellcolor[gray]{0.95} 0.675 (0.430)                           & \cellcolor[gray]{1.0} 0.680 (0.385) & \cellcolor[gray]{0.6} 0.357 (0.412)                           & \cellcolor[gray]{0.55} 0.268 (0.370)                           & \cellcolor[gray]{0.8} 0.540 (0.387)                           & \cellcolor[gray]{0.7} 0.537 (0.384)                           & \cellcolor[gray]{0.9} 0.577 (0.386)                           \\
				&                  & Training Time  & \cellcolor[gray]{0.95} 0.144 (0.019)                           & \cellcolor[gray]{0.8} 0.173 (0.007)                           & \cellcolor[gray]{1.0} 0.098 (0.012)                           & \cellcolor[gray]{0.9} 0.149 (0.040)                           & \cellcolor[gray]{0.7} 0.684 (0.026)                           & \cellcolor[gray]{0.55} 4.075 (7.105)                           & \cellcolor[gray]{0.6} 2.247 (2.489)                           \\
				&                  & \# of Nodes    & 4.3 (0.7)                               & 4.0 (0.0)                               & 20.6 (4.0)                              & 11.7 (3.3)                              & 60.5 (1.4)                              & 63.5 (0.9)                              & 62.5 (1.4)                              \\
				&                  & \# of Clusters & ---                                     & 2.0 (0.0)                               & 17.2 (4.1)                              & 9.6 (3.2)                               & 57.5 (2.5)                              & 62.9 (1.4)                              & 60.2 (2.7)                              \\ \hhline{~|-|-|-|-|-|-|-|-|-|}
				& COIL20           & Accuracy       & \cellcolor[gray]{0.8} 0.992 (0.007)                           & \cellcolor[gray]{0.7} 0.952 (0.024)                           & \cellcolor[gray]{0.6} 0.951 (0.018)                           & \cellcolor[gray]{0.55} 0.850 (0.050)                           & \cellcolor[gray]{1.0} 1.000 (0.000) & \cellcolor[gray]{1.0} 1.000 (0.000) & \cellcolor[gray]{0.9} 0.996 (0.007)                           \\
				&                  & NMI            & \cellcolor[gray]{0.8} 0.993 (0.007)                           & \cellcolor[gray]{0.6} 0.960 (0.018)                           & \cellcolor[gray]{0.7} 0.962 (0.014)                           & \cellcolor[gray]{0.55} 0.894 (0.032)                           & \cellcolor[gray]{1.0} 1.000 (0.000) & \cellcolor[gray]{1.0} 1.000 (0.000) & \cellcolor[gray]{0.9} 0.996 (0.007)                           \\
				&                  & ARI            & \cellcolor[gray]{0.8} 0.983 (0.017)                           & \cellcolor[gray]{0.6} 0.912 (0.044)                           & \cellcolor[gray]{0.7} 0.912 (0.037)                           & \cellcolor[gray]{0.55} 0.739 (0.089)                           & \cellcolor[gray]{1.0} 1.000 (0.000) & \cellcolor[gray]{1.0} 1.000 (0.000) & \cellcolor[gray]{0.9} 0.991 (0.016)                           \\
				&                  & Training Time  & \cellcolor[gray]{0.7} 3.106 (0.087)                           & \cellcolor[gray]{1.0} 0.545 (0.015)                           & \cellcolor[gray]{0.9} 1.603 (0.186)                           & \cellcolor[gray]{0.8} 2.938 (0.286)                           & \cellcolor[gray]{0.95} 1.409 (0.048)                           & \cellcolor[gray]{0.55} 11.990 (20.749)                         & \cellcolor[gray]{0.6} 5.679 (8.093)                           \\
				&                  & \# of Nodes    & 326.5 (14.6)                            & 103.7 (5.7)                             & 318.7 (14.9)                            & 98.2 (18.4)                             & 1069.4 (14.1)                           & 1055.6 (10.7)                           & 1001.0 (0.9)                            \\
				&                  & \# of Clusters & ---                                     & 24.4 (2.0)                              & 155.0 (15.5)                            & 34.0 (7.8)                              & 965.1 (19.9)                            & 940.1 (18.2)                            & 974.1 (1.5)                            \\ \hhline{~|-|-|-|-|-|-|-|-|-|}
				& Iris             & Accuracy       & \cellcolor[gray]{0.9} 0.973 (0.040)                           & \cellcolor[gray]{1.0} 0.977 (0.039) & \cellcolor[gray]{0.55} 0.937 (0.063)                           & \cellcolor[gray]{0.8} 0.967 (0.055)                           & \cellcolor[gray]{0.7} 0.963 (0.046)                           & \cellcolor[gray]{0.7} 0.963 (0.046)                           & \cellcolor[gray]{1.0} 0.977 (0.039) \\
				&                  & NMI            & \cellcolor[gray]{0.9} 0.934 (0.097)                           & \cellcolor[gray]{1.0} 0.944 (0.090) & \cellcolor[gray]{0.55} 0.854 (0.145)                           & \cellcolor[gray]{0.8} 0.921 (0.129)                           & \cellcolor[gray]{0.7} 0.917 (0.097)                           & \cellcolor[gray]{0.7} 0.917 (0.097)                           & \cellcolor[gray]{0.95} 0.942 (0.096)                           \\
				&                  & ARI            & \cellcolor[gray]{0.9} 0.922 (0.115)                           & \cellcolor[gray]{0.95} 0.923 (0.128)                           & \cellcolor[gray]{0.55} 0.823 (0.183)                           & \cellcolor[gray]{0.8} 0.899 (0.170)                           & \cellcolor[gray]{0.7} 0.888 (0.138)                           & \cellcolor[gray]{0.7} 0.888 (0.138)                           & \cellcolor[gray]{1.0} 0.932 (0.115) \\
				&                  & Training Time  & \cellcolor[gray]{0.9} 0.022 (0.001)                           & \cellcolor[gray]{1.0} 0.004 (0.000)                           & \cellcolor[gray]{0.7} 0.128 (0.030)                           & \cellcolor[gray]{0.55} 0.214 (0.050)                           & \cellcolor[gray]{0.95} 0.006 (0.000)                           & \cellcolor[gray]{0.8} 0.052 (0.148)                           & \cellcolor[gray]{0.6} 0.182 (0.267)                           \\
				&                  & \# of Nodes    & 12.1 (1.5)                              & 25.7 (2.6)                              & 37.1 (6.6)                              & 22.1 (5.2)                              & 134.0 (1.1)                             & 134.1 (0.9)                             & 21.5 (1.9)                              \\
				&                  & \# of Clusters & ---                                     & 5.5 (1.4)                               & 19.9 (5.8)                              & 8.7 (4.1)                               & 133.7 (1.7)                             & 133.7 (1.7)                             & 7.2 (1.7)                               \\ \hhline{~|-|-|-|-|-|-|-|-|-|}
				& Isolet           & Accuracy       & \cellcolor[gray]{0.95} 0.904 (0.010)                           & \cellcolor[gray]{1.0} 0.909 (0.012) & \cellcolor[gray]{0.6} 0.823 (0.015)                           & \cellcolor[gray]{0.55} 0.739 (0.029)                           & \cellcolor[gray]{0.8} 0.852 (0.014)                           & \cellcolor[gray]{0.7} 0.842 (0.013)                           & \cellcolor[gray]{0.9} 0.881 (0.012)                           \\
				&                  & NMI            & \cellcolor[gray]{0.95} 0.898 (0.010)                           & \cellcolor[gray]{1.0} 0.900 (0.010) & \cellcolor[gray]{0.6} 0.832 (0.012)                           & \cellcolor[gray]{0.55} 0.778 (0.022)                           & \cellcolor[gray]{0.8} 0.858 (0.010)                           & \cellcolor[gray]{0.7} 0.848 (0.010)                           & \cellcolor[gray]{0.9} 0.880 (0.009)                           \\
				&                  & ARI            & \cellcolor[gray]{0.95} 0.821 (0.021)                           & \cellcolor[gray]{1.0} 0.829 (0.023) & \cellcolor[gray]{0.6} 0.695 (0.025)                           & \cellcolor[gray]{0.55} 0.593 (0.041)                           & \cellcolor[gray]{0.8} 0.740 (0.021)                           & \cellcolor[gray]{0.7} 0.722 (0.021)                           & \cellcolor[gray]{0.9} 0.784 (0.024)                           \\
				&                  & Training Time  & \cellcolor[gray]{0.7} 13.601 (0.430)                          & \cellcolor[gray]{1.0} 2.191 (0.080)                           & \cellcolor[gray]{0.95} 4.358 (0.333)                           & \cellcolor[gray]{0.8} 11.282 (4.310)                          & \cellcolor[gray]{0.9} 7.468 (0.068)                           & \cellcolor[gray]{0.6} 19.951 (23.957)                         & \cellcolor[gray]{0.55} 29.610 (17.282)                         \\
				&                  & \# of Nodes    & 401.6 (22.4)                            & 164.9 (6.6)                             & 718.8 (40.8)                            & 189.4 (52.1)                            & 2003.8 (18.7)                           & 1556.9 (47.6)                           & 2449.2 (22.8)                           \\
				&                  & \# of Clusters & ---                                     & 34.8 (3.2)                              & 462.5 (27.3)                            & 94.6 (28.4)                             & 1574.6 (17.3)                           & 1198.1 (47.5)                           & 1444.7 (32.8)                          \\ \hhline{~|-|-|-|-|-|-|-|-|-|}
				& OptDigits        & Accuracy       & \cellcolor[gray]{1.0} 0.988 (0.005) & \cellcolor[gray]{0.95} 0.982 (0.005)                           & \cellcolor[gray]{0.6} 0.969 (0.009)                           & \cellcolor[gray]{0.55} 0.959 (0.008)                           & \cellcolor[gray]{0.8} 0.974 (0.007)                           & \cellcolor[gray]{0.7} 0.972 (0.005)                           & \cellcolor[gray]{0.9} 0.975 (0.009)                           \\
				&                  & NMI            & \cellcolor[gray]{1.0} 0.975 (0.009) & \cellcolor[gray]{0.95} 0.962 (0.011)                           & \cellcolor[gray]{0.6} 0.938 (0.015)                           & \cellcolor[gray]{0.55} 0.921 (0.013)                           & \cellcolor[gray]{0.8} 0.946 (0.013)                           & \cellcolor[gray]{0.7} 0.943 (0.010)                           & \cellcolor[gray]{0.9} 0.949 (0.016)                           \\
				&                  & ARI            & \cellcolor[gray]{1.0} 0.974 (0.010) & \cellcolor[gray]{0.95} 0.960 (0.012)                           & \cellcolor[gray]{0.6} 0.932 (0.019)                           & \cellcolor[gray]{0.55} 0.913 (0.017)                           & \cellcolor[gray]{0.8} 0.943 (0.015)                           & \cellcolor[gray]{0.7} 0.938 (0.012)                           & \cellcolor[gray]{0.9} 0.945 (0.019)                           \\
				&                  & Training Time  & \cellcolor[gray]{0.9} 1.286 (0.028)                           & \cellcolor[gray]{1.0} 0.643 (0.011)                           & \cellcolor[gray]{0.7} 4.932 (0.289)                           & \cellcolor[gray]{0.6} 6.160 (0.695)                           & \cellcolor[gray]{0.95} 0.650 (0.011)                           & \cellcolor[gray]{0.8} 4.042 (4.549)                           & \cellcolor[gray]{0.55} 6.645 (10.283)                          \\
				&                  & \# of Nodes    & 638.0 (21.2)                            & 1241.2 (13.7)                           & 487.3 (31.8)                            & 222.2 (34.0)                            & 819.3 (12.9)                            & 719.9 (22.6)                            & 769.3 (16.1)                            \\
				&                  & \# of Clusters & ---                                     & 36.8 (5.8)                              & 176.5 (16.3)                            & 56.9 (10.9)                             & 453.2 (15.4)                            & 463.2 (16.9)                            & 352.3 (16.5)                            \\ \hhline{~|-|-|-|-|-|-|-|-|-|}
				& Seeds            & Accuracy       & \cellcolor[gray]{0.55} 0.895 (0.061)                           & \cellcolor[gray]{0.7} 0.912 (0.054)                           & \cellcolor[gray]{0.6} 0.900 (0.055)                           & \cellcolor[gray]{0.95} 0.924 (0.064)                           & \cellcolor[gray]{0.9} 0.921 (0.062)                           & \cellcolor[gray]{0.8} 0.921 (0.068)                           & \cellcolor[gray]{1.0} 0.926 (0.063) \\
				&                  & NMI            & \cellcolor[gray]{0.6} 0.760 (0.126)                           & \cellcolor[gray]{0.7} 0.781 (0.120)                           & \cellcolor[gray]{0.55} 0.758 (0.139)                           & \cellcolor[gray]{0.9} 0.811 (0.143)                           & \cellcolor[gray]{0.8} 0.806 (0.142)                           & \cellcolor[gray]{0.95} 0.815 (0.143)                           & \cellcolor[gray]{1.0} 0.815 (0.141) \\
				&                  & ARI            & \cellcolor[gray]{0.55} 0.715 (0.160)                           & \cellcolor[gray]{0.6} 0.750 (0.141)                           & \cellcolor[gray]{0.7} 0.750 (0.143)                           & \cellcolor[gray]{0.9} 0.788 (0.173)                           & \cellcolor[gray]{0.8} 0.782 (0.164)                           & \cellcolor[gray]{1.0} 0.794 (0.155) & \cellcolor[gray]{0.95} 0.790 (0.167)                           \\
				&                  & Training Time  & \cellcolor[gray]{0.9} 0.025 (0.001)                           & \cellcolor[gray]{1.0} 0.005 (0.000)                           & \cellcolor[gray]{0.6} 0.213 (0.041)                           & \cellcolor[gray]{0.55} 0.328 (0.066)                           & \cellcolor[gray]{0.95} 0.009 (0.000)                           & \cellcolor[gray]{0.7} 0.132 (0.281)                           & \cellcolor[gray]{0.8} 0.050 (0.136)                           \\
				&                  & \# of Nodes    & 16.5 (2.1)                              & 34.0 (4.2)                              & 40.8 (6.9)                              & 25.2 (7.3)                              & 164.5 (3.9)                             & 159.7 (4.1)                             & 120.9 (0.9)                             \\
				&                  & \# of Clusters & ---                                     & 5.9 (1.7)                               & 18.8 (5.6)                              & 6.7 (4.1)                               & 156.4 (6.0)                             & 148.8 (7.0)                             & 75.8 (2.8)                              \\ \hhline{~|-|-|-|-|-|-|-|-|-|}
				& Semeion          & Accuracy       & \cellcolor[gray]{1.0} 0.924 (0.023) & \cellcolor[gray]{0.95} 0.916 (0.018)                           & \cellcolor[gray]{0.6} 0.819 (0.032)                           & \cellcolor[gray]{0.55} 0.720 (0.038)                           & \cellcolor[gray]{0.9} 0.889 (0.033)                           & \cellcolor[gray]{0.8} 0.886 (0.027)                           & \cellcolor[gray]{0.7} 0.884 (0.029)                           \\
				&                  & NMI            & \cellcolor[gray]{1.0} 0.887 (0.031) & \cellcolor[gray]{0.95} 0.873 (0.027)                           & \cellcolor[gray]{0.6} 0.753 (0.034)                           & \cellcolor[gray]{0.55} 0.660 (0.039)                           & \cellcolor[gray]{0.9} 0.838 (0.041)                           & \cellcolor[gray]{0.8} 0.834 (0.037)                           & \cellcolor[gray]{0.7} 0.831 (0.036)                           \\
				&                  & ARI            & \cellcolor[gray]{1.0} 0.841 (0.045) & \cellcolor[gray]{0.95} 0.826 (0.035)                           & \cellcolor[gray]{0.6} 0.650 (0.050)                           & \cellcolor[gray]{0.55} 0.512 (0.051)                           & \cellcolor[gray]{0.9} 0.772 (0.062)                           & \cellcolor[gray]{0.8} 0.768 (0.057)                           & \cellcolor[gray]{0.7} 0.761 (0.056)                           \\
				&                  & Training Time  & \cellcolor[gray]{0.9} 0.679 (0.022)                           & \cellcolor[gray]{1.0} 0.267 (0.010)                           & \cellcolor[gray]{0.8} 0.824 (0.066)                           & \cellcolor[gray]{0.7} 0.854 (0.142)                           & \cellcolor[gray]{0.95} 0.394 (0.009)                           & \cellcolor[gray]{0.55} 3.514 (4.117)                           & \cellcolor[gray]{0.6} 0.896 (0.894)                           \\
				&                  & \# of Nodes    & 234.8 (11.0)                            & 195.9 (7.4)                             & 231.2 (20.8)                            & 86.5 (19.0)                             & 596.7 (13.4)                            & 586.2 (13.9)                            & 579.2 (9.1)                             \\
				&                  & \# of Clusters & ---                                     & 21.8 (2.7)                              & 144.9 (15.9)                            & 45.8 (12.0)                             & 432.1 (18.2)                            & 402.5 (16.4)                            & 534.5 (12.6)                           \\ \hhline{~|-|-|-|-|-|-|-|-|-|}
				& Sonar            & Accuracy       & \cellcolor[gray]{0.8} 0.803 (0.080)                           & \cellcolor[gray]{0.6} 0.739 (0.126)                           & \cellcolor[gray]{0.55} 0.702 (0.108)                           & \cellcolor[gray]{0.7} 0.753 (0.074)                           & \cellcolor[gray]{0.95} 0.827 (0.090)                           & \cellcolor[gray]{0.9} 0.822 (0.117)                           & \cellcolor[gray]{1.0} 0.836 (0.077) \\
				&                  & NMI            & \cellcolor[gray]{0.8} 0.329 (0.187)                           & \cellcolor[gray]{0.7} 0.246 (0.207)                           & \cellcolor[gray]{0.55} 0.179 (0.160)                           & \cellcolor[gray]{0.6} 0.235 (0.154)                           & \cellcolor[gray]{0.9} 0.385 (0.217)                           & \cellcolor[gray]{1.0} 0.438 (0.253) & \cellcolor[gray]{0.95} 0.421 (0.223)                           \\
				&                  & ARI            & \cellcolor[gray]{0.8} 0.358 (0.208)                           & \cellcolor[gray]{0.7} 0.257 (0.229)                           & \cellcolor[gray]{0.55} 0.172 (0.183)                           & \cellcolor[gray]{0.6} 0.240 (0.161)                           & \cellcolor[gray]{0.9} 0.425 (0.235)                           & \cellcolor[gray]{0.95} 0.443 (0.270)                           & \cellcolor[gray]{1.0} 0.448 (0.226) \\
				&                  & Training Time  & \cellcolor[gray]{0.9} 0.030 (0.001)                           & \cellcolor[gray]{1.0} 0.007 (0.000)                           & \cellcolor[gray]{0.6} 0.184 (0.034)                           & \cellcolor[gray]{0.55} 0.189 (0.069)                           & \cellcolor[gray]{0.95} 0.021 (0.001)                           & \cellcolor[gray]{0.8} 0.120 (0.223)                           & \cellcolor[gray]{0.7} 0.148 (0.307)                           \\
				&                  & \# of Nodes    & 34.4 (4.2)                              & 7.1 (0.9)                               & 36.0 (8.8)                              & 27.9 (10.3)                             & 116.8 (27.3)                            & 70.5 (2.8)                              & 84.3 (5.0)                              \\
				&                  & \# of Clusters & ---                                     & 2.1 (0.3)                               & 19.4 (7.0)                              & 13.2 (6.1)                              & 99.1 (26.5)                             & 58.5 (4.2)                              & 40.5 (7.3)                              \\ \hhline{~|-|-|-|-|-|-|-|-|-|}
				& TOX171           & Accuracy       & \cellcolor[gray]{0.8} 0.760 (0.104)                           & \multirow{6}{*}{N/A}                                    & \cellcolor[gray]{0.7} 0.600 (0.138)                           & \cellcolor[gray]{0.6} 0.596 (0.101)                           & \cellcolor[gray]{0.95} 0.857 (0.077)                           & \cellcolor[gray]{1.0} 0.860 (0.074) & \cellcolor[gray]{0.9} 0.851 (0.082)                           \\
				&                  & NMI            & \cellcolor[gray]{0.8} 0.627 (0.143)                           &                                      & \cellcolor[gray]{0.7} 0.481 (0.144)                           & \cellcolor[gray]{0.6} 0.465 (0.100)                           & \cellcolor[gray]{1.0} 0.769 (0.111) & \cellcolor[gray]{1.0} 0.769 (0.111) & \cellcolor[gray]{0.9} 0.761 (0.115)                           \\
				&                  & ARI            & \cellcolor[gray]{0.8} 0.450 (0.206)                           &                                     & \cellcolor[gray]{0.7} 0.252 (0.179)                           & \cellcolor[gray]{0.6} 0.237 (0.141)                           & \cellcolor[gray]{0.95} 0.668 (0.168)                           & \cellcolor[gray]{1.0} 0.671 (0.167) & \cellcolor[gray]{0.9} 0.649 (0.183)                           \\
				&                  & Training Time  & \cellcolor[gray]{0.9} 0.614 (0.046)                           &                           & \cellcolor[gray]{1.0} 0.244 (0.030)                           & \cellcolor[gray]{0.95} 0.604 (0.180)                           & \cellcolor[gray]{0.8} 1.437 (0.111)                           & \cellcolor[gray]{0.6} 3.249 (1.746)                           & \cellcolor[gray]{0.7} 1.561 (0.330)                           \\
				&                  & \# of Nodes    & 40.0 (5.1)                              &                                      & 44.3 (4.6)                              & 27.0 (7.4)                              & 145.4 (1.1)                             & 145.0 (1.1)                             & 153.9 (0.3)                             \\
				&                  & \# of Clusters & ---                                     &                                      & 29.1 (3.8)                              & 14.3 (5.0)                              & 143.3 (1.9)                             & 141.5 (2.3)                             & 153.9 (0.3)                             \\
			% \bottomrule
			\hhline{----------}
			\hhline{----------}
		\end{tabularx}
		}
	\end{adjustwidth}
	\footnotesize \raggedright 
	\hspace{-23mm}The best value in each metric is indicated by bold. The standard deviation is indicated in parentheses. Training time is in [sec].
	
	\hspace{-23mm}The brighter the cell color, the better the performance.
	
	\hspace{-23mm}N/A indicates that the corresponding algorithm could not build a predictive model.
\end{table}

As an overall trend, ASC, CAEAC, CAEAC-I, and CAEAC-C show better classification performance than SOINN+C, and GSOINN+. FTCAC showed the best classification performance on several datasets whereas FTCAC cannot build a predictive model for TOX171. Regarding training time, ASC, FTCAC, and CAEAC are shorter than SOINN+C, GSOINN+, CAEAC-I, and CAEAC-C. With respect to generated nodes and clusters, CAEAC and its variants tend to generate a large number of nodes and clusters than those of compared algorithms.

\begin{figure}[htbp]
	\centering
	\includegraphics[width=4.0in]{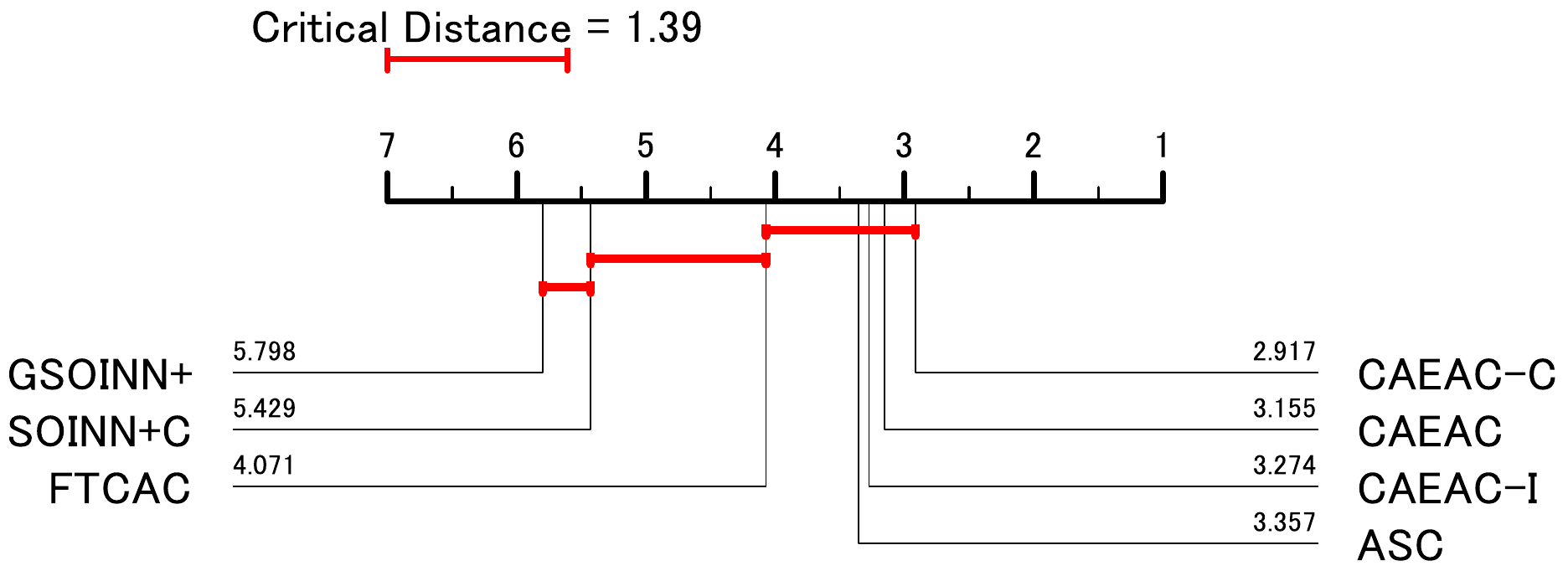}
	\caption{Critical difference diagram of classification tasks.}
	\label{fig:CriticalDifference}
	% \vspace{-1mm}
\end{figure}

Here, the null hypothesis is rejected on the Friedman test over all the evaluation metrics and datasets. Thus, we apply the Nemenyi post-hoc analysis. Fig. \ref{fig:CriticalDifference} shows a critical difference diagram based on the classification performance including all the evaluation metrics and datasets. Better performance is shown by lower average ranks, i.e., on the right side of a critical distance diagram. In theory, different algorithms within a critical distance (i.e., a red line) do not have a statistically significance difference \cite{demvsar06}. In Fig. \ref{fig:CriticalDifference}, CAEAC-C shows the lowest rank value, but there is no statistically significant difference from CAEAC, CAEAC-I, ASC, and FTCAC.

Among the compared algorithms, FTCAC cannot build a predictive model for TOX171 as shown in Table \ref{tab:Classification}. This is a critical problem of FTCAC. SOINN+C and GSOINN+ show inferior classification performance than ASC, CAEAC, CAEAC-I, and CAEAC-C as shown in Fig. \ref{fig:CriticalDifference} (and Table \ref{tab:Classification}). Although ASC shows comparable classification performance to CAEAC, CAEAC-I, and CAEAC-C, ASC deletes some generated nodes and edges during its learning process. Therefore, it may be difficult for ASC to maintain continual learning capability when learning additional data points after the deletion process. This can be regarded as the functional limitation of ASC.

The above observations suggest that CAEAC, CAEAC-I, and CAEAC-C have several advantages over compared algorithms not only in classification performance but also functional perspectives. The characteristics of CAEAC and its variants are summarized as follows:

\begin{itemize}
	\vspace{-1mm}
	\item \textbf{CAEAC}
	
	This algorithm shows stable and good classification performance with maintaining fast computation.
	
	\vspace{1mm}
	\item \textbf{CAEAC-I}
	
	The classification performance and computation time of this algorithm are both inferior to CAEAC and CAEAC-C. However, in Table \ref{tab:Classification}, it has the highest number of best evaluation metric values among CAEAC and its variants. Therefore, this algorithm is worth trying when high classification performance is desired.
	
	\vspace{1mm}
	\item \textbf{CAEAC-C}
	
	This algorithm can be the first-choice algorithm because it shows superior classification performance than CAEAC and CAEAC-I.
\end{itemize}

\subsection{Sensitivity of Parameters}
\label{sec:paramSensitiveity}
\begin{figure}[htbp]
	\begin{adjustwidth}{-\extralength}{0cm}
	\centering
	\subfloat[Aggregation]{
		\includegraphics[width=1.35in]{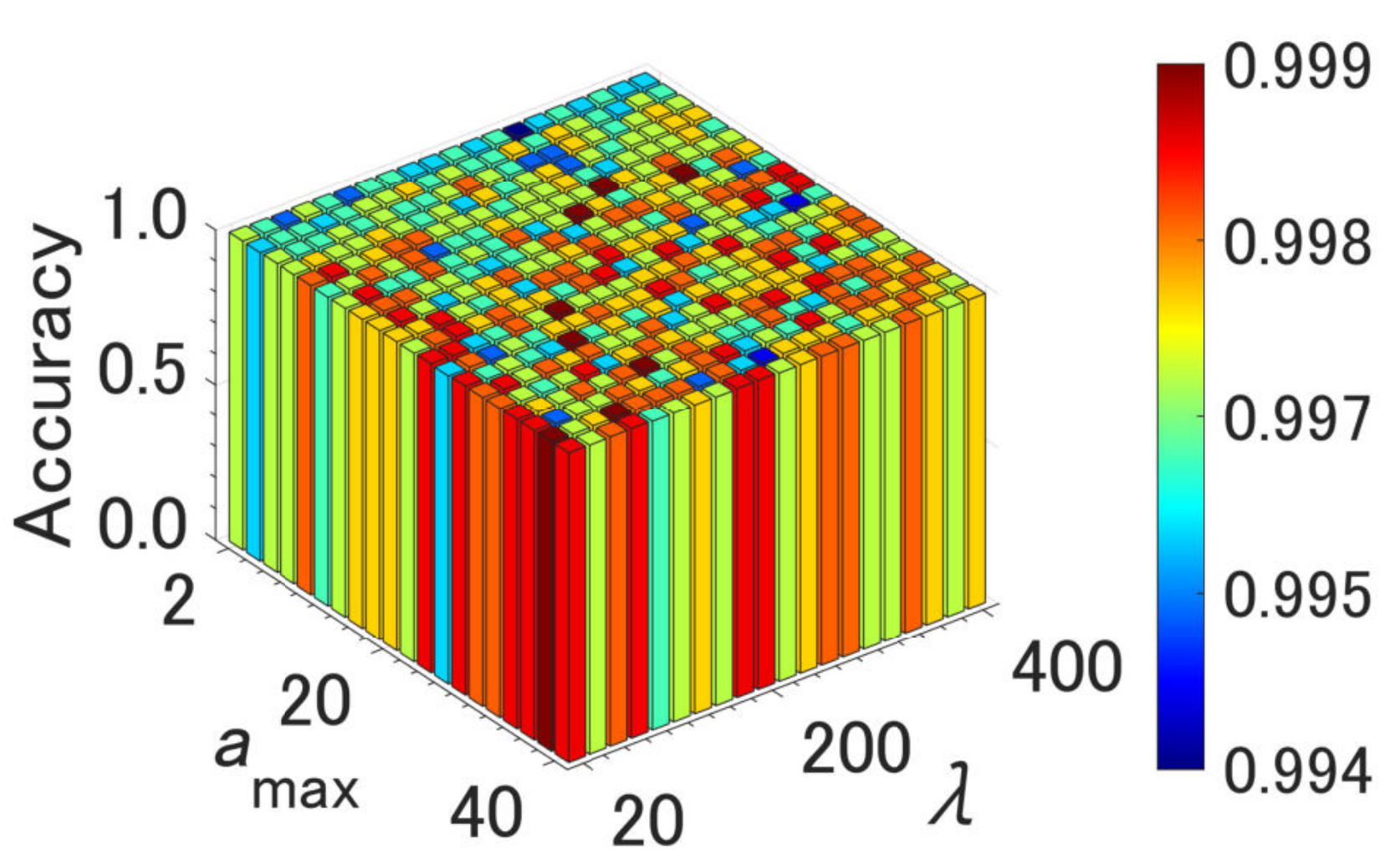}
		\label{fig:PS_Aggregation_ASC}
	}
	% \hspace{-1.4mm}
	% \hfil
	\subfloat[Compound]{
		\includegraphics[width=1.35in]{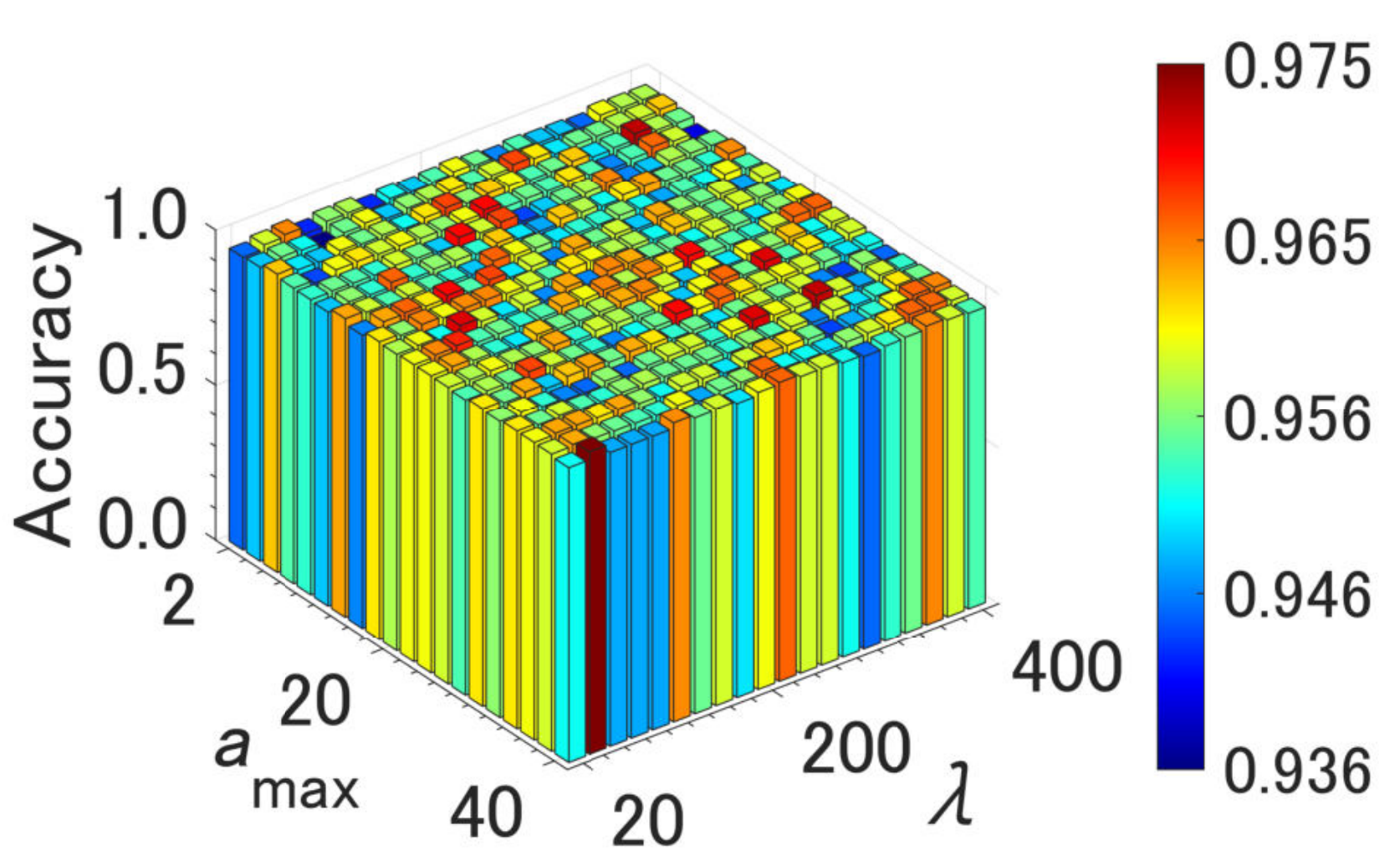}
		\label{fig:PS_Compound_ASC}
	}
	% \hspace{-1.4mm}
	% \hfil
	\subfloat[Hard Distribution]{
		\includegraphics[width=1.35in]{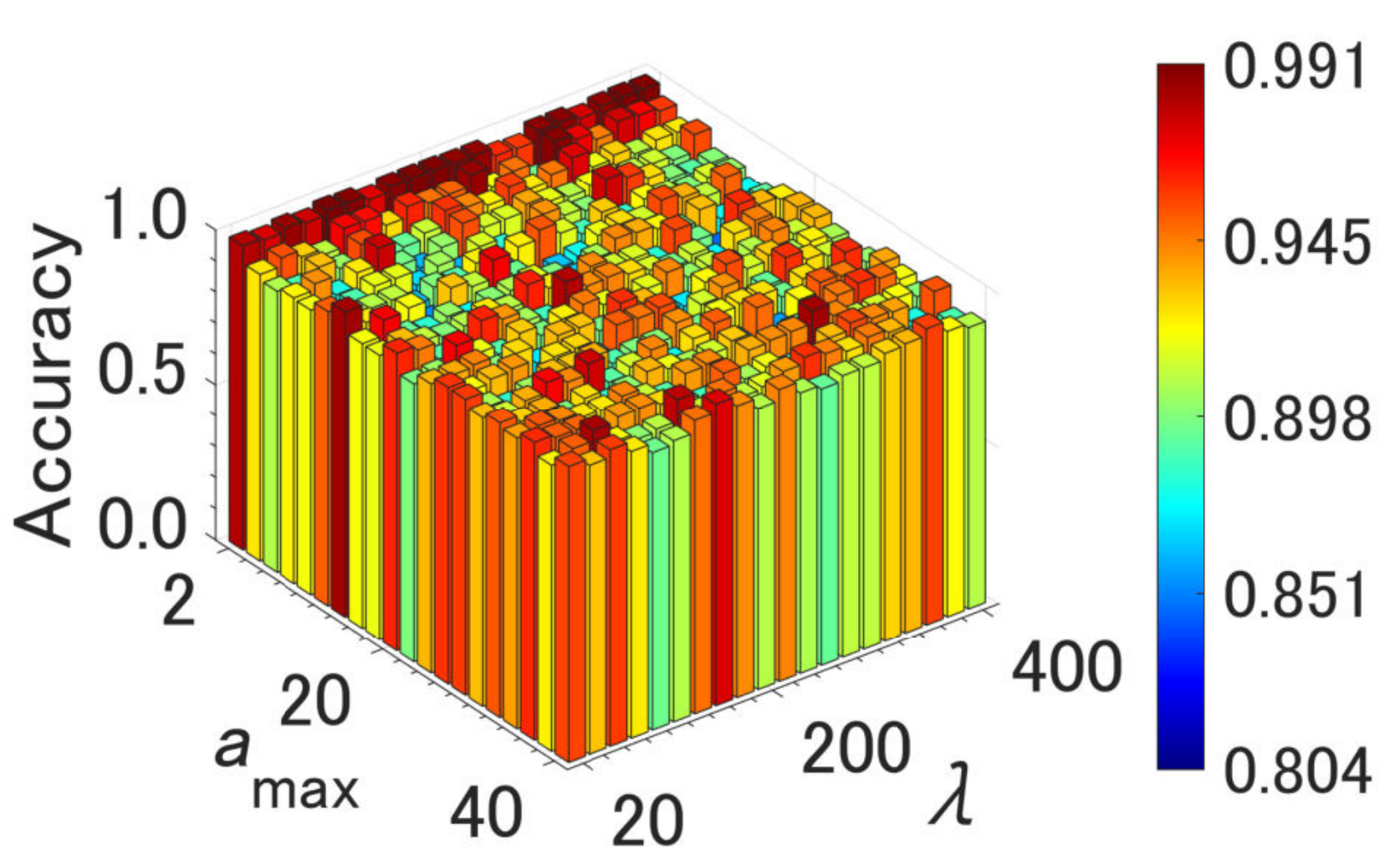}
		\label{fig:PS_HardDistribution_ASC}
	}
	% \hspace{-1.4mm}
	% \hfil
	\subfloat[Jain]{
		\includegraphics[width=1.35in]{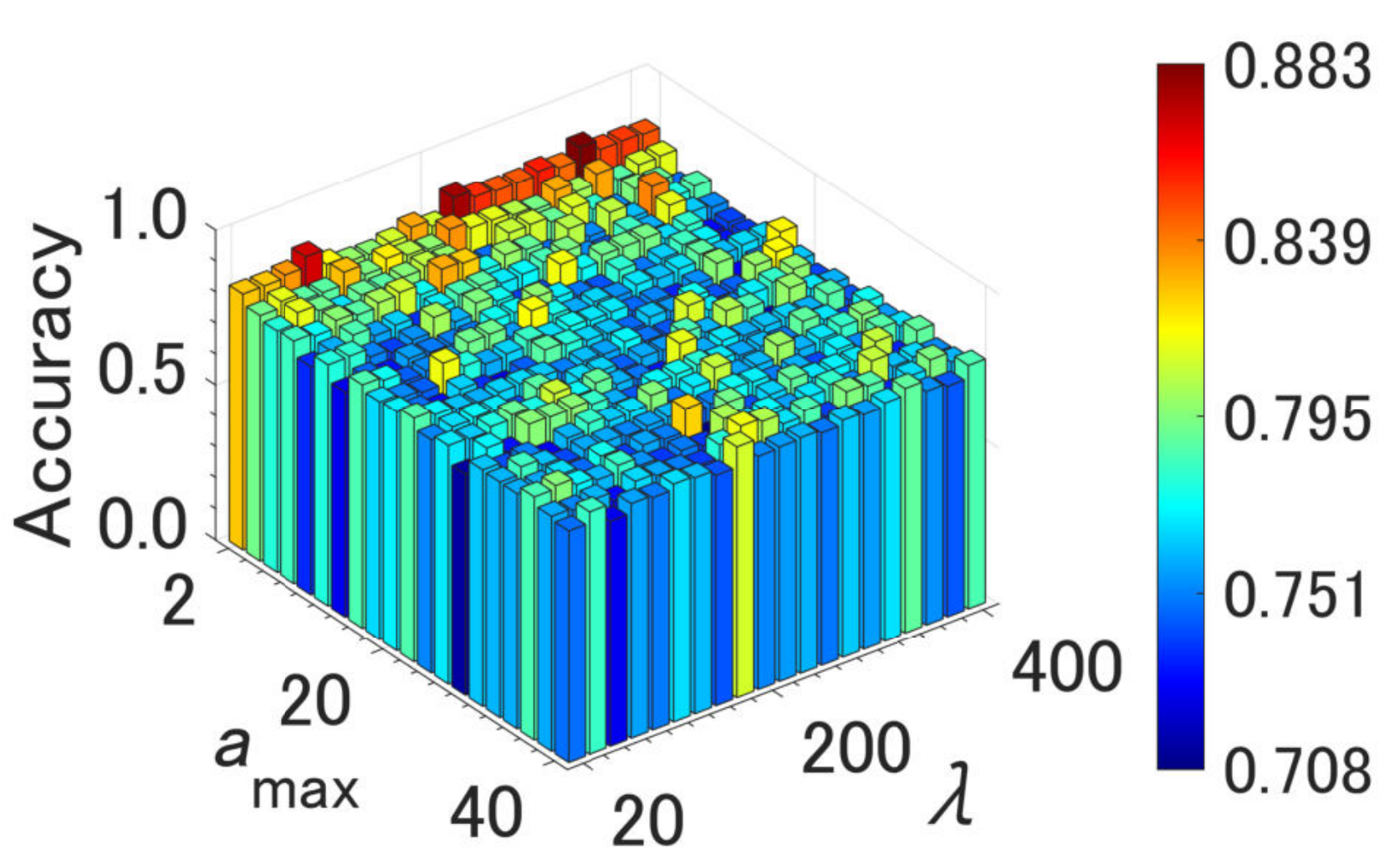}
		\label{fig:PS_Jain_ASC}
	}
	% \hspace{-1.4mm}
	% \hfil
	\subfloat[Pathbased]{
		\includegraphics[width=1.35in]{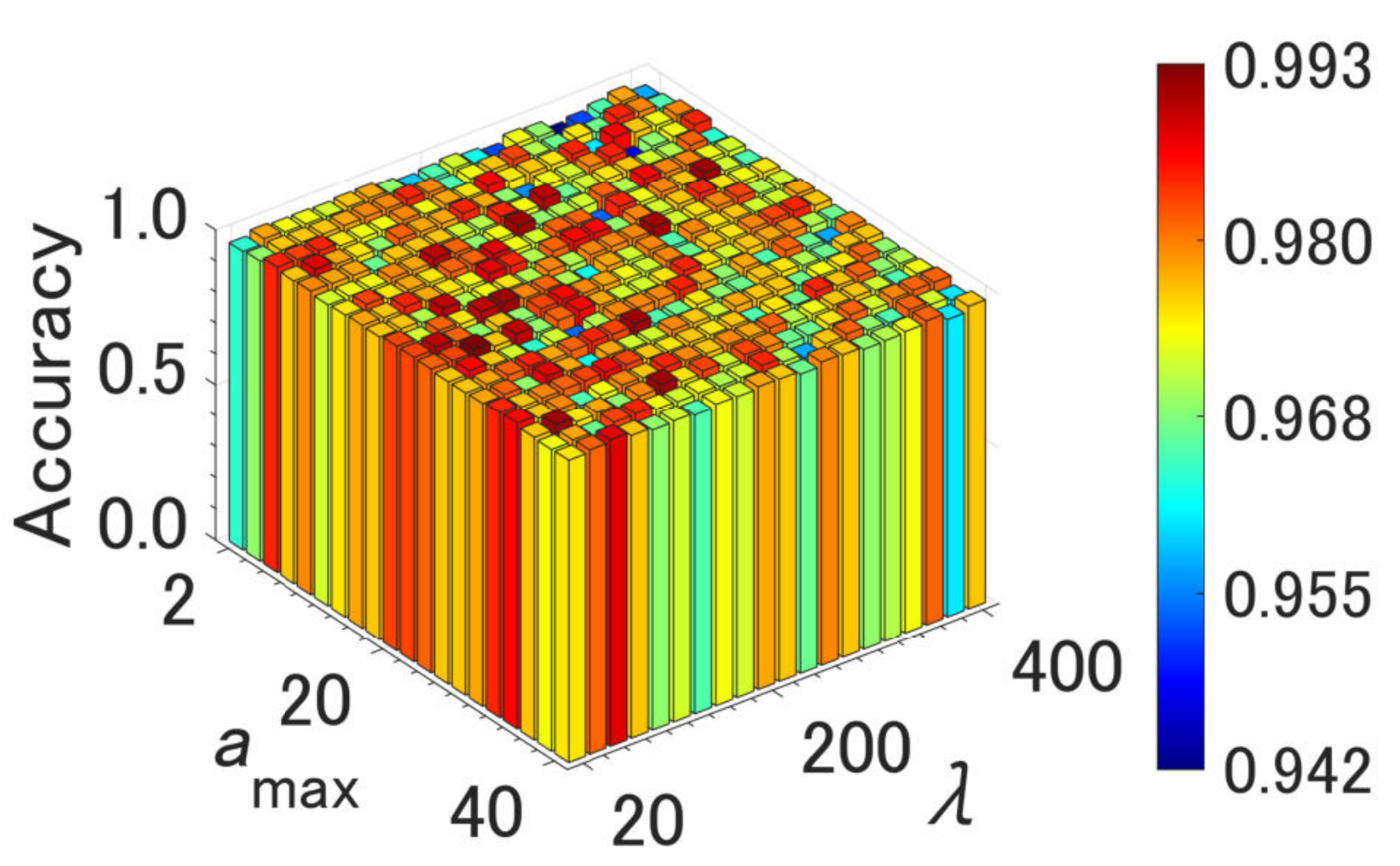}
		\label{fig:PS_Pathbased_ASC}
	}
	\\
	% \vspace{-2.5mm}
	%	\hfil
	% \hspace{-1.4mm}
	\subfloat[ALLAML]{
		\includegraphics[width=1.35in]{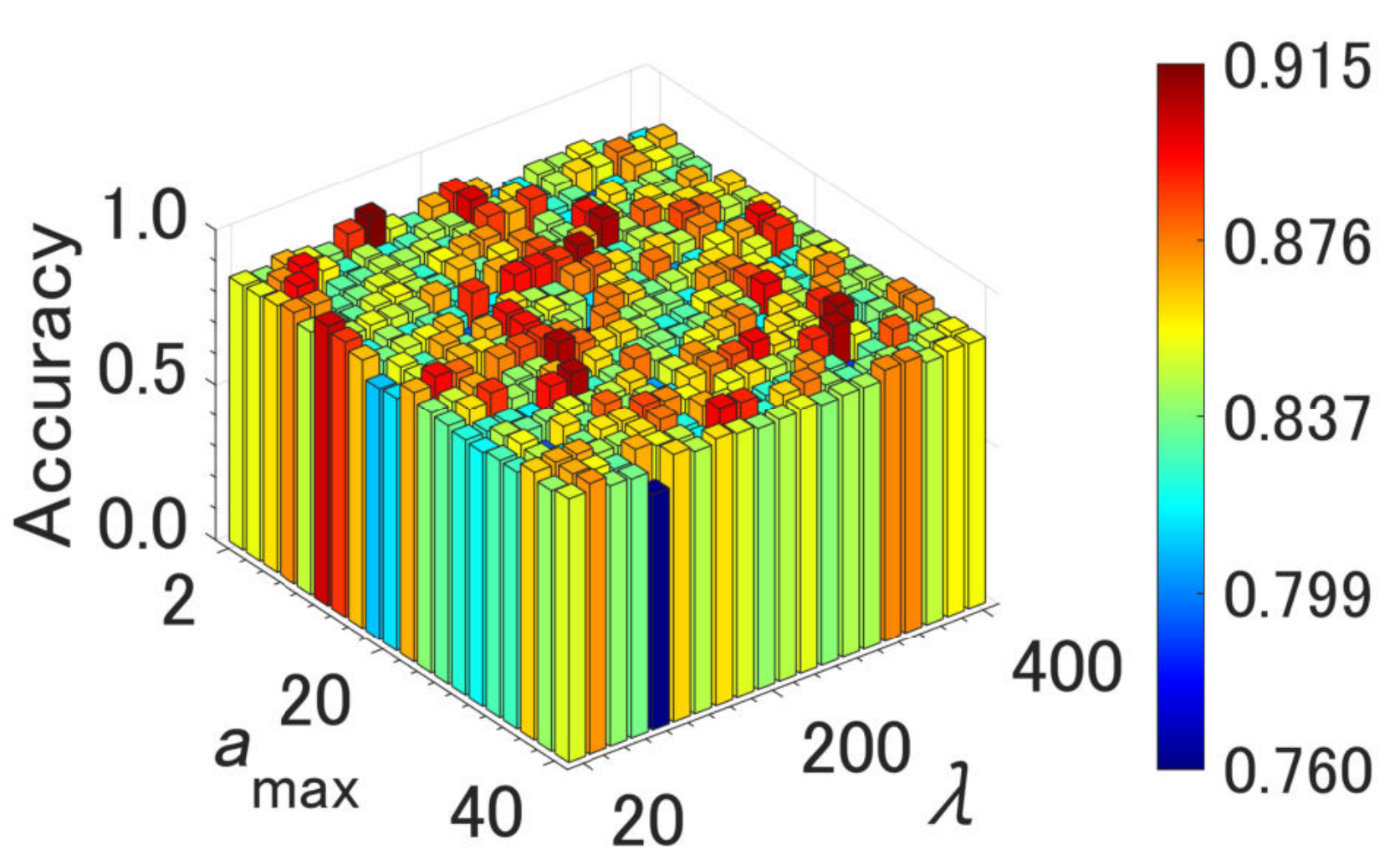}
		\label{fig:PS_ALLAML_ASC}
	}
	% \hspace{-1.4mm}
	% \hfil
	\subfloat[COIL20]{
		\includegraphics[width=1.35in]{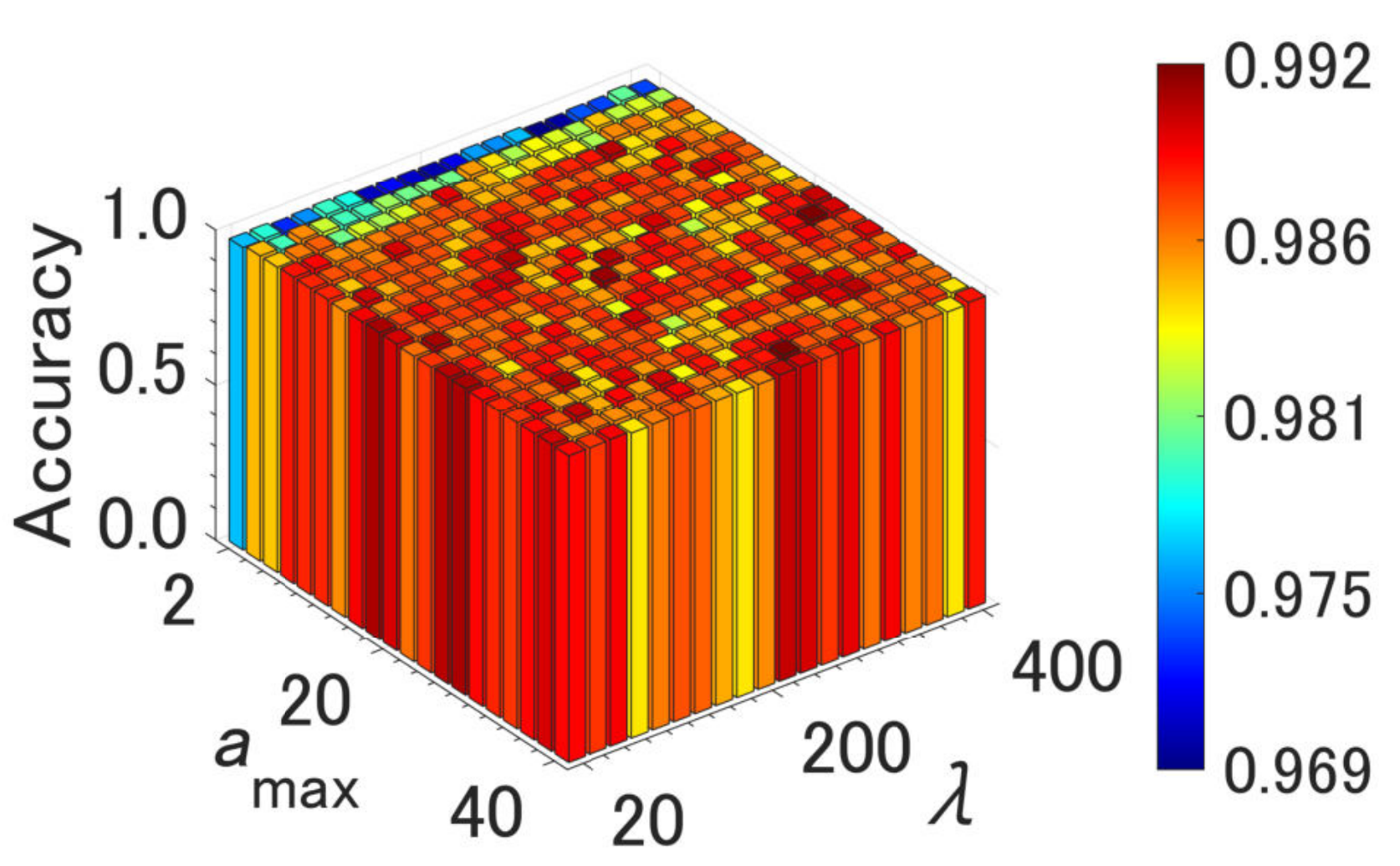}
		\label{fig:PS_COIL20_ASC}
	}
	% \hspace{-1.4mm}
	% \hfil
	\subfloat[Iris]{
		\includegraphics[width=1.35in]{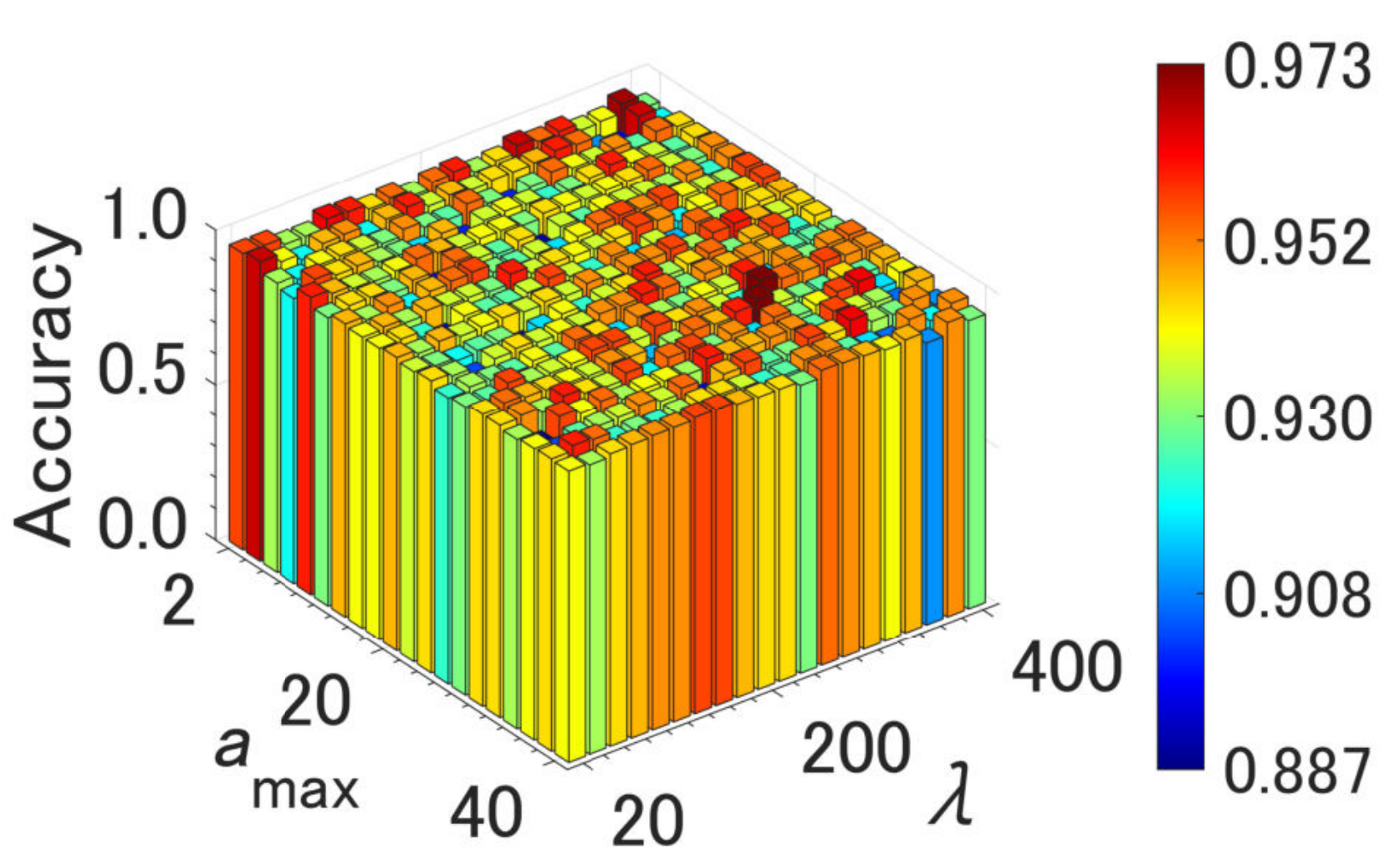}
		\label{fig:PS_Iris_ASC}
	}
	% \hspace{-1.4mm}
	% \hfil
	\subfloat[Isolet]{
		\includegraphics[width=1.35in]{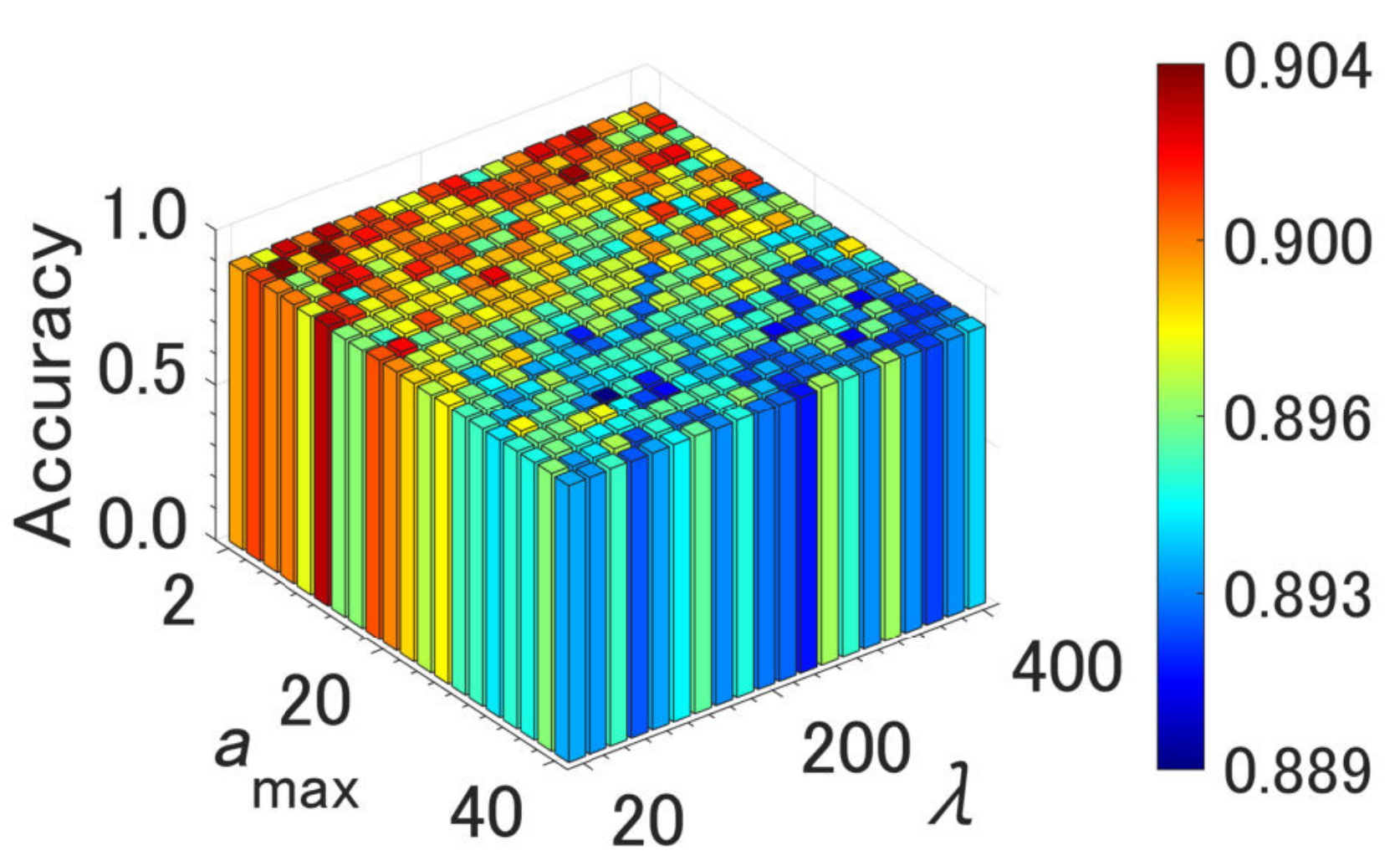}
		\label{fig:PS_Isolet_ASC}
	}
	% \hspace{-1.4mm}
	% \hfil
	\subfloat[OptDigits]{
		\includegraphics[width=1.35in]{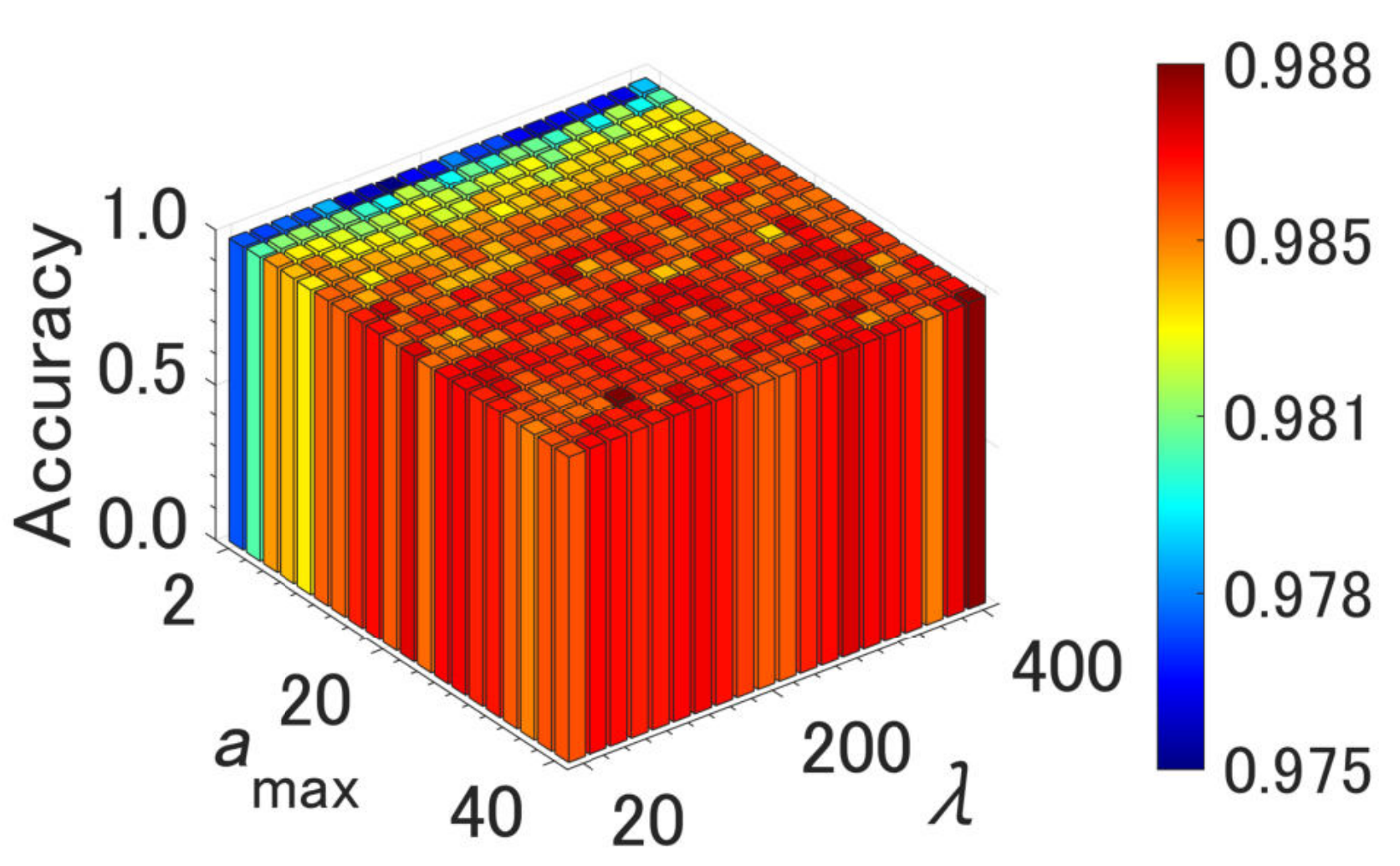}
		\label{fig:PS_OptDigits_ASC}
	}
	\\
	% \vspace{-2.5mm}
	%	\hfil
	\subfloat[Seeds]{
		\includegraphics[width=1.35in]{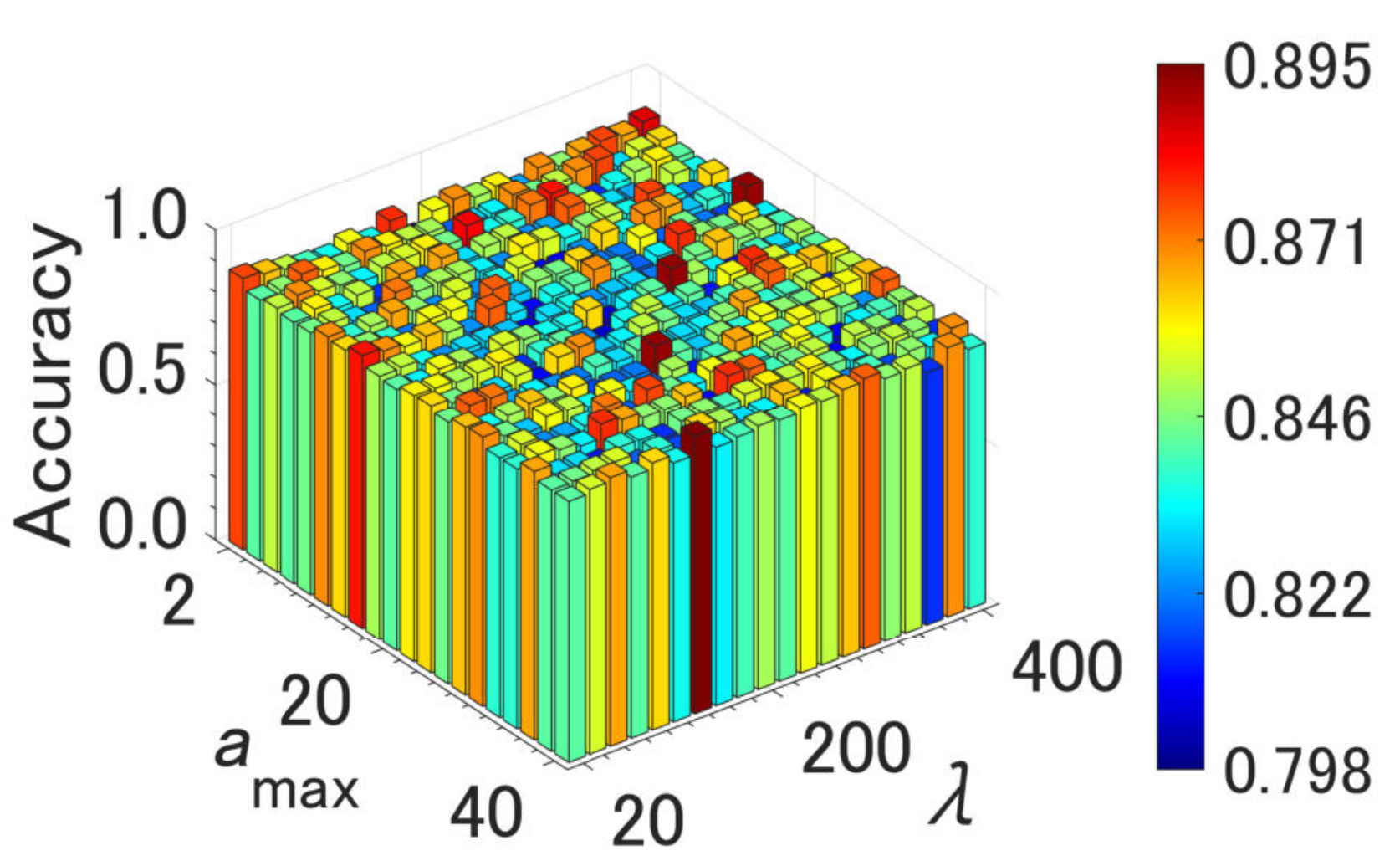}
		\label{fig:PS_Seeds_ASC}
	}
	% \hspace{2mm}
	%	\hfil
	\subfloat[Semeion]{
		\includegraphics[width=1.35in]{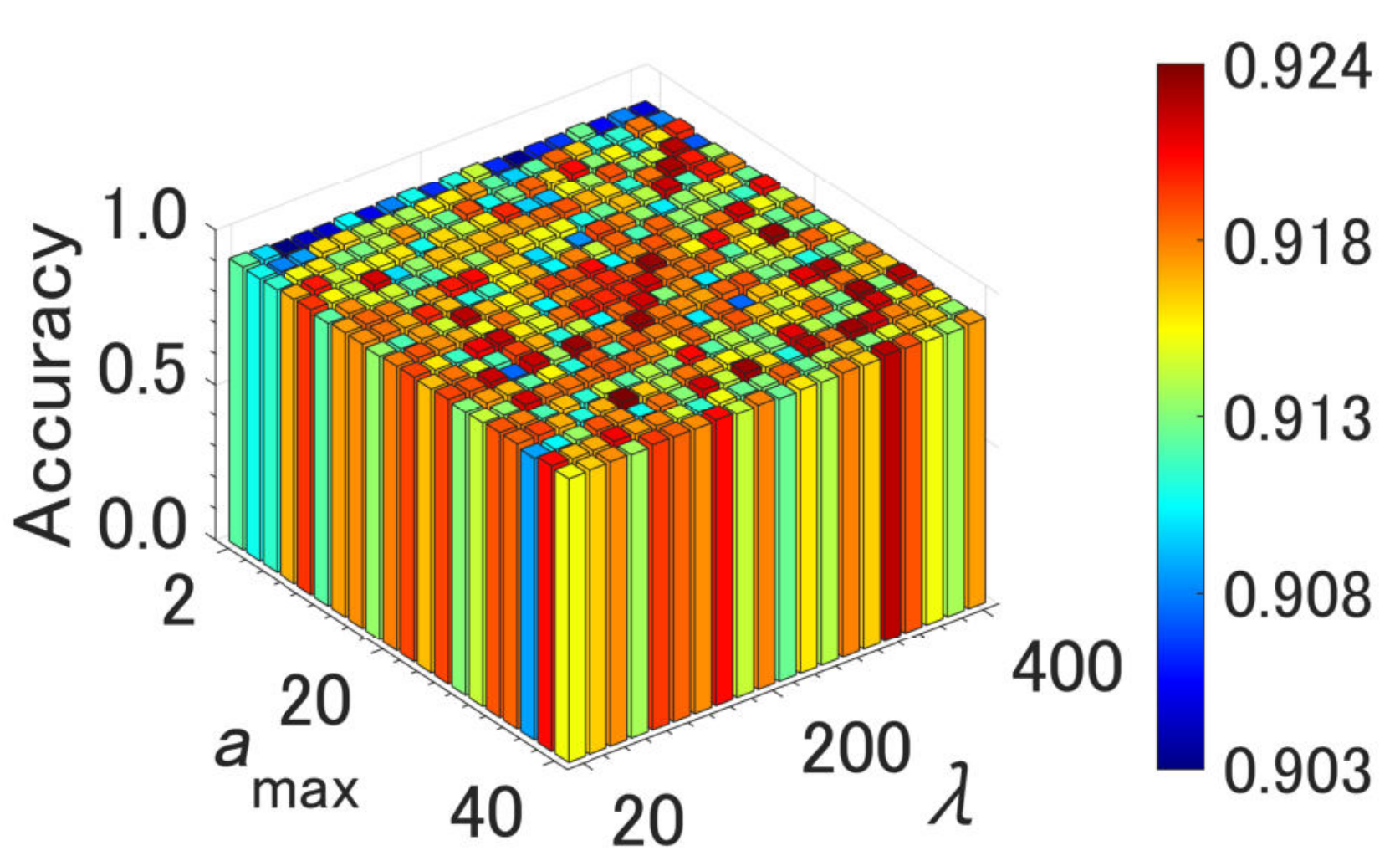}
		\label{fig:PS_Semeion_ASC}
	}
	% \hspace{2mm}
	%	\hfil
	\subfloat[Sonar]{
		\includegraphics[width=1.35in]{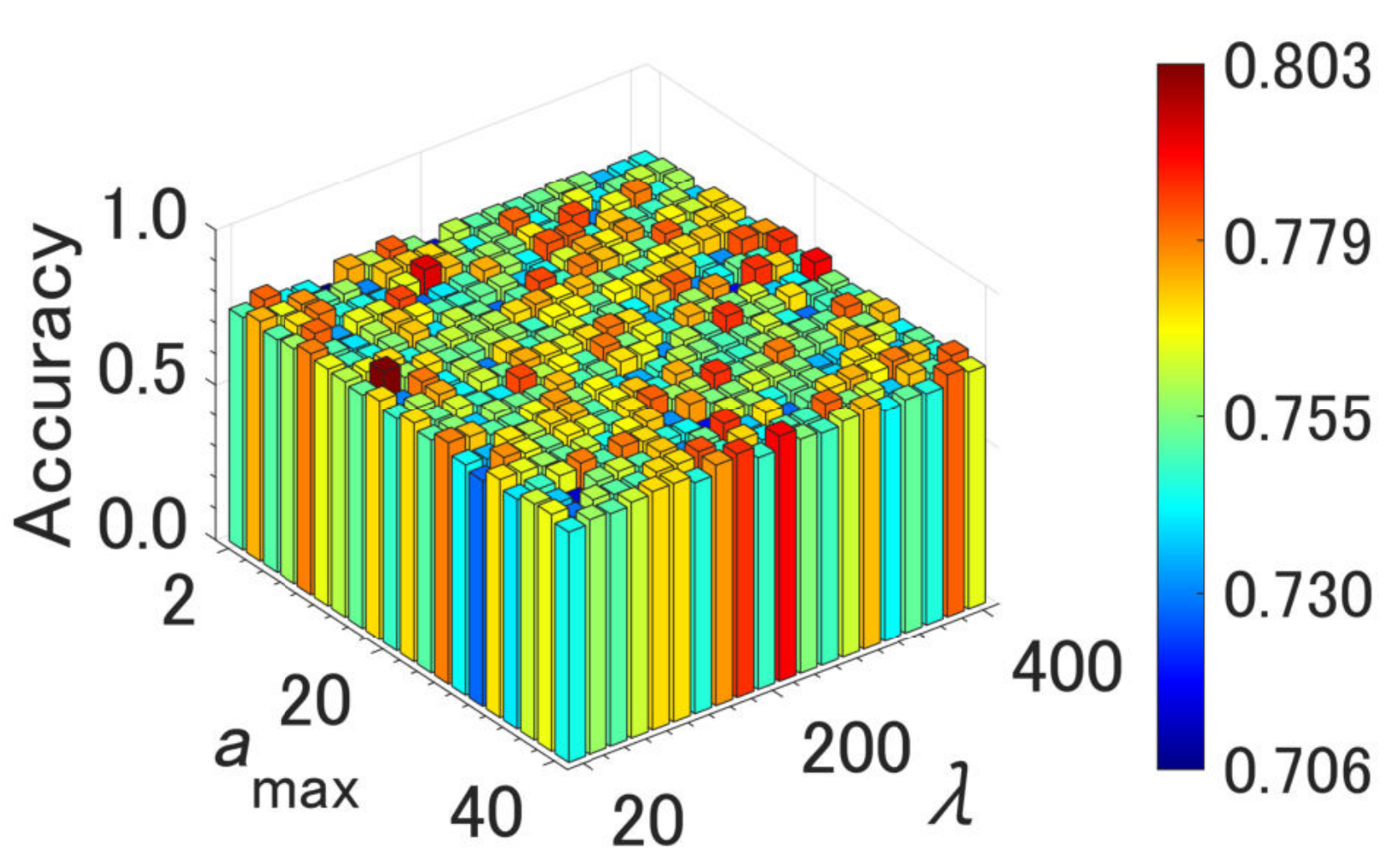}
		\label{fig:PS_Sonar_ASC}
	}
	% \hspace{2mm}
	%	\hfil
	\subfloat[TOX171]{
		\includegraphics[width=1.35in]{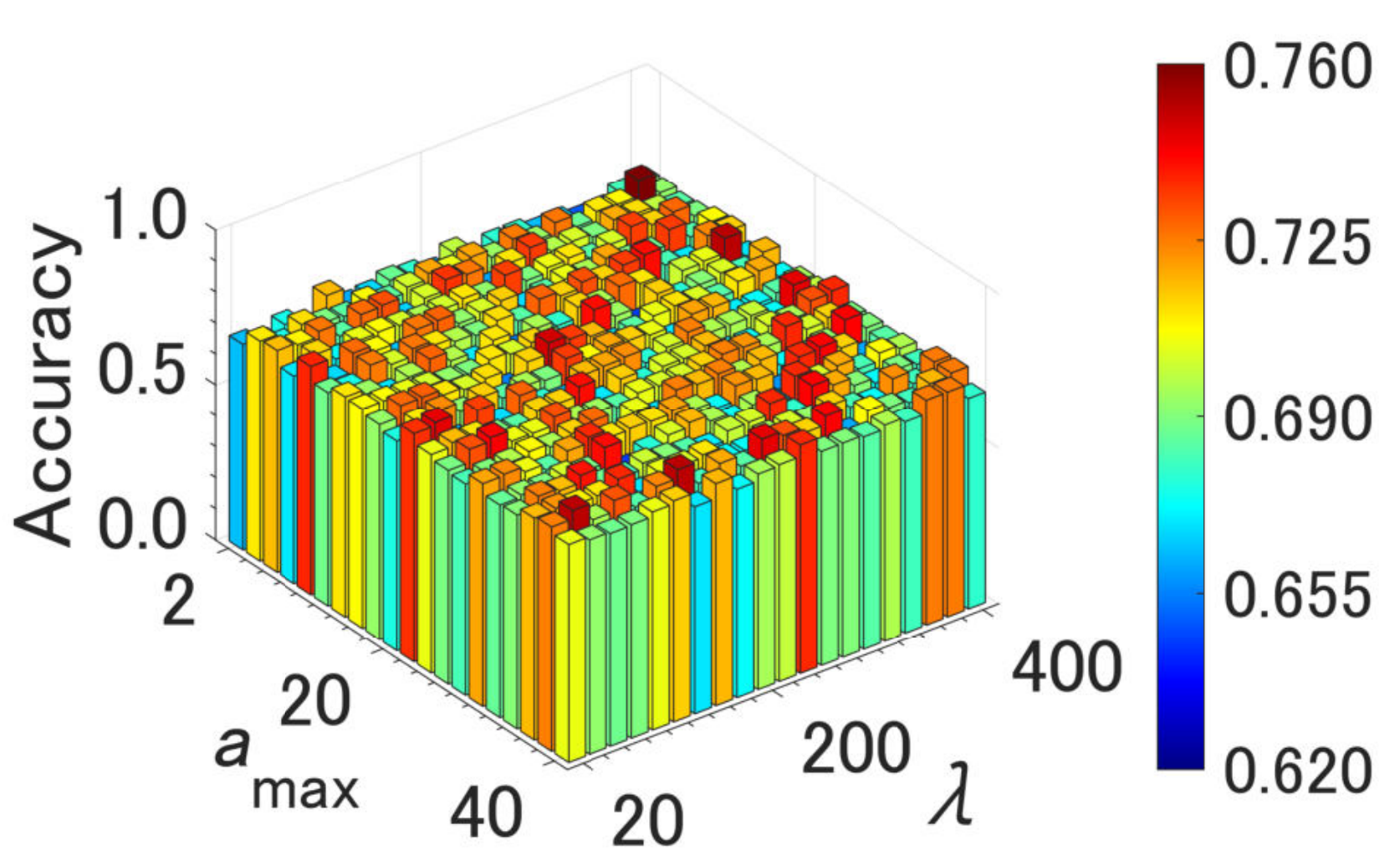}
		\label{fig:PS_TOX171_ASC}
	}
	% \vspace{-1mm}
	\end{adjustwidth}
	\caption{Effects of the parameter specifications of ASC on Accuracy.}
	\label{fig:paramSensitivity_ASC}
\end{figure}

\begin{figure}[htbp]
	\begin{adjustwidth}{-\extralength}{0cm}
	\centering
	\subfloat[Aggregation]{
		\includegraphics[width=1.35in]{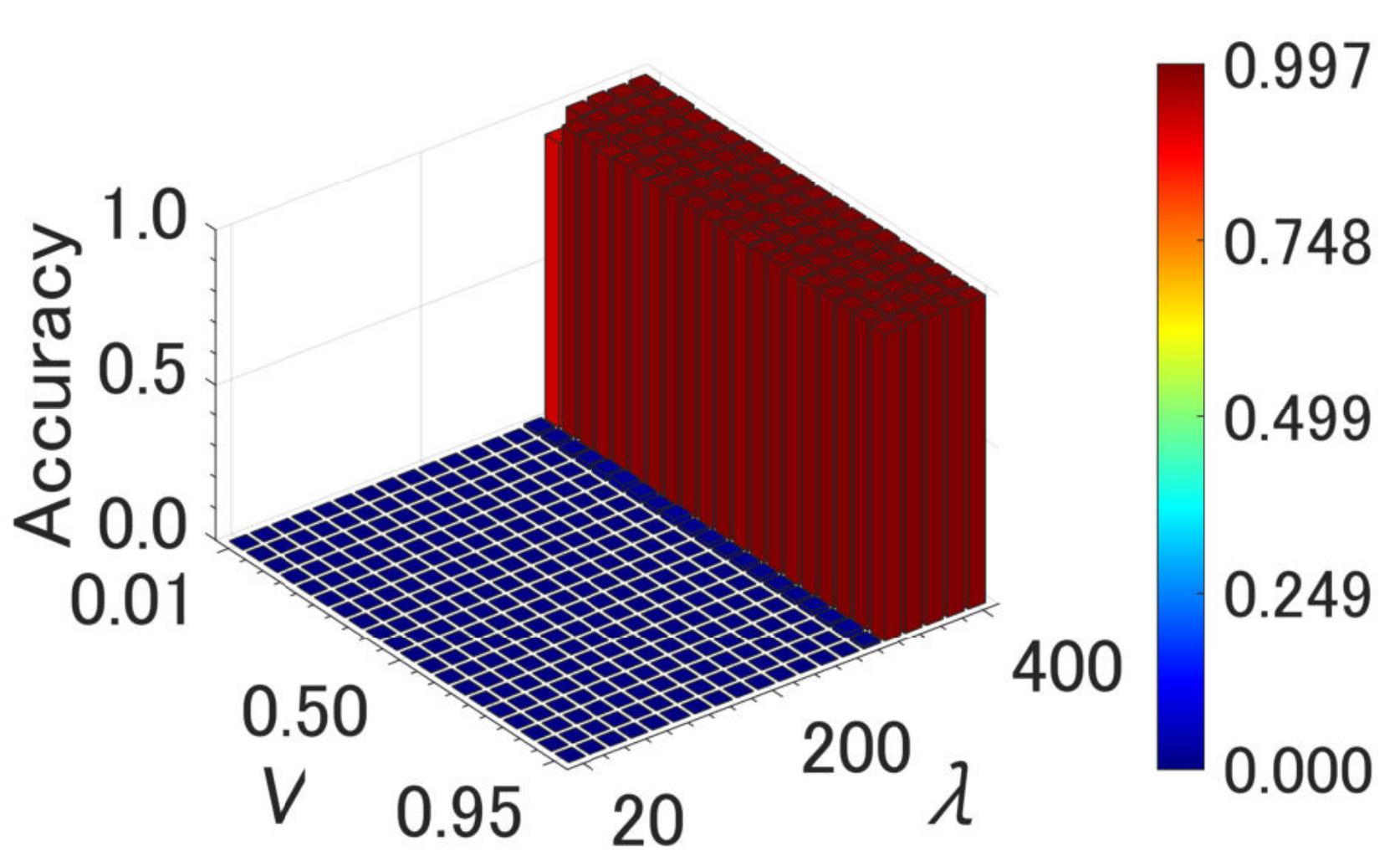}
		\label{fig:PS_Aggregation_FTCAC}
	}
	% \hspace{-1.4mm}
	% \hfil
	\subfloat[Compound]{
		\includegraphics[width=1.35in]{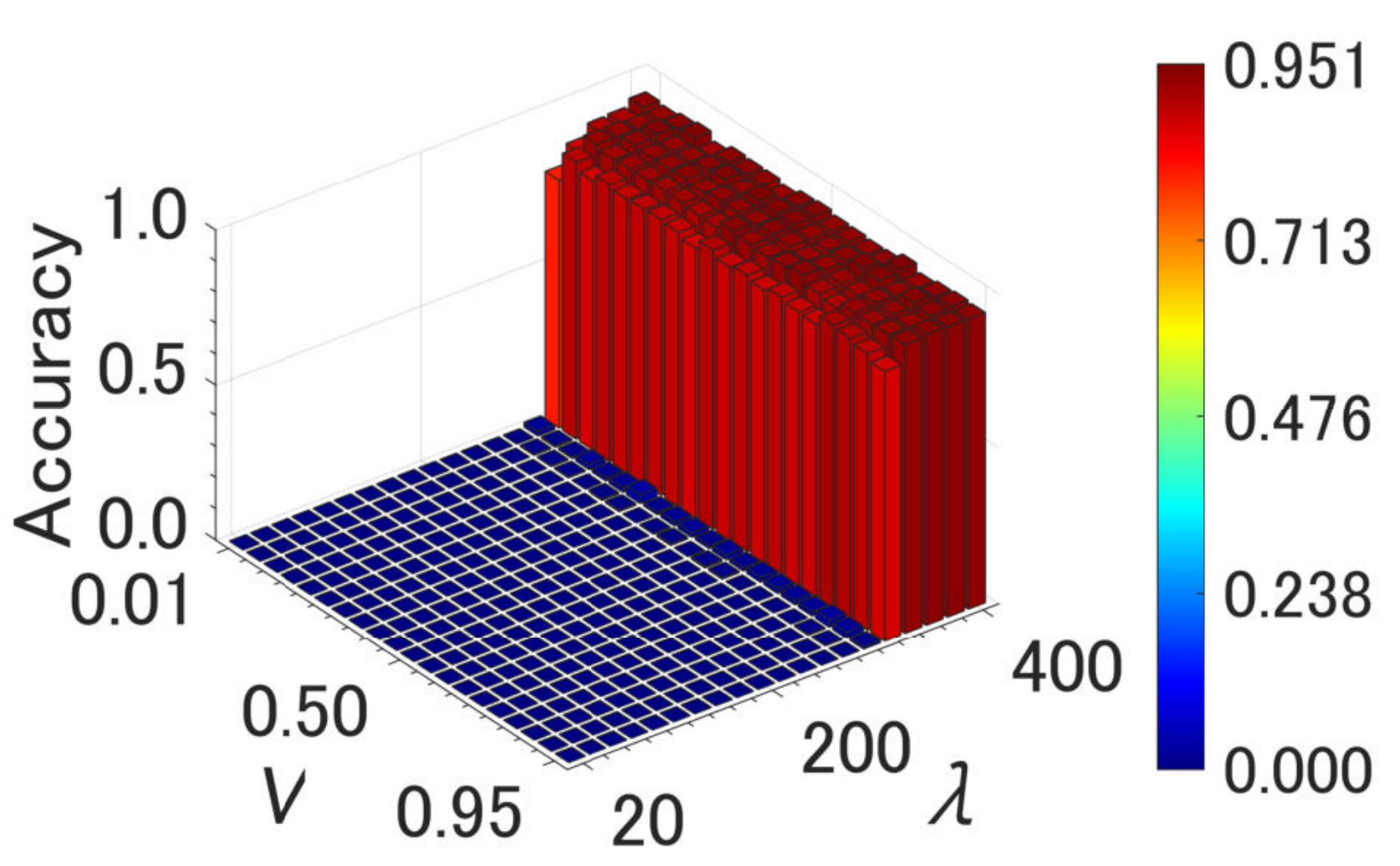}
		\label{fig:PS_Compound_FTCAC}
	}
	% \hspace{-1.4mm}
	% \hfil
	\subfloat[Hard Distribution]{
		\includegraphics[width=1.35in]{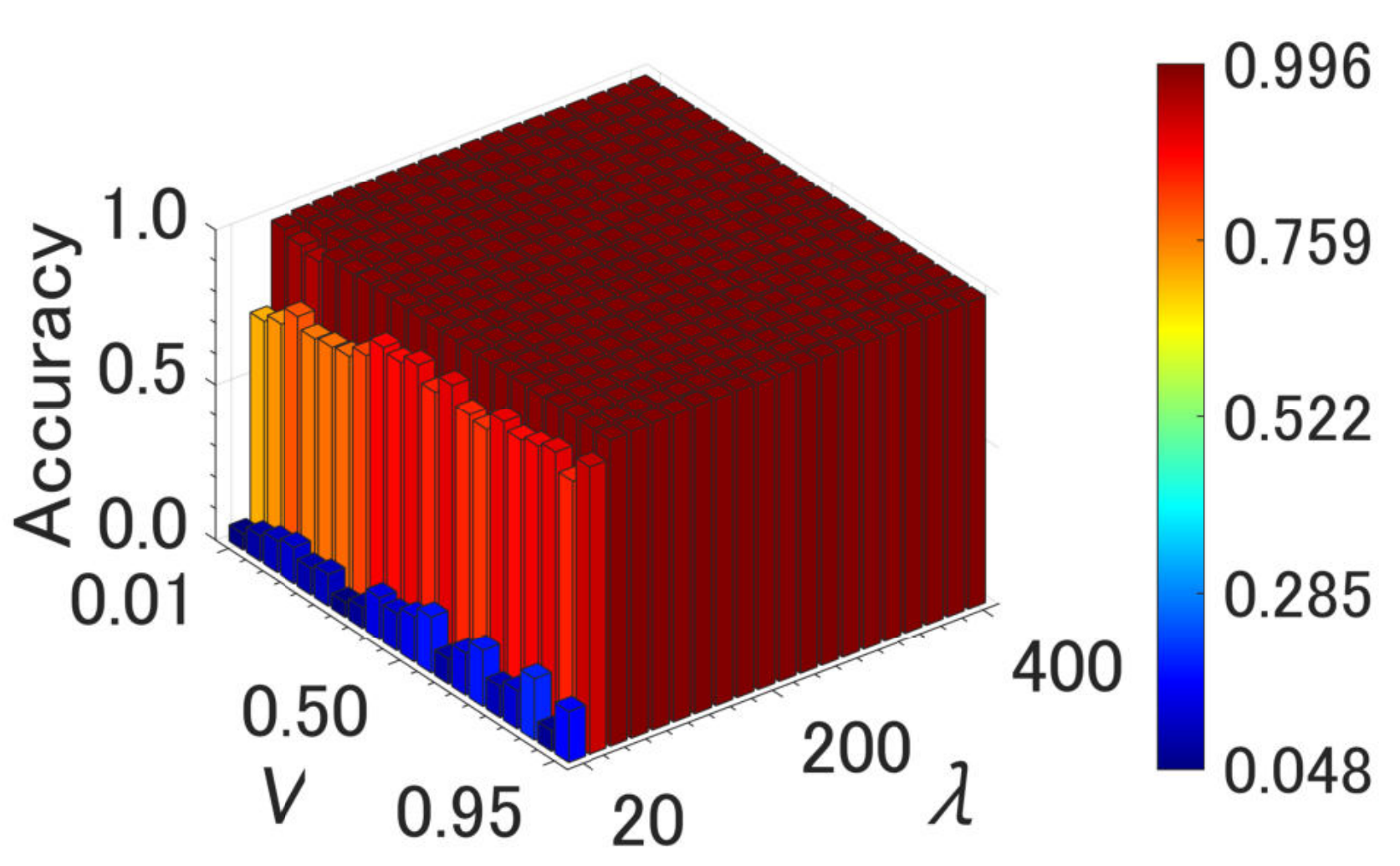}
		\label{fig:PS_HardDistribution_FTCAC}
	}
	% \hspace{-1.4mm}
	% \hfil
	\subfloat[Jain]{
		\includegraphics[width=1.35in]{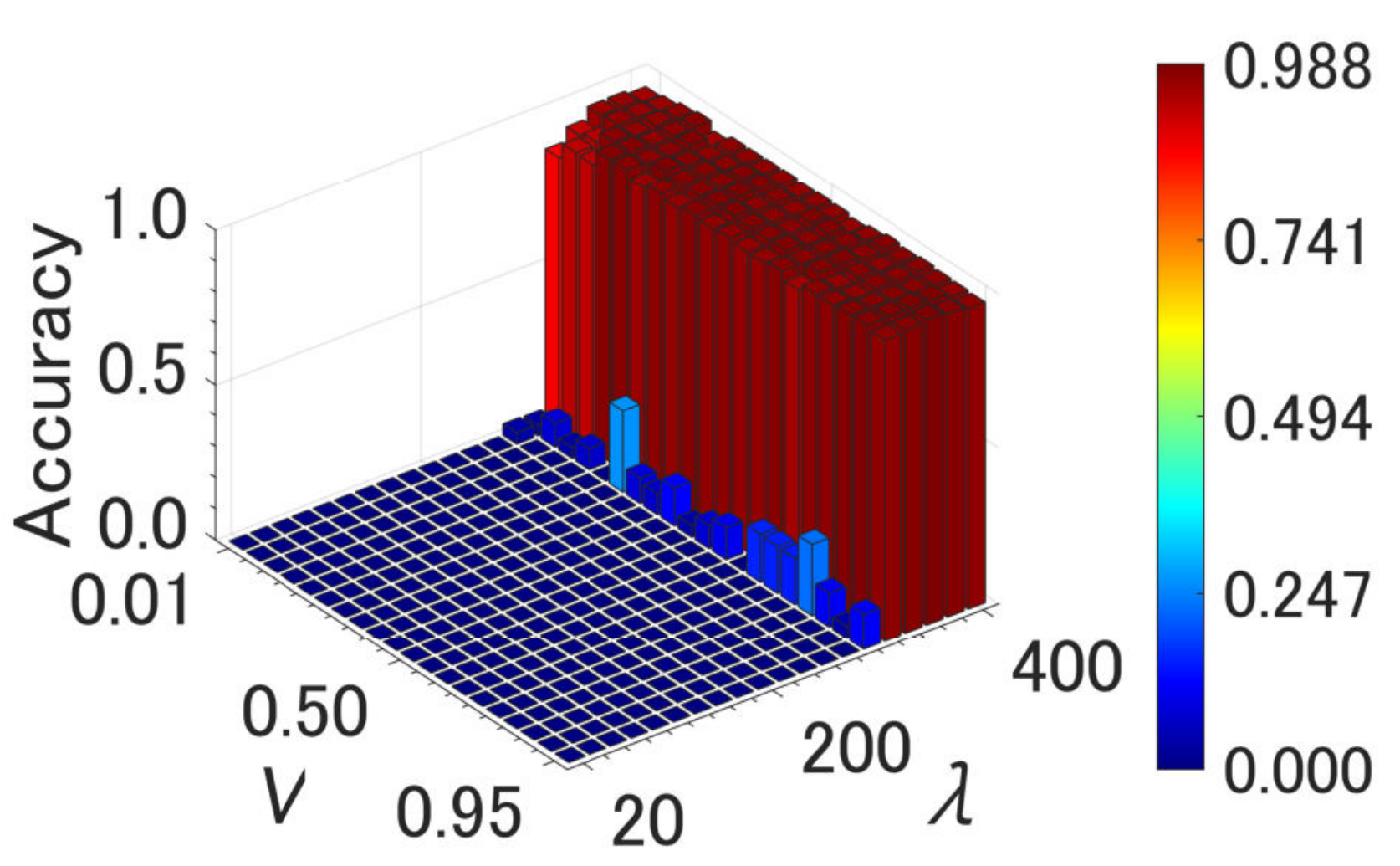}
		\label{fig:PS_Jain_FTCAC}
	}
	% \hspace{-1.4mm}
	% \hfil
	\subfloat[Pathbased]{
		\includegraphics[width=1.35in]{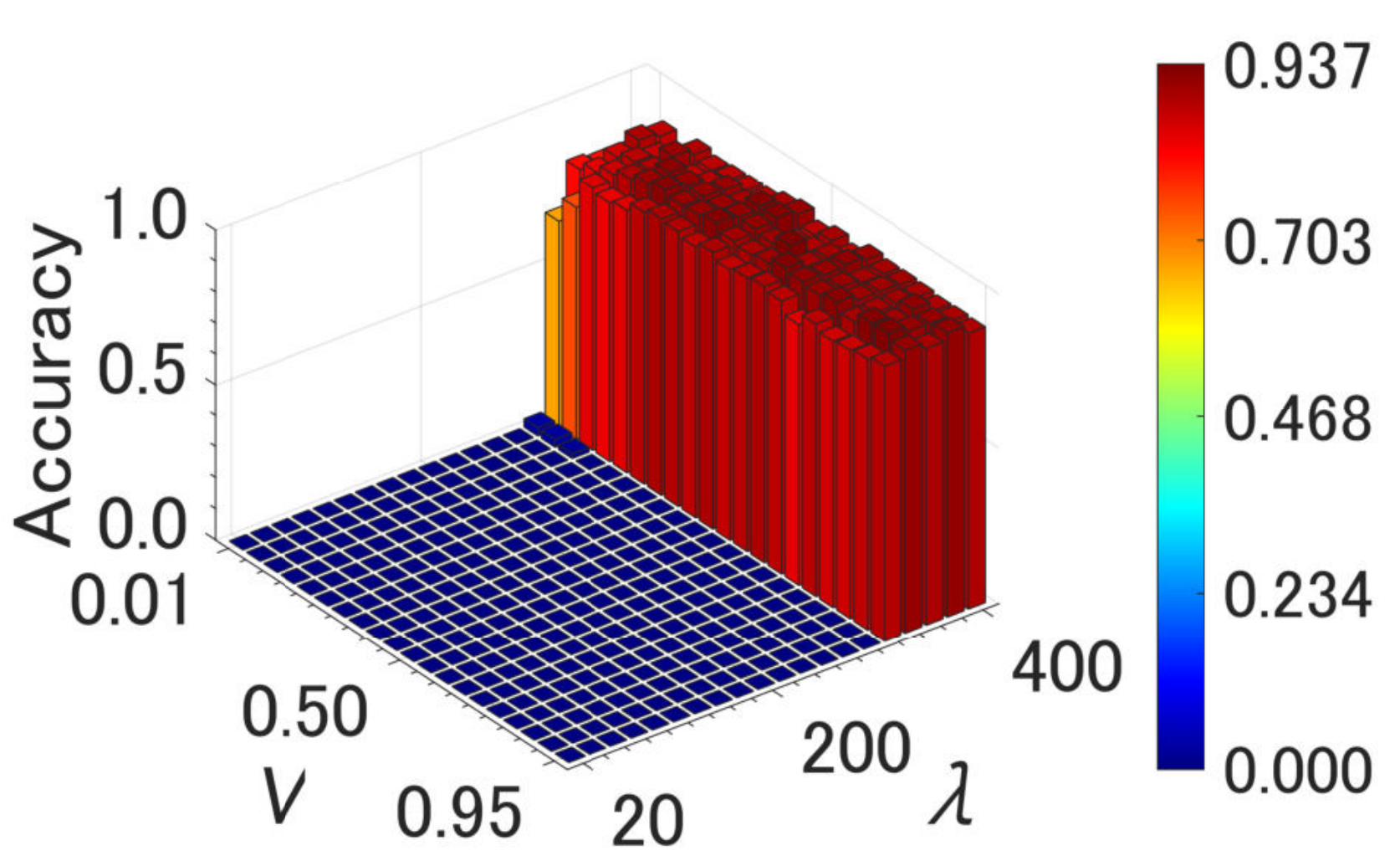}
		\label{fig:PS_Pathbased_FTCAC}
	}
	\\
	% \vspace{-2.5mm}
	%	\hfil
	% \hspace{-1.4mm}
	\subfloat[ALLAML]{
		\includegraphics[width=1.35in]{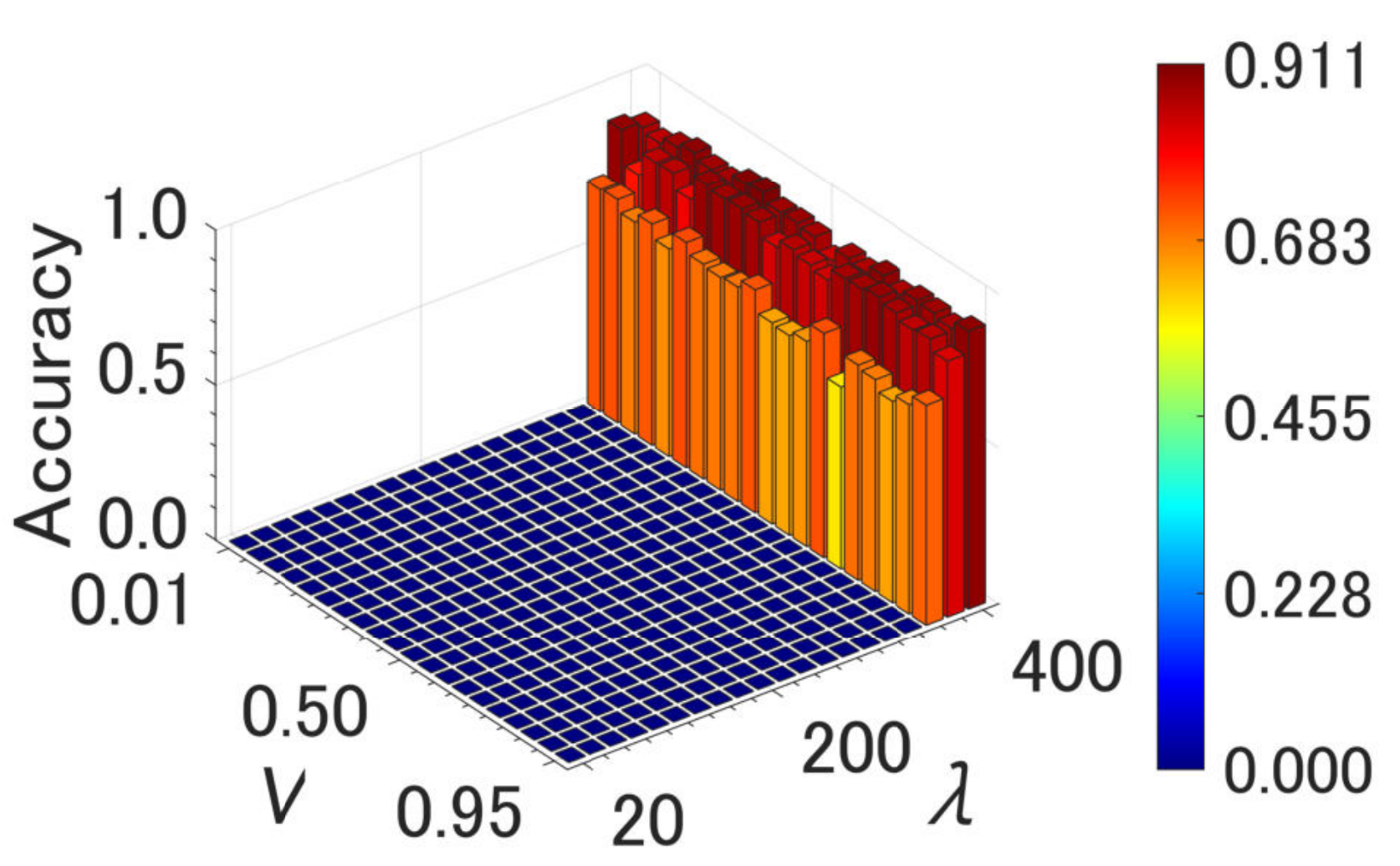}
		\label{fig:PS_ALLAML_FTCAC}
	}
	% \hspace{-1.4mm}
	% \hfil
	\subfloat[COIL20]{
		\includegraphics[width=1.35in]{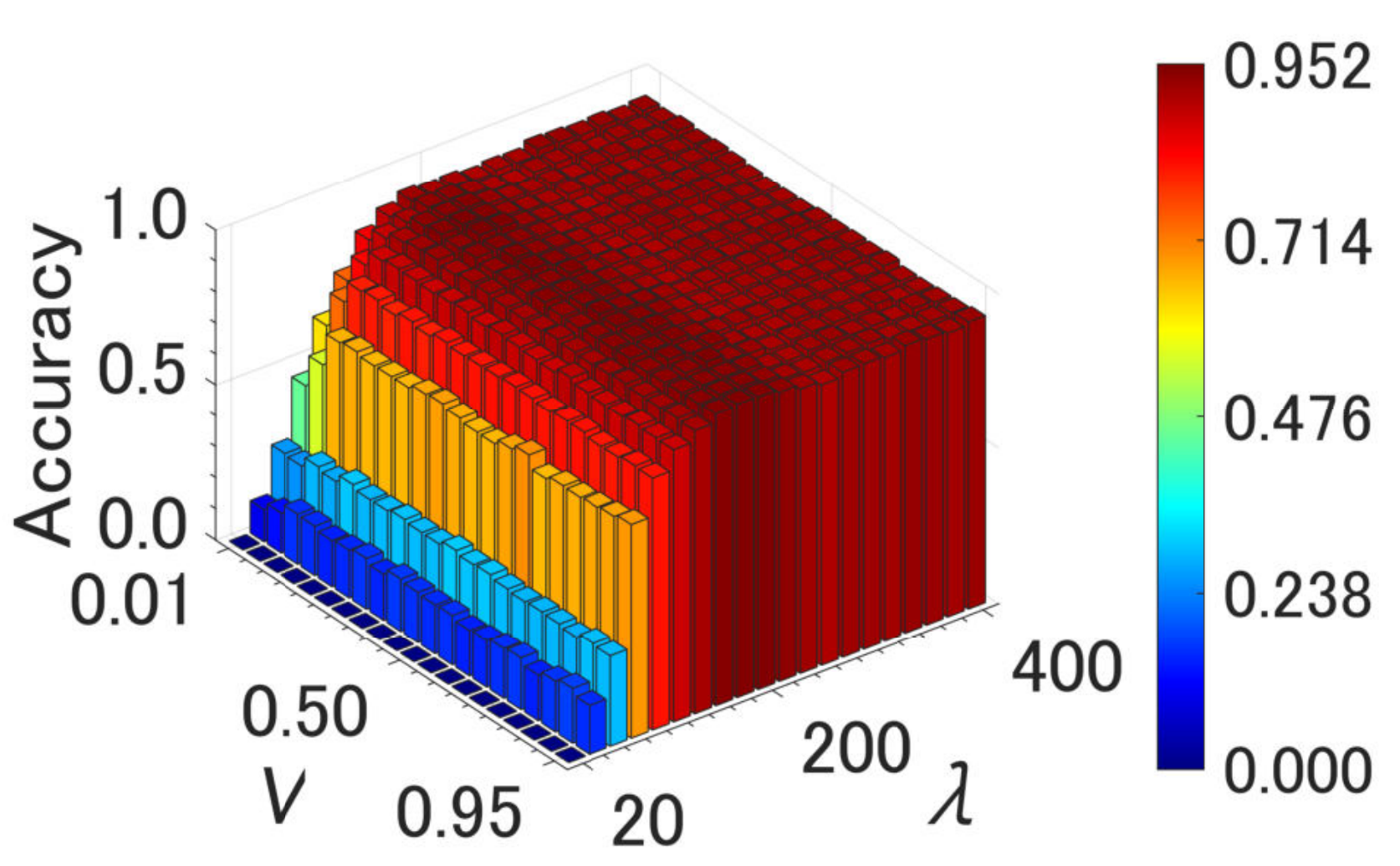}
		\label{fig:PS_COIL20_FTCAC}
	}
	% \hspace{-1.4mm}
	% \hfil
	\subfloat[Iris]{
		\includegraphics[width=1.35in]{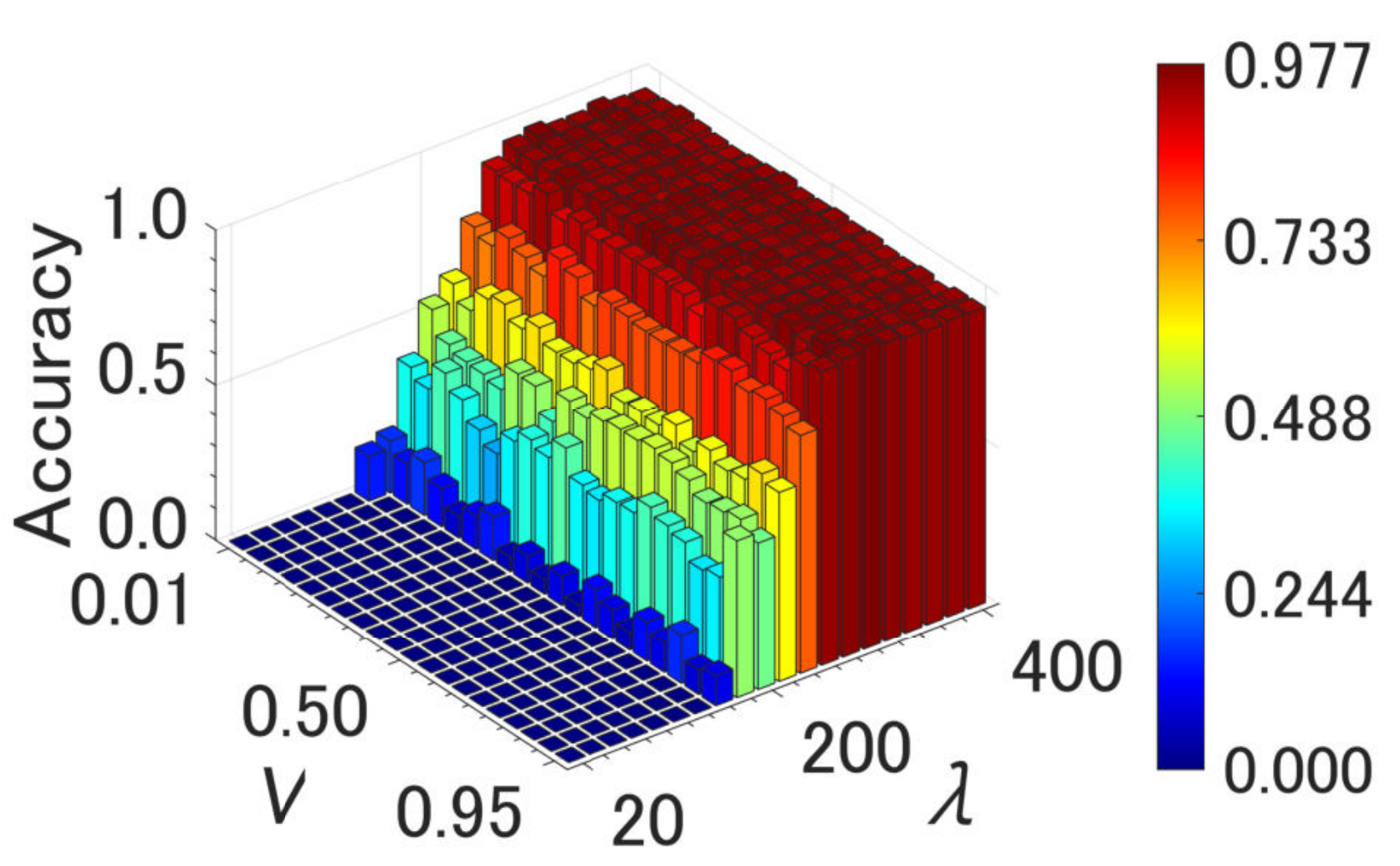}
		\label{fig:PS_Iris_FTCAC}
	}
	% \hspace{-1.4mm}
	\hfil
	\subfloat[Isolet]{
		\includegraphics[width=1.35in]{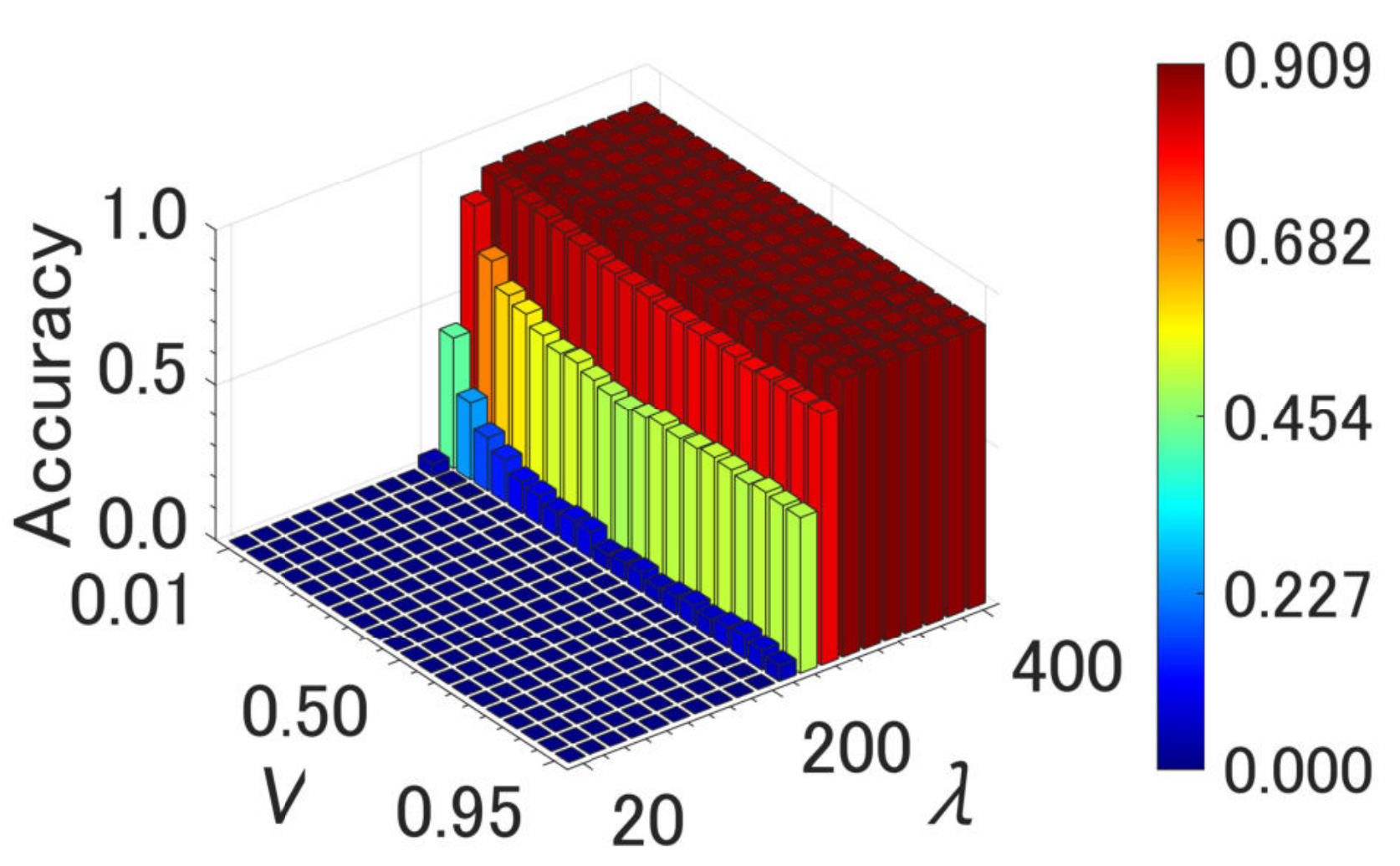}
		\label{fig:PS_Isolet_FTCAC}
	}
	% \hspace{-1.4mm}
	% \hfil
	\subfloat[OptDigits]{
		\includegraphics[width=1.35in]{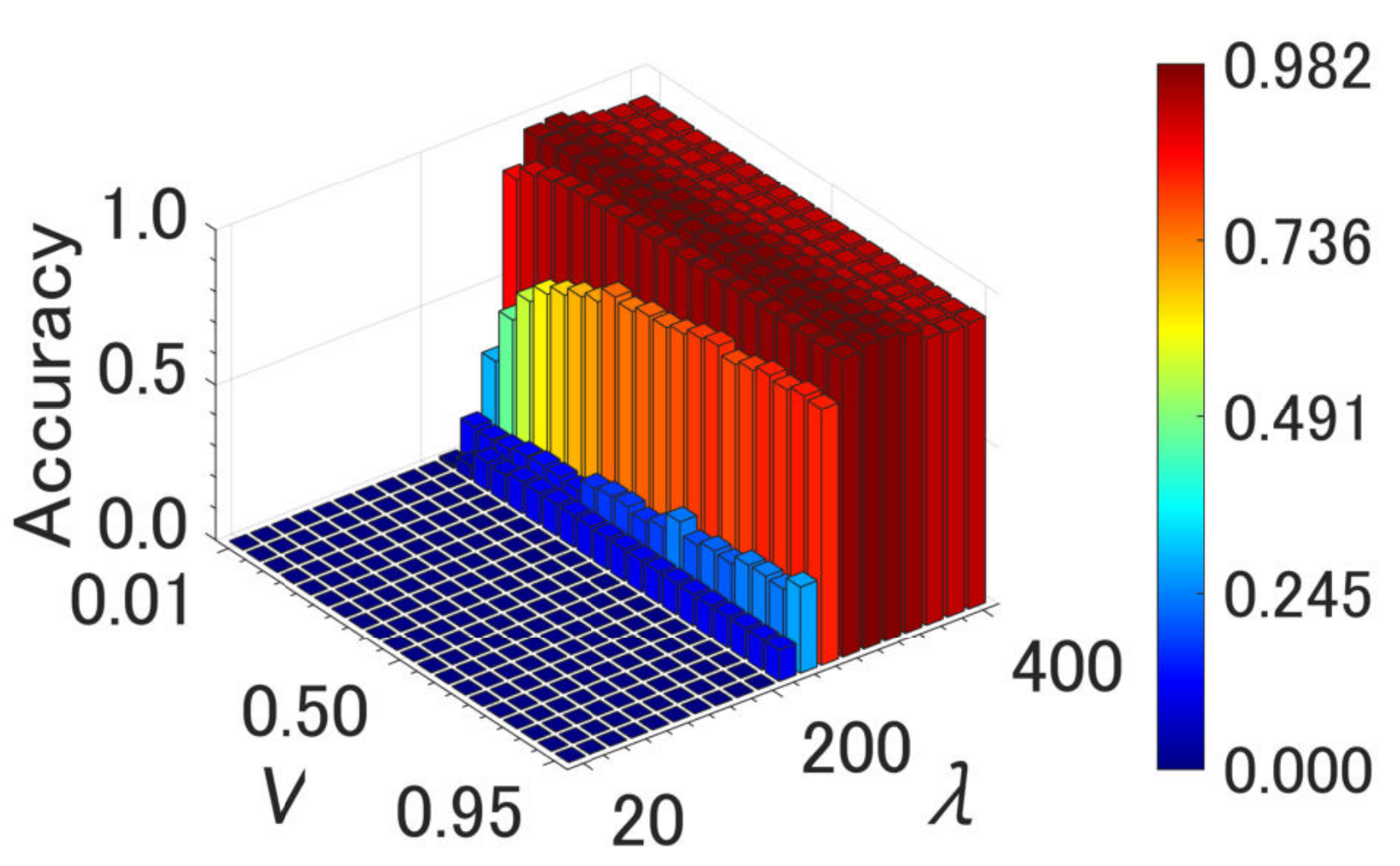}
		\label{fig:PS_OptDigits_FTCAC}
	}
	\\
	% \hspace{-2.5mm}
	% \vspace{-2.5mm}
	%	\hfil
	\subfloat[Seeds]{
		\includegraphics[width=1.35in]{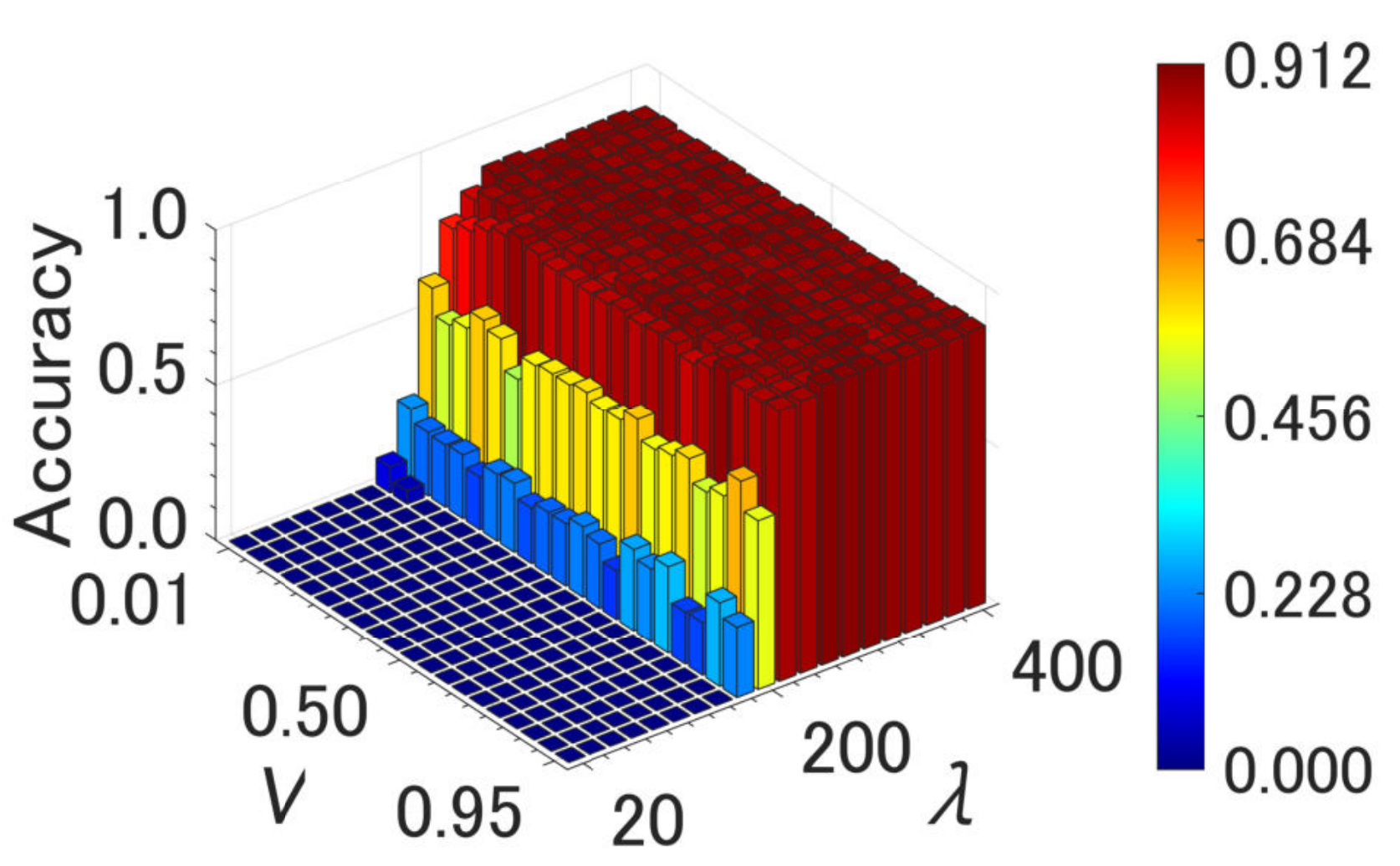}
		\label{fig:PS_Seeds_FTCAC}
	}
	% \hspace{1.7mm}
	%	\hfil
	\subfloat[Semeion]{
		\includegraphics[width=1.35in]{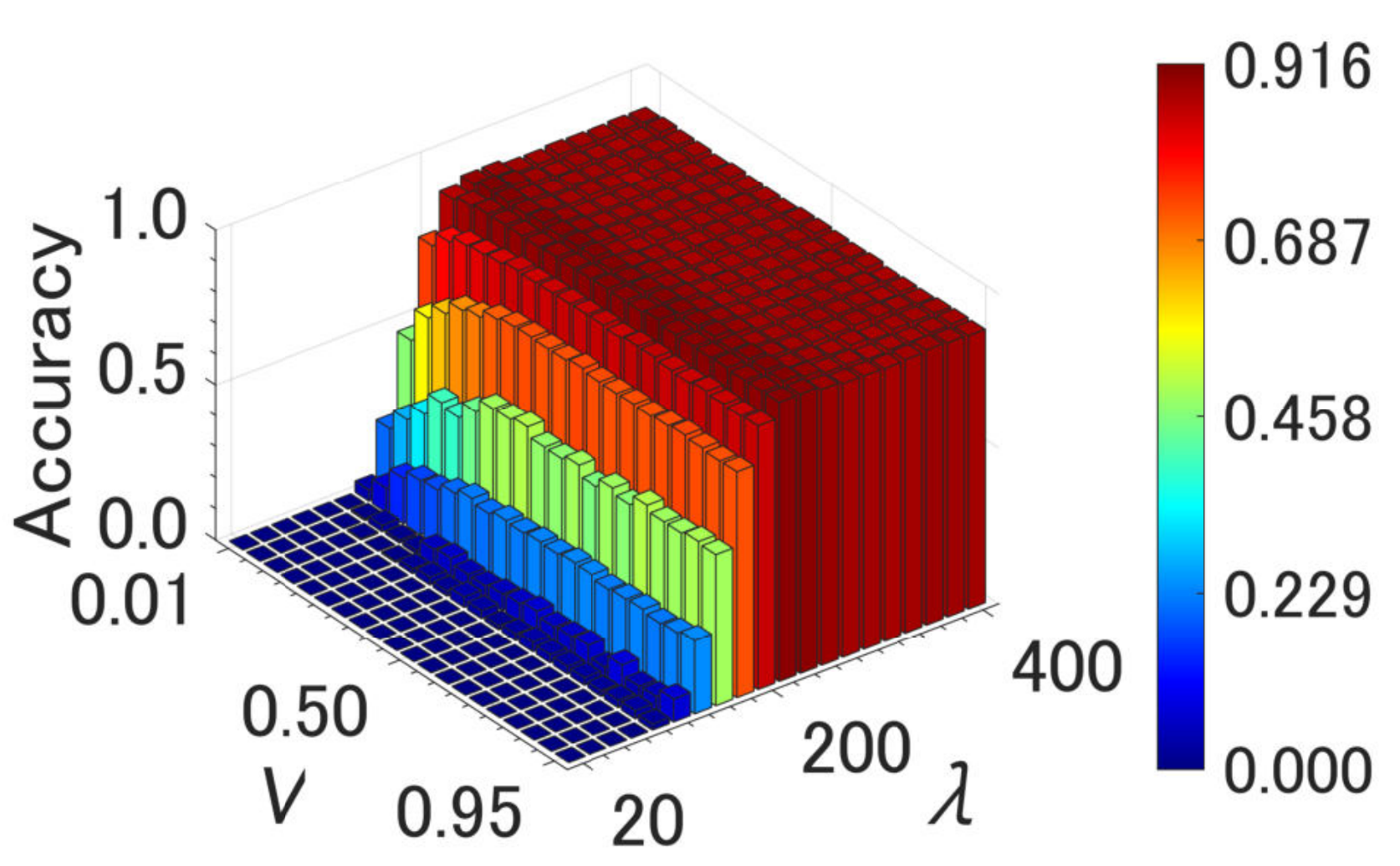}
		\label{fig:PS_Semeion_FTCAC}
	}
	% \hspace{1.7mm}
	%	\hfil
	\subfloat[Sonar]{
		\includegraphics[width=1.35in]{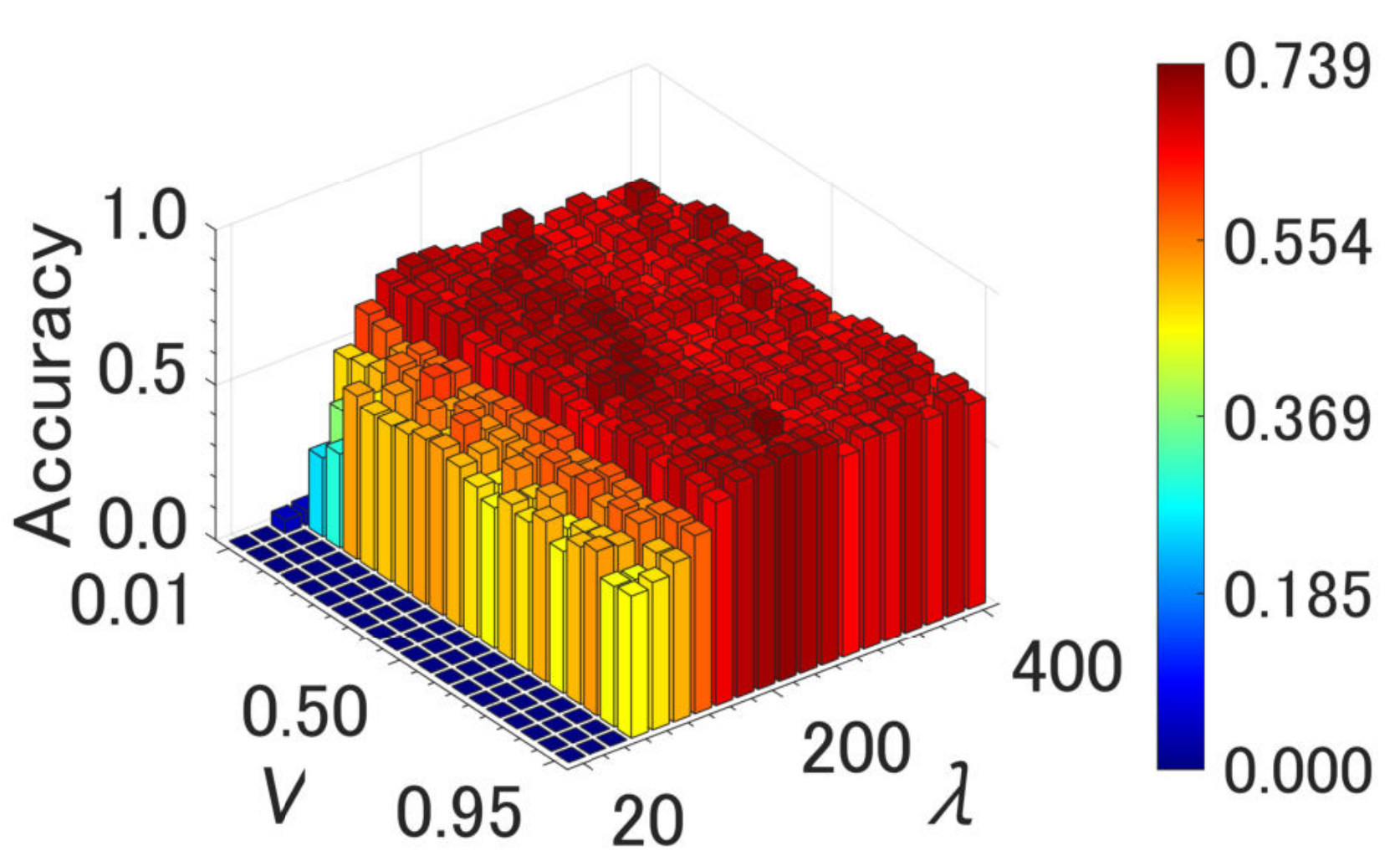}
		\label{fig:PS_Sonar_FTCAC}
	}
	% \vspace{-1mm}
	\end{adjustwidth}
	\caption{Effects of the parameter specifications of FTCAC on Accuracy (TOX171 is N/A). }
	\label{fig:paramSensitivity_FTCAC}
\end{figure}

\begin{figure}[htbp]
	\begin{adjustwidth}{-\extralength}{0cm}
	\centering
	\subfloat[Aggregation]{
		\includegraphics[width=1.35in]{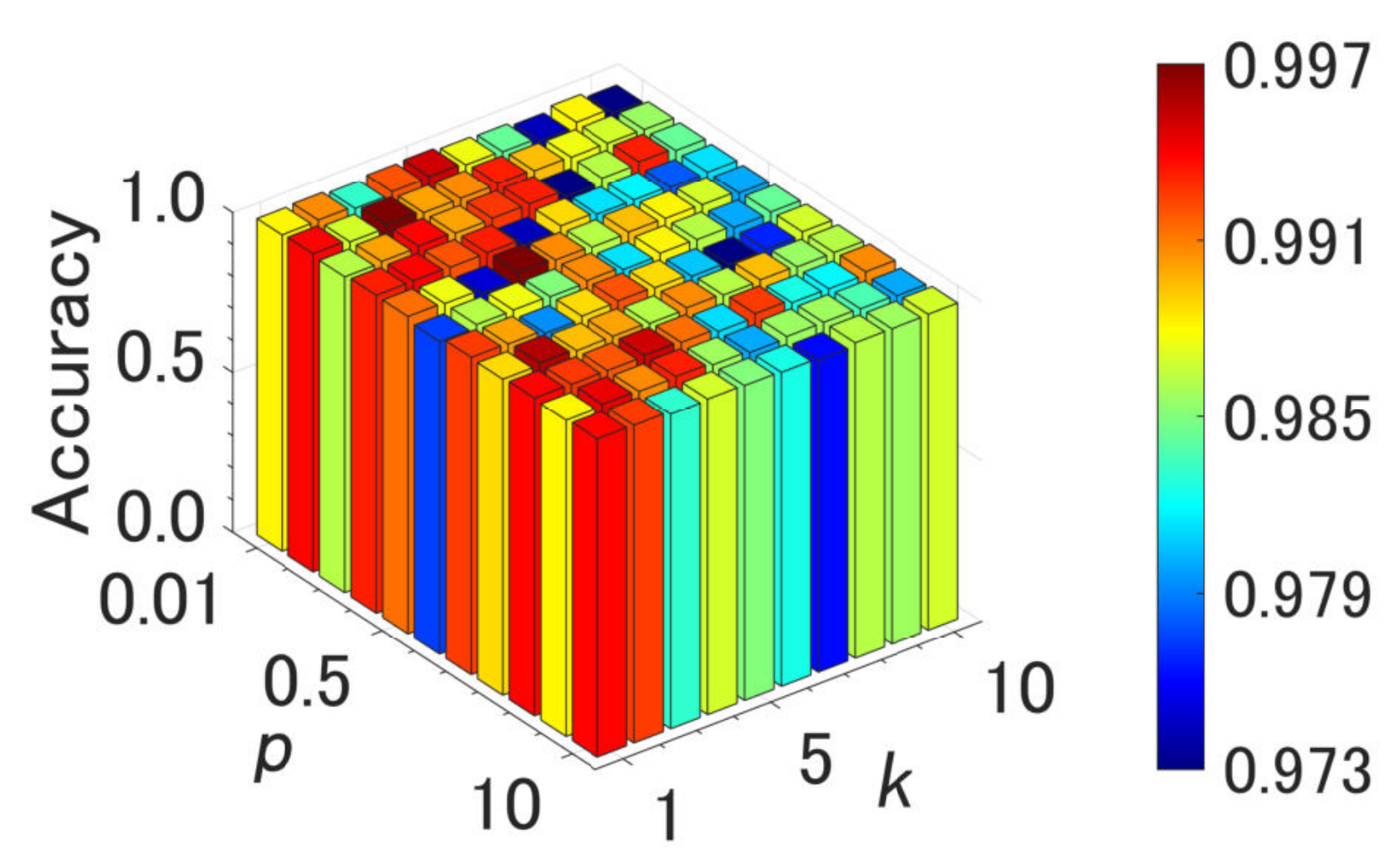}
		\label{fig:PS_Aggregation_GSp}
	}
	% \hspace{-1.4mm}
	% \hfil
	\subfloat[Compound]{
		\includegraphics[width=1.35in]{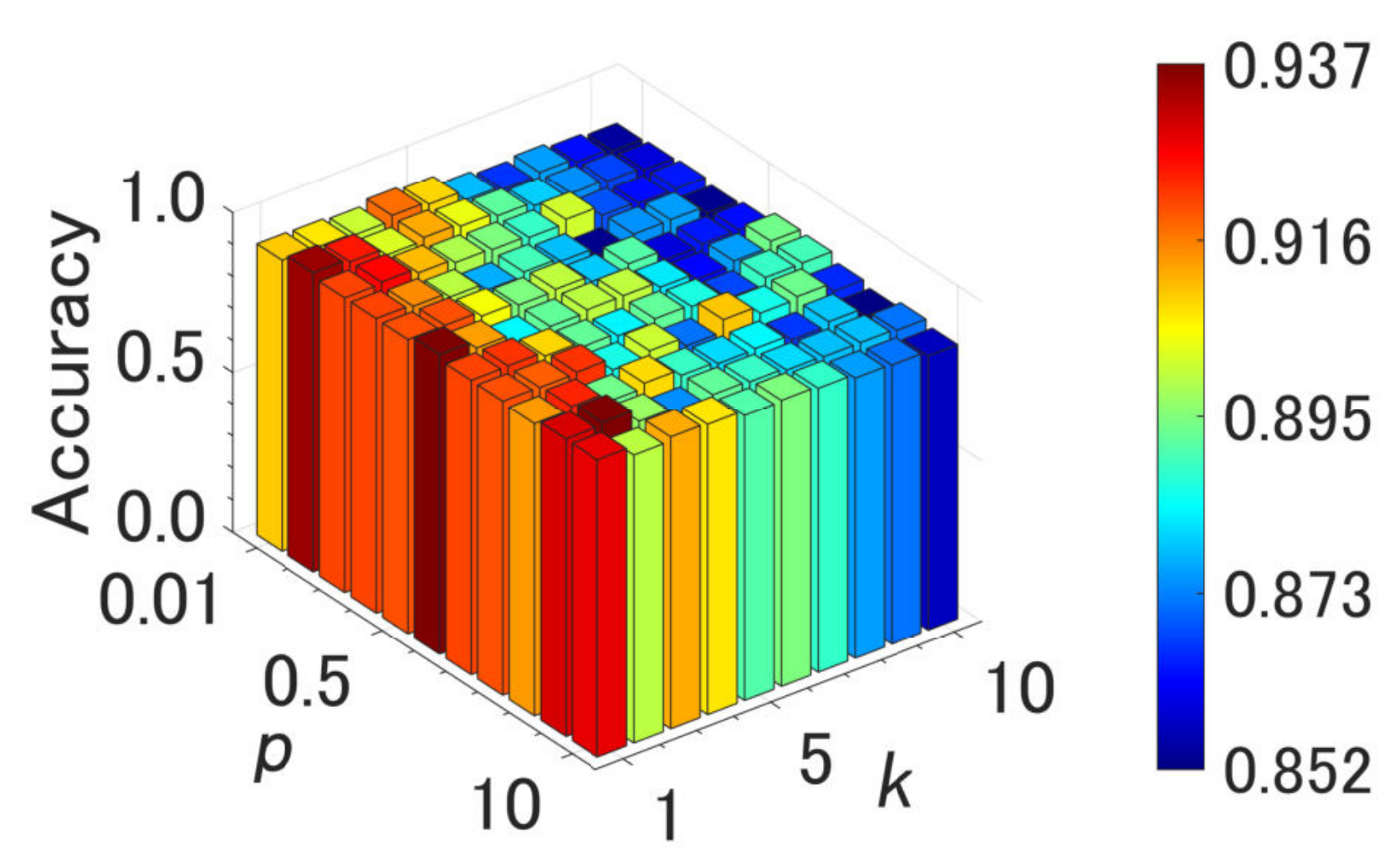}
		\label{fig:PS_Compound_GSp}
	}
	% \hspace{-1.4mm}
	% \hfil
	\subfloat[Hard Distribution]{
		\includegraphics[width=1.35in]{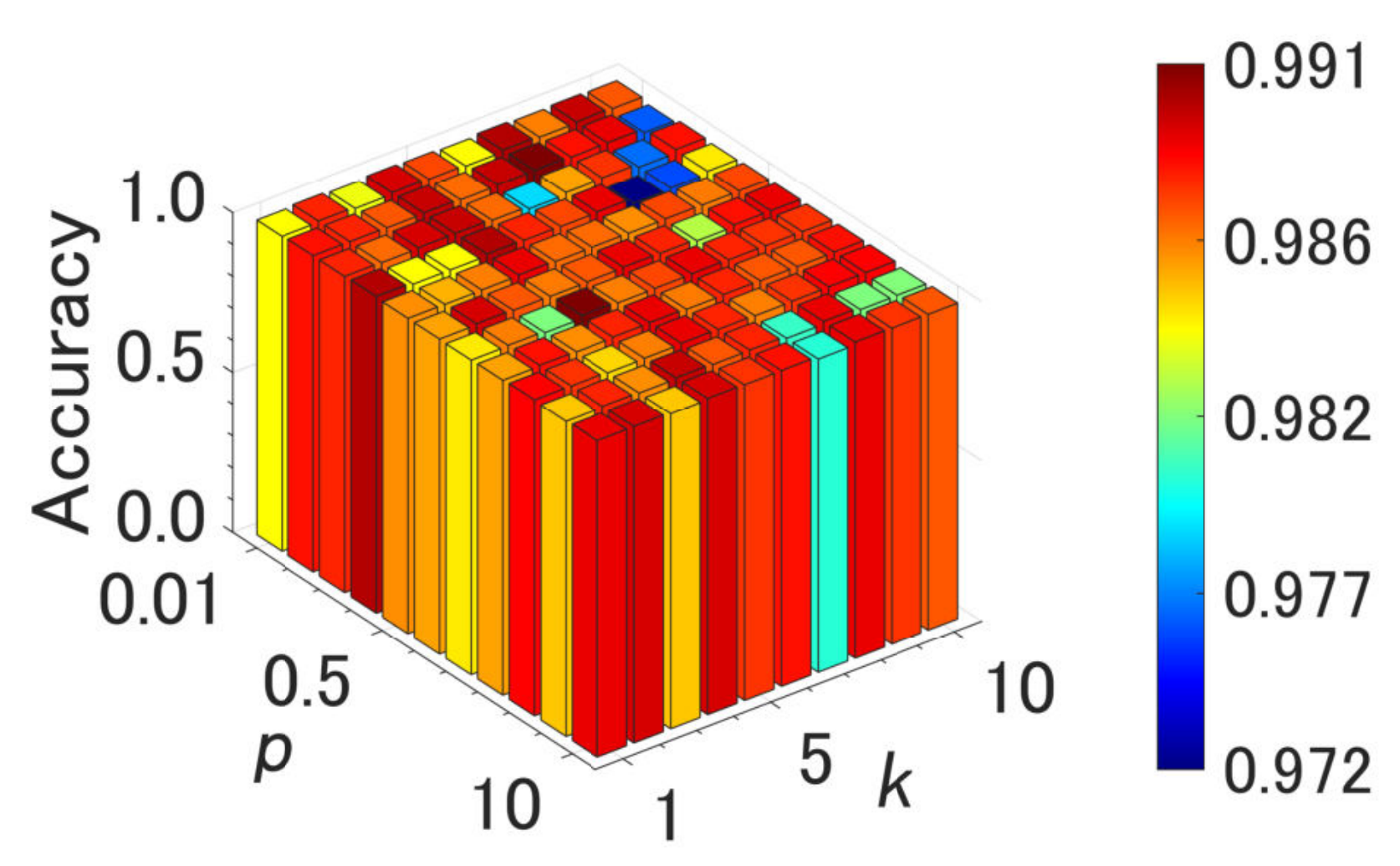}
		\label{fig:PS_HardDistribution_GSp}
	}
	% \hspace{-1.4mm}
	% \hfil
	\subfloat[Jain]{
		\includegraphics[width=1.35in]{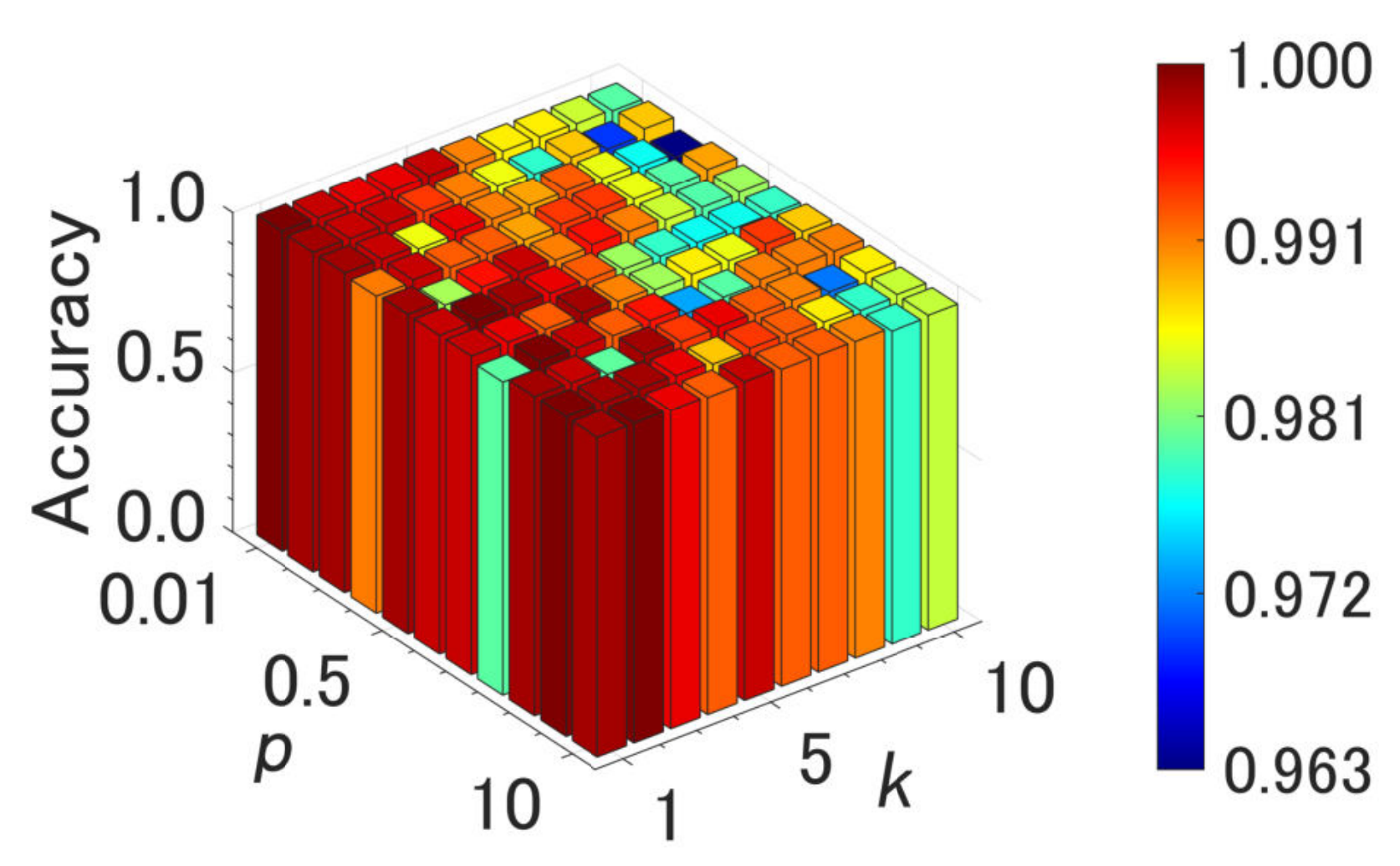}
		\label{fig:PS_Jain_GSp}
	}
	% \hspace{-1.4mm}
	% \hfil
	\subfloat[Pathbased]{
		\includegraphics[width=1.35in]{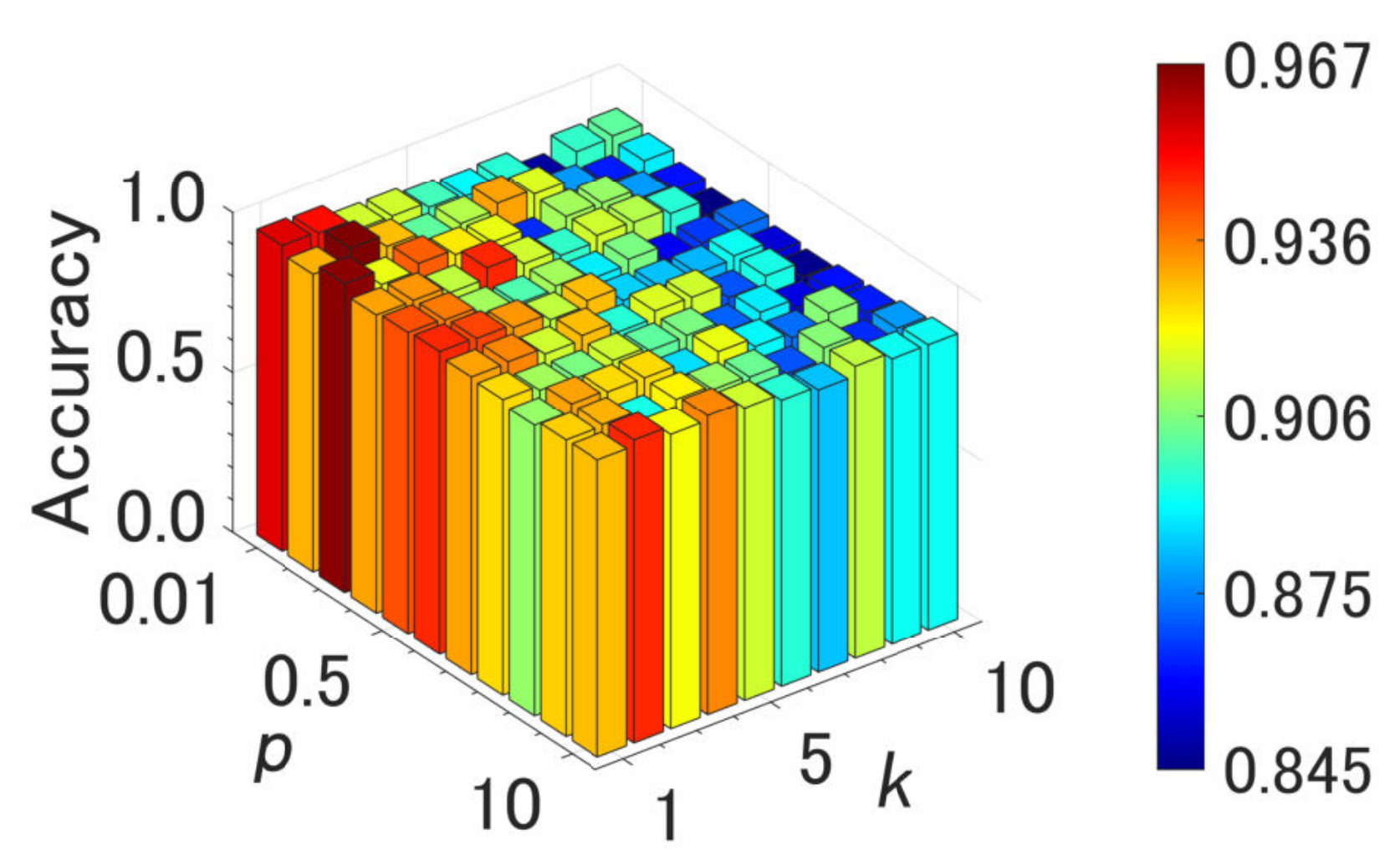}
		\label{fig:PS_Pathbased_GSp}
	}
	\\
	% \vspace{-2.5mm}
	%	\hfil
	% \hspace{-1.4mm}
	\subfloat[ALLAML]{
		\includegraphics[width=1.35in]{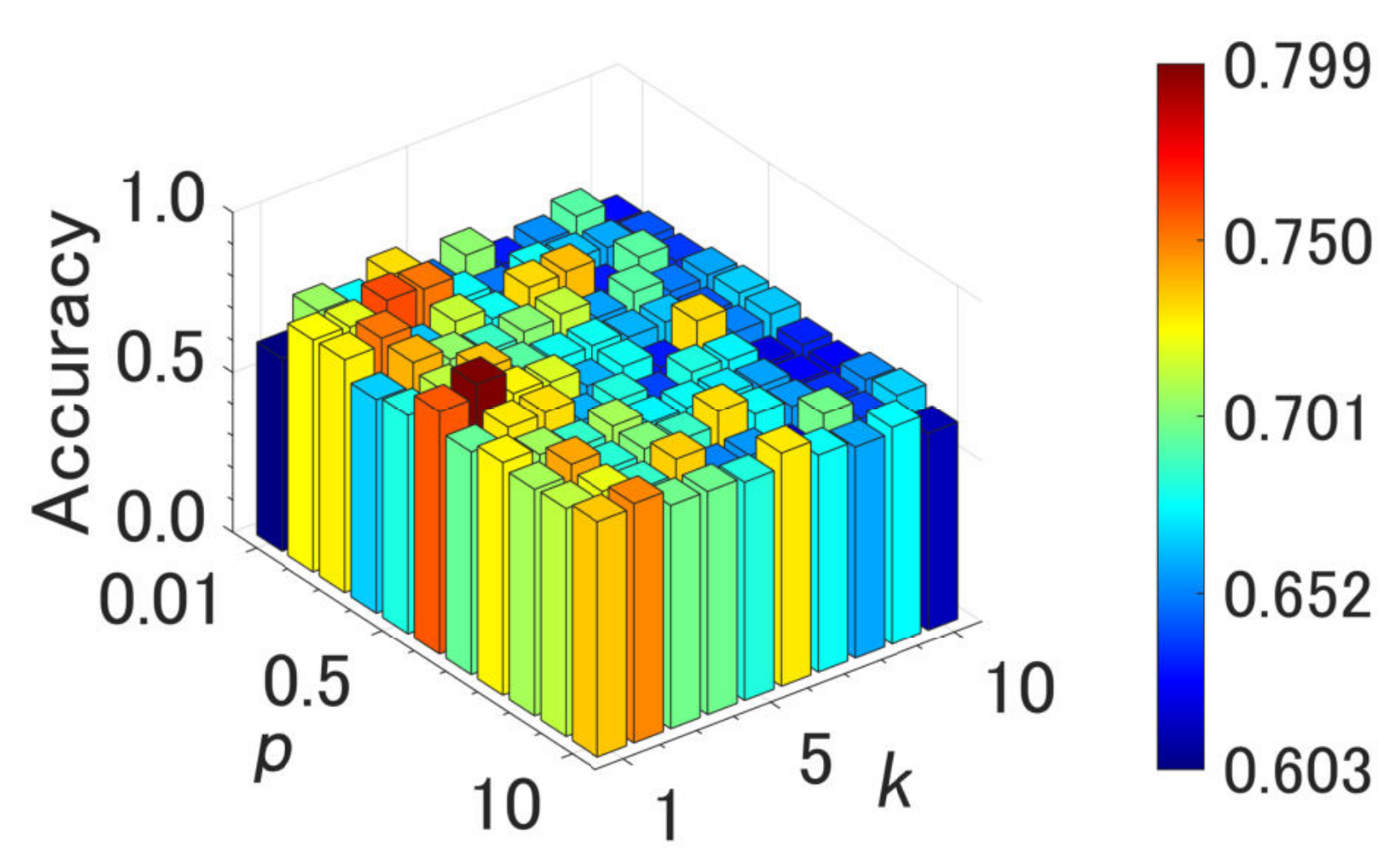}
		\label{fig:PS_ALLAML_GSp}
	}
	% \hspace{-1.4mm}
	% \hfil
	\subfloat[COIL20]{
		\includegraphics[width=1.35in]{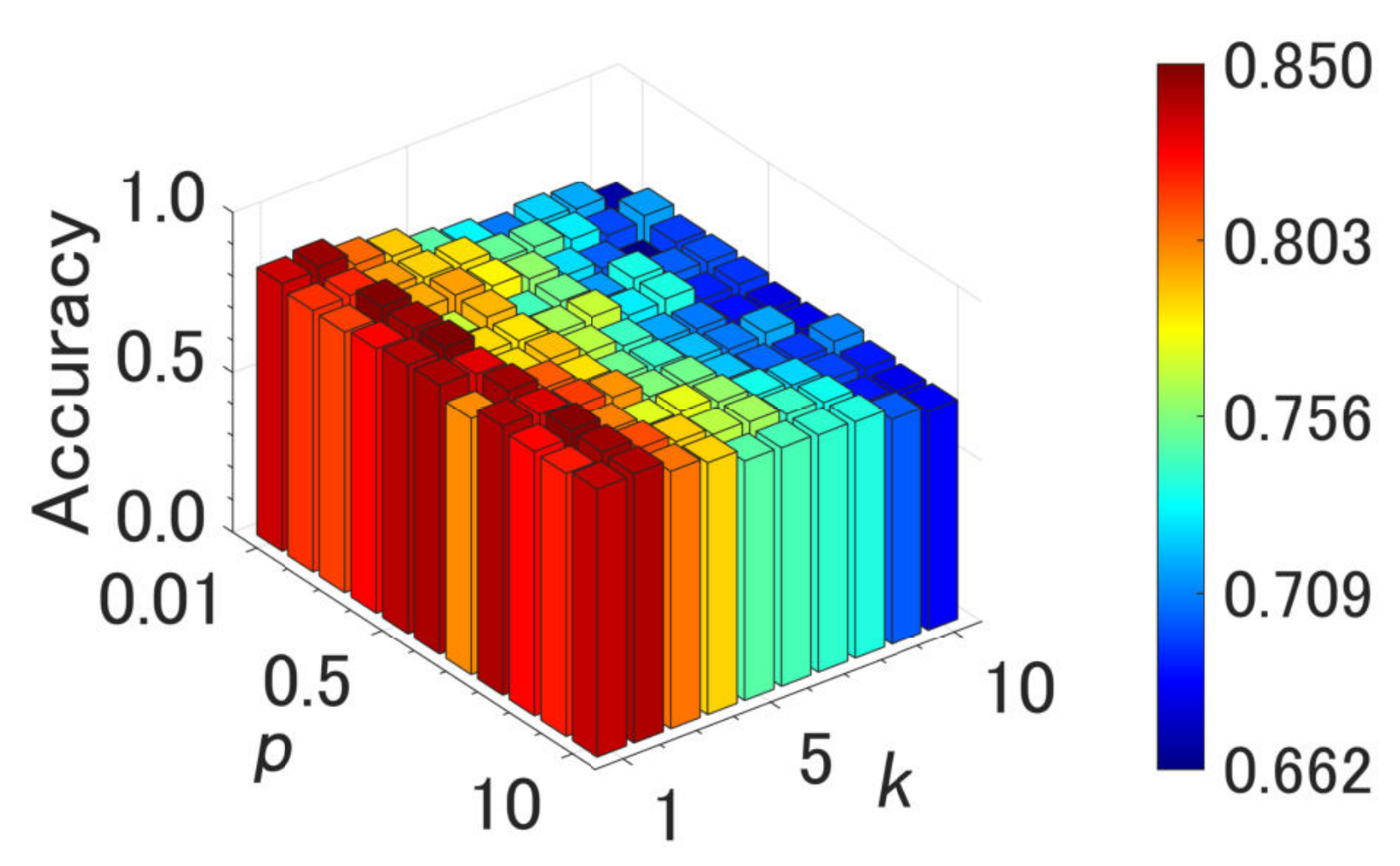}
		\label{fig:PS_COIL20_GSp}
	}
	% \hspace{-1.4mm}
	% \hfil
	\subfloat[Iris]{
		\includegraphics[width=1.35in]{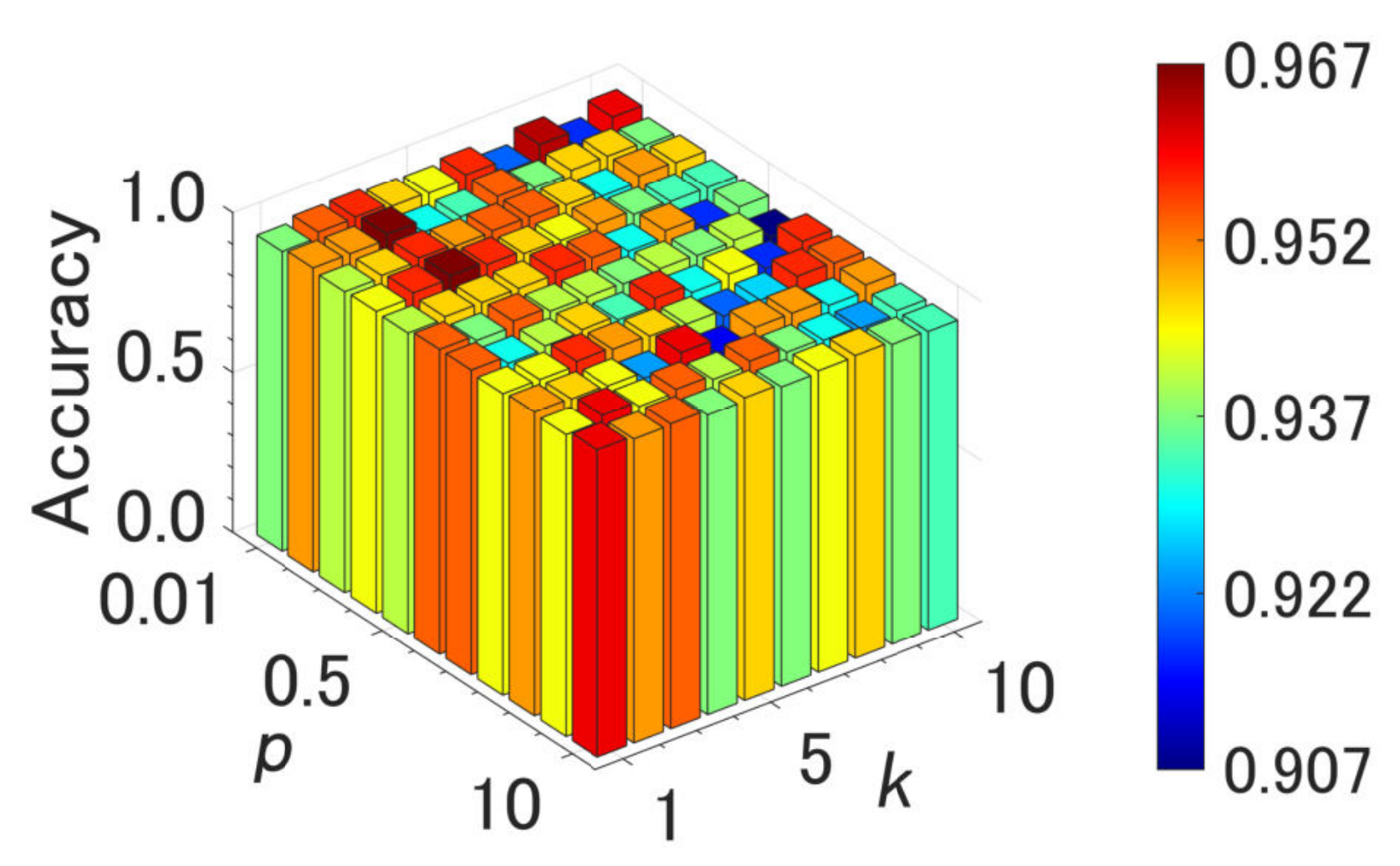}
		\label{fig:PS_Iris_GSp}
	}
	% \hspace{-1.4mm}
	% \hfil
	\subfloat[Isolet]{
		\includegraphics[width=1.35in]{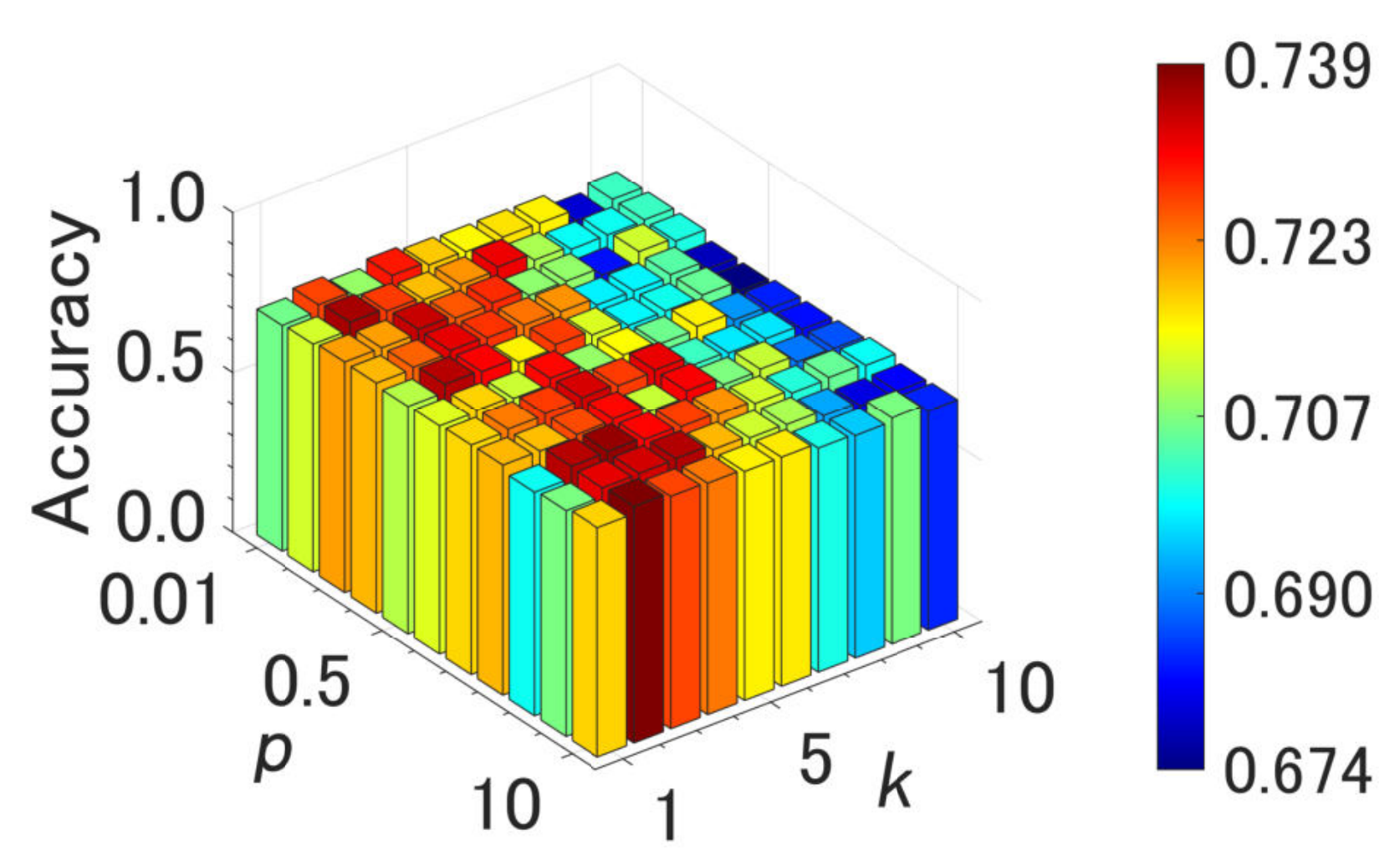}
		\label{fig:PS_Isolet_GSp}
	}
	% \hspace{-1.4mm}
	% \hfil
	\subfloat[OptDigits]{
		\includegraphics[width=1.35in]{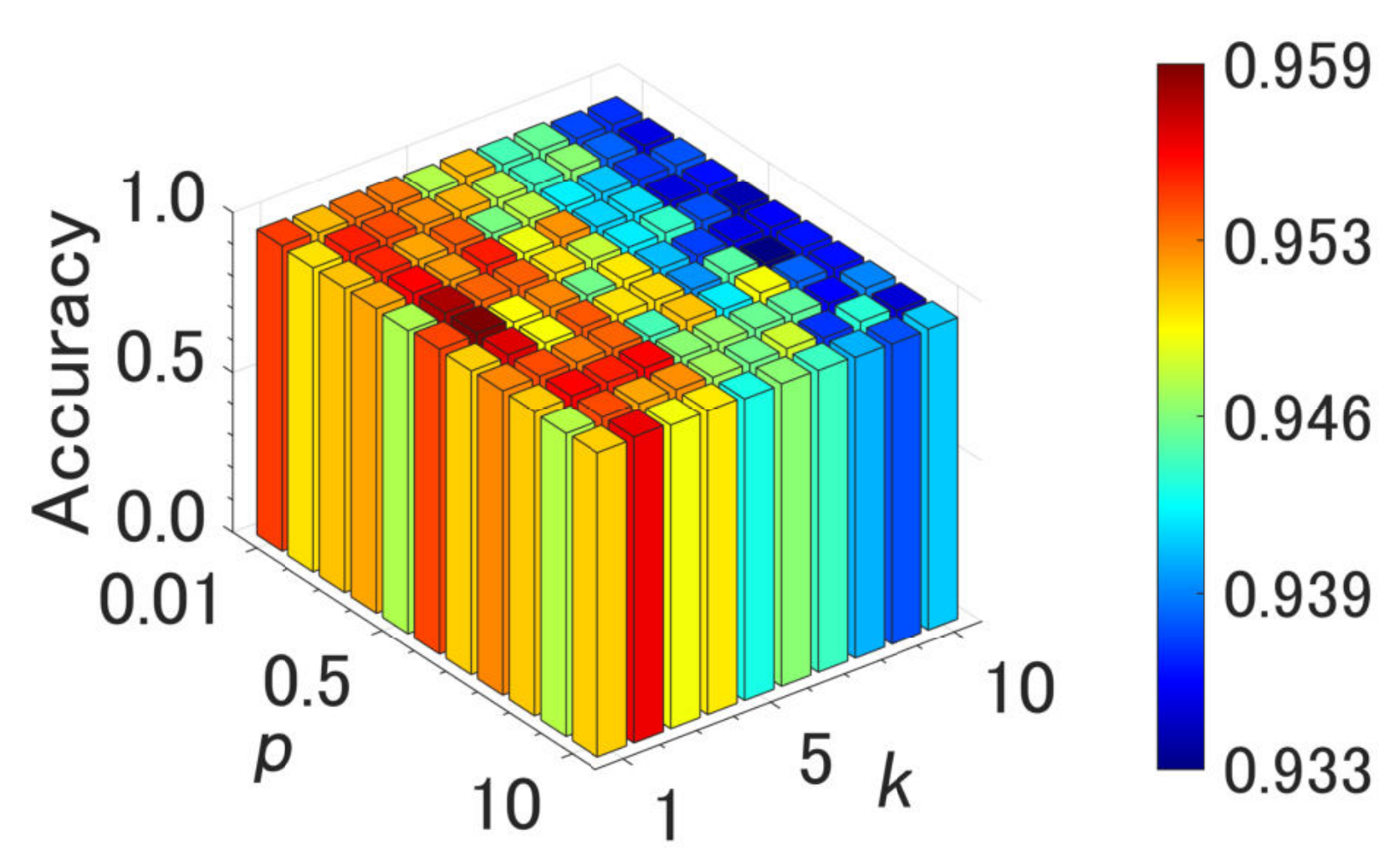}
		\label{fig:PS_OptDigits_GSp}
	}
	\\
	% \vspace{-2.5mm}
	%	\hfil
	\subfloat[Seeds]{
		\includegraphics[width=1.35in]{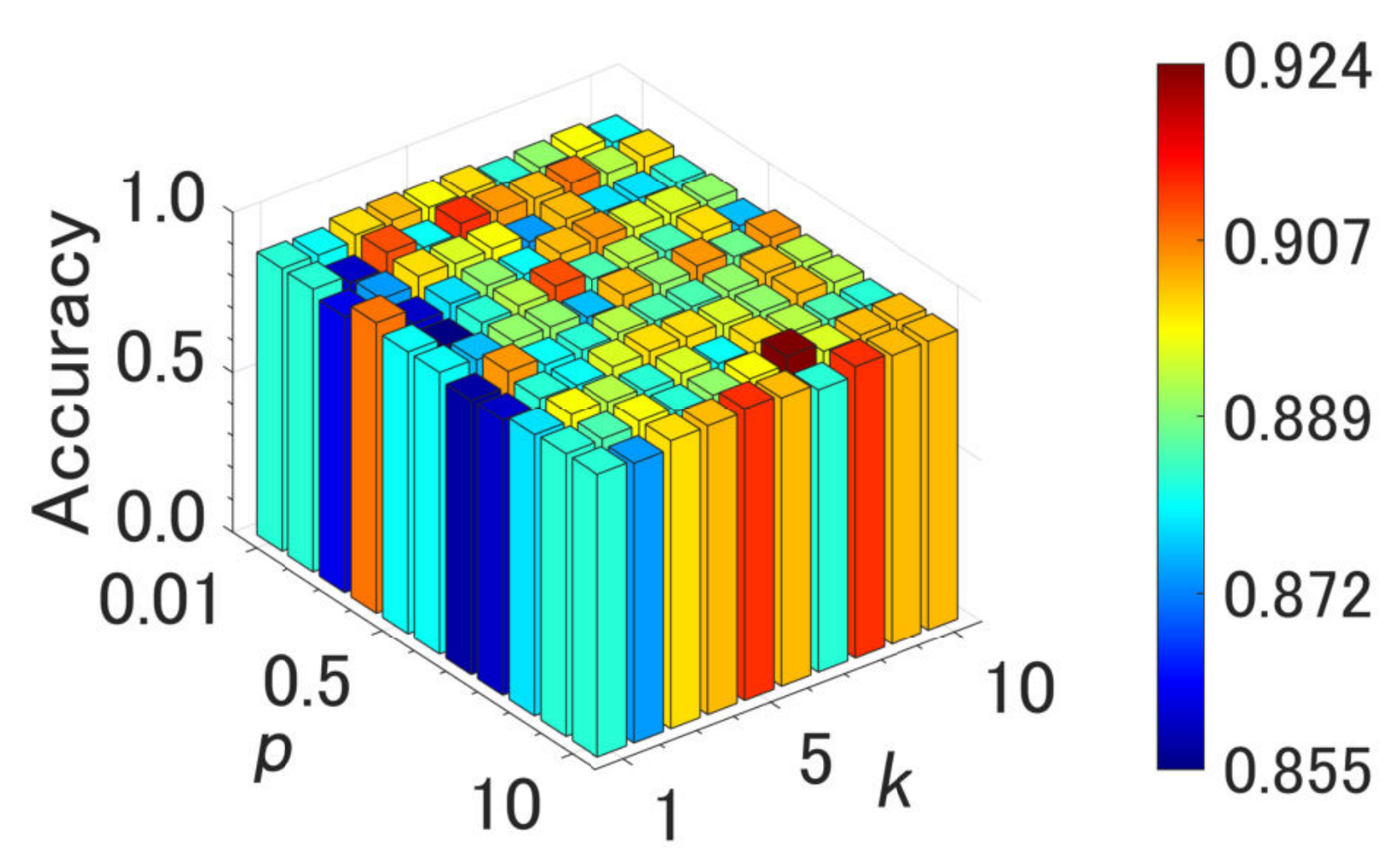}
		\label{fig:PS_Seeds_GSp}
	}
	% \hspace{2mm}
	%	\hfil
	\subfloat[Semeion]{
		\includegraphics[width=1.35in]{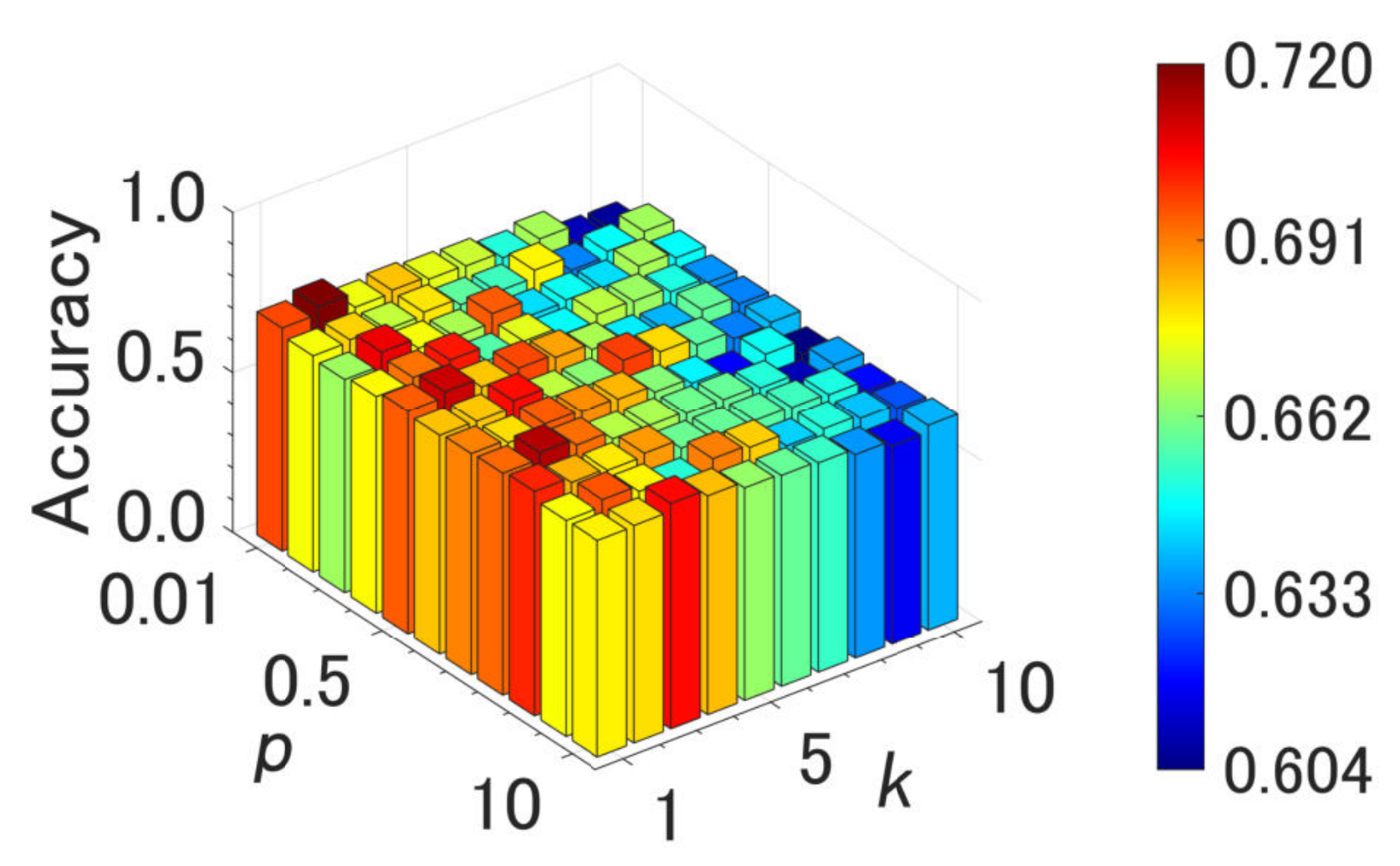}
		\label{fig:PS_Semeion_GSp}
	}
	% \hspace{2mm}
	%	\hfil
	\subfloat[Sonar]{
		\includegraphics[width=1.35in]{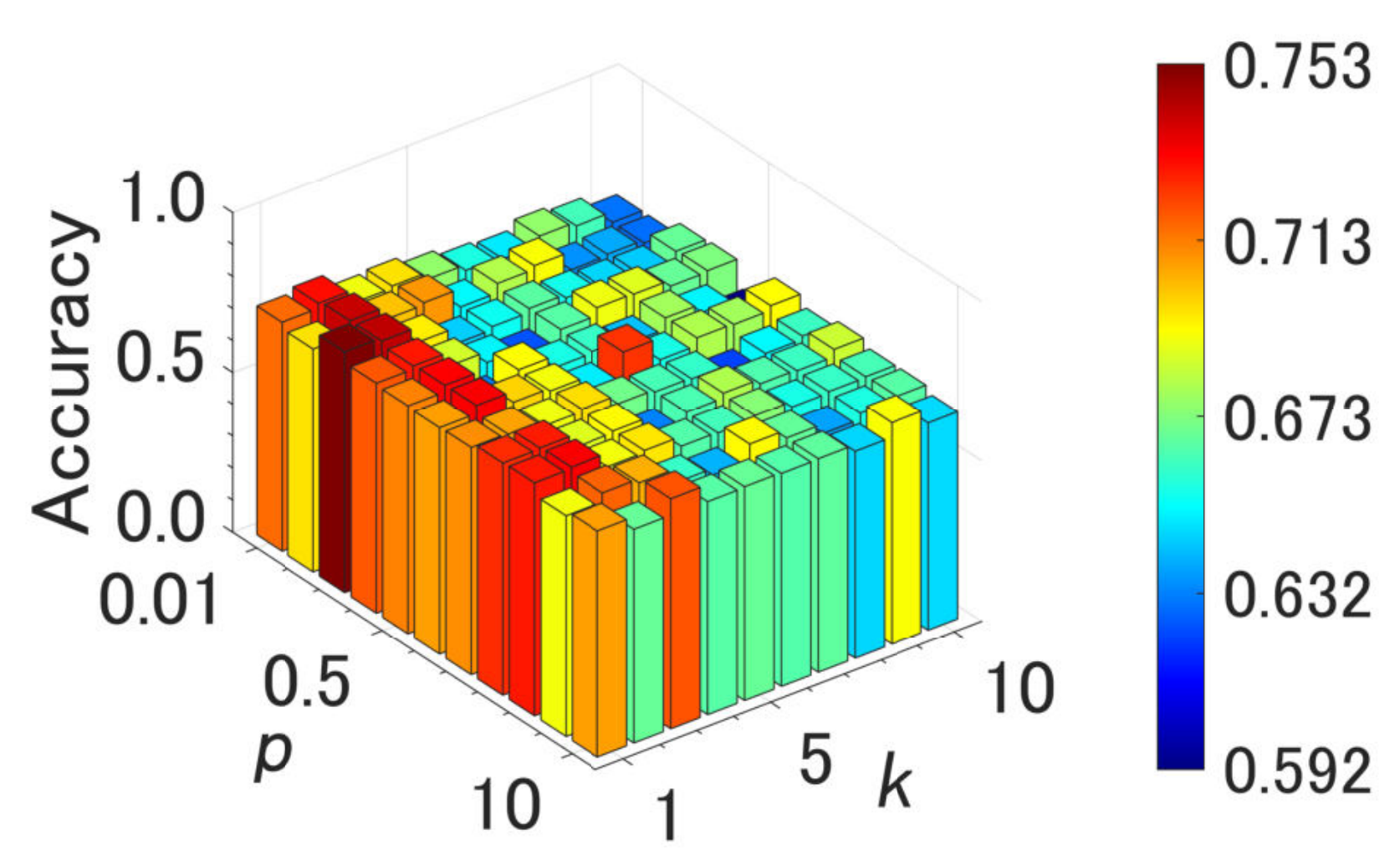}
		\label{fig:PS_Sonar_GSp}
	}
	% \hspace{2mm}
	%	\hfil
	\subfloat[TOX171]{
		\includegraphics[width=1.35in]{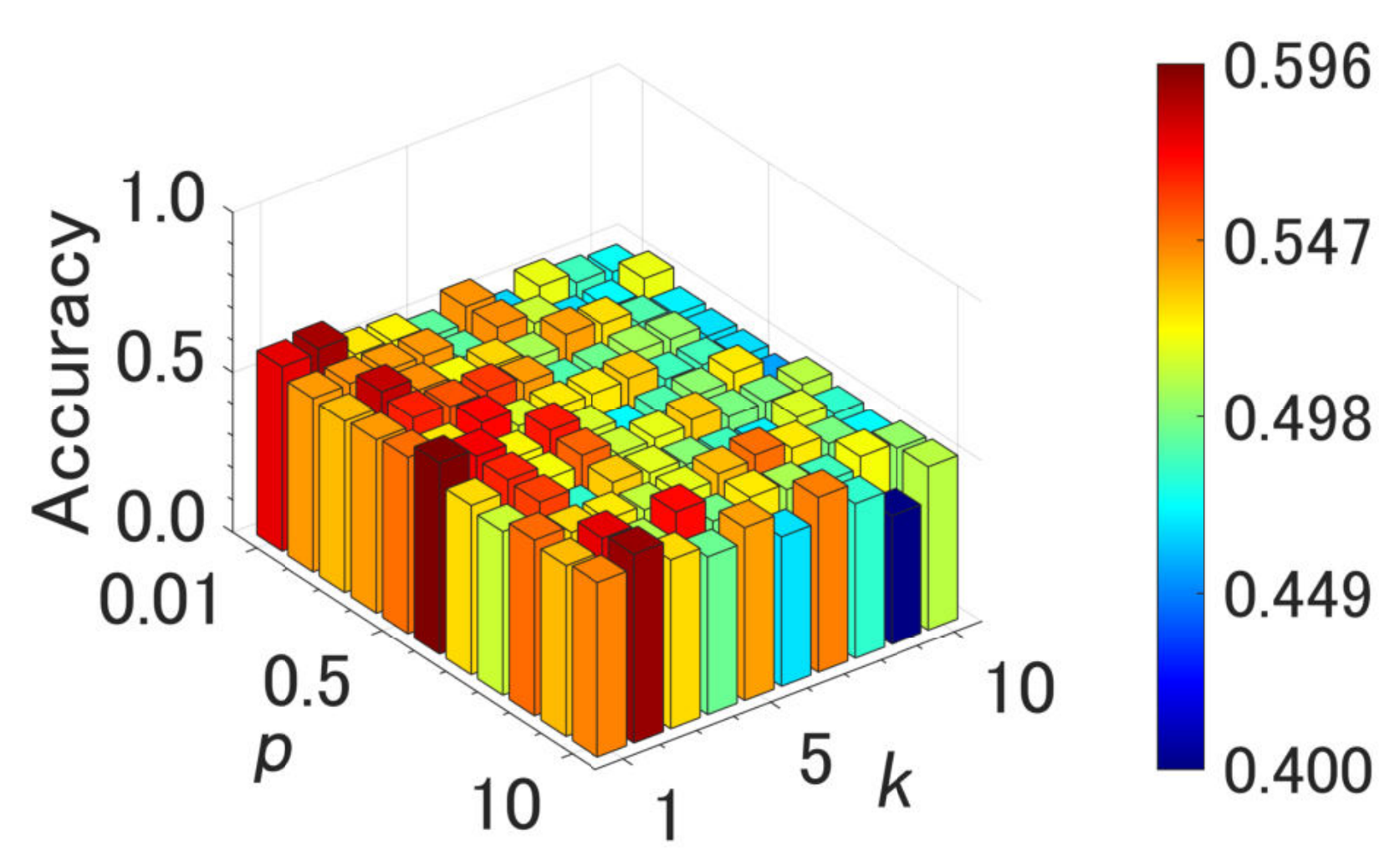}
		\label{fig:PS_TOX171_GSp}
	}
	% \vspace{-1mm}
	\end{adjustwidth}
	\caption{Effects of the parameter specifications of GSOINN+ on Accuracy.}
	\label{fig:paramSensitivity_GSp}
\end{figure}

Figs. \ref{fig:paramSensitivity_ASC}-\ref{fig:paramSensitivity_GSp} show effects of the parameter settings on Accuracy for ASC, FTCAC, and GSOINN+, respectively. For FTCAC, each parameter must be carefully specified to obtain high Accuracy for classification tasks. If the parameters of FTCAC are not specified to an appropriate range, either the classification performance deteriorates dramatically or a predictive model cannot be built. On the other hand, parameter specifications of ASC and GSOINN+ are easier than FTCAC because the range of parameters that provide high Accuracy is wider.

\begin{figure}[htbp]
	\begin{adjustwidth}{-\extralength}{0cm}
	\centering
	\subfloat[Aggregation]{
		\includegraphics[width=1.35in]{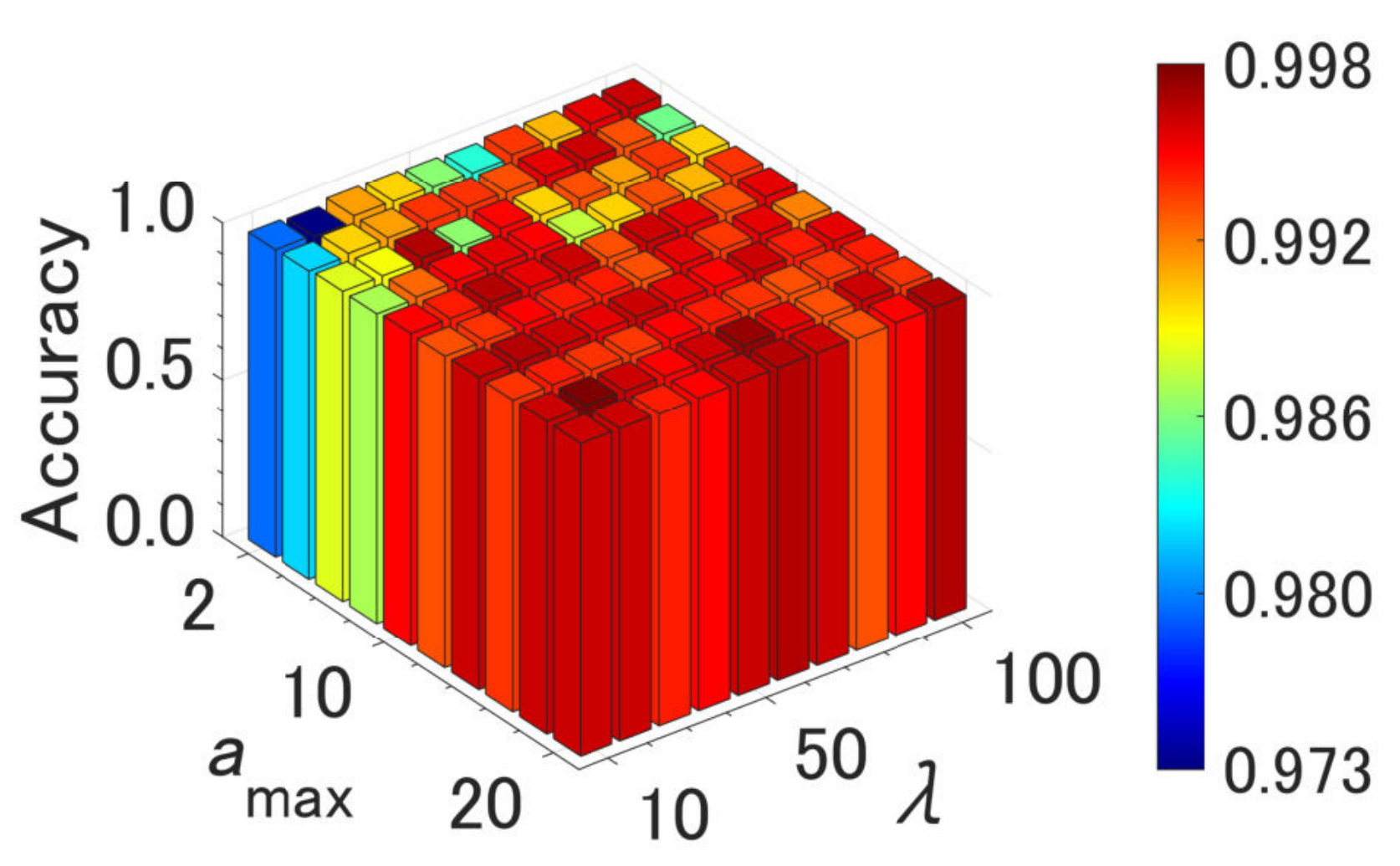}
		\label{fig:PS_Aggregation_G}
	}
	% \hspace{-1.4mm}
	% \hfil
	\subfloat[Compound]{
		\includegraphics[width=1.35in]{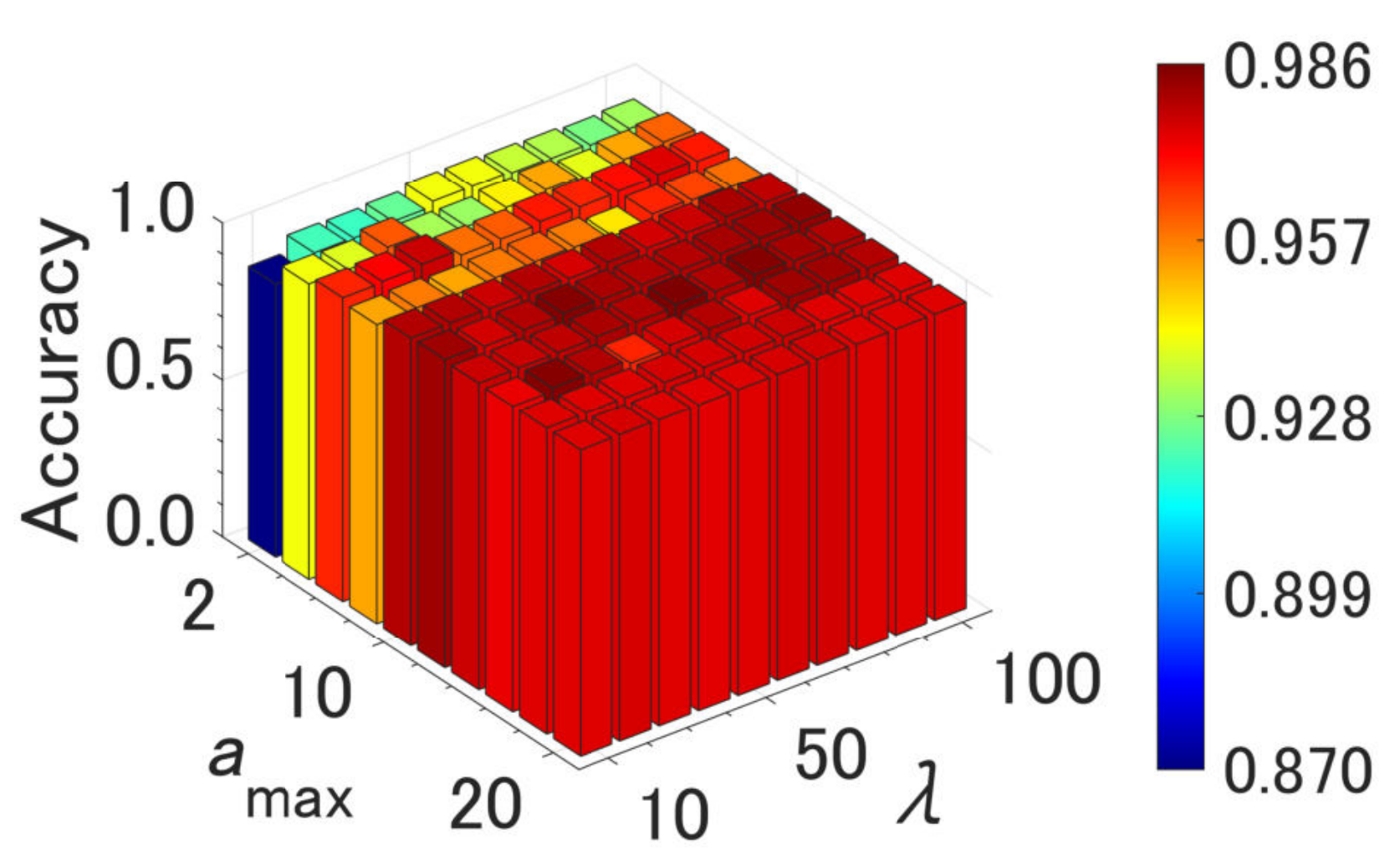}
		\label{fig:PS_Compound_G}
	}
	% \hspace{-1.4mm}
	% \hfil
	\subfloat[Hard Distribution]{
		\includegraphics[width=1.35in]{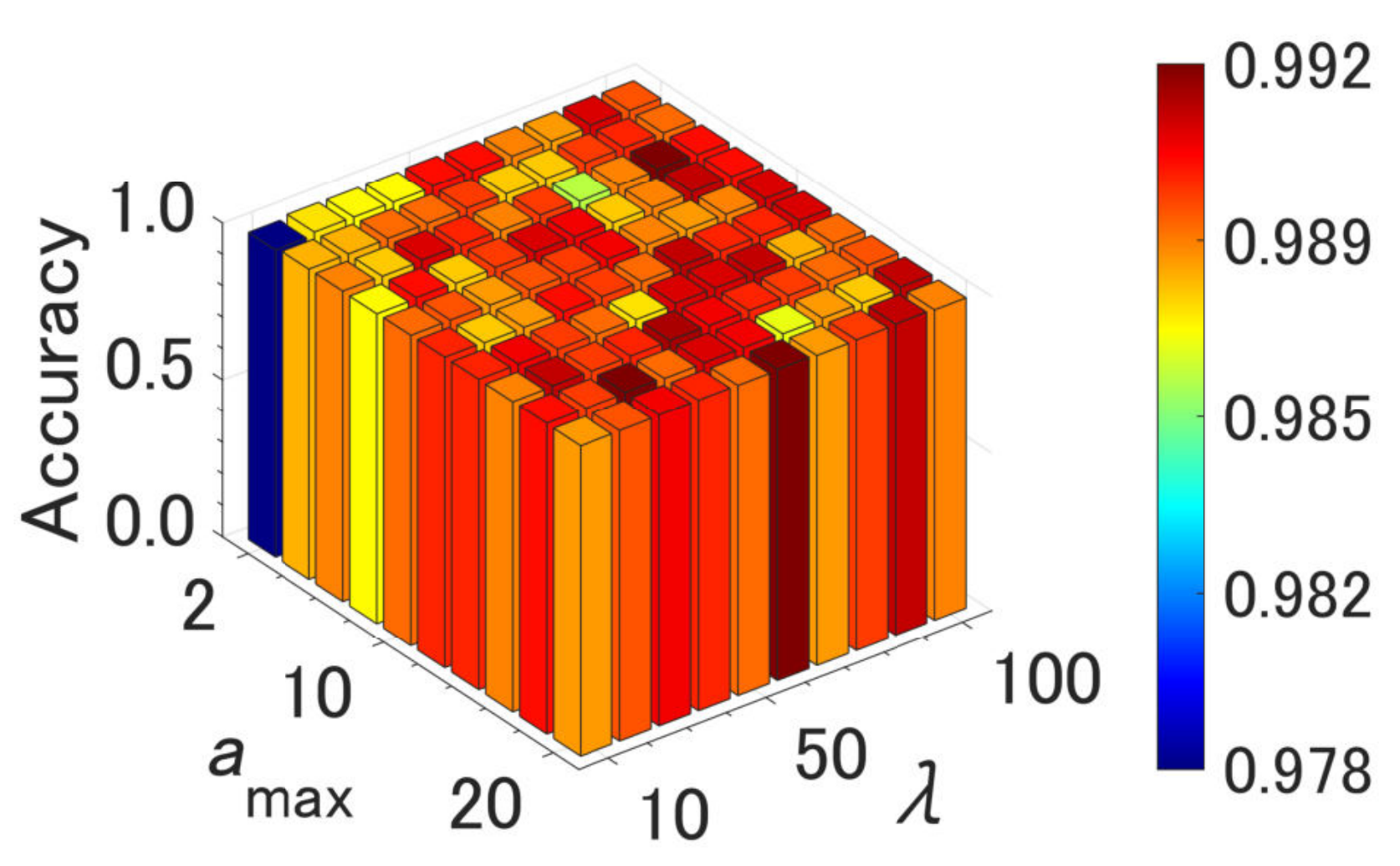}
		\label{fig:PS_HardDistribution_G}
	}
	% \hspace{-1.4mm}
	% \hfil
	\subfloat[Jain]{
		\includegraphics[width=1.35in]{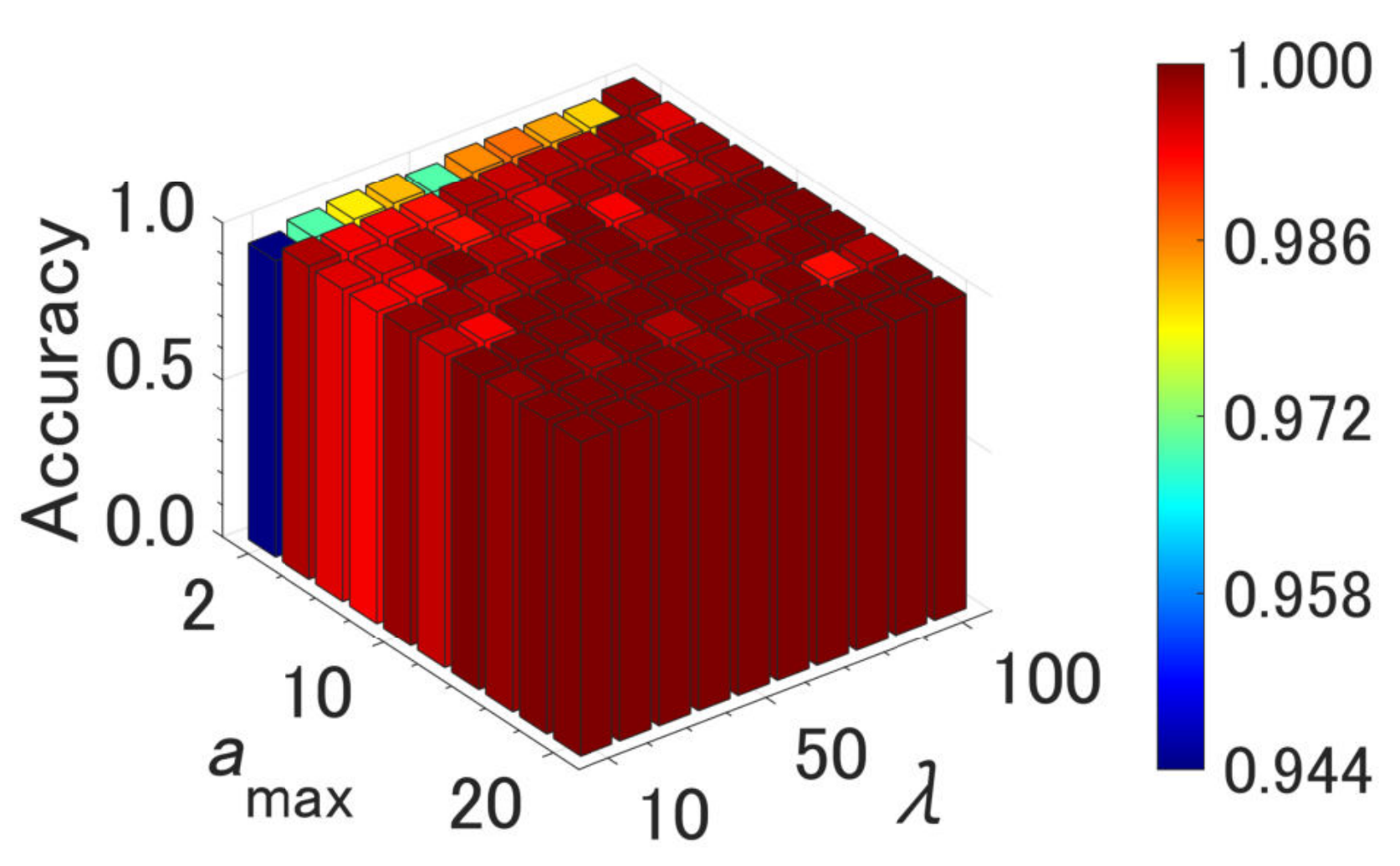}
		\label{fig:PS_Jain_G}
	}
	% \hspace{-1.4mm}
	% \hfil
	\subfloat[Pathbased]{
		\includegraphics[width=1.35in]{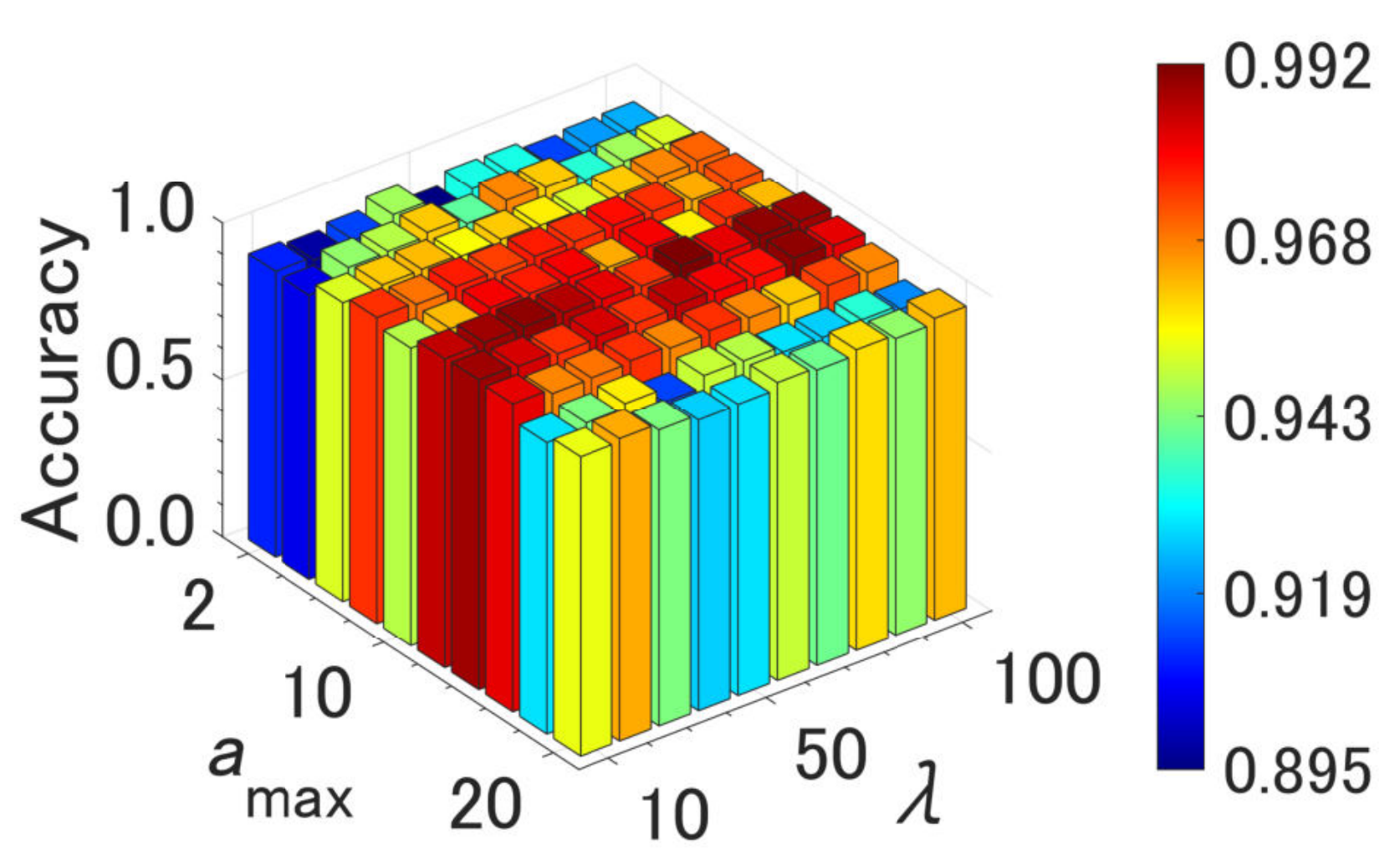}
		\label{fig:PS_Pathbased_G}
	}
	\\
	% \vspace{-2.5mm}
	%	\hfil
	% \hspace{-1.4mm}
	\subfloat[ALLAML]{
		\includegraphics[width=1.35in]{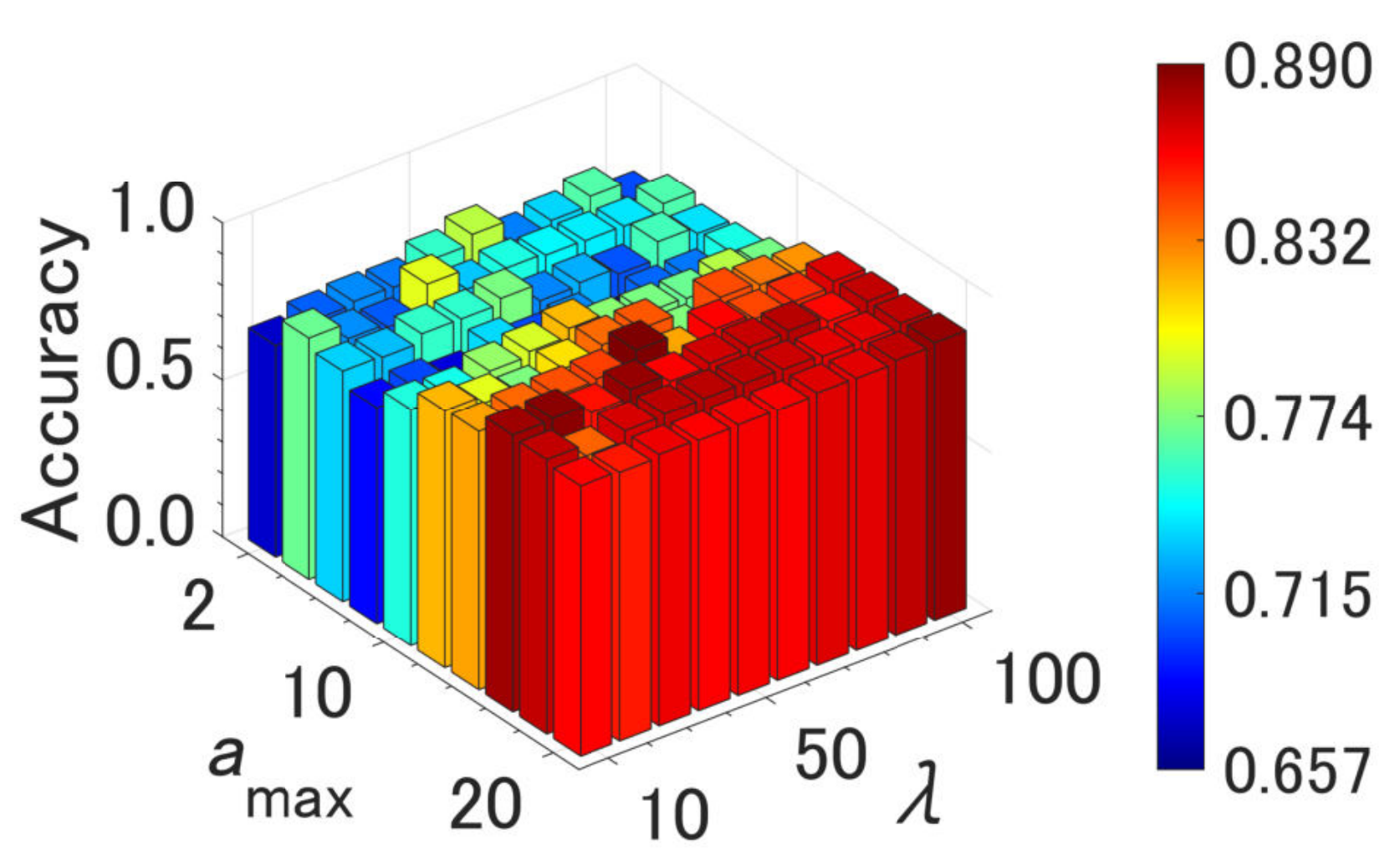}
		\label{fig:PS_ALLAML_G}
	}
	% \hspace{-1.4mm}
	% \hfil
	\subfloat[COIL20]{
		\includegraphics[width=1.35in]{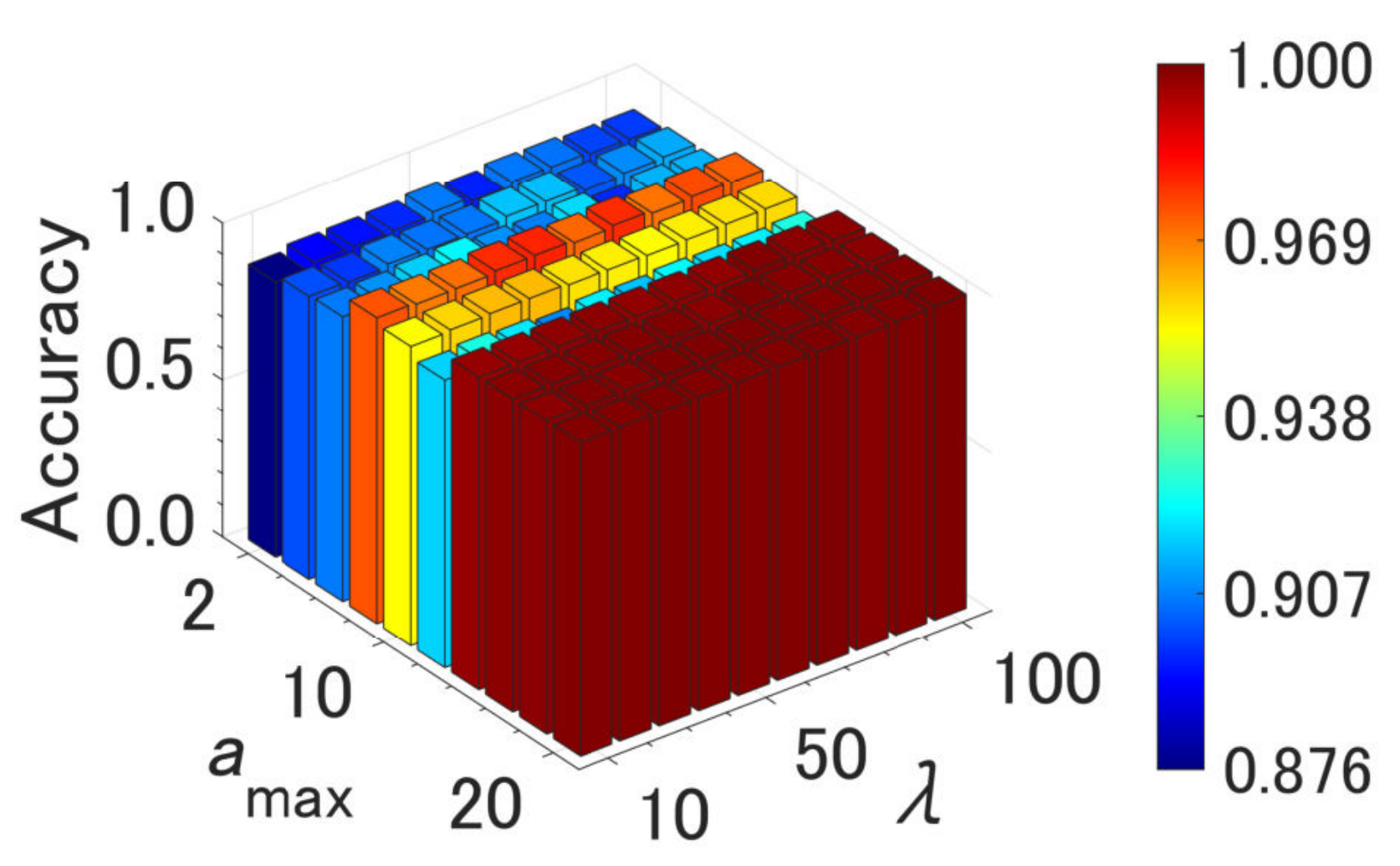}
		\label{fig:PS_COIL20_G}
	}
	% \hspace{-1.4mm}
	% \hfil
	\subfloat[Iris]{
		\includegraphics[width=1.35in]{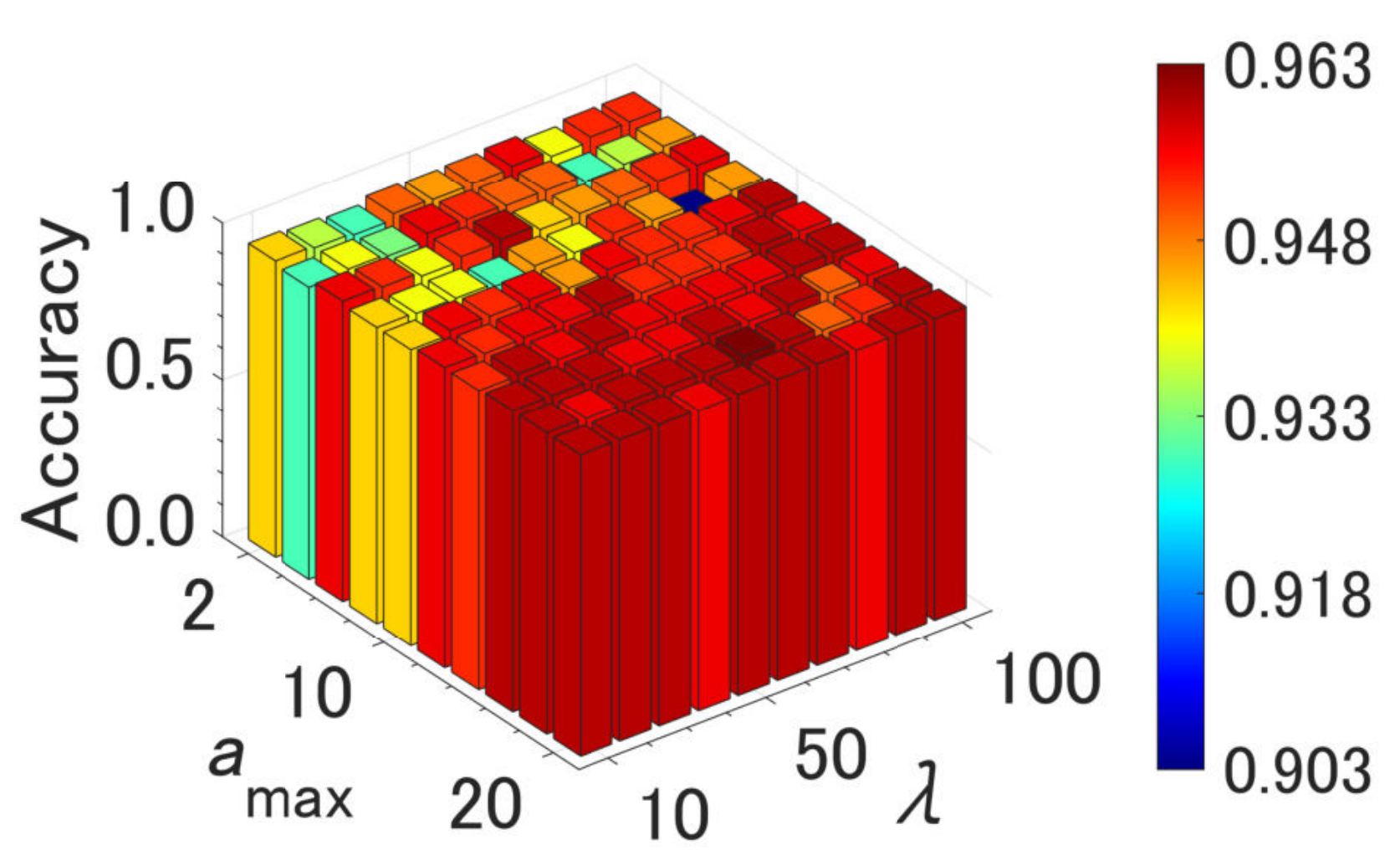}
		\label{fig:PS_Iris_G}
	}
	% \hspace{-1.4mm}
	% \hfil
	\subfloat[Isolet]{
		\includegraphics[width=1.35in]{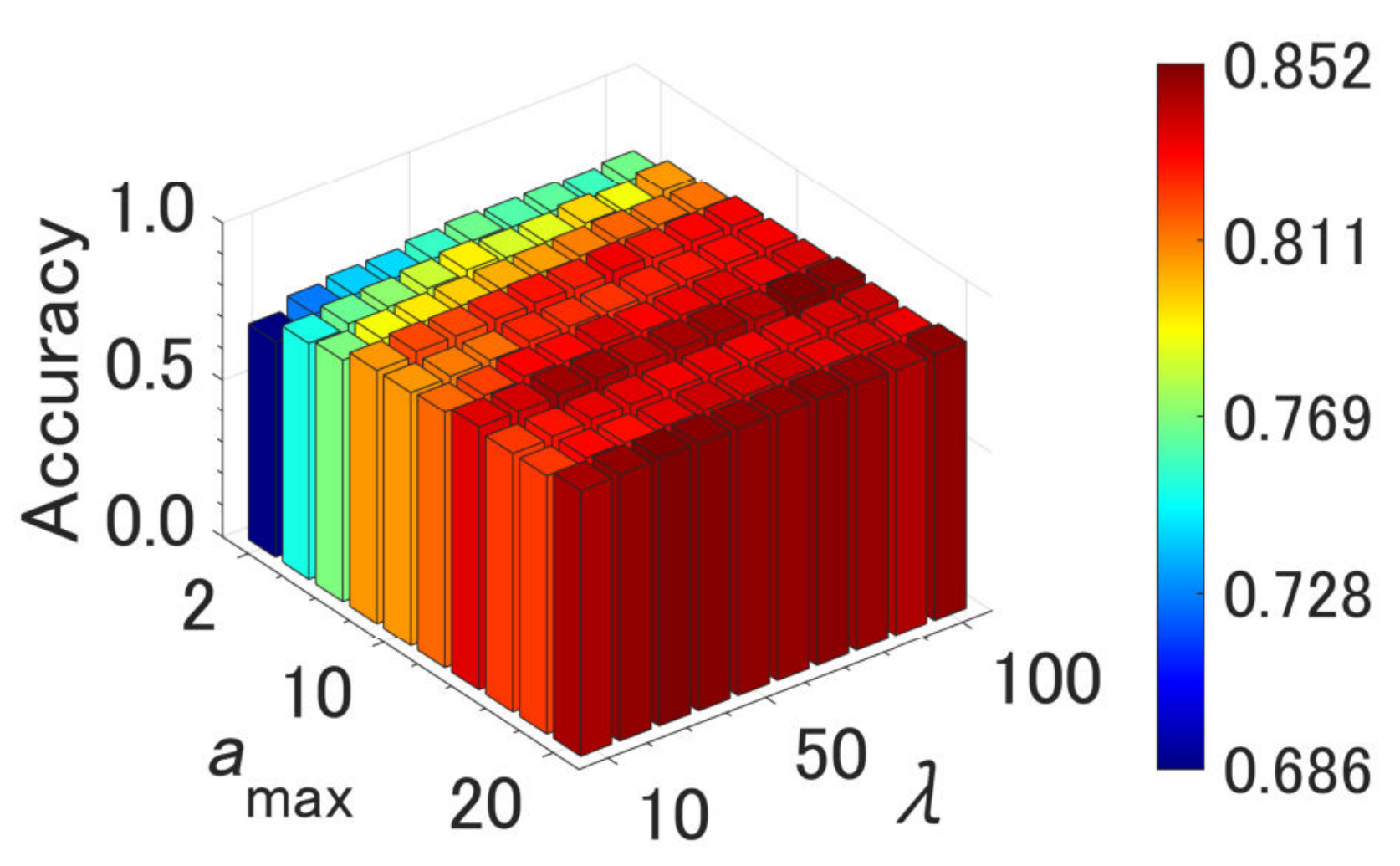}
		\label{fig:PS_Isolet_G}
	}
	% \hspace{-1.4mm}
	% \hfil
	\subfloat[OptDigits]{
		\includegraphics[width=1.35in]{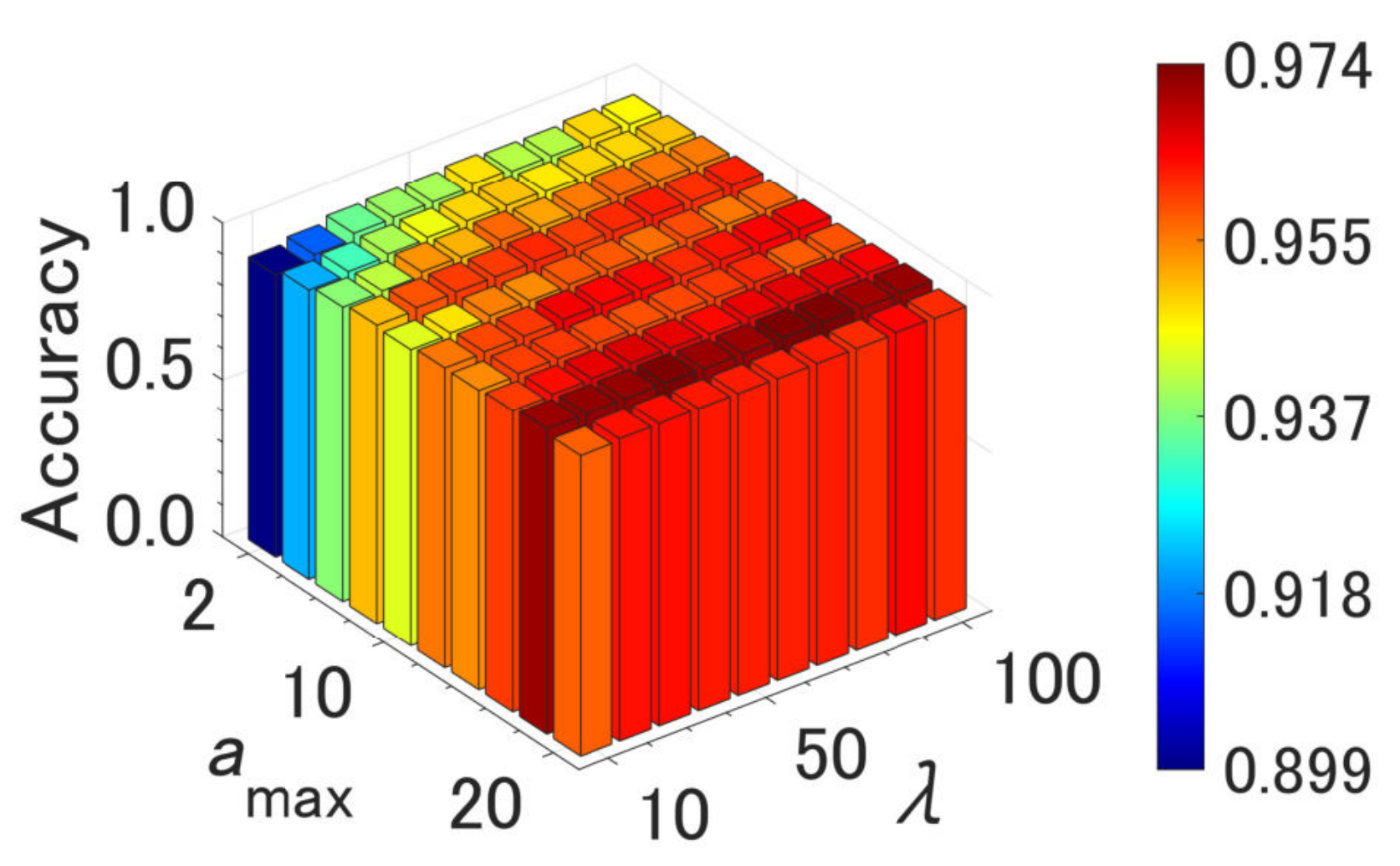}
		\label{fig:PS_OptDigits_G}
	}
	\\
	% \vspace{-2.5mm}
	%	\hfil
	\subfloat[Seeds]{
		\includegraphics[width=1.35in]{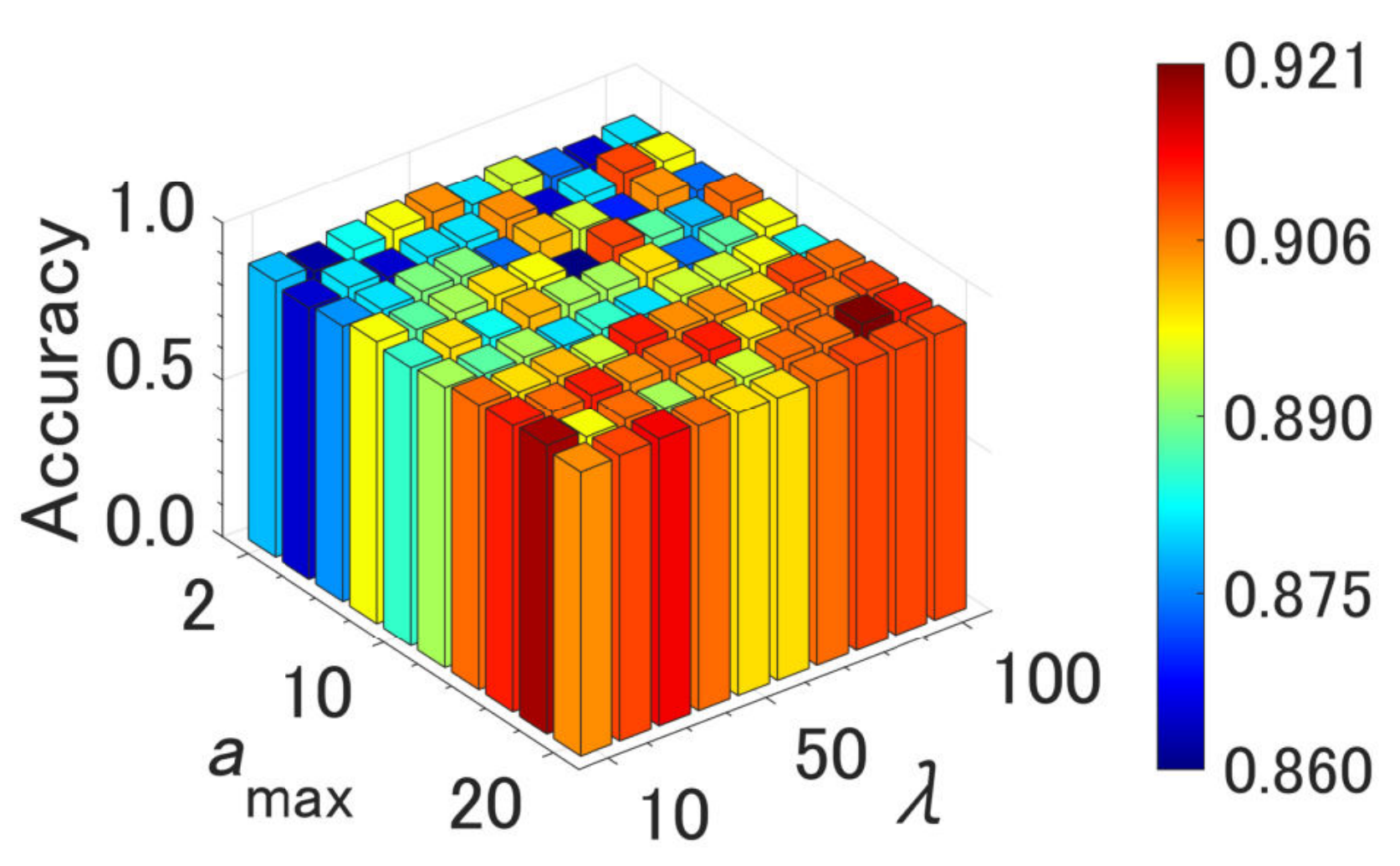}
		\label{fig:PS_Seeds_G}
	}
	% \hspace{2mm}
	%	\hfil
	\subfloat[Semeion]{
		\includegraphics[width=1.35in]{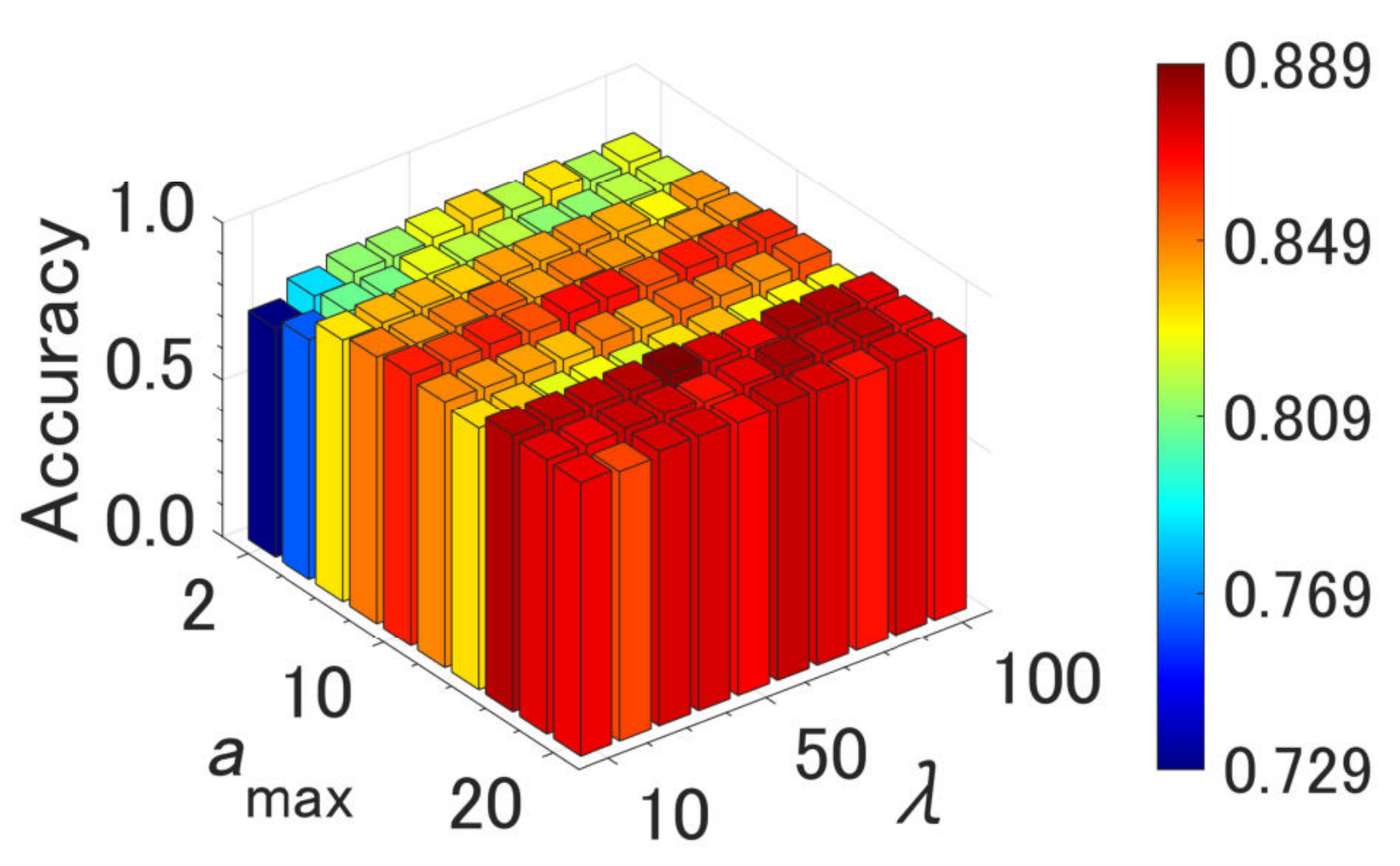}
		\label{fig:PS_Semeion_G}
	}
	% \hspace{2mm}
	%	\hfil
	\subfloat[Sonar]{
		\includegraphics[width=1.35in]{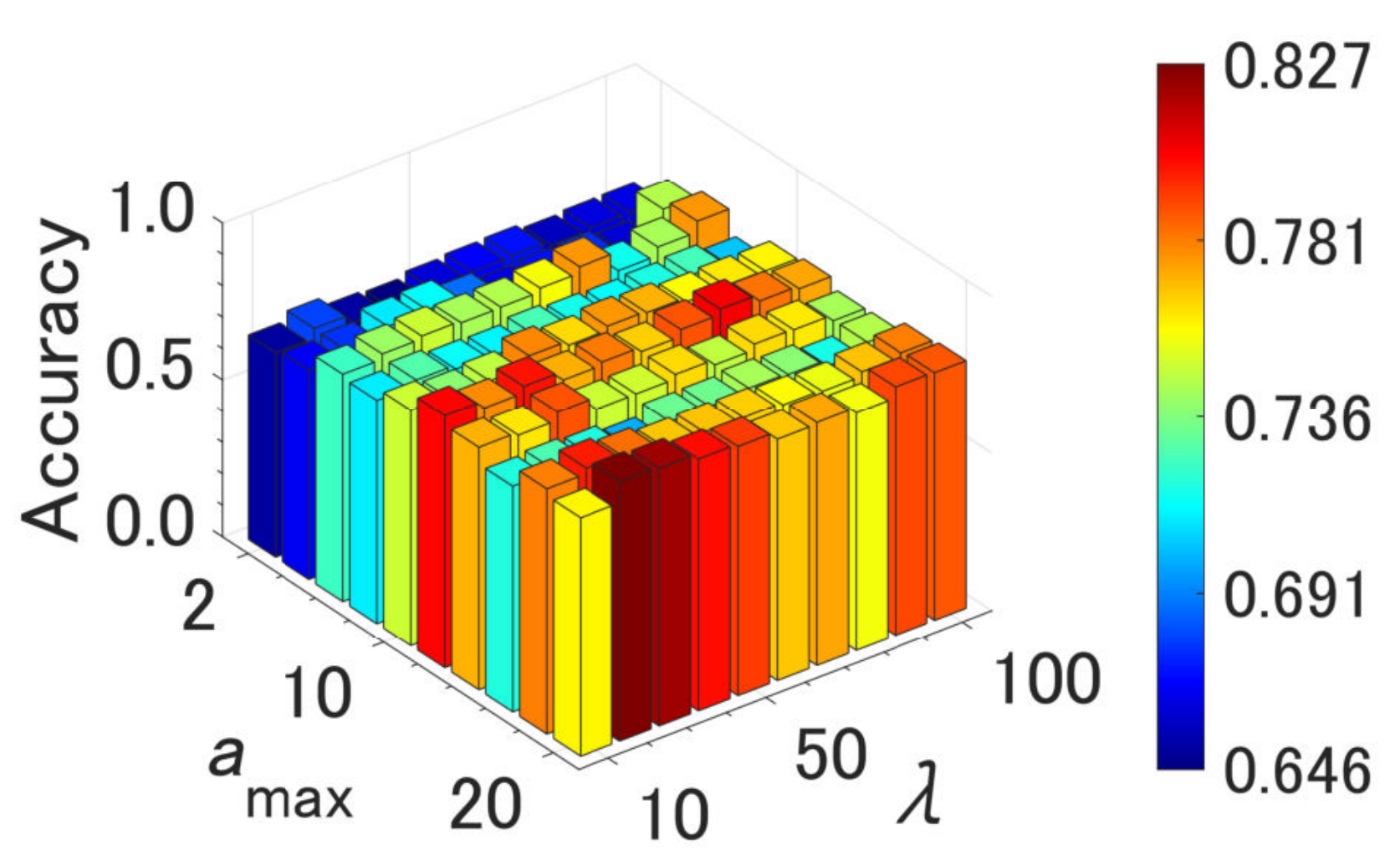}
		\label{fig:PS_Sonar_G}
	}
	% \hspace{2mm}
	%	\hfil
	\subfloat[TOX171]{
		\includegraphics[width=1.35in]{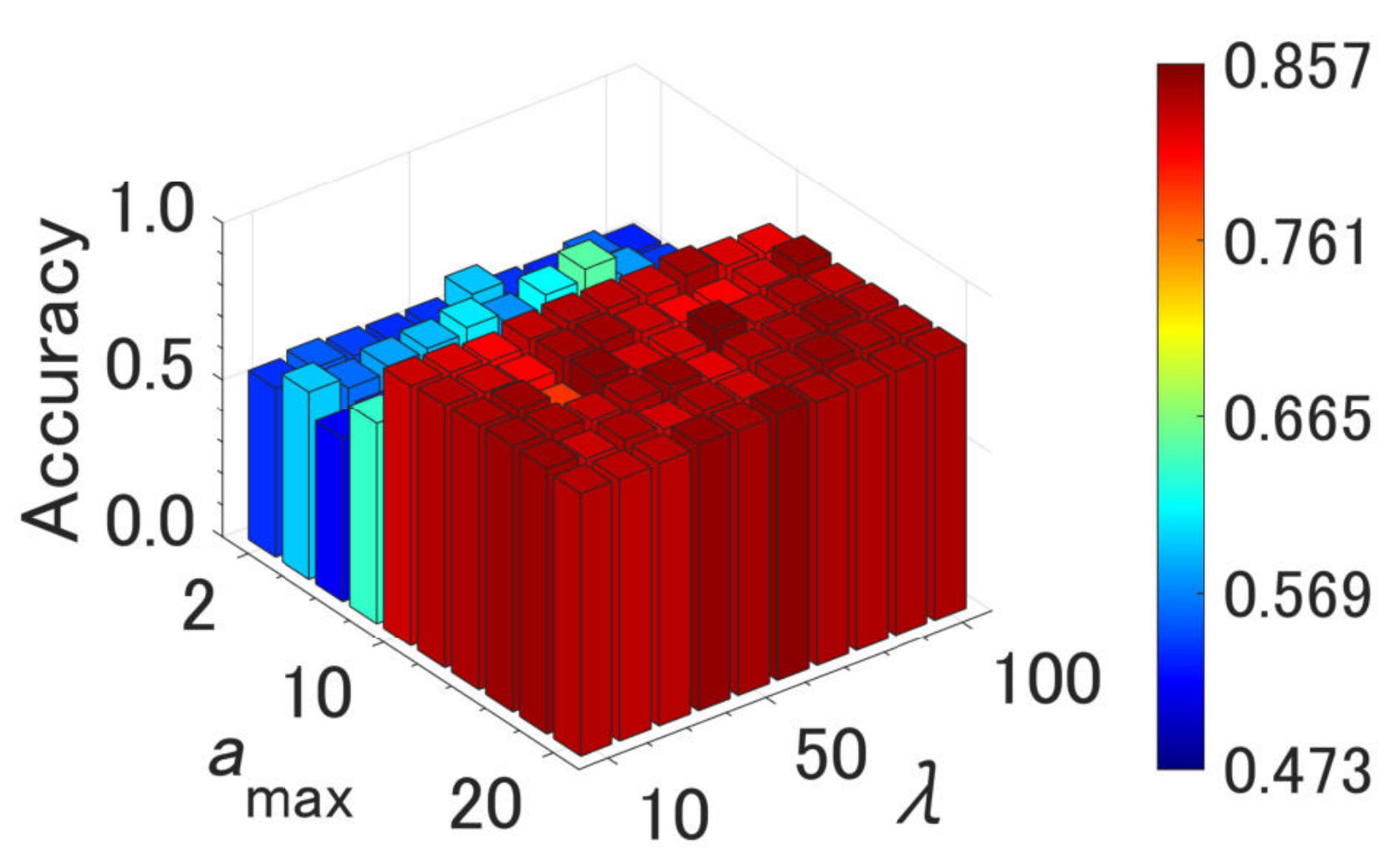}
		\label{fig:PS_TOX171_G}
	}
	% \vspace{-1mm}
	\end{adjustwidth}
	\caption{Effects of the parameter specifications of CAEAC on Accuracy.}
	\label{fig:paramSensitivity_G}
\end{figure}

\begin{figure}[htbp]
	\begin{adjustwidth}{-\extralength}{0cm}
	\centering
	\subfloat[Aggregation]{
		\includegraphics[width=1.35in]{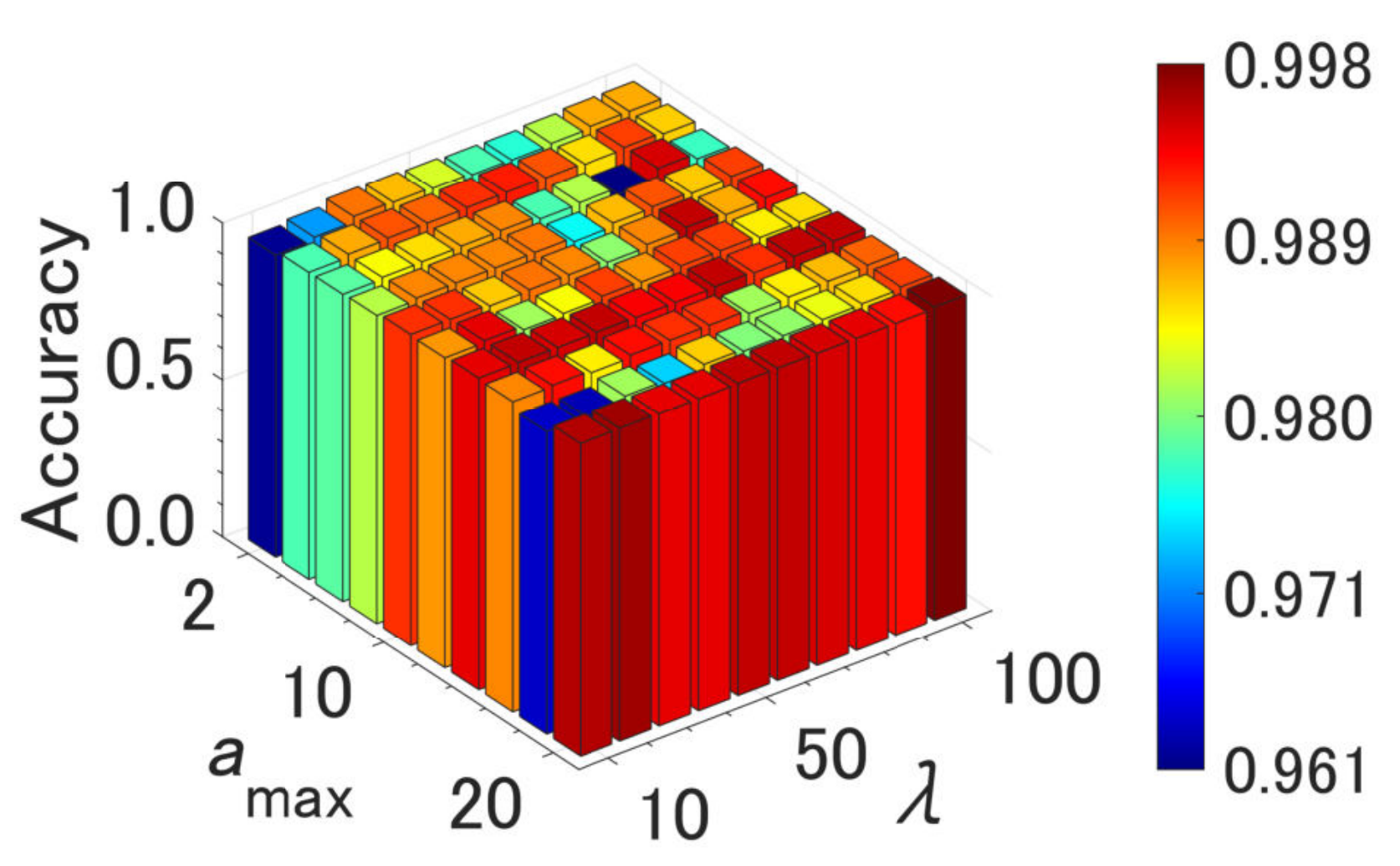}
		\label{fig:PS_Aggregation_I}
	}
	% \hspace{-1.4mm}
	% \hfil
	\subfloat[Compound]{
		\includegraphics[width=1.35in]{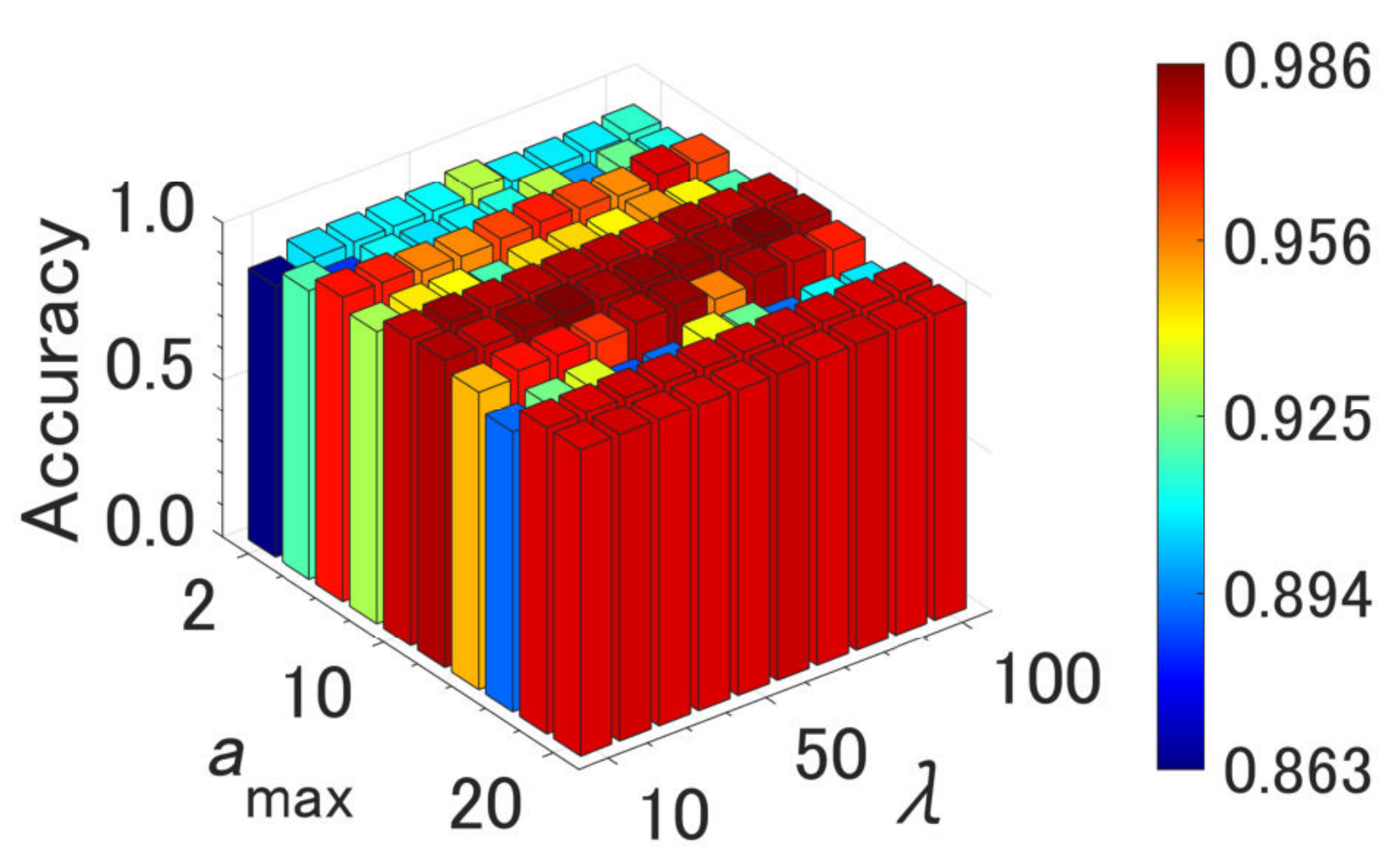}
		\label{fig:PS_Compound_I}
	}
	% \hspace{-1.4mm}
	% \hfil
	\subfloat[Hard Distribution]{
		\includegraphics[width=1.35in]{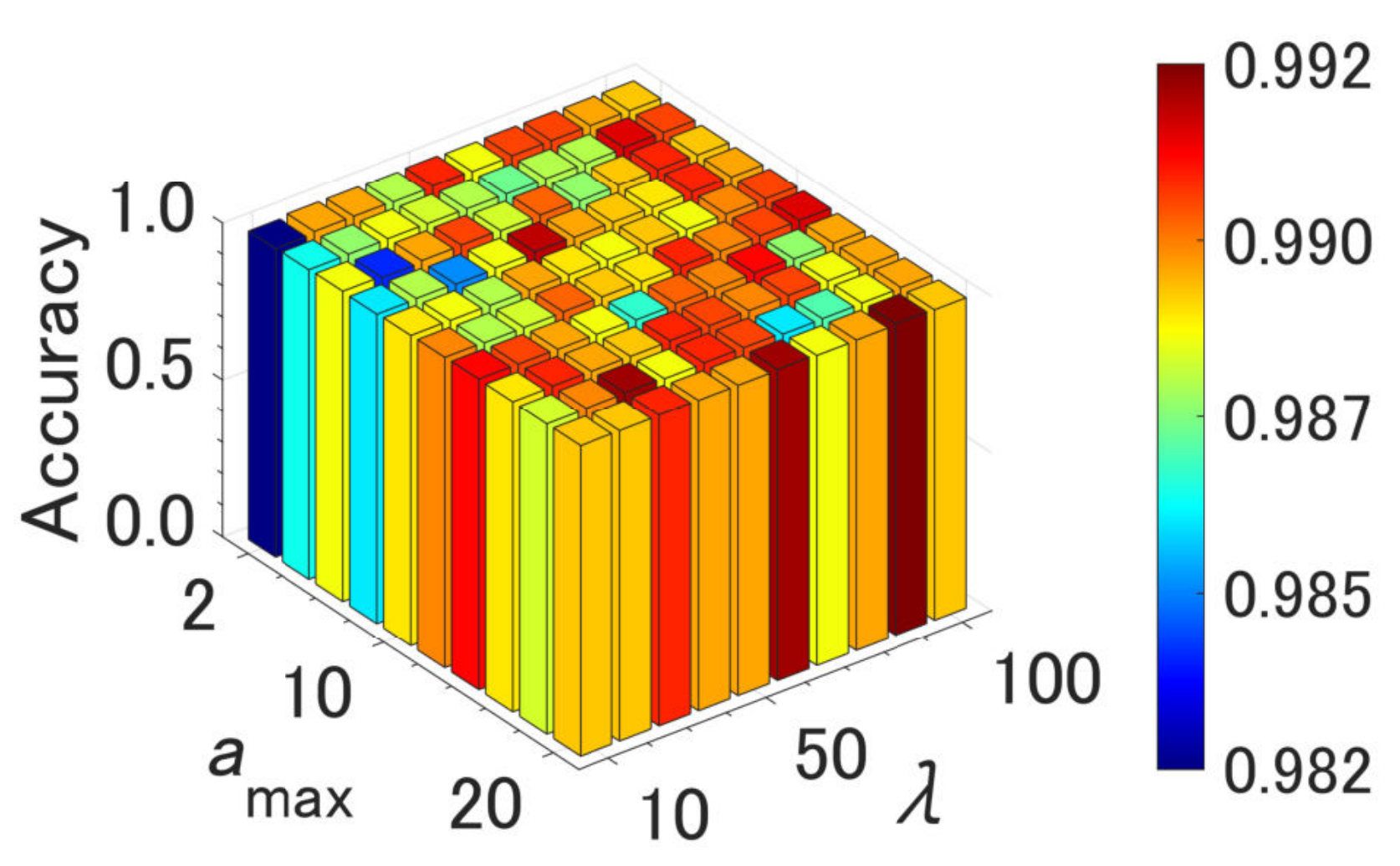}
		\label{fig:PS_HardDistribution_I}
	}
	% \hspace{-1.4mm}
	% \hfil
	\subfloat[Jain]{
		\includegraphics[width=1.35in]{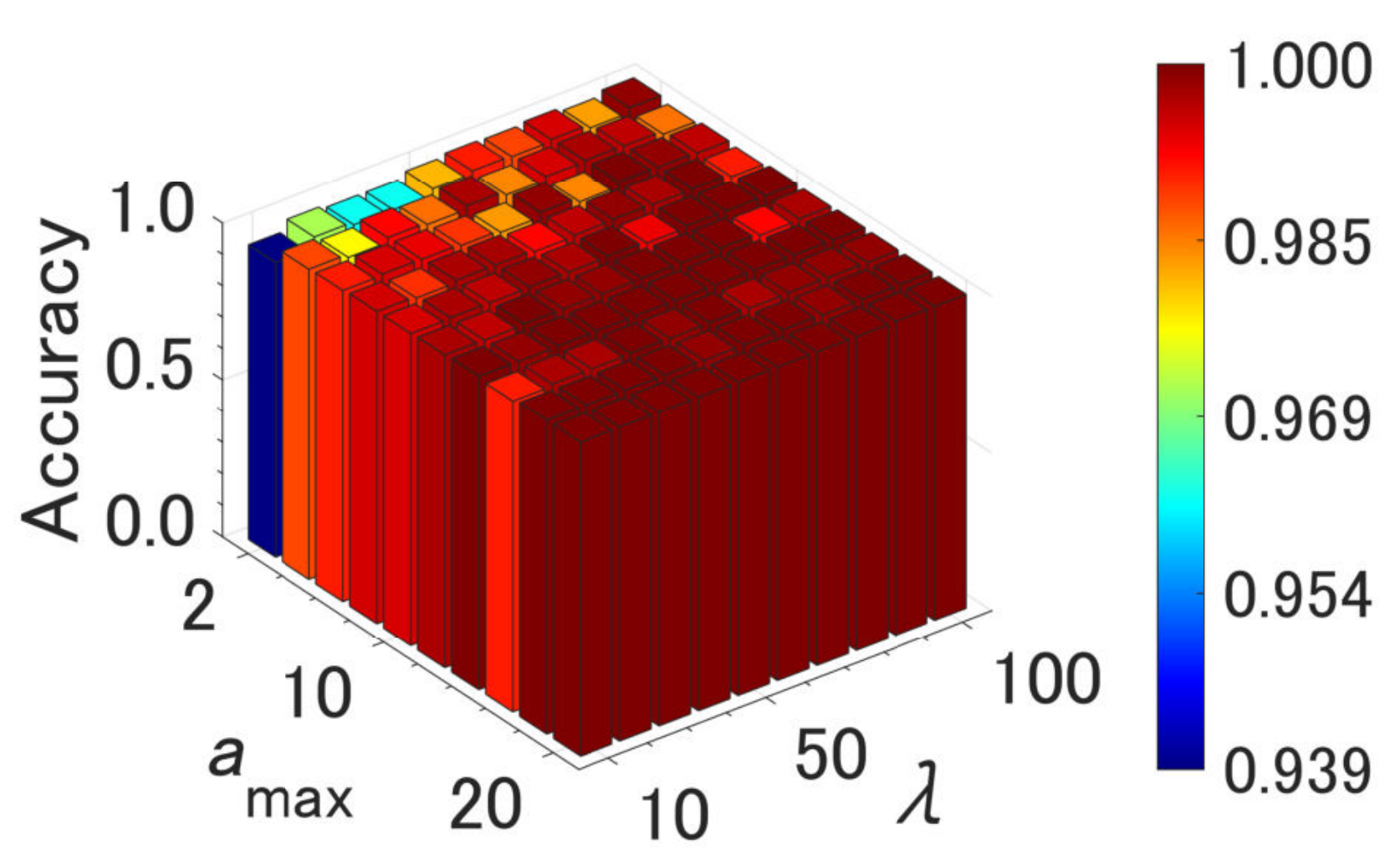}
		\label{fig:PS_Jain_I}
	}
	% \hspace{-1.4mm}
	% \hfil
	\subfloat[Pathbased]{
		\includegraphics[width=1.35in]{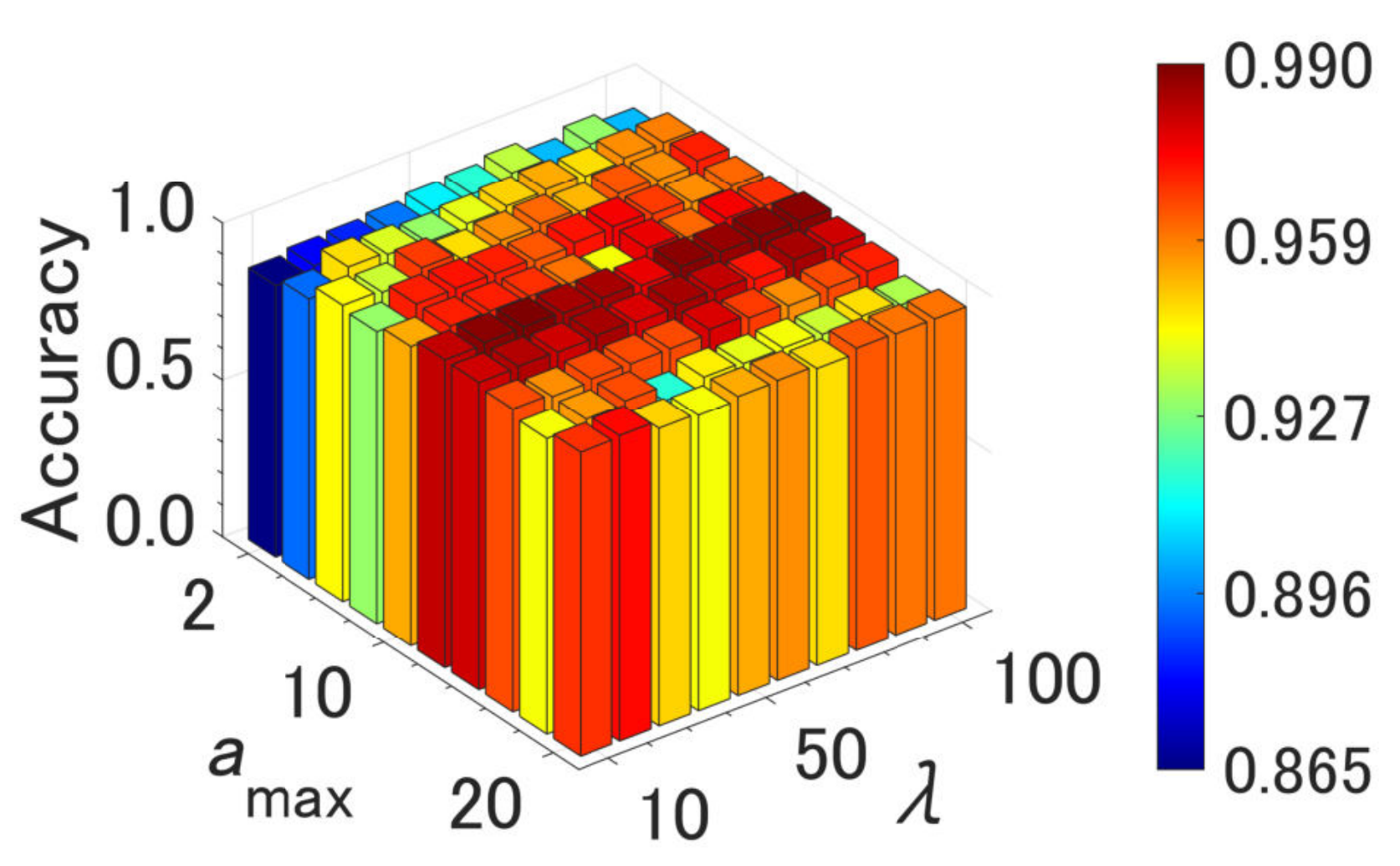}
		\label{fig:PS_Pathbased_I}
	}
	\\
	% \vspace{-2.5mm}
	%	\hfil
	% \hspace{-1.4mm}
	\subfloat[ALLAML]{
		\includegraphics[width=1.35in]{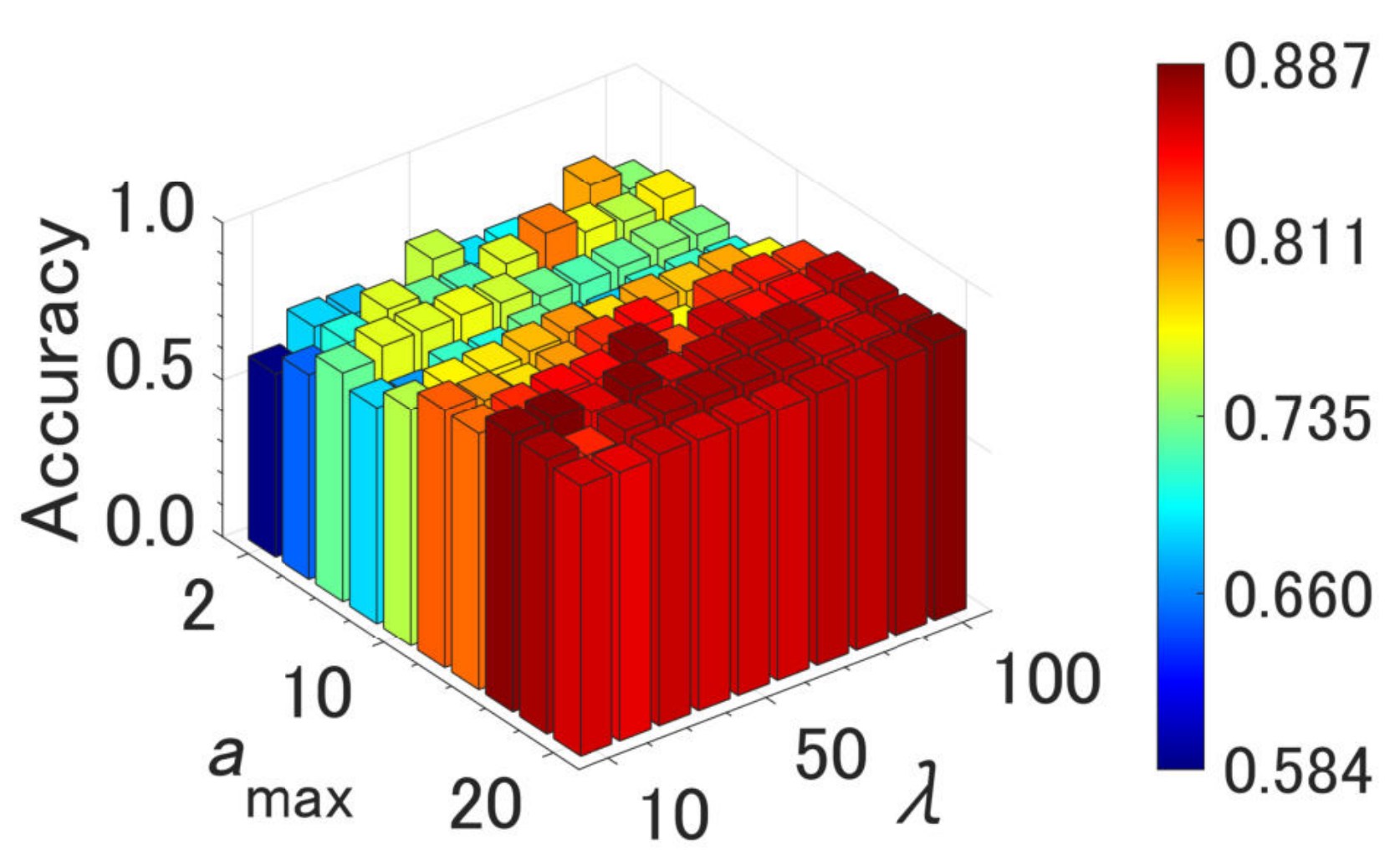}
		\label{fig:PS_ALLAML_I}
	}
	% \hspace{-1.4mm}
	% \hfil
	\subfloat[COIL20]{
		\includegraphics[width=1.35in]{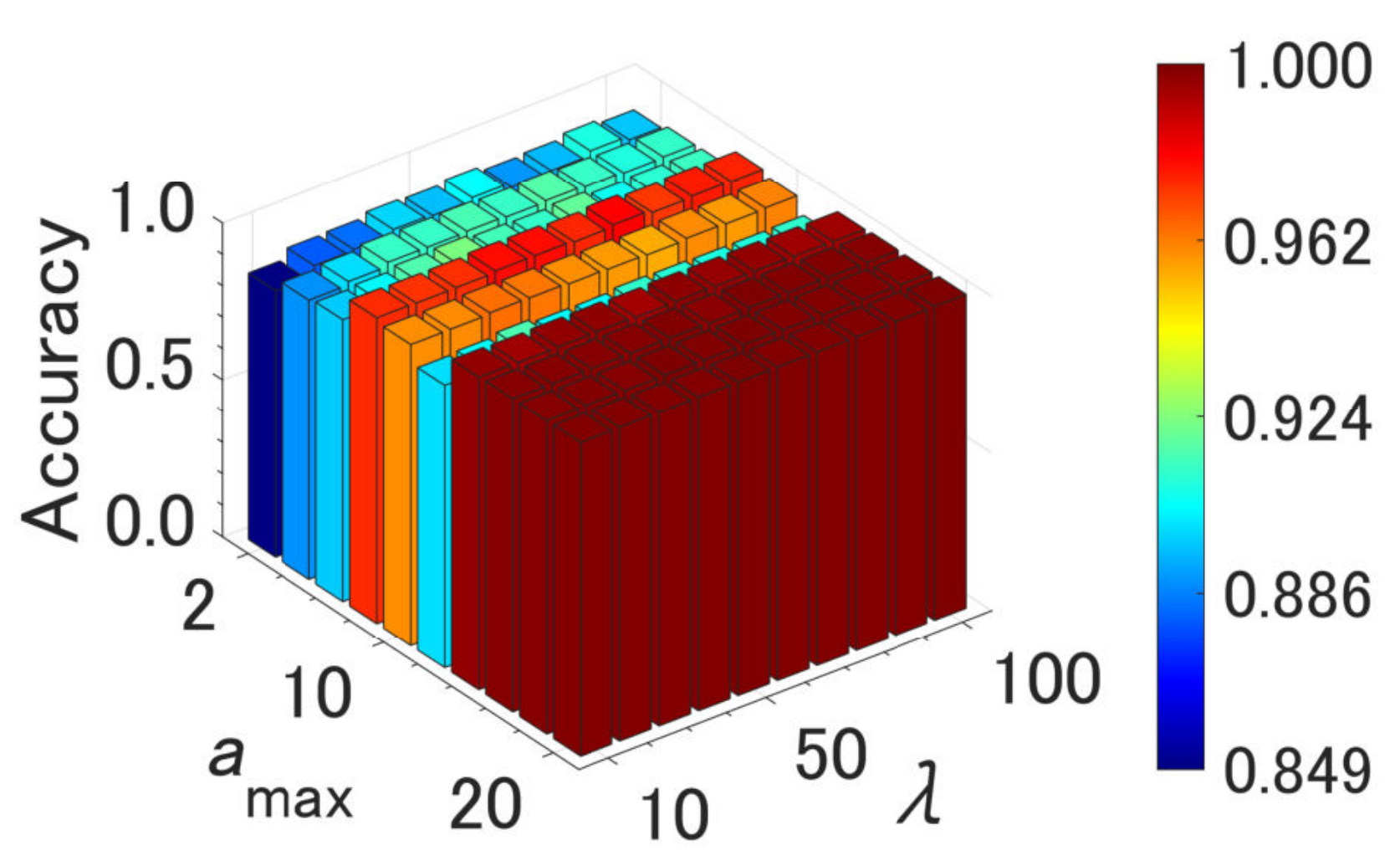}
		\label{fig:PS_COIL20_I}
	}
	% \hspace{-1.4mm}
	% \hfil
	\subfloat[Iris]{
		\includegraphics[width=1.35in]{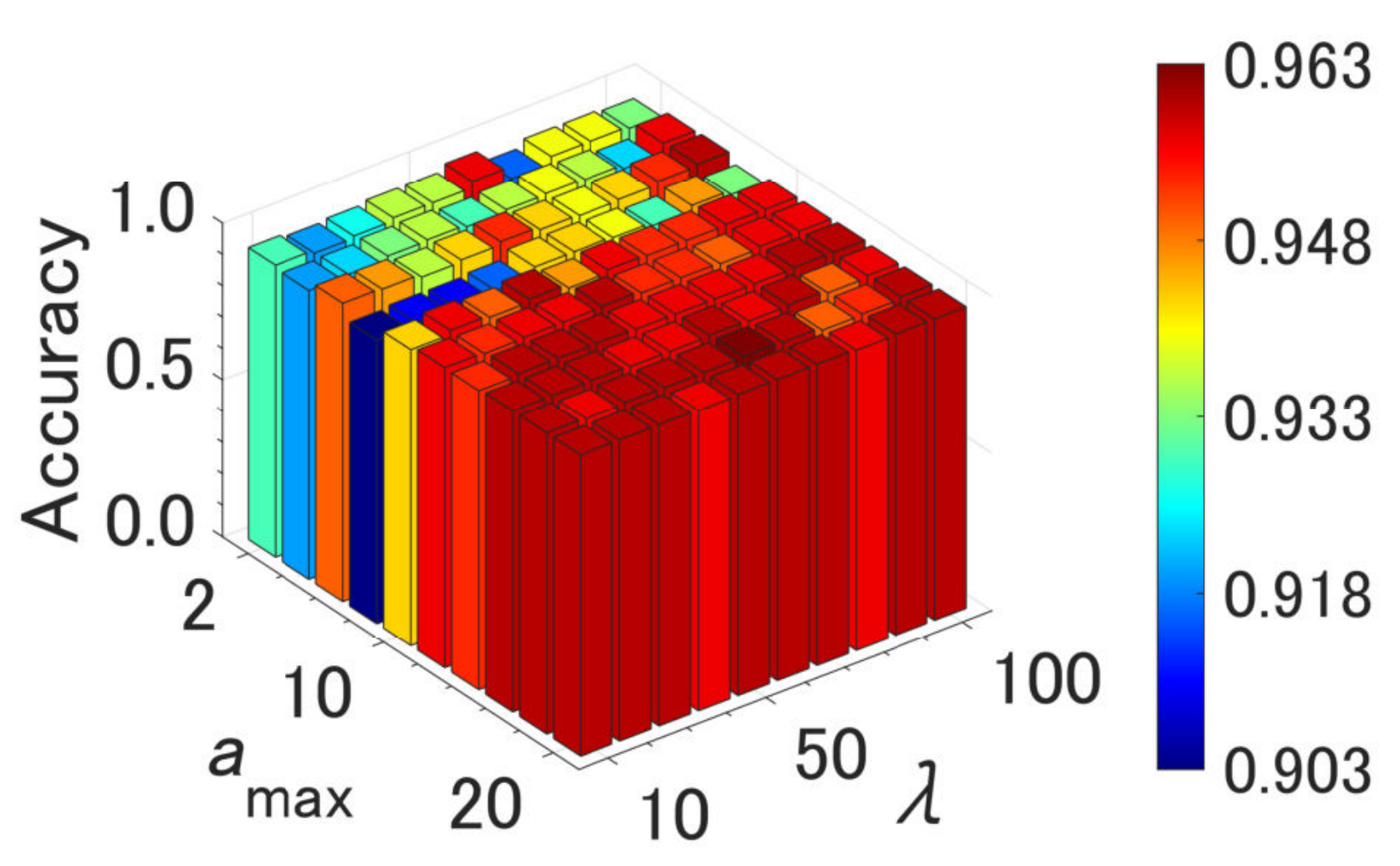}
		\label{fig:PS_Iris_I}
	}
	% \hspace{-1.4mm}
	% \hfil
	\subfloat[Isolet]{
		\includegraphics[width=1.35in]{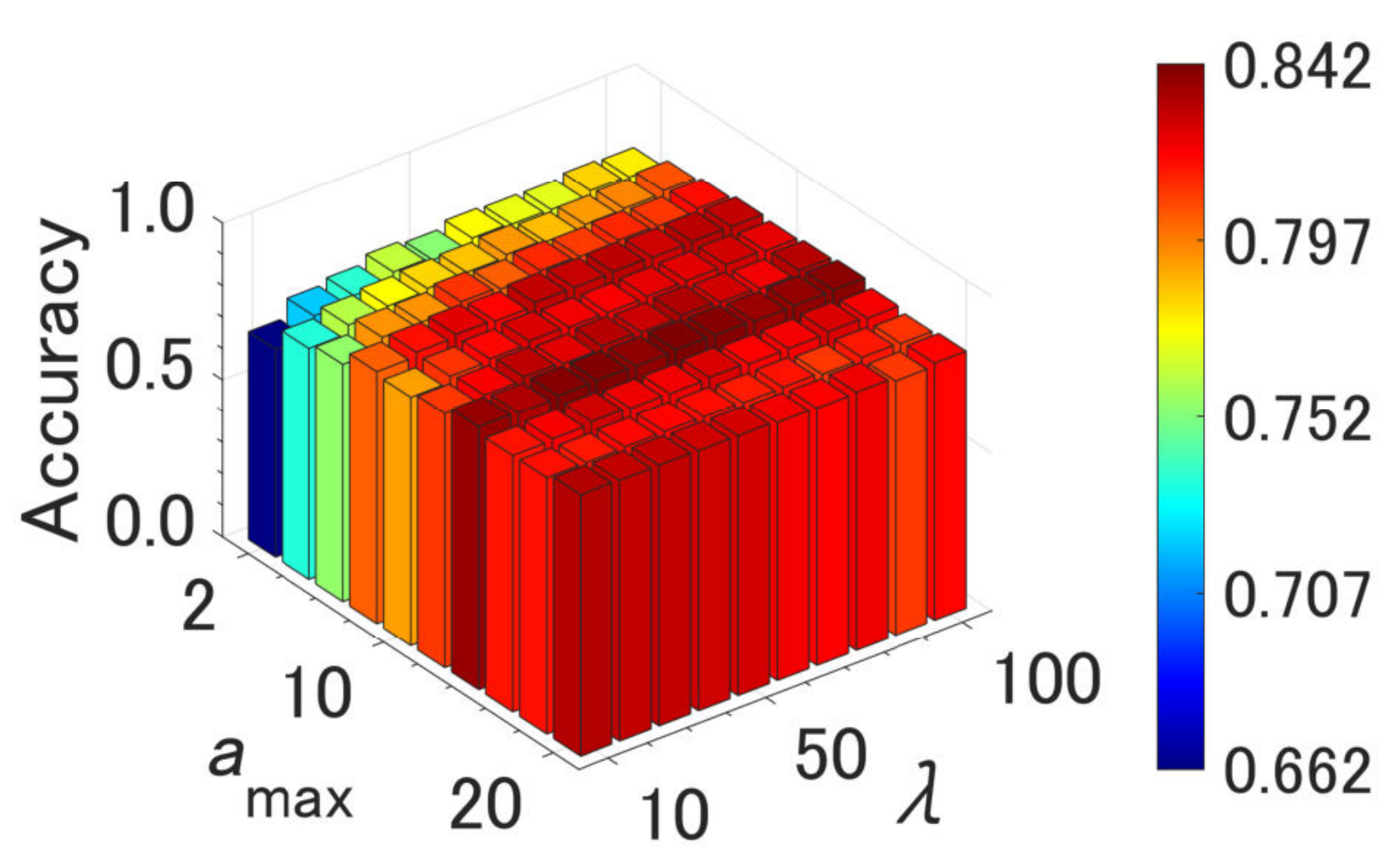}
		\label{fig:PS_Isolet_I}
	}
	% \hspace{-1.4mm}
	% \hfil
	\subfloat[OptDigits]{
		\includegraphics[width=1.35in]{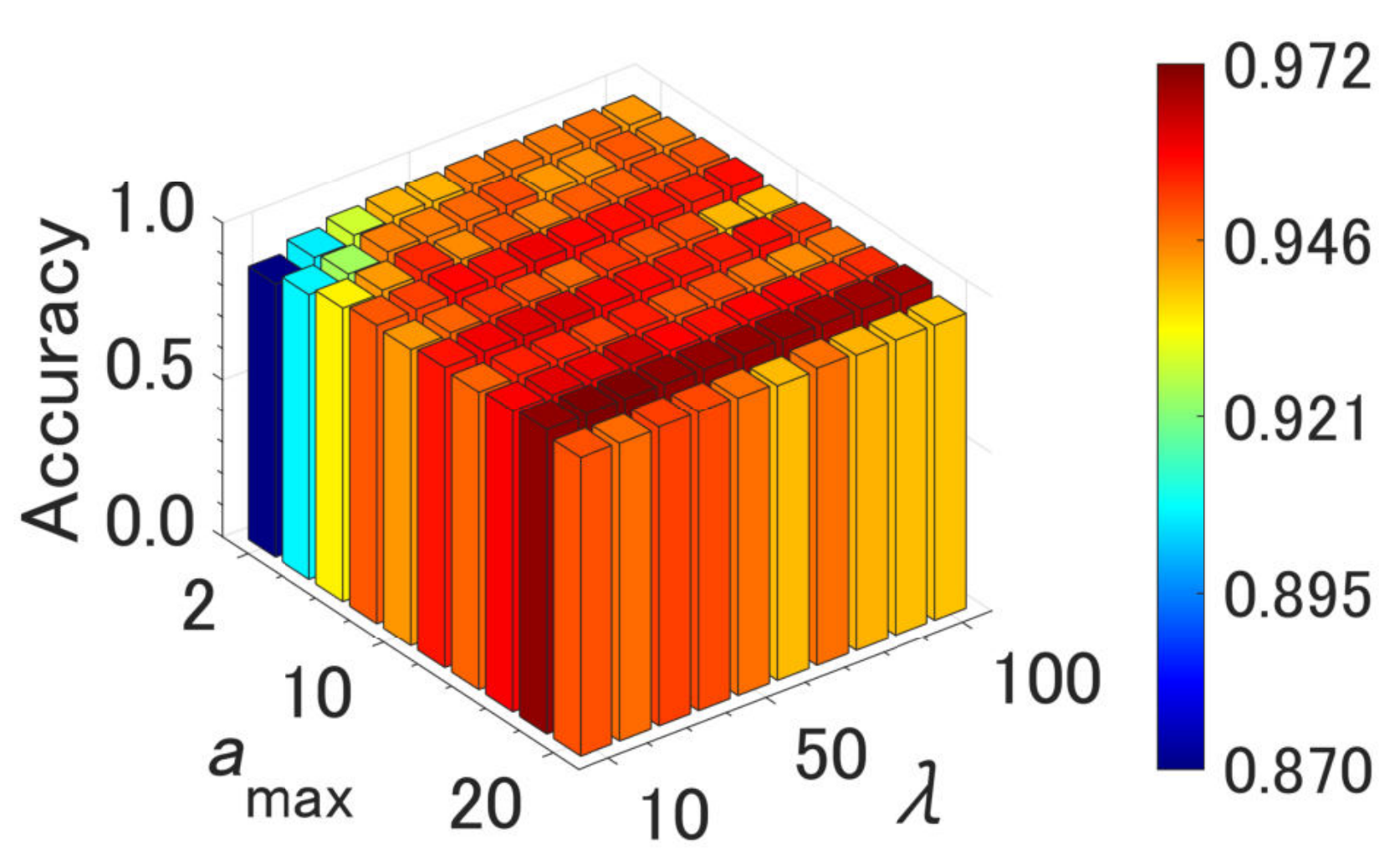}
		\label{fig:PS_OptDigits_I}
	}
	\\
	% \vspace{-2.5mm}
	%	\hfil
	\subfloat[Seeds]{
		\includegraphics[width=1.35in]{figures/CAEAC_Seeds_ACC.pdf}
		\label{fig:PS_Seeds_I}
	}
	% \hspace{2mm}
	%	\hfil
	\subfloat[Semeion]{
		\includegraphics[width=1.35in]{figures/CAEAC_Semeion_ACC.pdf}
		\label{fig:PS_Semeion_I}
	}
	% \hspace{2mm}
	%	\hfil
	\subfloat[Sonar]{
		\includegraphics[width=1.35in]{figures/CAEAC_Sonar_ACC.pdf}
		\label{fig:PS_Sonar_I}
	}
	% \hspace{2mm}
	%	\hfil
	\subfloat[TOX171]{
		\includegraphics[width=1.35in]{figures/CAEAC_TOX171_ACC.pdf}
		\label{fig:PS_TOX171_I}
	}
	% \vspace{-1mm}
	\end{adjustwidth}
	\caption{Effects of the parameter specifications of CAEAC-I on Accuracy.}
	\label{fig:paramSensitivity_I}
\end{figure}

\begin{figure}[htbp]
	\begin{adjustwidth}{-\extralength}{0cm}
	\centering
	\subfloat[Aggregation]{
		\includegraphics[width=1.35in]{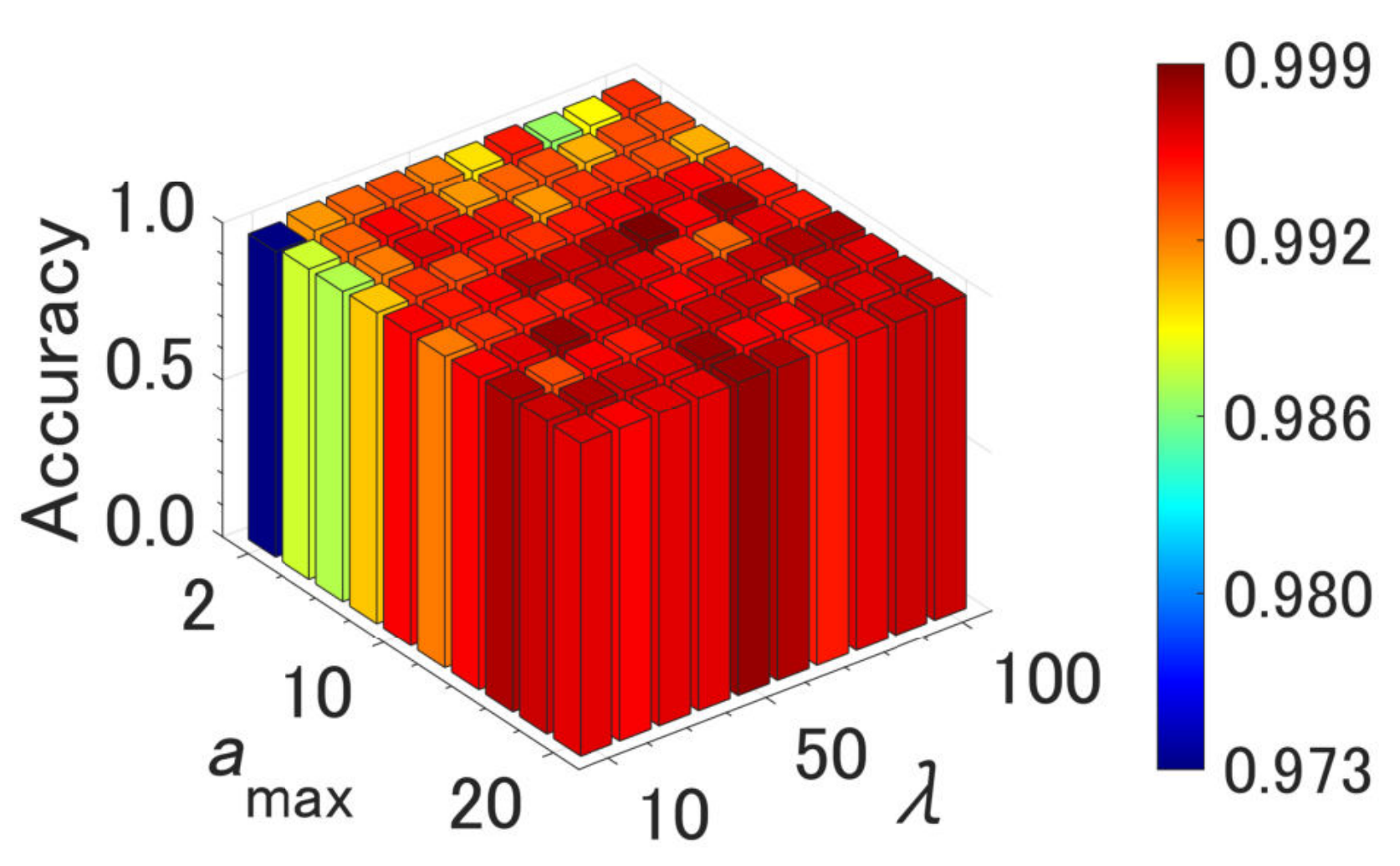}
		\label{fig:PS_Aggregation_C}
	}
	% \hspace{-1.4mm}
	% \hfil
	\subfloat[Compound]{
		\includegraphics[width=1.35in]{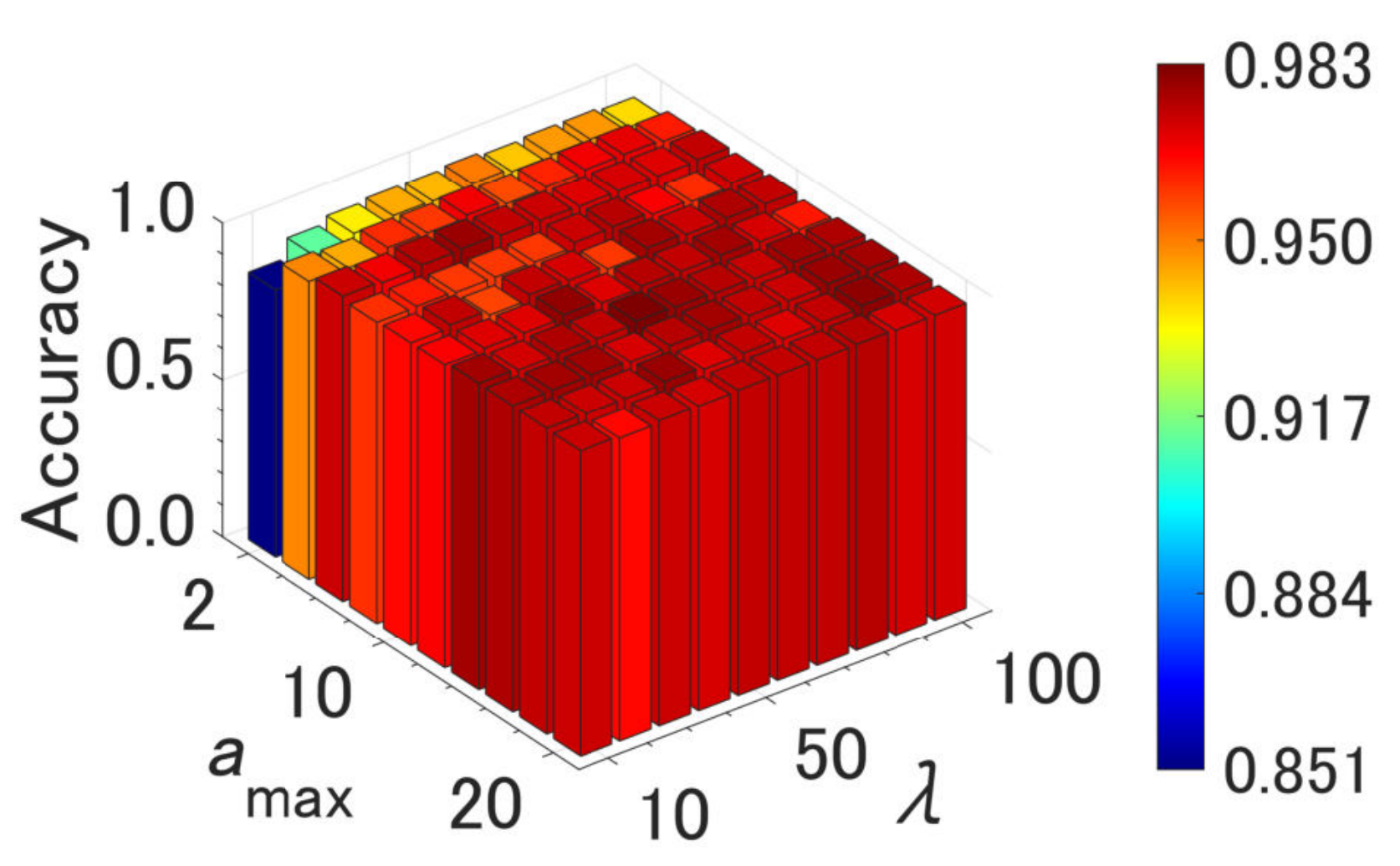}
		\label{fig:PS_Compound_C}
	}
	% \hspace{-1.4mm}
	% \hfil
	\subfloat[Hard Distribution]{
		\includegraphics[width=1.35in]{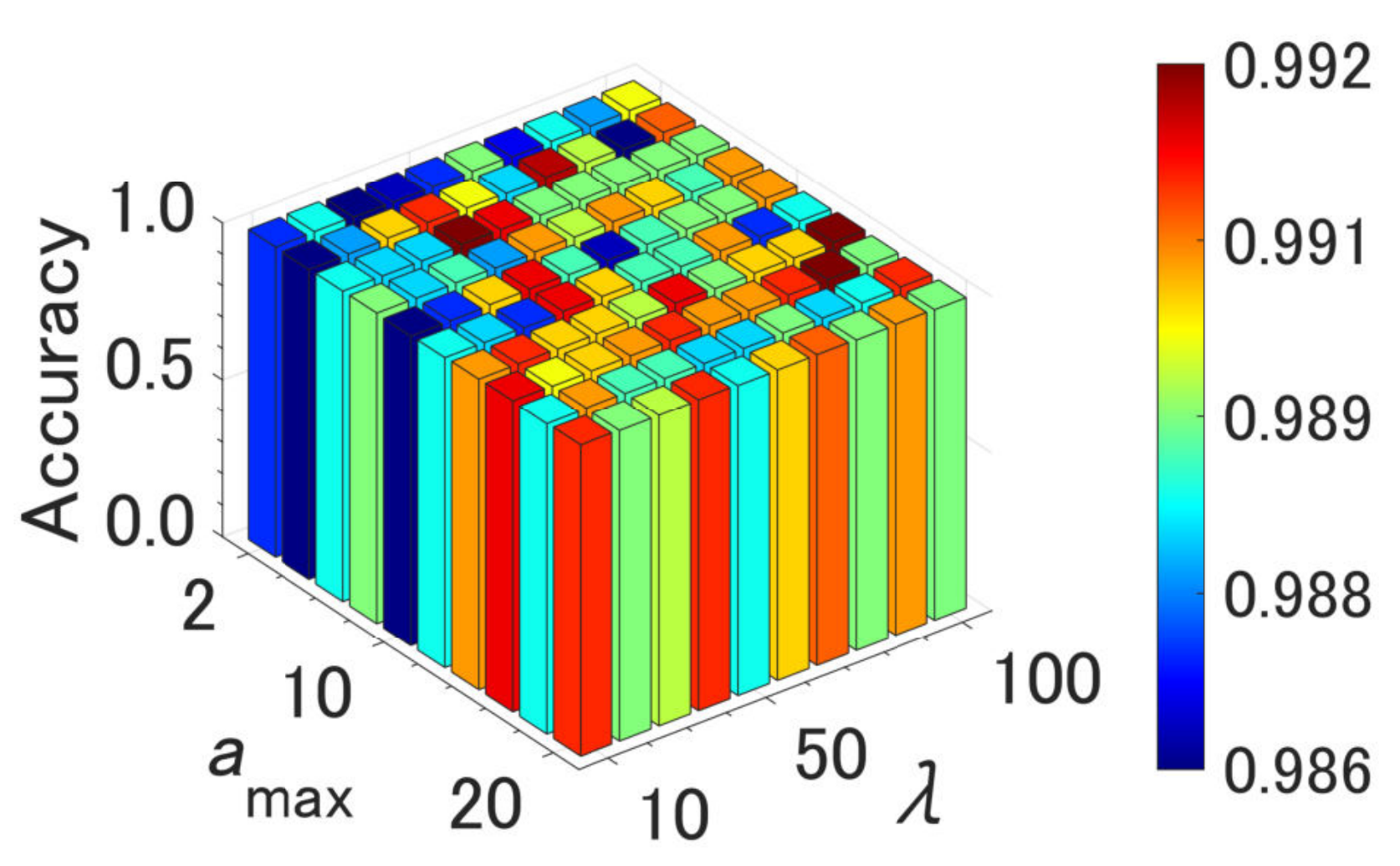}
		\label{fig:PS_HardDistribution_C}
	}
	% \hspace{-1.4mm}
	% \hfil
	\subfloat[Jain]{
		\includegraphics[width=1.35in]{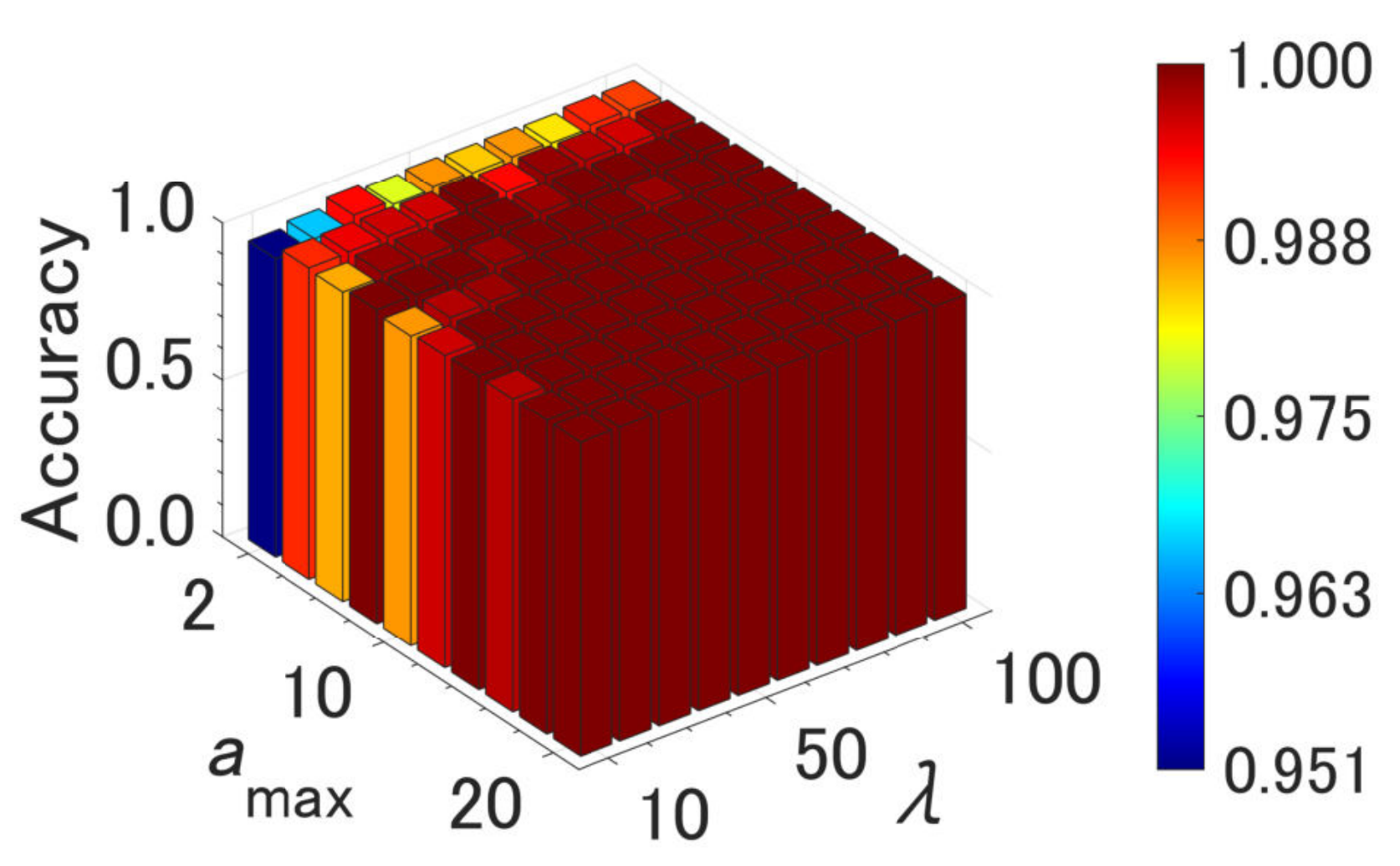}
		\label{fig:PS_Jain_C}
	}
	% \hspace{-1.4mm}
	% \hfil
	\subfloat[Pathbased]{
		\includegraphics[width=1.35in]{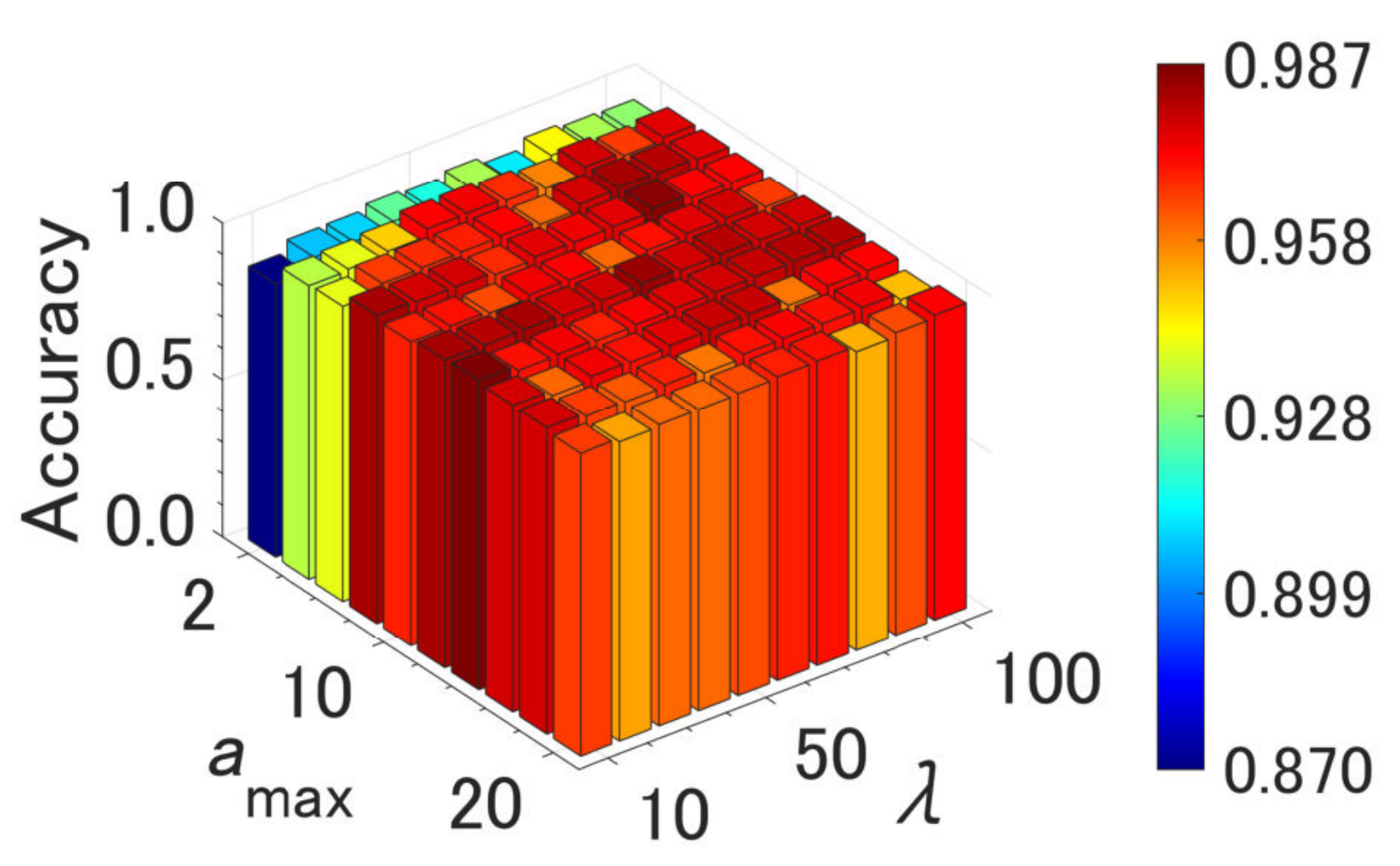}
		\label{fig:PS_Pathbased_C}
	}
	\\
	% \vspace{-2.5mm}
	%	\hfil
	% \hspace{-1.4mm}
	\subfloat[ALLAML]{
		\includegraphics[width=1.35in]{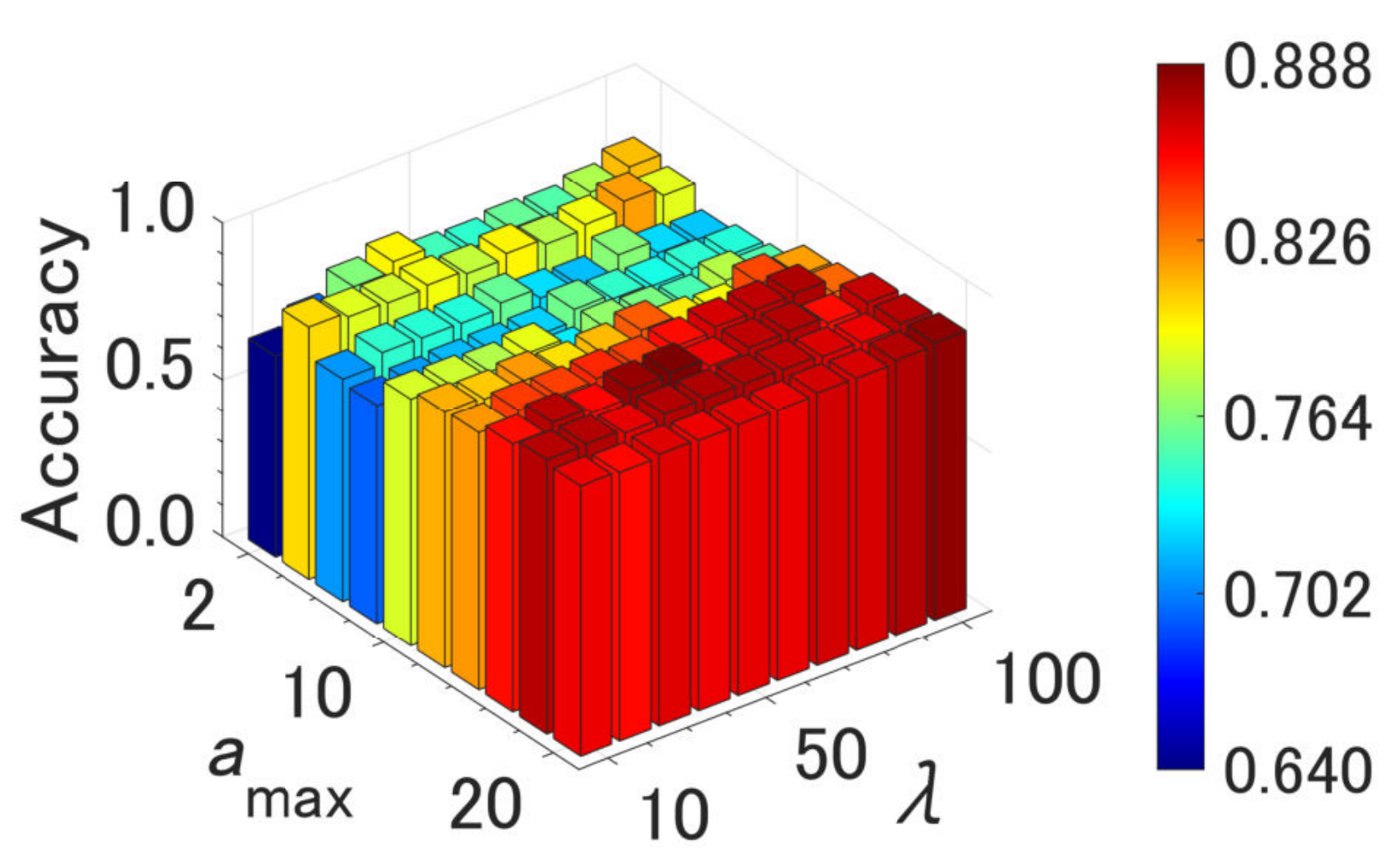}
		\label{fig:PS_ALLAML_C}
	}
	% \hspace{-1.4mm}
	% \hfil
	\subfloat[COIL20]{
		\includegraphics[width=1.35in]{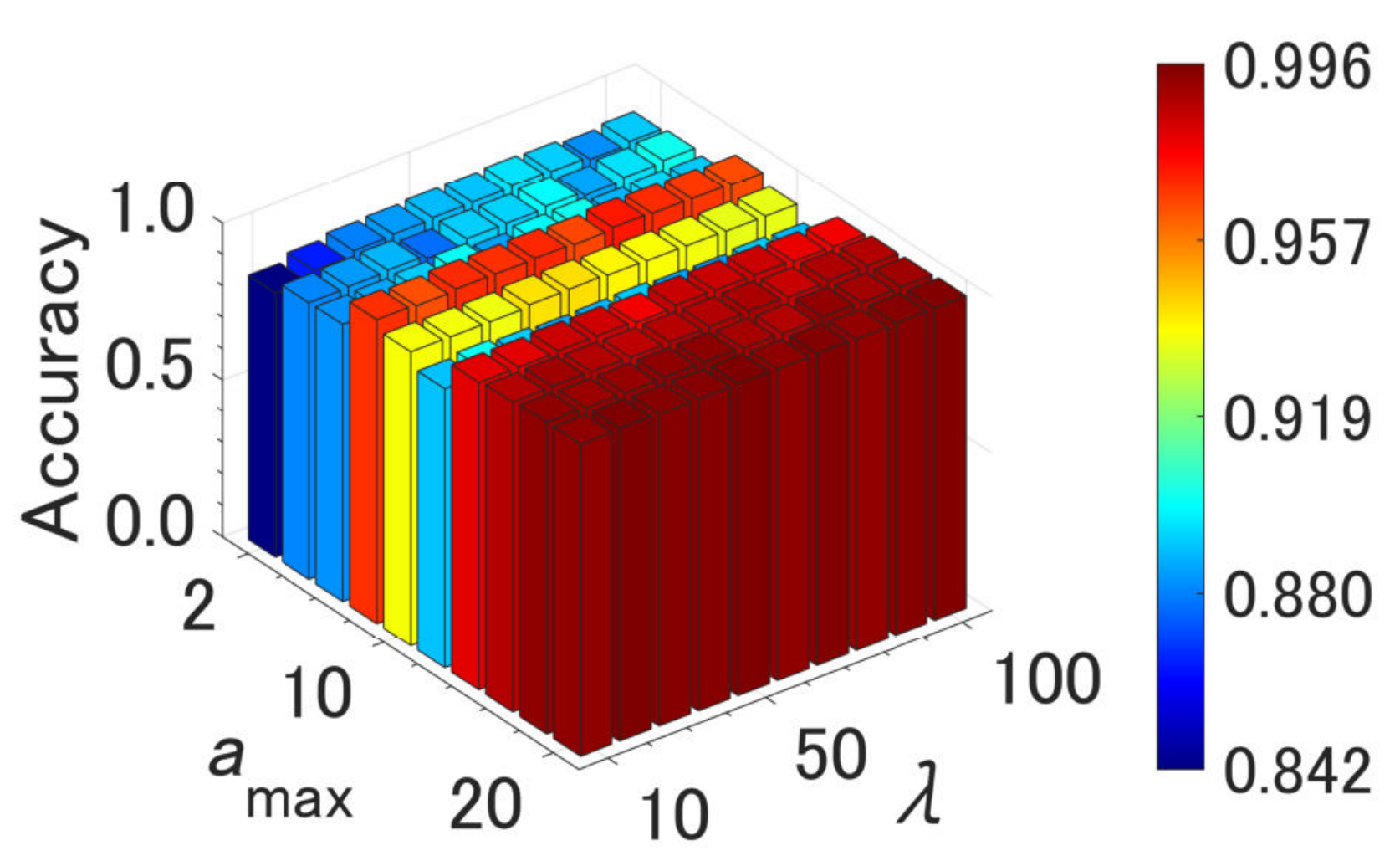}
		\label{fig:PS_COIL20_C}
	}
	% \hspace{-1.4mm}
	% \hfil
	\subfloat[Iris]{
		\includegraphics[width=1.35in]{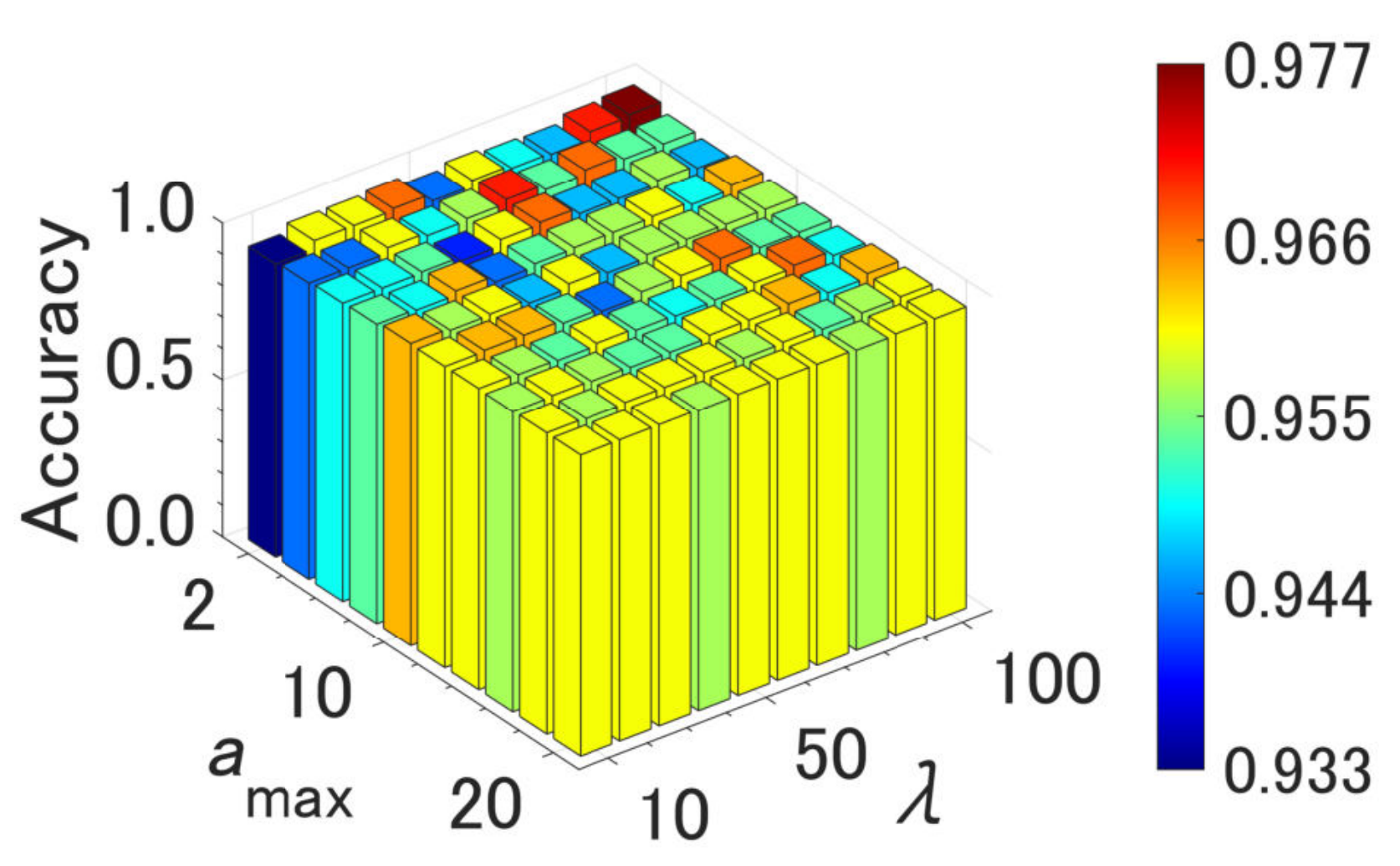}
		\label{fig:PS_Iris_C}
	}
	% \hspace{-1.4mm}
	% \hfil
	\subfloat[Isolet]{
		\includegraphics[width=1.35in]{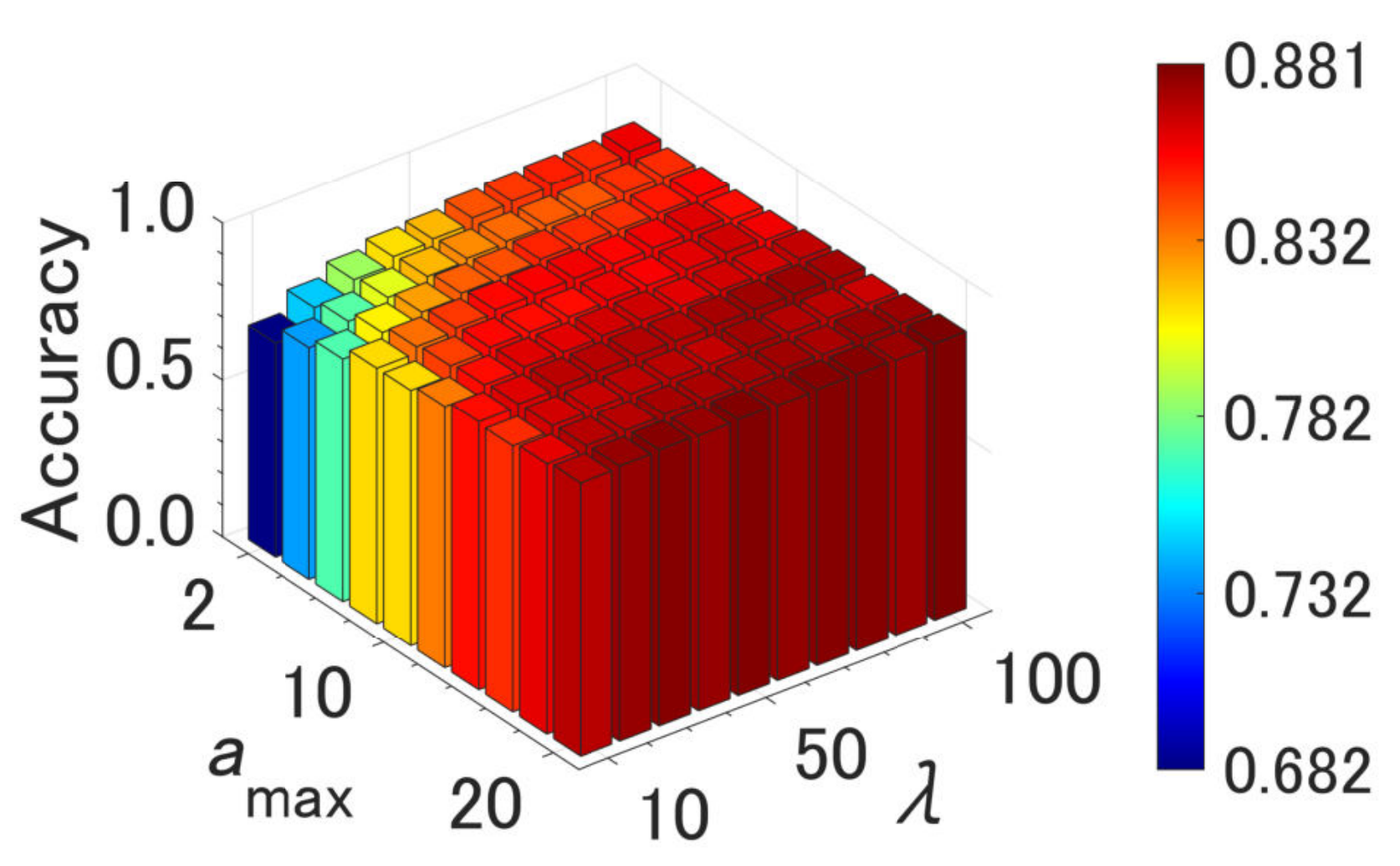}
		\label{fig:PS_Isolet_C}
	}
	% \hspace{-1.4mm}
	% \hfil
	\subfloat[OptDigits]{
		\includegraphics[width=1.35in]{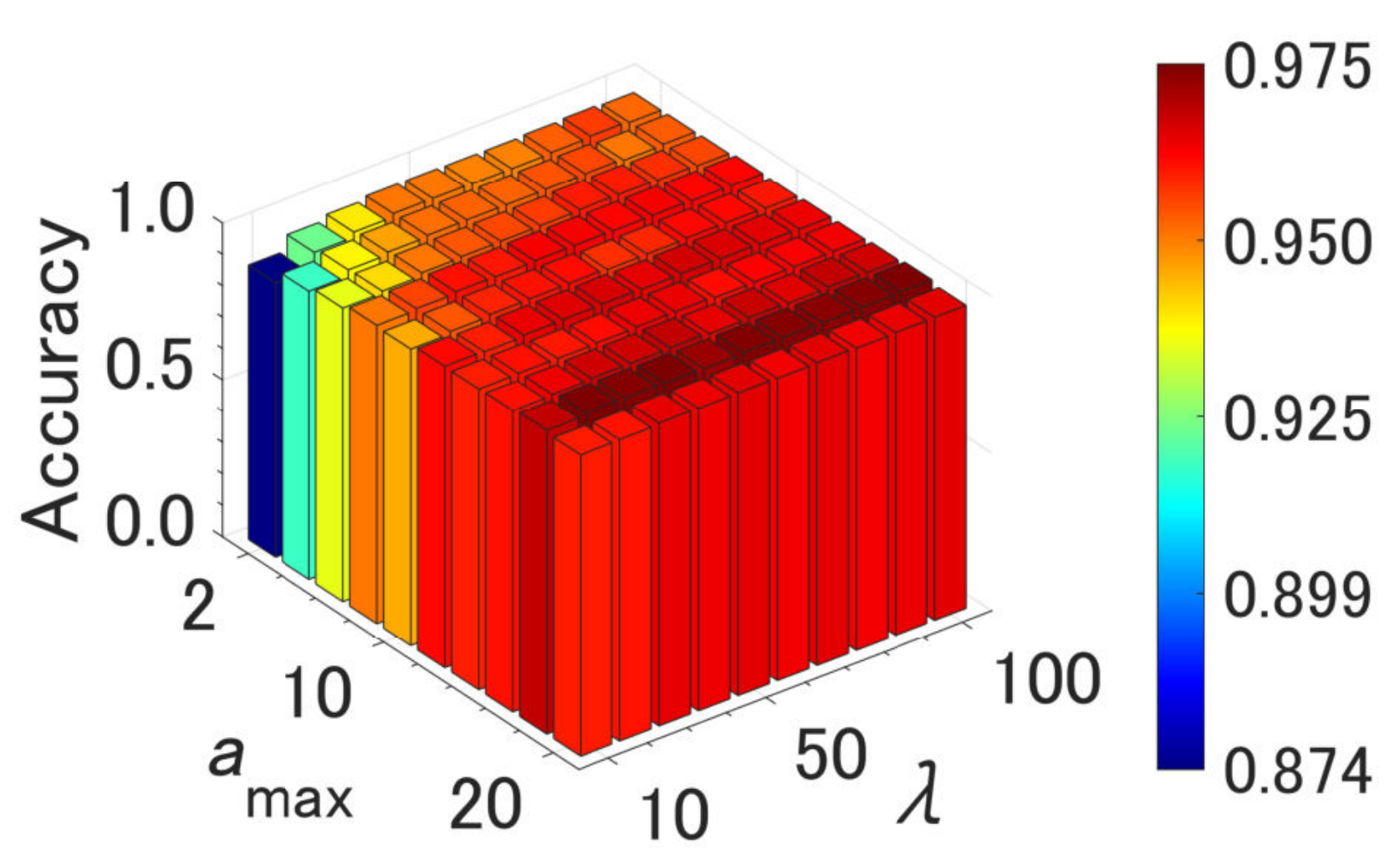}
		\label{fig:PS_OptDigits_C}
	}
	\\
	% \vspace{-2.5mm}
	%	\hfil
	\subfloat[Seeds]{
		\includegraphics[width=1.35in]{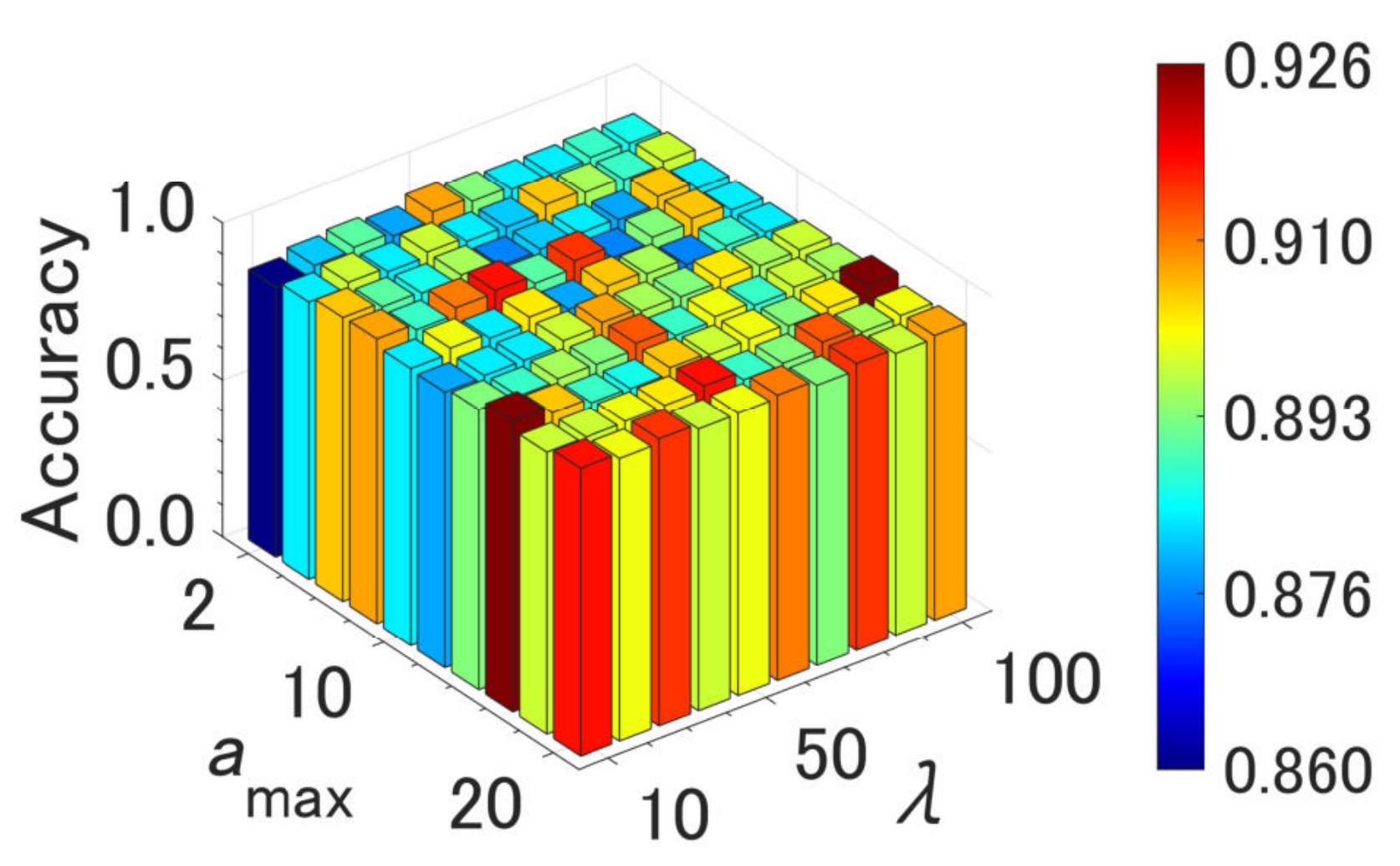}
		\label{fig:PS_Seeds_C}
	}
	% \hspace{2mm}
	%	\hfil
	\subfloat[Semeion]{
		\includegraphics[width=1.35in]{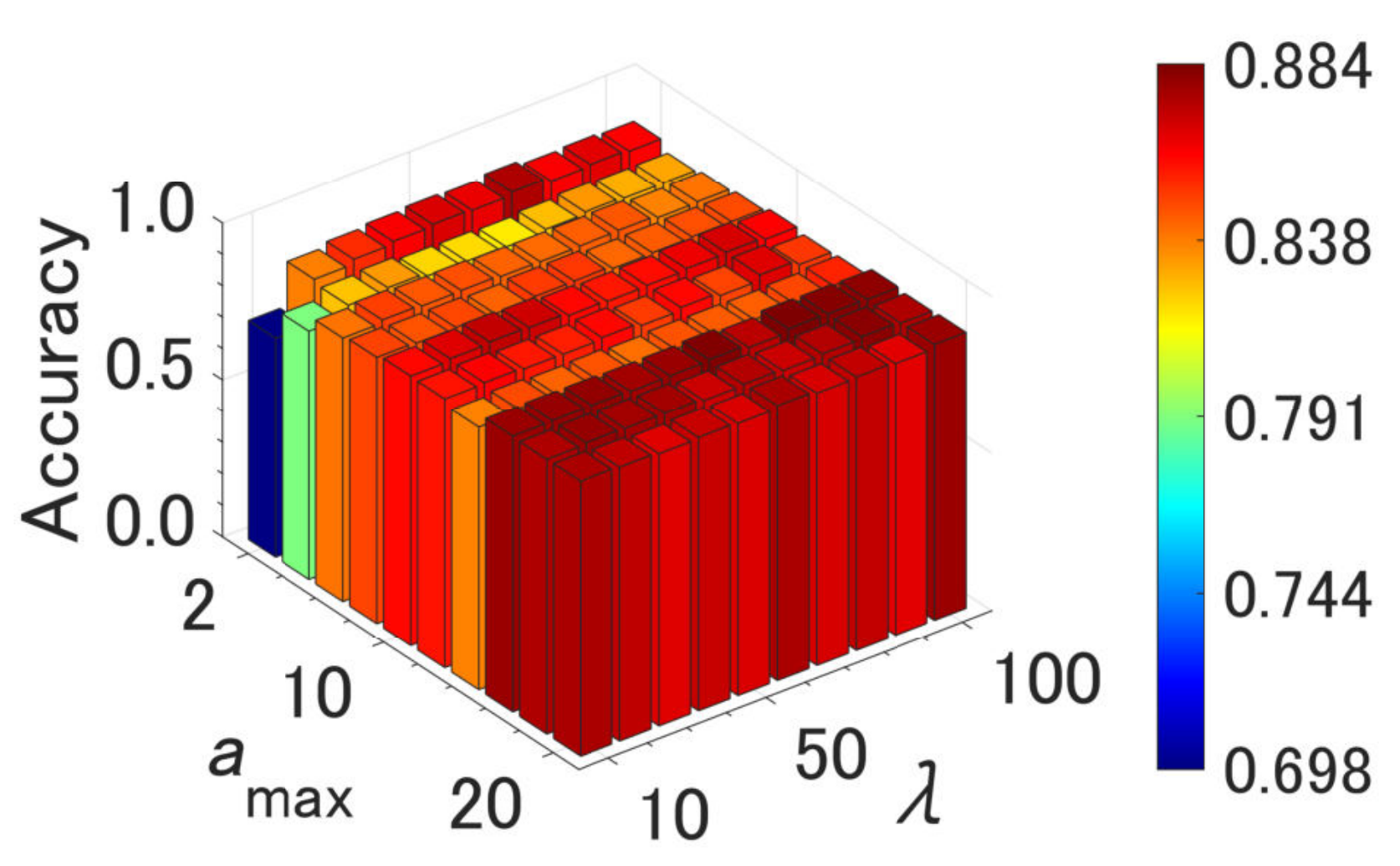}
		\label{fig:PS_Semeion_C}
	}
	% \hspace{2mm}
	%	\hfil
	\subfloat[Sonar]{
		\includegraphics[width=1.35in]{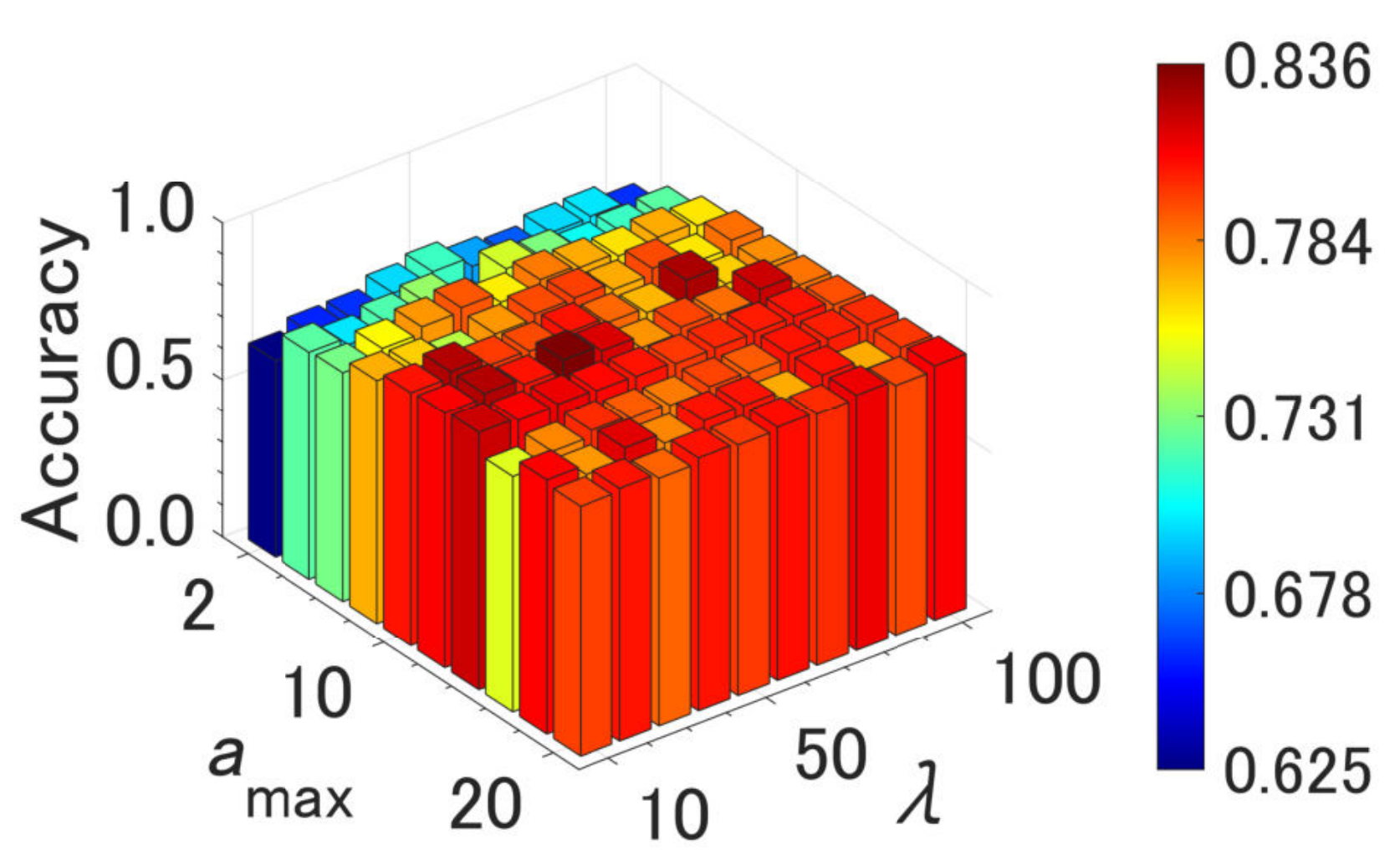}
		\label{fig:PS_Sonar_C}
	}
	% \hspace{2mm}
	%	\hfil
	\subfloat[TOX171]{
		\includegraphics[width=1.35in]{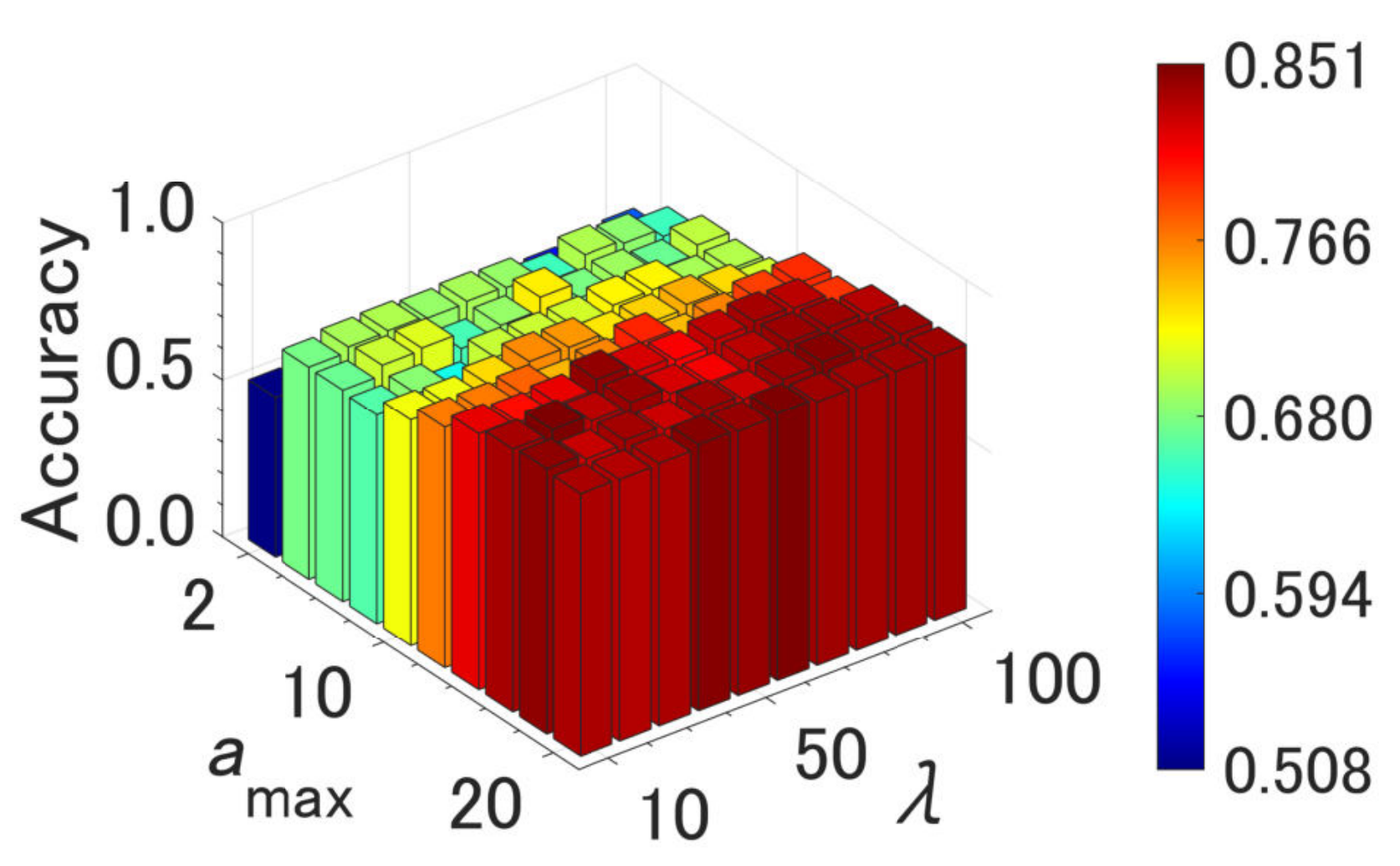}
		\label{fig:PS_TOX171_C}
	}
	% \vspace{-1mm}
	\end{adjustwidth}
	\caption{Effects of the parameter specifications of CAEAC-C on Accuracy.}
	\label{fig:paramSensitivity_C}
\end{figure}

Figs. \ref{fig:paramSensitivity_G}-\ref{fig:paramSensitivity_C} show effects of the parameter settings on Accuracy for CAEAC, CAEAC-I, and CAEAC-C, respectively. $ \lambda $ is the predefined interval for computing $ \sigma $ and deleting an isolated node, and $ a_{\text{max}}$ is the predefined threshold of an age of edge. As a general trend among CAEAC and its variants, there are no large undulations in the bar graphs for each parameter like FTCAC (i.e., Fig. \ref{fig:paramSensitivity_FTCAC}).

From the above observations, the sensitivity of the parameters of ASC, GSOINN+, CAEAC, CAEAC-I, and CAEAC-C are lower than FTCAC. Moreover, it can be considered that ASC, GSOINN+, CAEAC, CAEAC-I, and CAEAC-C can utilize the common parameter setting to a wide variety of datasets without difficulty of parameter specifications.

\section{Conclusion}
\label{sec:conclusion}
This paper proposed a supervised classification algorithm capable of continual learning by an ART-based growing self-organizing clustering algorithm, called CAEAC. In addition, its two variants, called CAEAC-I and CAEAC-C, are also proposed by modifying the CIM computation. Experimental results showed that CAEAC and its variants have the superior classification performance compared with the state-of-the-art clustering-based classification algorithms.

In regard to continual learning, the concepts of learned information sometimes change over time. This is known as concept drift \cite{lu18}. Although handling concept drift is an important capability in classification algorithms capable of continual learning, CAEAC focuses only on avoiding catastrophic forgetting. Thus, as a future research topic, we will focus on handling concept drift in CAEAC and its variants in order to improve the functionality of the algorithms.

\vspace{6pt} 

\authorcontributions{Conceptualization, N. Masuyama; methodology, N. Masuyama and Y. Nojima; software, N. Masuyama and F. Dawood; validation, N. Masuyama, N. Nojima, and F. Dawood; formal analysis, N. Masuyama and N. Nojima; investigation, N. Masuyama, F. Dawood, and Z. Liu; resources, N. Masuyama, F. Dawood, and Z. Liu; data curation, N. Masuyama, F. Dawood, and Z. Liu; writing---original draft preparation, N. Masuyama; writing---review and editing, N. Masuyama, and N. Nojima; visualization, N. Masuyama and N. Nojima; supervision, N. Masuyama; project administration, N. Masuyama; funding acquisition, N. Masuyama and N. Nojima. All authors have read and agreed to the published version of the manuscript.}

\funding{This work was supported by the Japan Society for the Promotion of Science (JSPS) KAKENHI Grant Number JP19K20358 and 22H03664.}

\acknowledgments{We would like to thank Mr. I. Tsubota (Fujitsu Ltd.) for his advice on the experimental design in carrying out this study.}

\begin{adjustwidth}{-\extralength}{0cm}

\reftitle{References}


\bibliography{myref}

\begin{thebibliography}{999}

\bibitem[McCloskey and Cohen(1989)]{mccloskey89}
McCloskey, M.; Cohen, N.J.
\newblock Catastrophic interference in connectionist networks: {T}he sequential
  learning problem.
\newblock {\em Psychology of Learning and Motivation} {\bf 1989}, {\em
  24},~109--165.

\bibitem[Parisi \em{et~al.}(2019)Parisi, Kemker, Part, Kanan, and
  Wermter]{parisi19}
Parisi, G.I.; Kemker, R.; Part, J.L.; Kanan, C.; Wermter, S.
\newblock Continual lifelong learning with neural networks: A review.
\newblock {\em Neural Networks} {\bf 2019}, {\em 113},~57--71.

\bibitem[Van~de Ven and Tolias(2019)]{van19}
Van~de Ven, G.M.; Tolias, A.S.
\newblock Three scenarios for continual learning.
\newblock {\em arXiv preprint arXiv:1904.07734} {\bf 2019}.

\bibitem[Wiewel and Yang(2019)]{wiewel19}
Wiewel, F.; Yang, B.
\newblock Localizing catastrophic forgetting in neural networks.
\newblock {\em arXiv preprint arXiv:1906.02568} {\bf 2019}.

\bibitem[Kirkpatrick \em{et~al.}(2017)Kirkpatrick, Pascanu, Rabinowitz, Veness,
  Desjardins, Rusu, Milan, Quan, Ramalho, Grabska-Barwinska,
  et~al.]{kirkpatrick17}
Kirkpatrick, J.; Pascanu, R.; Rabinowitz, N.; Veness, J.; Desjardins, G.; Rusu,
  A.A.; Milan, K.; Quan, J.; Ramalho, T.; Grabska-Barwinska, A.;  et~al.
\newblock Overcoming catastrophic forgetting in neural networks.
\newblock {\em Proceedings of the National Academy of Sciences} {\bf 2017},
  {\em 114},~3521--3526.

\bibitem[Furao and Hasegawa(2006)]{furao06}
Furao, S.; Hasegawa, O.
\newblock {An incremental network for on-line unsupervised classification and
  topology learning}.
\newblock {\em Neural Networks} {\bf 2006}, {\em 19},~90--106.

\bibitem[Fritzke(1995)]{fritzke95}
Fritzke, B.
\newblock {A growing neural gas network learns topologies}.
\newblock {\em Advances in Neural Information Processing Systems} {\bf 1995},
  {\em 7},~625--632.

\bibitem[Shen and Hasegawa(2008)]{shen08}
Shen, F.; Hasegawa, O.
\newblock {A fast nearest neighbor classifier based on self-organizing
  incremental neural network}.
\newblock {\em Neural Networks} {\bf 2008}, {\em 21},~1537--1547.

\bibitem[Parisi \em{et~al.}(2017)Parisi, Tani, Weber, and Wermter]{parisi17}
Parisi, G.I.; Tani, J.; Weber, C.; Wermter, S.
\newblock Lifelong learning of human actions with deep neural network
  self-organization.
\newblock {\em Neural Networks} {\bf 2017}, {\em 96},~137--149.

\bibitem[Wiwatcharakoses and Berrar(2021)]{wiwatcharakoses21}
Wiwatcharakoses, C.; Berrar, D.
\newblock A self-organizing incremental neural network for continual supervised
  learning.
\newblock {\em Expert Systems with Applications} {\bf 2021}, {\em 185},~115662.

\bibitem[Grossberg(1987)]{grossberg87}
Grossberg, S.
\newblock {Competitive learning: From interactive activation to adaptive
  resonance}.
\newblock {\em Cognitive Science} {\bf 1987}, {\em 11},~23--63.

\bibitem[Liu \em{et~al.}(2007)Liu, Pokharel, and Pr{\'\i}ncipe]{liu07}
Liu, W.; Pokharel, P.P.; Pr{\'\i}ncipe, J.C.
\newblock {Correntropy: Properties and applications in non-{G}aussian signal
  processing}.
\newblock {\em IEEE Transactions on Signal Processing} {\bf 2007}, {\em
  55},~5286--5298.

\bibitem[Chalasani and Pr{\'\i}ncipe(2015)]{chalasani15}
Chalasani, R.; Pr{\'\i}ncipe, J.C.
\newblock {Self-organizing maps with information theoretic learning}.
\newblock {\em Neurocomputing} {\bf 2015}, {\em 147},~3--14.

\bibitem[Masuyama \em{et~al.}(2019{\natexlab{a}})Masuyama, Loo, Ishibuchi,
  Kubota, Nojima, and Liu]{masuyama19b}
Masuyama, N.; Loo, C.K.; Ishibuchi, H.; Kubota, N.; Nojima, Y.; Liu, Y.
\newblock Topological Clustering via Adaptive Resonance Theory With Information
  Theoretic Learning.
\newblock {\em IEEE Access} {\bf 2019}, {\em 7},~76920--76936.

\bibitem[Masuyama \em{et~al.}(2019{\natexlab{b}})Masuyama, Amako, Nojima, Liu,
  Loo, and Ishibuchi]{masuyamaFTCA}
Masuyama, N.; Amako, N.; Nojima, Y.; Liu, Y.; Loo, C.K.; Ishibuchi, H.
\newblock Fast Topological Adaptive Resonance Theory Based on Correntropy
  Induced Metric.
\newblock In Proceedings of the Proceedings of IEEE Symposium Series on
  Computational Intelligence,  2019, pp. 2215--2221.

\bibitem[Masuyama \em{et~al.}(2022)Masuyama, Amako, Yamada, Nojima, and
  Ishibuchi]{masuyama22a}
Masuyama, N.; Amako, N.; Yamada, Y.; Nojima, Y.; Ishibuchi, H.
\newblock Adaptive resonance theory-based topological clustering with a
  divisive hierarchical structure capable of continual learning.
\newblock {\em arXiv preprint arXiv:2201.10713} {\bf 2022}.

\bibitem[McLachlan \em{et~al.}(2019)McLachlan, Lee, and
  Rathnayake]{mclachlan19}
McLachlan, G.J.; Lee, S.X.; Rathnayake, S.I.
\newblock Finite mixture models.
\newblock {\em Annual Review of Statistics and its Application} {\bf 2019},
  {\em 6},~355--378.

\bibitem[Lloyd(1982)]{lloyd82}
Lloyd, S.
\newblock {Least squares quantization in {PCM}}.
\newblock {\em IEEE Transactions on Information Theory} {\bf 1982}, {\em
  28},~129--137.

\bibitem[Carpenter and Grossberg(1988)]{carpenter88}
Carpenter, G.A.; Grossberg, S.
\newblock {The ART of adaptive pattern recognition by a self-organizing neural
  network}.
\newblock {\em Computer} {\bf 1988}, {\em 21},~77--88.

\bibitem[Wiwatcharakoses and Berrar(2020)]{wiwatcharakoses20}
Wiwatcharakoses, C.; Berrar, D.
\newblock {SOINN}+, a self-organizing incremental neural network for
  unsupervised learning from noisy data streams.
\newblock {\em Expert Systems with Applications} {\bf 2020}, {\em 143},~113069.

\bibitem[Marsland \em{et~al.}(2002)Marsland, Shapiro, and Nehmzow]{marsland02}
Marsland, S.; Shapiro, J.; Nehmzow, U.
\newblock {A self-organising network that grows when required}.
\newblock {\em Neural Networks} {\bf 2002}, {\em 15},~1041--1058.

\bibitem[Tan \em{et~al.}(2014)Tan, Watada, Ibrahim, and Khalid]{tan14}
Tan, S.C.; Watada, J.; Ibrahim, Z.; Khalid, M.
\newblock Evolutionary fuzzy {ARTMAP} neural networks for classification of
  semiconductor defects.
\newblock {\em IEEE Transactions on Neural Networks and Learning Systems} {\bf
  2014}, {\em 26},~933--950.

\bibitem[Matias and Neto(2018)]{matias18}
Matias, A.L.; Neto, A.R.R.
\newblock On{ARTMAP}: A fuzzy {ARTMAP}-based architecture.
\newblock {\em Neural Networks} {\bf 2018}, {\em 98},~236--250.

\bibitem[Matias \em{et~al.}(2021)Matias, Neto, Mattos, and Gomes]{matias21}
Matias, A.L.; Neto, A.R.R.; Mattos, C.L.C.; Gomes, J.P.P.
\newblock A novel fuzzy {ARTMAP} with area of influence.
\newblock {\em Neurocomputing} {\bf 2021}, {\em 432},~80--90.

\bibitem[Carpenter \em{et~al.}(1991)Carpenter, Grossberg, and
  Rosen]{carpenter91b}
Carpenter, G.A.; Grossberg, S.; Rosen, D.B.
\newblock {Fuzzy {ART}: Fast stable learning and categorization of analog
  patterns by an adaptive resonance system}.
\newblock {\em Neural Networks} {\bf 1991}, {\em 4},~759--771.

\bibitem[Vigdor and Lerner(2007)]{vigdor07}
Vigdor, B.; Lerner, B.
\newblock {The {B}ayesian {ARTMAP}}.
\newblock {\em IEEE Transactions on Neural Networks} {\bf 2007}, {\em
  18},~1628--1644.

\bibitem[Wang \em{et~al.}(2019)Wang, Zhu, Meng, and He]{wang19}
Wang, L.; Zhu, H.; Meng, J.; He, W.
\newblock Incremental Local Distribution-Based Clustering Using {B}ayesian
  Adaptive Resonance Theory.
\newblock {\em IEEE Transactions on Neural Networks and Learning Systems} {\bf
  2019}, {\em 30},~3496--3504.

\bibitem[da~Silva \em{et~al.}(2020)da~Silva, Elnabarawy, and Wunsch~II]{da20}
da~Silva, L.E.B.; Elnabarawy, I.; Wunsch~II, D.C.
\newblock Distributed dual vigilance fuzzy adaptive resonance theory learns
  online, retrieves arbitrarily-shaped clusters, and mitigates order
  dependence.
\newblock {\em Neural Networks} {\bf 2020}, {\em 121},~208--228.

\bibitem[Masuyama \em{et~al.}(2018)Masuyama, Loo, and Dawood]{masuyama18}
Masuyama, N.; Loo, C.K.; Dawood, F.
\newblock {Kernel {B}ayesian {ART} and {ARTMAP}}.
\newblock {\em Neural Networks} {\bf 2018}, {\em 98},~76--86.

\bibitem[Masuyama \em{et~al.}(2019)Masuyama, Loo, and Wermter]{masuyama19a}
Masuyama, N.; Loo, C.K.; Wermter, S.
\newblock A Kernel {B}ayesian Adaptive Resonance Theory with a Topological
  Structure.
\newblock {\em International Journal of Neural Systems} {\bf 2019}, {\em
  29},~1850052 (20 pages).

\bibitem[da~Silva \em{et~al.}(2019)da~Silva, Elnabarawy, and Wunsch~II]{da19}
da~Silva, L.E.B.; Elnabarawy, I.; Wunsch~II, D.C.
\newblock Dual vigilance fuzzy adaptive resonance theory.
\newblock {\em Neural Networks} {\bf 2019}, {\em 109},~1--5.

\bibitem[Belouadah \em{et~al.}(2021)Belouadah, Popescu, and
  Kanellos]{belouadah20}
Belouadah, E.; Popescu, A.; Kanellos, I.
\newblock A comprehensive study of class incremental learning algorithms for
  visual tasks.
\newblock {\em Neural Networks} {\bf 2021}, {\em 135},~38--54.

\bibitem[Zenke \em{et~al.}(2017)Zenke, Poole, and Ganguli]{zenke17}
Zenke, F.; Poole, B.; Ganguli, S.
\newblock Continual learning through synaptic intelligence.
\newblock In Proceedings of the Proceedings of International Conference on
  Machine Learning,  2017, pp. 3987--3995.

\bibitem[Shin \em{et~al.}(2017)Shin, Lee, Kim, and Kim]{shin2017}
Shin, H.; Lee, J.K.; Kim, J.; Kim, J.
\newblock Continual learning with deep generative replay.
\newblock In Proceedings of the Proceedings of the 31st International
  Conference on Neural Information Processing Systems,  2017, pp. 2994--3003.

\bibitem[Nguyen \em{et~al.}(2018)Nguyen, Li, Bui, and Turner]{nguyen2018}
Nguyen, C.V.; Li, Y.; Bui, T.D.; Turner, R.E.
\newblock Variational continual learning.
\newblock In Proceedings of the Proceedings of International Conference on
  Learning Representations,  2018, pp. 1--18.

\bibitem[Tahir and Loo(2020)]{tahir20}
Tahir, G.A.; Loo, C.K.
\newblock An open-ended continual learning for food recognition using class
  incremental extreme learning machines.
\newblock {\em IEEE Access} {\bf 2020}, {\em 8},~82328--82346.

\bibitem[Kongsorot \em{et~al.}(2020)Kongsorot, Horata, and
  Musikawan]{kongsorot20}
Kongsorot, Y.; Horata, P.; Musikawan, P.
\newblock An incremental kernel extreme learning machine for multi-label
  learning with emerging new labels.
\newblock {\em IEEE Access} {\bf 2020}, {\em 8},~46055--46070.

\bibitem[Parisi \em{et~al.}(2018)Parisi, Tani, Weber, and Wermter]{parisi18}
Parisi, G.I.; Tani, J.; Weber, C.; Wermter, S.
\newblock Lifelong learning of spatiotemporal representations with dual-memory
  recurrent self-organization.
\newblock {\em Frontiers in Neurorobotics} {\bf 2018}, {\em 12},~78.

\bibitem[Carpenter \em{et~al.}(1992)Carpenter, Grossberg, Markuzon, Reynolds,
  and Rosen]{carpenter92}
Carpenter, G.A.; Grossberg, S.; Markuzon, N.; Reynolds, J.H.; Rosen, D.B.
\newblock {Fuzzy {ARTMAP}: A neural network architecture for incremental
  supervised learning of analog multidimensional maps}.
\newblock {\em IEEE Transactions on Neural Networks} {\bf 1992}, {\em
  3},~698--713.

\bibitem[Henderson and Parmeter(2012)]{henderson12}
Henderson, D.J.; Parmeter, C.F.
\newblock Normal reference bandwidths for the general order, multivariate
  kernel density derivative estimator.
\newblock {\em Statistics \& Probability Letters} {\bf 2012}, {\em
  82},~2198--2205.

\bibitem[Silverman(2018)]{silverman18}
Silverman, B.W.
\newblock {\em Density estimation for statistics and data analysis}; Routledge,
   2018.

\bibitem[Masuyama \em{et~al.}(2020)Masuyama, Nojima, Loo, and
  Ishibuchi]{masuyama20a}
Masuyama, N.; Nojima, Y.; Loo, C.K.; Ishibuchi, H.
\newblock Multi-label classification based on adaptive resonance theory.
\newblock In Proceedings of the Proceedins of 2020 IEEE Symposium Series on
  Computational Intelligence,  2020, pp. 1913--1920.

\bibitem[Fr{\"a}nti and Sieranoja(2018)]{franti18}
Fr{\"a}nti, P.; Sieranoja, S.
\newblock K-means properties on six clustering benchmark datasets.
\newblock {\em Applied Intelligence} {\bf 2018}, {\em 48},~4743--4759.

\bibitem[Liu \em{et~al.}(2017)Liu, Wang, Liu, Zeng, Liu, and Alsaadi]{liu17}
Liu, W.; Wang, Z.; Liu, X.; Zeng, N.; Liu, Y.; Alsaadi, F.E.
\newblock {A survey of deep neural network architectures and their
  applications}.
\newblock {\em Neurocomputing} {\bf 2017}, {\em 234},~11--26.

\bibitem[Dua and Graff(2019)]{dua19}
Dua, D.; Graff, C.
\newblock {UCI} machine learning repository.
\newblock University of California, Irvine, School of Information and Computer
  Sciences,  2019.

\bibitem[Strehl and Ghosh(2002)]{strehl02}
Strehl, A.; Ghosh, J.
\newblock {Cluster ensembles---A knowledge reuse framework for combining
  multiple partitions}.
\newblock {\em Journal of Machine Learning Research} {\bf 2002}, {\em
  3},~583--617.

\bibitem[Hubert and Arabie(1985)]{hubert85}
Hubert, L.; Arabie, P.
\newblock {Comparing partitions}.
\newblock {\em Journal of Classification} {\bf 1985}, {\em 2},~193--218.

\bibitem[Dem{\v{s}}ar(2006)]{demvsar06}
Dem{\v{s}}ar, J.
\newblock Statistical comparisons of classifiers over multiple data sets.
\newblock {\em Journal of Machine Learning Research} {\bf 2006}, {\em
  7},~1--30.

\bibitem[Lu \em{et~al.}(2018)Lu, Liu, Dong, Gu, Gama, and Zhang]{lu18}
Lu, J.; Liu, A.; Dong, F.; Gu, F.; Gama, J.; Zhang, G.
\newblock Learning under concept drift: {A} review.
\newblock {\em IEEE Transactions on Knowledge and Data Engineering} {\bf 2018},
  {\em 31},~2346--2363.

\end{thebibliography}

\PublishersNote{}
\end{adjustwidth}
\end{document}